\documentclass{article}
\usepackage{scicite}
\usepackage{helvet}
\usepackage{times}
\usepackage[margin=1.25in]{geometry}
\usepackage{hyperref}
\usepackage{xcolor}
\definecolor{darkblue}{rgb}{0, 0, 0.5}
\hypersetup{colorlinks=true,citecolor=darkblue, linkcolor=darkblue, urlcolor=darkblue}
\usepackage{graphicx}
\usepackage{tikz}
\usetikzlibrary{shapes}
\usetikzlibrary{arrows.meta}
\usetikzlibrary{fadings}
\usepackage{pgfplots}
\usepackage{subcaption}
\usepackage{bm}
\usepackage{amsmath}
\usepackage{amsfonts}
\usepackage{pifont}
\usepackage{algorithm}
\usepackage{algorithmic}
\usepackage[percent]{overpic}
\usepackage{multirow}
\usepackage[above]{placeins}
\usepackage[font={footnotesize,sf},labelfont={bf}]{caption}
\usepackage[page]{appendix}
\usepackage{sectsty}
\usepackage{lineno}

\parttitlefont{\sffamily}
\chaptertitlefont{\sffamily}
\allsectionsfont{\sffamily}

\graphicspath{ {img/} }

\newcommand{\defeq}{\buildrel\rm {def}\over=}
\DeclareMathAlphabet{\mathpzc}{OT1}{pzc}{m}{it}

\newcommand{\R}{\mathbb{R}}

\newcommand{\X}{\mathbf{X}}

\newcommand{\G}{\mathbf{G}}

\newcommand{\Xprime}{\mathbf{X}'}

\newcommand{\F}{\mathcal{F}}
\newcommand{\D}{\mathcal{D}}
\newcommand{\x}{\mathbf{x}}
\newcommand{\xprime}{\mathbf{x'}}
\newcommand{\y}{\mathbf{y}}

\renewcommand{\b}{{\mathbf{b}}}

\renewcommand{\d}{{\mathbf{d}}}

\renewcommand{\t}{\mathbf{t}}

\renewcommand{\t}{\mathbf{t}}

\newcommand{\V}{\mathbf{V}}

\newcommand{\W}{\mathbf{W}}
\newcommand{\z}{\mathbf{z}}

\newcommand{\s}{\mathbf{s}}
\newcommand{\snull}{\mathbf{s_0}}
\newcommand{\fin}{f_{\text{in}}}
\newcommand{\uin}{\u_{\text{in}}}
\newcommand{\firf}{f_{\text{IRF}}}
\newcommand{\uirf}{\u_{\text{IRF}}}
\newcommand{\fff}{f_{\text{FF}}}
\newcommand{\frnn}{f_{\text{RNN}}}

\renewcommand{\u}{\mathbf{u}}
\newcommand{\vv}{\mathbf{v}}
\newcommand{\w}{\mathbf{w}}

\definecolor{5gram}{rgb}{1.0, 0.0, 0.16}
\definecolor{rate}{rgb}{1.0, 0.5161632220455751, 0.0}
\definecolor{wlen}{rgb}{0.8055113937466878, 1.0, 0.0}
\definecolor{unigram}{rgb}{0.12718600953895065, 1.0, 0.0}
\definecolor{sentpos}{rgb}{0.0, 1.0, 0.5481762597512125}
\definecolor{wdelta}{rgb}{0.0, 0.774722932651321, 1.0}
\definecolor{prevwasfix}{rgb}{0.0, 0.09271099744245515, 1.0}
\definecolor{trial}{rgb}{0.5893009377664112, 0.0, 1.0}

\definecolor{c1}{rgb}{1.0, 0.0, 0.16}
\definecolor{c2}{rgb}{1.0, 0.2829888712241654, 0.0}
\definecolor{c3}{rgb}{1.0, 0.7281399046104929, 0.0}
\definecolor{c4}{rgb}{0.8055113937466878, 1.0, 0.0}
\definecolor{c5}{rgb}{0.36036036036036034, 1.0, 0.0}
\definecolor{c6}{rgb}{0.0, 1.0, 0.08433480919249393}
\definecolor{c7}{rgb}{0.0, 1.0, 0.5481762597512125}
\definecolor{c8}{rgb}{0.0, 1.0, 0.9909340080118073}
\definecolor{c9}{rgb}{0.0, 0.5615942028985503, 1.0}
\definecolor{c10}{rgb}{0.0, 0.09271099744245515, 1.0}
\definecolor{c11}{rgb}{0.35485933503836337, 0.0, 1.0}
\definecolor{c12}{rgb}{0.8024296675191819, 0.0, 1.0}

\newcommand{\kone}{\tikz[baseline=-0.5ex]{\draw[c1,line width=1pt] (-0.2, 0) -- (0.2, 0); \node[c1] at (0, 0) {$\bullet$};}}
\newcommand{\ktwo}{\tikz[baseline=-0.5ex]{\draw[c2,line width=1pt] (-0.2, 0) -- (0.2, 0); \node[c2] at (0, 0) {};}}
\newcommand{\kthree}{\tikz[baseline=-0.5ex]{\draw[c3,line width=1pt] (-0.2, 0) -- (0.2, 0); \node[draw=c3,fill=c3,circle,minimum
width=0.15cm,minimum height=0.15cm,inner sep=0pt] at (0, 0) {};}}
\newcommand{\kfour}{\tikz[baseline=-0.5ex]{\draw[c4,line width=1pt] (-0.2, 0) -- (0.2, 0); \node[draw=c4,fill=c4,isosceles
triangle,isosceles triangle stretches,shape border rotate=90,minimum
width=0.15cm,minimum height=0.15cm,inner sep=0pt,rotate=180] at (0, 0) {};}}
\newcommand{\kfive}{\tikz[baseline=-0.5ex]{\draw[c5,line width=1pt] (-0.2, 0) -- (0.2, 0); \node[draw=c5,fill=c5,isosceles triangle,isosceles triangle stretches,shape border rotate=90,minimum
width=0.15cm,minimum height=0.15cm,inner sep=0pt] at (0, 0) {};}}
\newcommand{\ksix}{\tikz[baseline=-0.5ex]{\draw[c6,line width=1pt] (-0.2, 0) -- (0.2, 0); \node[draw=c6,fill=c6,isosceles triangle,isosceles triangle stretches,shape border rotate=90,minimum
width=0.2cm,minimum height=0.2cm,inner sep=0pt, rotate=90] at (0, 0) {};}}
\newcommand{\kseven}{\tikz[baseline=-0.5ex]{\draw[c7,line width=1pt] (-0.2, 0) -- (0.2, 0); \node[draw=c7,fill=c7,isosceles triangle,isosceles triangle stretches,shape border rotate=90,minimum
width=0.2cm,minimum height=0.2cm,inner sep=0pt,rotate=-90] at (0, 0) {};}}
\newcommand{\keight}{\tikz[baseline=-0.5ex]{\draw[c8,line width=1pt] (-0.2, 0) -- (0.2, 0); \draw[c8,line width=0.75pt] (0, 0) -- (0.10825, 0.0625);\draw[c8,line width=0.75pt] (0, 0) -- (-0.10825, 0.0625);\draw[c8,line width=0.75pt] (0, 0) -- (0, -0.125);}}
\newcommand{\knine}{\tikz[baseline=-0.5ex]{\draw[c9,line width=1pt] (-0.2, 0) -- (0.2, 0); \draw[c9,line width=0.75pt] (0, 0) -- (0.10825, -0.0625);\draw[c9,line width=0.75pt] (0, 0) -- (-0.10825, -0.0625);\draw[c9,line width=0.75pt] (0, 0) -- (0, 0.125);}}
\newcommand{\kten}{\tikz[baseline=-0.5ex]{\draw[c10,line width=1pt] (-0.2, 0) -- (0.2, 0); \draw[c10,line width=0.75pt] (0, 0) -- (0.0625, 0.10825);\draw[c10,line width=0.75pt] (0, 0) -- (0.0625, -0.10825);\draw[c10,line width=0.75pt] (0, 0) -- (-0.125, 0);}}
\newcommand{\keleven}{\tikz[baseline=-0.5ex]{\draw[c11,line width=1pt] (-0.2, 0) -- (0.2, 0); \draw[c11,line width=0.75pt] (0, 0) -- (-0.0625, 0.10825);\draw[c11,line width=0.75pt] (0, 0) -- (-0.0625, -0.10825);\draw[c11,line width=0.75pt] (0, 0) -- (0.125, 0);}}
\newcommand{\ktwelve}{\tikz[baseline=-0.5ex]{\draw[c12,line width=1pt] (-0.2, 0) -- (0.2, 0); \node[draw=c12,fill=c12,regular polygon,regular polygon sides=8,minimum
width=0.15cm,minimum height=0.15cm,inner sep=0pt] at (0, 0) {};}}

\begin{document}

\begin{tabular}{rp{.7\textwidth}}
    \textbf{Title:} & A Deep Learning Approach to Analyzing Continuous-Time Systems\newline\\
    \textbf{Short Title:} & Deep Learning Analysis of Continuous Time Series\newline\\
    \textbf{Authors:} & Cory Shain\textsuperscript{1} and William Schuler\textsuperscript{2}\newline\\
    \textbf{Affiliations:} & \textsuperscript{1}Massachusetts Institute of Technology
        \newline Department of Brain \& Cognitive Sciences
        \newline 43 Vassar St.
        \newline Cambridge, MA 02139 USA\newline
        \newline\textsuperscript{2}The Ohio State University
        \newline Department of Linguistics
        \newline 1217 Neil Ave.
        \newline Columbus, OH 43210 USA\newline\\
    \textbf{Corresponding Author:} & Cory Shain\newline
        \newline Massachusetts Institute of Technology
        \newline Department of Brain \& Cognitive Sciences
        \newline 43 Vassar St.
        \newline Cambridge, MA 02139 USA\newline
        \newline +1 (617) 253-5748
        \newline \url{cshain@mit.edu}\newline\\
    \textbf{Keywords:} & Deep learning; Time series; Data analysis; Nonlinear regression; Human language processing
\end{tabular}
\pagebreak

\tableofcontents
\pagebreak

\section*{Abstract}

Scientists often use observational time series data to study complex natural processes, but regression analyses often assume simplistic dynamics.
Recent advances in deep learning have yielded startling improvements to the performance of models of complex processes, but deep learning is generally not used for scientific analysis.
Here we show that deep learning can be used to analyze complex processes, providing flexible function approximation while preserving interpretability.
Our approach relaxes standard simplifying assumptions (e.g., linearity, stationarity, and homoscedasticity) that are implausible for many natural systems and may critically affect the interpretation of data.
We evaluate our model on incremental human language processing, a domain with complex continuous dynamics.
We demonstrate substantial improvements on behavioral and neuroimaging data, and we show that our model enables discovery of novel patterns in exploratory analyses, controls for diverse confounds in confirmatory analyses, and opens up research questions that are otherwise hard to study.

\pagebreak

\section*{Teaser}

We propose a flexible and interpretable way of using deep neural networks for scientific analyses of time series data.
\pagebreak

\section{Introduction}
Nature is full of complex ripple effects for which experimental intervention is impossible, impractical, or potentially distortionary.
How does deforestation affect carbon levels \cite{houghtonetal00}?
Do economic shocks lead to civil conflict \cite{migueletal04}?
How are the meanings of words in a podcast processed by the brain \cite{huthetal16}?
To address such questions, scientists collect and analyze \textit{observational time series data} that encode the response of a complex system to external forces (predictors).
This paradigm shifts the burden of design effort from experimental setup to data analysis.
Critical variables must be appropriately coded, control variables must cover plausible confounds, and statistical models must be expressive enough to capture the structure of the underlying process.

In practice, regression analyses of observational time series data are overwhelmingly based on linear models or generalizations thereof \cite{galton1886,sims80,hamilton94,wood06,batesetal15}.
These approaches make the following simplifying assumptions in some combination: time passes in discrete steps, effects are linear, and the response is stationary (time-invariant) and homoscedastic (constant variance).
Are these assumptions always appropriate for observational time series?
One reason for doubt comes from startling recent advances using deep neural networks (DNNs, i.e.\ artificial neural networks with at least one hidden layer; \textit{\citen{lecunetal15}}) to capture complex processes, from speech comprehension \cite{gravesetal13} to fusion reactions \cite{degraveetal22} to StarCraft II gameplay \cite{vinyalsetal19}.
These gains come from DNNs' ability to integrate information nonlinearly, discover hidden structure, and adapt to context \cite{lecunetal15}, abilities which are absent from standard regression analyses but are plausibly present in the modeled processes.
Nevertheless, DNNs are rarely used for scientific data analysis because they are ``black boxes'': they can accurately map inputs to outputs, but the computations they use to do so are opaque and therefore of limited value for understanding the modeled system.

Here we show that an appropriate combination of DNN design and black box interpretation can overcome this issue, synthesizing the flexibility of deep learning with the interpretability of linear regression.
Our approach---the continuous-time deconvolutional regressive neural network (CDRNN)---uses deep learning to relax the key simplifying assumptions above (discrete time, linearity, stationarity, and homoscedasticity) in order to estimate, visualize, and test properties of the response structure of a complex process from data.
We evaluate this proposal on data from real-time naturalistic language processing, a continuous-time process in which the mind is managing a large space of variables (including not only linguistic information, like semantic, syntactic, lexical, and phonological structure, but also audiovisual perception, world knowledge, episodic memory, and social information), each with myriad structural and statistical relationships to aspects of the local context \cite{taylor53,warren70,ehrlichrayner81,nicolswinney89,gibson00,hale01,lewisvasishth05,frankgoodman12}.
Language processing happens so rapidly that responses to words likely overlap in time \cite{mitchell84,smithkutas15,shainschuler19}.
In addition, patterns in language processing can change moment-by-moment due to e.g., task habituation, attentional fluctuation, and fatigue \cite{baayenetal18,prasadlinzen19,christiansonetal22}.
Thus, studies of language processing might be particularly ill served by analyses that rely on stationary linear models: poor alignment to the underlying process can lead to misleading inferences and limit the range of questions that can be investigated \cite[see \textbf{Supplementary Information \ref{sect:si-motivation}} for review]{baayenetal18,shainschuler19}.
We show that CDRNNs yield large improvements to out-of-sample model fit over alternative approaches on both behavioral and neuroimaging data.
We further show that CDRNNs enable both flexible discovery of novel structure in exploratory analyses and control of diverse potential confounds in confirmatory analyses, and thus constitute an important advance for both goals.

\begin{table}
    \centering
    \footnotesize
    \sffamily
    \begin{tabular}{rr|ccccc}
        & Feature & \rotatebox[origin=lb]{90}{LM} & \rotatebox[origin=lb]{90}{GAM} & \rotatebox[origin=lb]{90}{GAMLSS} & \rotatebox[origin=lb]{90}{CDR} & \rotatebox[origin=lb]{90}{CDRNN}\\
        \hline
        \hline
        \multirow{2}{*}{\shortstack[r]{Impulse\\response}} & Discrete-time & $\checkmark$ & $\checkmark$ & $\checkmark$ & $\checkmark$ & $\checkmark$ \\
        & Continuous-time & & & & $\checkmark$ & $\checkmark$ \\
        \hline
        \multirow{2}{*}{Effects} & Linear & $\checkmark$ & $\checkmark$ & $\checkmark$ & $\checkmark$ & $\checkmark$ \\
        & Nonlinear & & $\checkmark$ & $\checkmark$ & & $\checkmark$ \\
        \hline
        \multirow{3}{*}{Interactions} & Linear sparse & $\checkmark$ & $\checkmark$ & $\checkmark$ & $\checkmark$ & $\checkmark$ \\
        & Nonlinear sparse & & $\checkmark$ & $\checkmark$ & & $\checkmark$ \\
        & Nonlinear dense & & & & & $\checkmark$ \\
        \hline
        \multirow{2}{*}{Nonstationarity} & Linear & $\checkmark$ & $\checkmark$ & $\checkmark$ & $\checkmark$ & $\checkmark$ \\
        & Nonlinear & & $\checkmark$ & $\checkmark$ & & $\checkmark$ \\
        \hline
        \multirow{2}{*}{\shortstack[r]{Predictive\\distribution}} & Homoscedastic & $\checkmark$ & $\checkmark$ & $\checkmark$ & $\checkmark$ & $\checkmark$ \\
        & Heteroscedastic & & & $\checkmark$ & & $\checkmark$ \\
    \end{tabular}
    \caption{Comparison of key features of the solution spaces defined by linear models (LMs), generalized additive models (GAMs), generalized additive models for location, scale, and shape (GAMLSS), continuous-time deconvolutional regression (CDR), and continuous-time deconvolutional regressive neural networks (CDRNNs).
    The sparse/dense distinction under \textit{Interactions} concerns whether analysts must explicitly add interactions to the model (sparse) or whether the model considers all possible interactions (dense).
    Features absent from a model type cannot be directly modeled when using it.}
    \label{tab:feature_comparison}
\end{table}

\subsection{The CDRNN Model}
\label{sect:model}

Building on work from our group \cite{shain21}, our core proposal (CDRNN) is a deep neural generalization of several existing techniques for time series regression, including linear models (\textit{LMs, \citen{galton1886}}), linear mixed effects models (\textit{LMEs, \citen{batesetal15}}), generalized additive models (\textit{GAMs, \citen{wood06}}), generalized additive models for location, scale, and shape (\textit{GAMLSS, \citen{rigbystasinopoulos05}}), and continuous-time deconvolutional regression (\textit{CDR, \citen{shainschuler19}}).
A comparison of the solution spaces defined by each of these models is shown in \textbf{Table~\ref{tab:feature_comparison}} (see \textbf{Supplementary Information~\ref{sect:si-motivation}} for detailed discussion).

To define the regression problem, let $\y \in \R^{Y}$ be a single sample from the $Y$-dimensional dependent variable that we seek to model (the response, e.g., an fMRI BOLD measure), taken at time $\tau$ (e.g., seconds elapsed between the start of the experiment and the acquisition time of the fMRI volume).
Let $\F$ be a probability distribution with $S$-dimensional parameter vector $\s \in \R^S$ (e.g., the mean and variance of a Normal distribution over the response) such that $\y \sim \F(\s)$.
Let $\X \in \R^{N \times K}$ be a matrix of $N$ $K$-dimensional predictor vectors $\x_n, 1 \leq n \leq N$ (e.g., the duration and relative frequency of each of the $N$ words in a story).
Let $\t \in \R^N$ be the vector of predictor timestamps $t_1, ..., t_N$ such that $t_n$ is the timestamp of $\x_n$ (e.g., seconds elapsed between the start of the experiment and the onset of a word in a story).
Let $\d \in \R^N$ be the vector of temporal offsets $d_1, ..., d_N$ such that $d_n = \tau - t_n$, i.e.\ the signed distance in time between $\y$ and  $\x_n$ (e.g., the time in seconds of an fMRI volume minus the time in seconds of the word onsets in a story).
A CDRNN defines a function from $\langle \X, \t, \tau \rangle$ to $\s$, that is, from the predictors and their timestamps to the parameters of the predictive distribution.

The CDRNN computation involves three stages (see \textbf{Supplementary Information~\ref{sect:si-buildup}} for detailed motivation).
The first stage is \textbf{input processing}: the vector of timestamps $\t$ is horizontally concatenated with the predictor matrix $\X$ to form the input to the input processing function $\fin \in \R^{N \times (K + 1)} \rightarrow \R^{N \times J}$ with parameters $\uin \in \R^{U_{\text{in}}}$.
The output $\Xprime \in \R^{N \times J}$ is a matrix of $N$ $J$-dimensional impulse vectors $\xprime_n, 1 \leq n \leq N$ computed independently by $\fin$:
\begin{equation}
    \xprime_n \defeq \fin\left(\begin{bmatrix}t_n \\ \x_n \end{bmatrix}; \uin \right)
\end{equation}

The second stage is the \textbf{impulse response function (IRF)}: the processed inputs $\Xprime$ are horizontally concatenated with $\d$ and $\t$ to yield the inputs to IRF $\firf \in \R^{N \times (J + 2)} \rightarrow \R^{N \times S \times (J + 1)}$ with parameters $\uirf \in \R^{U_{\text{IRF}}}$.
The output of the IRF is a sequence of convolution weight matrices $\G_n \in \R^{S \times (J + 1)}, 1 \leq n \leq N$, where each $\G_n$ describes the effect of impulse $\xprime_n$ (e.g., a word in a story) on the predictive distribution over the response (e.g., a BOLD measure) at time $\tau$ (or, equivalently, at delay $d_n$):
\begin{equation}
    \label{eq:Gn}
    \G_n \defeq \firf \left(\begin{bmatrix} d_n \\ t_n \\ \xprime_n \end{bmatrix}; \uirf\right)
\end{equation}

The third stage is \textbf{convolution}: the parameters $\s$ for the predictive distribution $\F$ are computed as the sum of (i) the temporal convolution of $\Xprime$ with $\G_1, \ldots, \G_N$ and (ii) learned bias vector $\snull$, where each transposed row $\xprime_n, 1 \leq n \leq N$ of $\Xprime$ is vertically concatenated with a bias (here called \textit{rate}, a ``deconvolutional intercept'' capturing general effects of event timing) and weighted by learned coefficient vector $\b \in \R^{J + 1}$:
\begin{equation}
    \label{eq:cdrnn_main}
    \s \defeq \snull + \sum_{n=1}^N \G_n \text{diag}(\b)\begin{bmatrix}1\\\xprime_n\end{bmatrix}
\end{equation}
The impulses $\xprime_n$ appear both in the inputs to the convolution weights $\G_n$ (eq.~\ref{eq:Gn}) and in the convolution itself (eq.~\ref{eq:cdrnn_main}) in order to allow $\G_n$ to be either nonlinear or linear on dimensions of $\xprime_n$, depending on the goals of the analyst (see \textbf{Supplementary Information~\ref{sect:si-buildup}} for details).

Mixed effects CDRNN models can be defined by letting the parameter vector $\vv \in \R^V$:
\begin{equation}
    \vv \defeq \begin{bmatrix} \uin \\ \uirf \\ \b \\ \snull\end{bmatrix}
\end{equation}
be the sum of a fixed part $\vv_0 \in \R^V$ and random part $\V\z$, where $\V \in \R^{V \times Z}$ is a random effects matrix whose rows sum to 0 and $\z \in \{0, 1\}^{Z}$ indicates which of $Z$ random effects levels apply to $\y$:
\begin{equation}
    \vv \defeq \vv_0 + \V\z
\end{equation}
The parameters of the model are therefore $\vv_0$ and $\V$, which may be fitted via maximum likelihood or Bayesian inference.
This definition assumes a singleton dataset $\D = \left\{\left<\X, \t, \y, \tau\right>\right\}$, but it extends without loss of generality to any finite dataset by applying eq.~\ref{eq:cdrnn_main} independently to each of $M$ elements in $\D = \left\{\left<\X_m, \t_m, \y_m, \tau_m\right> \mid 1 \leq m \leq M\right\}$.

These equations generalize multiple existing time series models.
If $\fin$ is set to be identity and $\firf$ is set to be a Dirac $\delta$ on $\d$, the result is a linear model.
If $\fin$ is set to be a parametric spline function and $\firf$ is set to be a Dirac $\delta$ on $\d$, the result is a GAM.
If, in addition, $\fin$ has vector-valued output that defines all parameters of the predictive distribution, the result is a GAMLSS.
If $\fin$ is set to be identity and $\firf$ is set to be a parametric kernel function, the result is a CDR model.

However, in this work, motivated by evidence that deep neural networks enable high-accuracy non-linear function approximation across domains and tasks \cite{lecunetal15}, we propose and evaluate CDRNNs, which we define as any model that instantiates $\fin$ or $\firf$ as a deep neural network.
Implementing CDRNNs requires a novel neural network architecture.
To see why, note that the convolution over time in eq.~\ref{eq:cdrnn_main} imposes an important constraint on the regression problem, namely, that the contributions of $\firf$ at timepoints $1 \leq n \leq N$ are \textit{additive}.
This constraint is central to CDRNNs' interpretability, since it allows $\firf$ to define a valid impulse response function, such that evaluating $\G_n$ yields a complete description of the causal contribution of input $n$ to the distribution $\F(\s)$ generated by the model.
Time series models widely used in deep learning---including recurrent neural networks \cite{elman91}, convolutional neural networks \cite{lecunetal89}, and transformers \cite{vaswanietal17}---violate this constraint by integrating over time in a non-linear fashion and therefore do not implement deconvolutional regression, continuous-time or otherwise.

\begin{figure}
\centering
\sffamily
\resizebox{0.65\textwidth}{!}{
\begin{tikzpicture}
        
    \draw[draw=white,fill=white] (-10.5, 14.25) rectangle ++(1, 5.5);
    \draw[line width=1,color=gray!50] (-12, 13.9) to  (16, 13.9);
    \draw[line width=1,color=gray!50] (-12, 8) to  (10, 8);
    \draw[arrows={-Triangle},line width=2,color=gray] (-12, 2.75) to  (10, 2.75);
    \draw[draw=lime,fill=lime!30,rounded corners=2pt,line width=2] (12.5, 14.5) rectangle ++(0.5, 3.75);
    \draw[draw=brown,fill=brown!30,rounded corners=2pt,line width=2] (15, 14.5) rectangle ++(0.5, 3.75);
    \node[scale=2] at (0, 1.75) {\textbf{Time}};
    \node[scale=3] at (-3, 16.25) {$+$};
    \node[scale=3] at (3, 16.25) {$+$};
    \node[scale=3] at (11.5, 16.25) {$+$};
    \node[scale=3] at (14, 16.25) {$=$};

    \node[rotate=90,scale=2] at (-11.5, 16.25) {\textbf{Convolution}};
    \node[rotate=90,scale=2] at (-11.5, 10.9) {\parbox{1.7cm}{\centering\textbf{Impulse Response}}};
    \node[rotate=90,scale=2] at (-11.5, 5.3) {\parbox{2cm}{\centering\textbf{Input Processing}}};
    
    \node[anchor=west,scale=1.5] at (8, 17.25) {Rate};
    \node[anchor=west,scale=1.5] at (8, 15.975) {Impulses $\xprime$};
    \node[anchor=west,scale=1.5] at (8, 12.3) {Time Offset $d$};
    \node[anchor=west,scale=1.5] at (8, 11.6) {Timestamp $t$};
    \node[anchor=west,scale=1.5] at (8, 10.175) {Impulses $\xprime$};
    \node[anchor=west,scale=1.5] at (8, 6.5) {Timestamp $t$};
    \node[anchor=west,scale=1.5] at (8, 5.075) {Predictors $\x$};
    \node[rotate=90,scale=1.5] at (12.75, 16.375) {Bias $\snull$};
    \node[rotate=90,scale=1.5] at (15.25, 16.375) {Output $\s$};

    \begin{scope}[shift={(-6, 0)}]
        \begin{scope}[shift={(-2,15)}]
            \draw[draw=yellow,fill=yellow!30,rounded corners=2pt,line width=2] (0, 0.1) rectangle ++(2.5, 2.5);
            \node[scale=3] at (1.25, 1.25) {$\G$};
            \filldraw[color=black] (2.75, 1.25) circle (0.05);
        \end{scope}
        \begin{scope}[shift={(1.2, 15)}]
            \draw[draw=gray!60,fill=gray!10,rounded corners=2pt,line width=2] (-0.2, -0.1) rectangle ++(0.9,2.85);
            \draw[draw=purple!60,fill=purple!20,rounded corners=2pt,line width=2] (0, 0.1) rectangle ++(0.5,1.75);
            \draw[draw=teal!60,fill=teal!20,line width=2] (0.25, 2.25) circle(0.25);
            \node at (0.25, 2.25) {\large$1$};
        \end{scope}

        \begin{scope}[shift={(-0.75, 9.1)}]
            \draw[draw=gray!60,fill=gray!10,rounded corners=2pt,line width=2] (-0.5,-0.3) rectangle ++(1,4);
            \draw[draw=red!60,fill=red!20,line width=2] (0,3.2) circle(0.25);
            \draw[draw=blue!60,fill=blue!20,line width=2] (0,2.5) circle(0.25);
            \draw[draw=purple!60,fill=purple!20,rounded corners=2pt,line width=2] (-0.25,0.2) rectangle ++(0.5,1.75);
        \end{scope}
        
        \draw[arrows={-Triangle},line width=1.5,color=gray] (-0.75, 13) to [out=90, in=270]  (-0.75, 15);

        \begin{scope}[shift={(-0.75, 4)}]
            \draw[draw=gray!60,fill=gray!10,rounded corners=2pt,line width=2] (-0.5,0) rectangle ++(1,3);
            \draw[draw=blue!40,fill=blue!10,dotted,line width=2] (0,2.5) circle(0.25);
            \draw[draw=green!60,fill=green!20,rounded corners=2pt,line width=2] (-0.25,0.2) rectangle ++(0.5,1.75);
        \end{scope}
        
        \draw[arrows={-Triangle},line width=1.5,color=gray] (-0.75, 7.15) to  (-0.75, 9.15);
    \end{scope}

    \begin{scope}[shift={(0, 0)}]
        \begin{scope}[shift={(-2,15)}]
            \draw[draw=yellow,fill=yellow!30,rounded corners=2pt,line width=2] (0, 0.1) rectangle ++(2.5, 2.5);
            \node[scale=3] at (1.25, 1.25) {$\G$};
            \filldraw[color=black] (2.75, 1.25) circle (0.05);
        \end{scope}
        \begin{scope}[shift={(1.2, 15)}]
            \draw[draw=gray!60,fill=gray!10,rounded corners=2pt,line width=2] (-0.2, -0.1) rectangle ++(0.9,2.85);
            \draw[draw=purple!60,fill=purple!20,rounded corners=2pt,line width=2] (0, 0.1) rectangle ++(0.5,1.75);
            \draw[draw=teal!60,fill=teal!20,line width=2] (0.25, 2.25) circle(0.25);
            \node at (0.25, 2.25) {\large$1$};
        \end{scope}

        \begin{scope}[shift={(-0.75, 9.1)}]
            \draw[draw=gray!60,fill=gray!10,rounded corners=2pt,line width=2] (-0.5,-0.3) rectangle ++(1,4);
            \draw[draw=red!60,fill=red!20,line width=2] (0,3.2) circle(0.25);
            \draw[draw=blue!60,fill=blue!20,line width=2] (0,2.5) circle(0.25);
            \draw[draw=purple!60,fill=purple!20,rounded corners=2pt,line width=2] (-0.25,0.2) rectangle ++(0.5,1.75);
        \end{scope}
        
        \draw[arrows={-Triangle},line width=1.5,color=gray] (-0.75, 13) to [out=90, in=270]  (-0.75, 15);

        \begin{scope}[shift={(-0.75, 4)}]
            \draw[draw=gray!60,fill=gray!10,rounded corners=2pt,line width=2] (-0.5,0) rectangle ++(1,3);
            \draw[draw=blue!40,fill=blue!10,dotted,line width=2] (0,2.5) circle(0.25);
            \draw[draw=green!60,fill=green!20,rounded corners=2pt,line width=2] (-0.25,0.2) rectangle ++(0.5,1.75);
        \end{scope}
        
        \draw[arrows={-Triangle},line width=1.5,color=gray] (-0.75, 7.15) to  (-0.75, 9.15);
    \end{scope}

    \begin{scope}[shift={(6, 0)}]
        \begin{scope}[shift={(-2,15)}]
            \draw[draw=yellow,fill=yellow!30,rounded corners=2pt,line width=2] (0, 0.1) rectangle ++(2.5, 2.5);
            \node[scale=3] at (1.25, 1.25) {$\G$};
            \filldraw[color=black] (2.75, 1.25) circle (0.05);
        \end{scope}
        \begin{scope}[shift={(1.2, 15)}]
            \draw[draw=gray!60,fill=gray!10,rounded corners=2pt,line width=2] (-0.2, -0.1) rectangle ++(0.9,2.85);
            \draw[draw=purple!60,fill=purple!20,rounded corners=2pt,line width=2] (0, 0.1) rectangle ++(0.5,1.75);
            \draw[draw=teal!60,fill=teal!20,line width=2] (0.25, 2.25) circle(0.25);
            \node at (0.25, 2.25) {\large$1$};
        \end{scope}

        \begin{scope}[shift={(-0.75, 9.1)}]
            \draw[draw=gray!60,fill=gray!10,rounded corners=2pt,line width=2] (-0.5,-0.3) rectangle ++(1,4);
            \draw[draw=red!60,fill=red!20,line width=2] (0,3.2) circle(0.25);
            \draw[draw=blue!60,fill=blue!20,line width=2] (0,2.5) circle(0.25);
            \draw[draw=purple!60,fill=purple!20,rounded corners=2pt,line width=2] (-0.25,0.2) rectangle ++(0.5,1.75);
        \end{scope}
        
        \draw[arrows={-Triangle},line width=1.5,color=gray] (-0.75, 13) to [out=90, in=270]  (-0.75, 15);

        \begin{scope}[shift={(-0.75, 4)}]
            \draw[draw=gray!60,fill=gray!10,rounded corners=2pt,line width=2] (-0.5,0) rectangle ++(1,3);
            \draw[draw=blue!40,fill=blue!10,dotted,line width=2] (0,2.5) circle(0.25);
            \draw[draw=green!60,fill=green!20,rounded corners=2pt,line width=2] (-0.25,0.2) rectangle ++(0.5,1.75);
        \end{scope}
        
        \draw[arrows={-Triangle},line width=1.5,color=gray] (-0.75, 7.15) to  (-0.75, 9.15);
    \end{scope}
    
\end{tikzpicture}}

\caption{\textbf{CDRNN architecture.} A grapical depiction of the CDRNN forward pass for generating one prediction.
Scalars are shown as circles, vectors are shown as narrow boxes, matrices are shown as wider boxes, and deep neural network transformations are shown as arrows.
Computation proceeds in three stages (bottom-to-top): (i) processing the inputs, (ii) applying the impulse response, and (iii) convolving the impulses with the IRF (convolution weights) over time to generate a parameterization for the predictive distribution over the response.
At the convolution stage, the impulses are augmented with bias term (\textit{rate}) that allows the model to capture generalized effects of the rate of events in time.
Components shown with dotted lines are not used in the base CDRNN implementation in this study (although they are explored in the full set of analyses, see \textbf{Supplementary Information}).
}
\label{fig:model}

\end{figure}
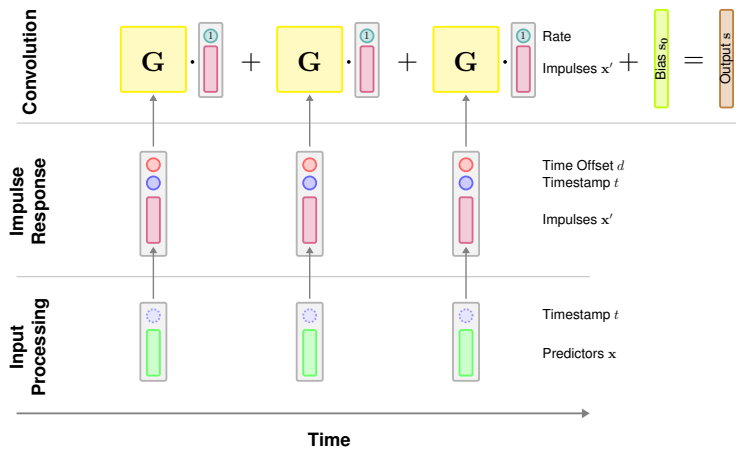

Our proposed CDRNN architecture is schematized in \textbf{Figure~\ref{fig:model}}, where arrows represent transformations implemented by a \textit{feedforward neural network} (FFN).
A feedforward neural network $\fff$ with $L$ layers contains weights $\W^{(\ell)} \in \R^{D_{\ell} \times D_{\ell - 1}}$ (where $D_0 = K$), biases $\b^{(\ell)} \in \R^{D_{\ell}}$ and activation functions $\sigma^{(\ell)}$, $1 \leq \ell \leq L$, and is defined recursively as follows (where $\fff^{(0)}(\x) = \x$, the input vector):
\begin{equation}
    \fff^{(\ell)}(\x) \defeq \sigma^{(\ell)}\left(\W^{(\ell)}\fff^{(\ell-1)}(\x) + \b^{(\ell)}\right)
\end{equation}
As shown in \textbf{Figure~\ref{fig:model}}, at each timepoint, our CDRNN transforms the predictor vectors $\x_n$ (green) along with their timestamps $t_n$ (blue) using an FFN implementing $\fin$ (input processing).
The outputs $\xprime_n$ (red) are then concatenated with their timestamps $t_n$ and their signed offset in time $\d_n$ from the prediction target and processed by a second FFN implementing $\firf$ (impulse response).
The IRF output $\G_n$ (yellow) is a matrix of convolution weights mapping $\xprime_n$ and a bias term (\textit{rate}) to the estimated contribution of timepoint $n$ to the predictive distribution over the response.
These estimated contributions are summed together with a bias term (an intercept, lime) to yield the output of the CDRNN: a vector $\s$ (brown) of parameters for the predictive distribution $\F$ over a single response measure at a point in time (convolution).
The parameters of the model are estimated using stochastic gradient descent, subject to standard deep neural network regularizers (e.g., weight penalties and dropout, see \textbf{Supplementary Information~\ref{sect:si-implementation})}.

Note that there is little reason in principle for both $\fin$ and $\firf$ to be deep neural networks in the general case, since effect nonlinearities and interactions can be directly estimated by $\firf$.
The primary interest of a deep neural $\fin$ is to allow GAM-like estimation of effect nonlinearities and interactions when the impulses $\Xprime$ are absent from the inputs to $\firf$, thus decoupling the shape of effects in predictor space and the shape of the response in time.
For this reason, in all analyses conducted in this study, $\fin$ is assumed to be identity.

This simple network respects the constraints imposed by eq.~\ref{eq:cdrnn_main} while also (i) modeling a continuous IRF via the dependence on $d_n$, (ii) relaxing temporal stationarity assumptions via the dependence on $t_n$, (iii) relaxing homoscedasticity assumptions by parameterizing the entire distribution $\F(\s)$ as a function of the inputs, and (iv) capturing arbitrary nonlinearities and effect interactions by processing the entire predictor vector with an FFN.
In \textbf{Supplementary Information~\ref{sect:si-extended}}, we additionally propose a generalization of this model that can capture non-independence between inputs in their effects on the response.

As deep neural networks, CDRNNs lack a transparent link between the values of parameters and their effects on the response.
This presents a challenge for effect estimation, which is of critical interest in scientific applications.
To address this, we employ a technique from deep learning known as \textit{perturbation analysis} \cite{ribeiroetal16,petsiuketal18}.
In brief, we quantify the effect on network outputs of manipulating network inputs, permitting analysis of the network's latent IRF.
Full technical details about CDRNN effect estimation and uncertainty quantification can be found in \textbf{Supplementary Information~\ref{sect:si-interpretation}}.
A documented software library for CDRNN regression is available at \url{https://github.com/coryshain/cdr}.

\section{Results}
\subsection{Model Validation}

We evaluate CDRNNs against LME, GAM, GAMLSS, and CDR baselines on three naturalistic language comprehension datasets: the Dundee eye-tracking (ET) dataset \cite{kennedyetal03}, the Natural Stories self-paced reading (SPR) dataset \cite{futrelletal20}, and the Natural Stories functional magnetic resonance imaging (fMRI) dataset (analyses restricted to responses in the ``language network'', i.e., a network of regions that specialize for language processing; \textit{\citen{shainetal20}}).
See \textbf{Methods} for description of datasets and predictors used in each analysis, and see \textbf{Supplementary Information~\ref{app:si-synth}} for validation on synthetic data with known ground truth IRFs.

\begin{figure}

    \centering
    \sffamily
    
    \includegraphics[width=\textwidth]{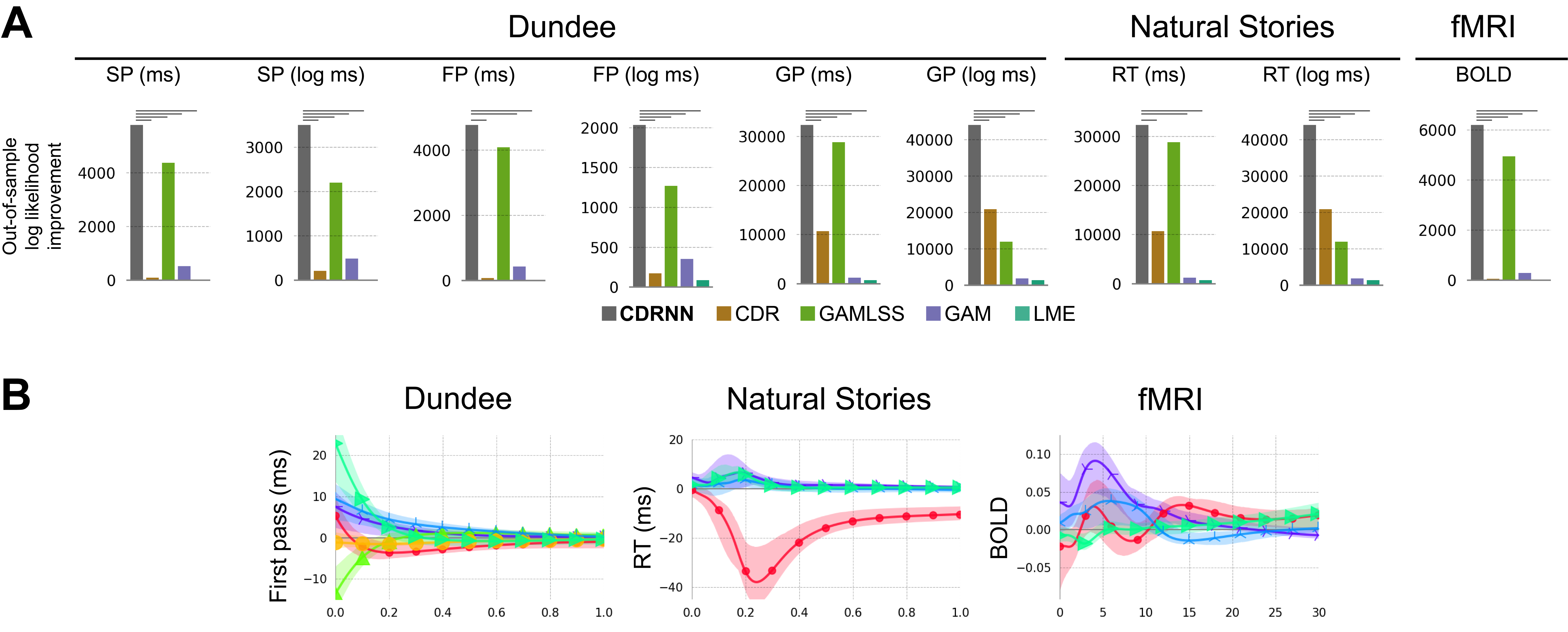}
    
    {\scriptsize
    {\kone~rate \hspace{1em}}
    {\kthree~saccade~length \hspace{1em}}
    {\kfive~previous~was~fixated \hspace{1em}}
    {\kseven~word~length/sound~power \hspace{1em}}
    {\knine~unigram~surprisal \hspace{1em}}\\
    {\keleven~5-gram~surprisal}\\
    }
    
    \caption{\textbf{Main result.} \textbf{A.} Out-of-sample (test set) log likelihood improvement of models over baseline using raw and log-transformed scan path (SP), first pass (FP), and go-past (GP) durations from the Dundee eye-tracking dataset, raw and log-transformed reading times (RT) from the Natural Stories self-paced reading dataset, and blood oxygen level dependent (BOLD) contrast from the Natural Stories functional magnetic resonance imaging (fMRI) dataset.
    See \textbf{Methods} for technical details.
    Significant improvements from CDRNNs over alternatives are indicated by horizontal lines in each subplot.
    CDRNNs generalize numerically better than all alternatives for all response variables, significantly so in all but two comarisons.
    \textbf{B.} CDRNN-estimated impulse response functions.
    Curves represent the estimated change in response ($y$-axis) from one standard deviation increase in each predictor (line color) as a function of delay from word onset ($x$-axis).
    }
    \label{fig:main}
    
\end{figure}

The key finding is shown in \textbf{Figure~\ref{fig:main}A}: CDRNNs generalize substantially better to unseen data than comparable LME, GAM, GAMLSS, or CDR baselines, numerically improving conditional out-of-sample log likelihood in each comparison (often by thousands of points, significant in all but two comparisons).
GAMLSS is the the best-performing alternative, suggesting that CDRNNs' gains in these analyses derive primarily from capturing heteroscedasticity, which the other models cannot do (see \textbf{Supplementary Information~\ref{app:cognitive}} for additional support for this conjecture).
Nonetheless, CDRNNs also yield consistent gains over GAMLSS, suggesting that CDRNNs' advantages go beyond heteroscedasticity.
For full results and analysis, including diverse evaluations on synthetic datasets and detailed exploration of model design choices, see \textbf{Supplementary Information~\ref{app:cognitive}}.

Not only do CDRNNs provide a generalizable description of complex processes, but their estimates are also richly detailed.
\textbf{Figure~\ref{fig:main}B} shows IRF plots describing the estimated change in the response associated with one standard deviation increase in a predictor, as a function of delay from word onset ($x$-axis).
As shown, most effects in reading (left and center) decay to near-zero within a one second window of stimulus presentation, whereas the fMRI response (right) is more diffuse, extending over 20-30s and showing the characteristic hemodynamic peak around 5s delay \cite{boyntonetal96}.
As has been previously reported, self-paced reading (center) is dominated by a large negative effect of \textit{rate} (the average effect of reading a word), suggesting that fast reading in the recent past engenders fast reading now, consistent with an inertia effect from repeated button pressing \cite{shainschuler19}.
All three modalities show a large positive effect of \textit{5-gram surprisal}, a measure of how predictable a word is from context.
This indicates an increase in both reading time and brain activity for less predictable words, consistent with predictive coding \cite{levy08,wilcoxetal20,shainetal20}.

\subsection{Pattern Discovery and Hypothesis Testing}

In this section, we exemplify applications of CDRNNs for both exploratory research (e.g., discovery and visualization of novel patterns) and scientific hypothesis testing.
In so doing, we show that, in addition to improving model fit and reducing dependence on standard simplifying assumptions, CDRNNs broaden the space of questions that are feasible to investigate using a given dataset.
We focus for simplicity on the Natural Stories fMRI dataset, since latencies are known to have a major influence on the fMRI BOLD signal \cite{boyntonetal96}.
To do so, we fit an ensemble of 10 CDRNN models using the base configuration for fMRI (see \textbf{Supplementary Information~\ref{sect:si-implementation}}).
From this single ensemble, we can obtain diverse estimates about the structure of the fMRI response (\textbf{Figure~\ref{fig:novel_main_effect}}), which we discuss below in five examples.
We test these estimates statistically using out-of-sample model comparison, thereby grounding results in the generalizability of findings (for details, see \textbf{Supplementary Information~\ref{sect:si-nhst}}).

\subsubsection{Example A: The Existence of Effects}
\label{sect:novel-a}

\textbf{Figure~\ref{fig:novel_main_effect}A} shows the estimated change in blood oxygen level dependent (BOLD) contrast as a surface relating predictor value and delay from stimulus onset (the line plots in \textbf{Figure~\ref{fig:main}B} are obtained by ``slicing'' these surfaces along the \textit{Delay} dimension at a fixed predictor value, i.e., one standard deviation above the training set mean).
These plots reveal how the response is expected to change at a given delay after observing a predictor of a given value, holding all other variables constant.
As shown, the language-selective regions whose activity is represented in this dataset are not very responsive to \textit{sound power} (n.s.), a low-level auditory feature: the uncertainty interval includes zero over nearly the entire surface.
There is an estimated effect of \textit{unigram surprisal} ($p < 0.0001$)---a measure of how frequently a word is used---that better matches the expected hemodynamic shape (peaking around 5s delay and then dipping), with an intriguing \textit{u}-shaped nonlinearity such that both low and high unigram surprisal (that is, respectively, highly frequent and highly infrequent) words yield an increase in BOLD relative words with average frequency.
The largest effects are associated with \textit{5-gram surprisal} ($p = 0.017$).

\subsubsection{Example B: Linearity of Effects}
\label{sect:novel-b}

The surfaces in \textbf{Figure~\ref{fig:novel_main_effect}A} suggest nonlinear effects of some predictors.
These can be visualized more clearly by ``slicing'' along the predictor dimension at a fixed delay (for simplicity, all plots at a fixed delay use 5s, the approximate location of the peak response).
Results are given in \textbf{Figure~\ref{fig:novel_main_effect}B}.
As shown, estimates support nonlinearities, especially a \textit{u}-shaped nonlinear effect of \textit{unigram surprisal} ($p < 0.0001$), and an inflection point in the \textit{5-gram surprisal} effect, which rises more steeply at higher values ($p < 0.0001$).

\subsubsection{Example C: Effect Interactions}
\label{sect:novel-c}

\begin{figure}
    \scriptsize
    \sffamily
    \centering
    
        
    
    \begin{subfigure}[t]{0.3\textwidth}
        \centering
        \begin{overpic}[trim=25 0 50 30,clip,width=\textwidth]{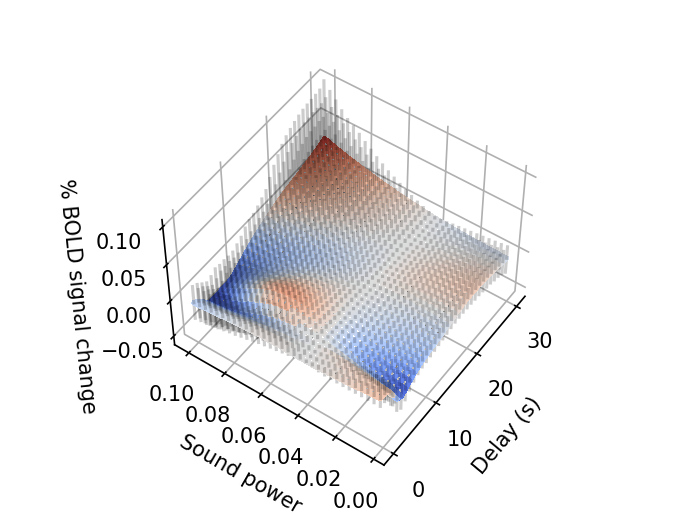}
            \put (-10,75) {\textbf{\Large A}}
        \end{overpic}
    \end{subfigure}
    \begin{subfigure}[t]{0.3\textwidth}
        \centering
        \includegraphics[trim=25 0 50 30,clip,width=\textwidth]{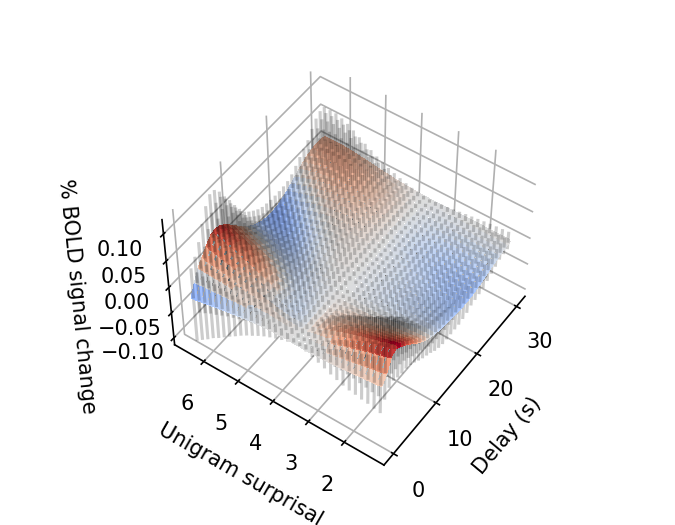}
    \end{subfigure}
    \begin{subfigure}[t]{0.3\textwidth}
        \centering
        \includegraphics[trim=25 0 50 30,clip,width=\textwidth]{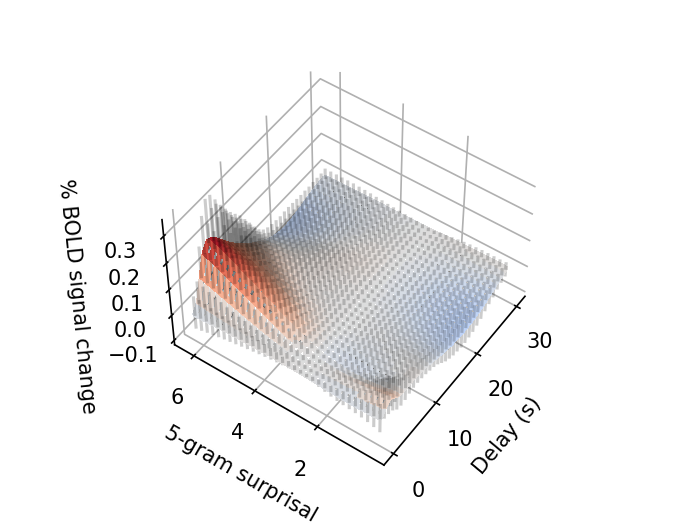}
    \end{subfigure}
    
    \begin{subfigure}[t]{0.3\textwidth}
        \centering
        \begin{overpic}[width=\textwidth]{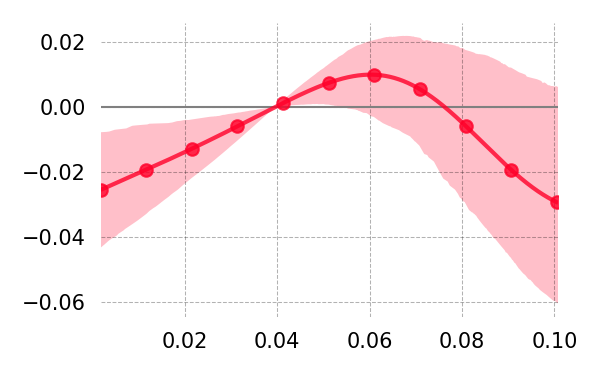}
            \put (-10,75) {\textbf{\Large B}}
            \put (-5,30) {\rotatebox[origin=c]{90}{\small BOLD}}
        \end{overpic}
        
        Sound power
    \end{subfigure}
    \begin{subfigure}[t]{0.3\textwidth}
        \centering
        \includegraphics[width=\textwidth]{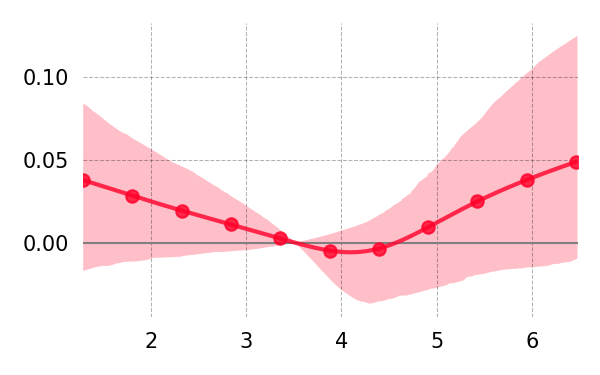}
        
        Unigram surprisal
    \end{subfigure}
    \begin{subfigure}[t]{0.3\textwidth}
        \centering
        \includegraphics[width=\textwidth]{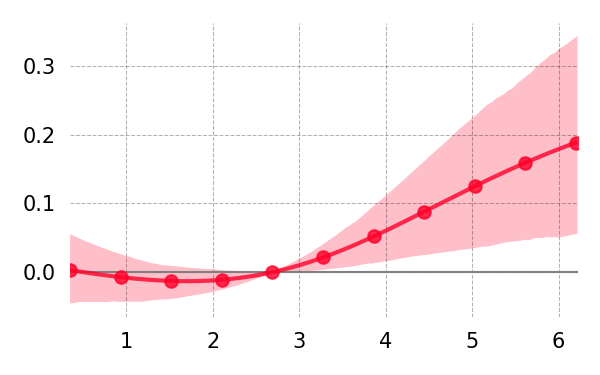}

        5-gram surprisal
    \end{subfigure}
    
    \begin{subfigure}[t]{0.3\textwidth}
        \centering
        \begin{overpic}[trim=25 0 50 30,clip,width=\textwidth]{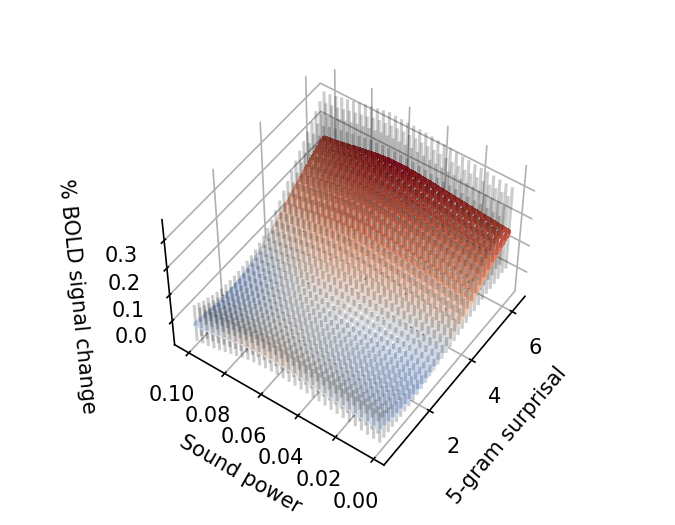}
            \put (-10,75) {\textbf{\Large C}}
        \end{overpic}
    \end{subfigure}
    \begin{subfigure}[t]{0.3\textwidth}
        \centering
        \includegraphics[trim=25 0 50 30,clip,width=\textwidth]{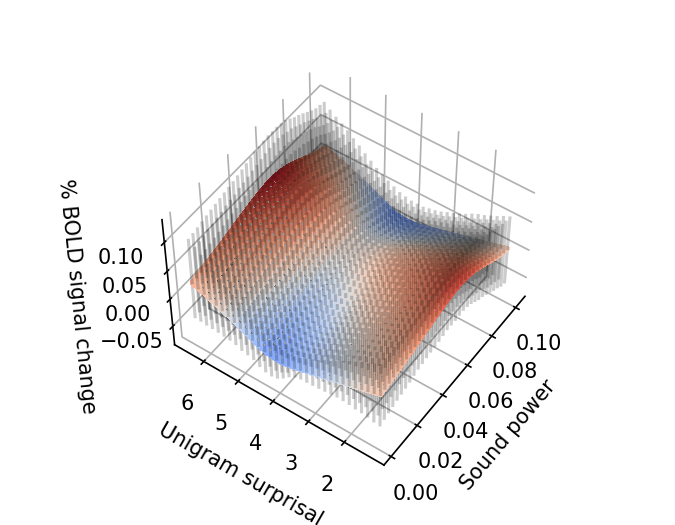}
    \end{subfigure}
    \begin{subfigure}[t]{0.3\textwidth}
        \centering
        \includegraphics[trim=25 0 50 30,clip,width=\textwidth]{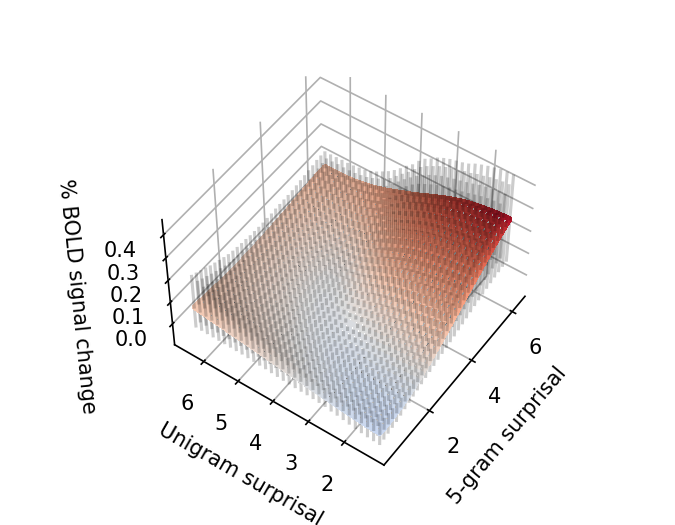}
    \end{subfigure}
    
    \begin{subfigure}[t]{0.3\textwidth}
        \centering
        \begin{overpic}[trim=25 0 50 30,clip,width=\textwidth]{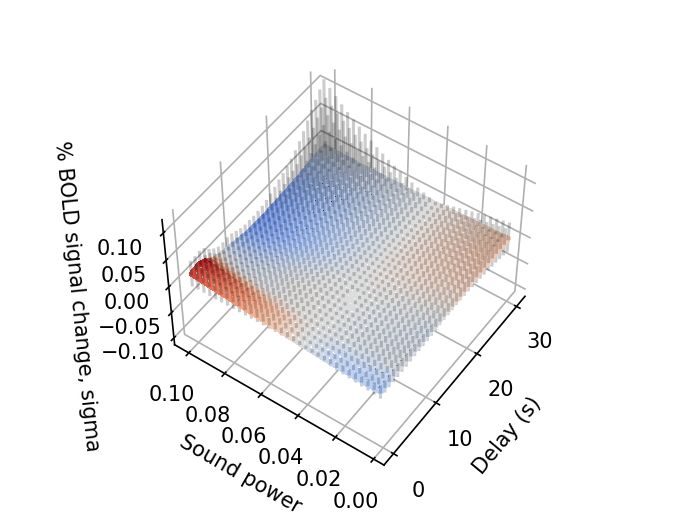}
            \put (-10,75) {\textbf{\Large D}}
        \end{overpic}
    \end{subfigure}
    \begin{subfigure}[t]{0.3\textwidth}
        \centering
        \includegraphics[trim=25 0 50 30,clip,width=\textwidth]{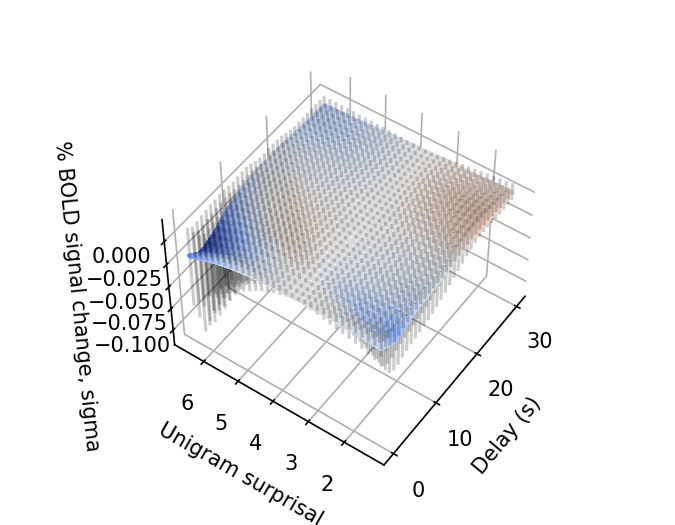}
    \end{subfigure}
    \begin{subfigure}[t]{0.3\textwidth}
        \centering
        \includegraphics[trim=25 0 50 30,clip,width=\textwidth]{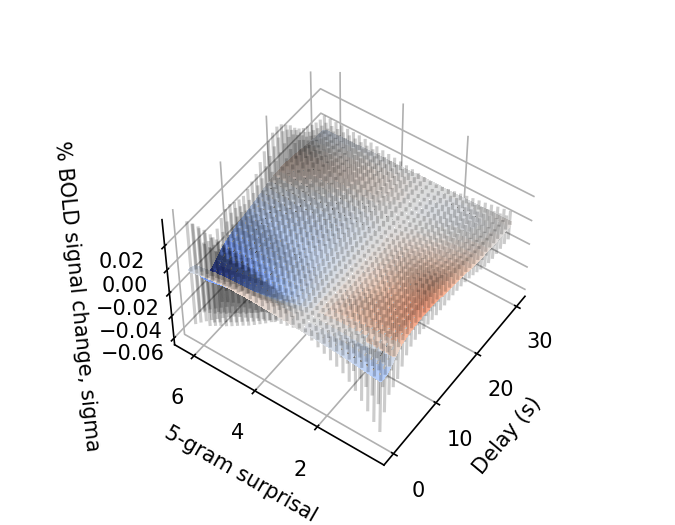}
    \end{subfigure}
    
    \begin{subfigure}[t]{0.3\textwidth}
        \centering
        \begin{overpic}[trim=25 0 50 30,clip,width=\textwidth]{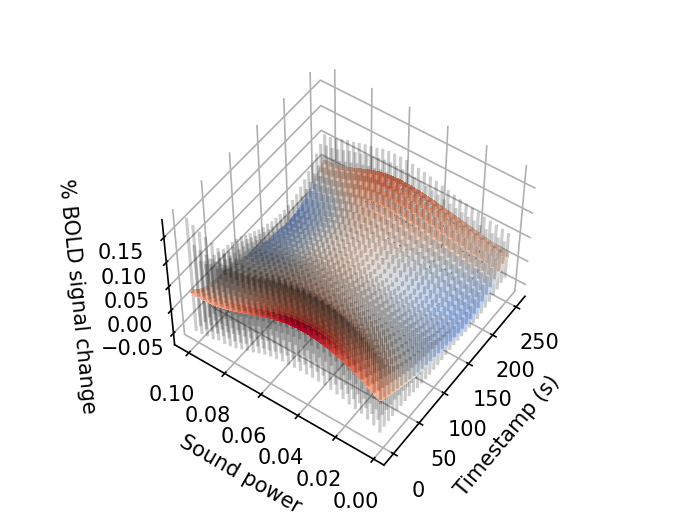}
            \put (-10,75) {\textbf{\Large E}}
        \end{overpic}
    \end{subfigure}
    \begin{subfigure}[t]{0.3\textwidth}
        \centering
        \includegraphics[trim=25 0 50 30,clip,width=\textwidth]{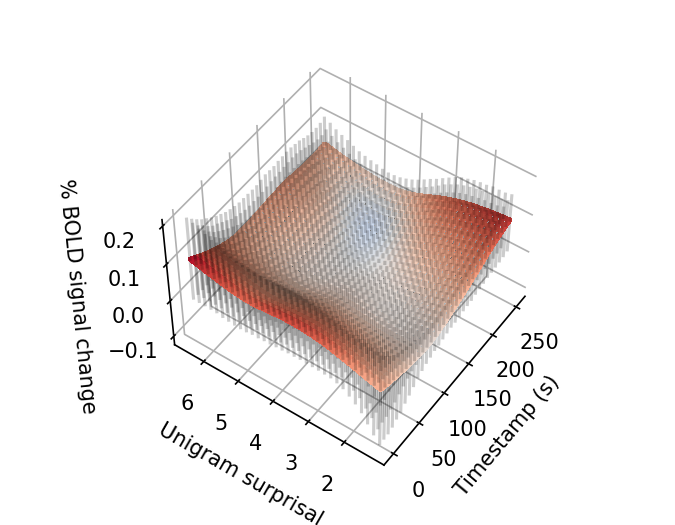}
    \end{subfigure}
    \begin{subfigure}[t]{0.3\textwidth}
        \centering
        \includegraphics[trim=25 0 50 30,clip,width=\textwidth]{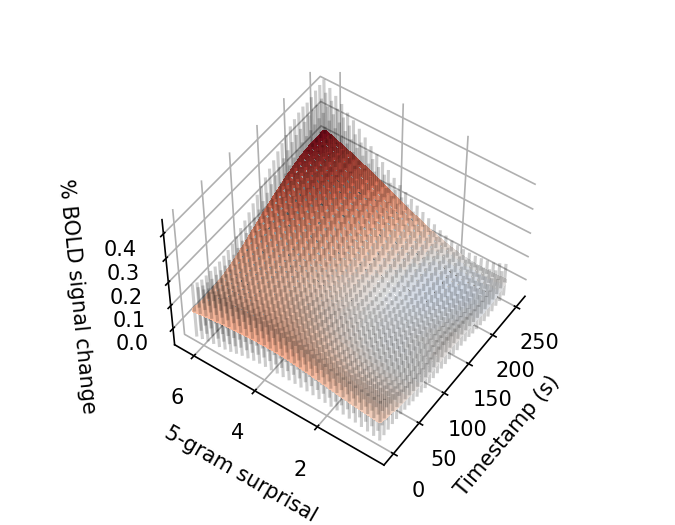}
    \end{subfigure}
    
    \caption{\textbf{CDRNN estimates derived from the Natural Stories fMRI dataset.}
    Colored bands (line plots) and vertical error bars (surface plots) show Monte Carlo estimated 95\% credible intervals.
    \textbf{A.} Univariate IRFs (hemodynamic responses).
    \textbf{B.} Functional form of effects at 5s delay.
    \textbf{C.} Effect interactions at 5s delay.
    \textbf{D.} Univariate IRFs of the $\sigma$ parameter of the predictive distribution.
    \textbf{E.} Nonstationarity at 5s delay.}
    \label{fig:novel_main_effect}
\end{figure}

CDRNNs implicitly model interactions between all predictors in the model.
The three pairwise effect interactions at 5s delay are plotted as surfaces in \textbf{Figure~\ref{fig:novel_main_effect}C}.
As shown, CDRNNs can discover both invariance and dependency between predictors.
The effects of \textit{unigram surprisal} and \textit{5-gram surprisal} are largely invariant to \textit{sound power}: the same basic response pattern holds regardless of the value of \textit{sound power}.
By contrast, \textit{unigram surprisal} and \textit{5-gram surprisal} appear to interact: the unigram effect flips direction (from increasing to decreasing) as one ascends the 5-gram continuum, and the 5-gram effect is more pronounced at lower values of unigram surprisal.
This interaction is nonlinear, since it could not be well approximated by a coefficient on the product of the two predictors.
However, this interaction does not improve test set likelihood and is therefore not significant.
This outcome demonstrates that increased representational capacity (i.e., ability to model an interaction) does not automatically lead to improved generalization performance.
This is an essential safeguard built into our proposed approach to hypothesis testing, which only admits effects that generalize robustly.

\subsubsection{Example D: Distributional Regression}
\label{sect:novel-ed}

Because CDRNNs can quantify the effect of a predictor on all parameters of the predictive distribution, not just the mean, they can be used to address questions about the distribution of a given response measure.
For example, word frequency and predictability have been argued to affect the location parameter but not the variance parameter of the distribution of fixation durations during reading \cite{staubetal10,staub2011:exgauss}.
The IRFs relating each predictor to $\sigma$ (the square root of the predictive variance) are plotted in \textbf{Figure~\ref{fig:novel_main_effect}D}.
As shown, \textit{sound power} has little effect on $\sigma$, and both \textit{unigram surprisal} and \textit{5-gram surprisal} are associated with a drop in $\sigma$ near the peak hemodynamic response of about 5s delay.
Of these, only the \textit{5-gram surprisal} effect is significant ($p < 0.0001$).

\subsubsection{Example E: Nonstationarity}
\label{sect:novel-e}

By conditioning the IRF on a representation of elapsed time, CDRNNs can capture nonstationarity in the response function.
For example, the effect of word predictability may change nonlinearly over the course of story listening.
Such nonstationarities are central to critical questions about adaptation and learning during language processing \cite{fineetal13,prasadlinzen19}.
Nonstationarity plots for the fMRI dataset are given in \textbf{Figure~\ref{fig:novel_main_effect}E}.
These plots represent the effect of a predictor at 5s delay as a function of its onset timestamp.
As shown, \textit{unigram surprisal} effects appear weaker toward the end of the story (but not significantly so, $p = 0.060$), whereas \textit{5-gram surprisal} effects are stronger toward the end of the story ($p = 0.0003$).

\subsubsection{Assumptions Influence Test Results}
\label{sect:assumptions-influence}

Avoiding assumptions of linearity, stationary, and homoscedasticity can be critical for hypothesis testing, even if the research hypothesis does not directly concern these assumptions.
For example, in Example C, we did not find a significant interaction between \textit{unigram surprisal} and \textit{5-gram surprisal}.
However, when we enforce a homoscedasticity assumption, the interaction becomes significant ($p < 0.0001$).
Likewise, in Example D, we did not find a significant effect of \textit{unigram surprisal} on the $\sigma$ parameter of the predictive distribution.
However, when we enforce a linearity assumption, then the effect of \textit{unigram surprisal} on $\sigma$ becomes significant ($p < 0.0001$).
These significant findings turn out to depend critically on implausible simplifying assumptions that CDRNNs can relax.
Precisely how a given simplifying assumption could affect a given experimental outcome is often difficult to anticipate.
CDRNNs help mitigate such concerns by avoiding these assumptions in the first place.

\section{Discussion}

We have proposed continuous-time deconvolutional regressive neural networks (CDRNNs), a new approach to analyzing observational time series data.
CDRNNs leverage the flexibility of deep learning to relax standard assumptions in regression analyses of time series (discrete time, linearity, stationarity, and homoscedasticity) while remaining interpretable.
This property enables flexible visualization and discovery of novel patterns in exploratory analyses and better control of confounds in confirmatory analyses.

We evaluated CDRNNs on data from the domain of human language processing and showed that they substantially improve fit to unseen data over established alternatives (\textbf{Figure~\ref{fig:main}A}) while providing detailed and plausible estimates of the dynamics of the modeled system (\textbf{Figure~\ref{fig:main}B}).
We then exemplified how a CDRNN can be used to visualize and test diverse properties of the response, including the existence of effects (\textbf{Figure~\ref{fig:novel_main_effect}A}), the functional form of effects (\textbf{Figure~\ref{fig:novel_main_effect}B}), the possibility of arbitrary nonlinear effect interactions (\textbf{Figure~\ref{fig:novel_main_effect}C}), effects on the probability distribution over the response (\textbf{Figure~\ref{fig:novel_main_effect}D}), and changes in effects over time (\textbf{Figure~\ref{fig:novel_main_effect}E}).
In fact, CDRNNs can be used to visualize \textit{any} property of the modeled system that can be cached out as a question about the response to input.
Likewise, CDRNNs can be used to test any null hypothesis that can be cached out as a model constraint (see \textbf{Supplementary Information~\ref{sect:si-nhst}} for discussion).
CDRNNs therefore constitute a highly general framework for estimating and testing the properties of continuous-time processes in nature.

The findings from these empirical evaluations also offer key insights about the brain response to language, some consistent with prior expectations, and some novel.
First, consistent with prior expectations \cite{fedorenkoetal10}, the high-level language network in the brain is sensitive to linguistic variables (\textit{unigram surprisal}, a measure of word frequency, and \textit{5-gram surprisal}, a measure of word predictability) but not a perceptual variable (\textit{sound power}).
This outcome also entails that effects of word frequency and predictability are at least partly dissociable in brain activity (since both effects together significantly improve on either effect individually), consistent with prior arguments for such a dissociation in reading \cite{staub15}.

Second, word frequency and predictability both have nonlinear effects on brain activity.
The predictability effect is superlinear, with an inflection point near the mean below which effects are weak, and above which effects are strong.
Although this is inconsistent with prior claims that processing cost is linear on our predictability measure \cite{smithlevy13,wilcoxetal20}, it should be taken with a grain of salt: BOLD is a complex signal that is not necessarily linear on neuronal activity \cite{logothetis08}, so it may be a problematic testbed for questions about functional form.
To our knowledge, the finding of a \textit{u}-shaped frequency effect (whereby BOLD increases when word frequency is \textit{either} low or high) is novel and warrants further investigation: although BOLD may not be linear on neuronal activity, evidence indicates that it is monotonic \cite{logothetisetal01}, and thus our findings plausibly reflect a \textit{u}-shaped effect at the neuronal level.

Third, results do not support a frequency-predictability interaction: although a non-linear interaction appears in the estimates (\textbf{Figure~\ref{fig:novel_main_effect}C}), it does not generalize.
Several prior studies have also failed to find such interactions, leading some to argue that frequency and predictability effects are driven by distinct cognitive mechanisms (for review, see \textit{\citen{staub15}}).

Fourth, results support an influence of word predictability on the scale parameter ($\sigma$) of the distribution over brain activity.
This finding is novel: in a prior reading study, predictability did not affect $\sigma$ \cite{staub2011:exgauss}.
Our results do not contradict this earlier work, since the fMRI BOLD response is a different measure than fixation durations during reading, with different distributional properties.
We speculate that predictability effects on the scale parameter may derive from less predictable words driving the BOLD response above the noise floor, thereby increasing model certainty about the expected BOLD value and thus decreasing $\sigma$.

Fifth, we find significant non-stationarity in the predictability response, such that less predictable words are associated with larger increases in BOLD as the story unfolds.
To our knowledge, this finding is novel.
It is possible that comprehenders increase their reliance on predictive processing later in the story, as they accumulate evidence toward a mental model of story content that might facilitate accurate prediction.
We leave detailed follow up of all of the above findings to future work.

Like all modeling approaches, CDRNNs have potential drawbacks.
\textit{First}, they often require more data and computation.
However, in practice, given the complexity of the problem they are tasked to solve (arbitrary nonlinear and nonstationary continuous-time influences of all possible sets of predictors on all parameters of the predictive distribution), they can be quite efficient.
For example, the Dundee model used in this study contained only 3484 trainable parameters---a tiny network by modern deep learning standards---and trained in a few hours on one GPU.
\textit{Second}, CDRNNs (like all deep neural networks) are vulnerable to local optima.
However, in an out-of-sample testing paradigm, training set likelihood maximization is not the goal, but rather \textit{generalization}.
To this end, CDRNNs can leverage the many existing techniques for robust generalization in deep neural networks \cite{srivastavaetal14}, and variability in performance can be mitigated through ensembling (see \textbf{Supplementary Information~\ref{sect:si-nhst}}).
Even in cases where lack of data prohibits out-of-sample evaluation, CDRNNs can be used to complement results from existing approaches, since they can visualize how estimates differ when simplifying assumptions are relaxed.
\textit{Third}, like all deep neural networks, there are many choice points in CDRNN model design, including the number of layers and hidden units, squashing functions, regularization and dropout strength, learning rate, and batch size.
Choices along any of these dimensions can materially impact effect estimates and generalization performance.
However, supplemental analyses show that model estimates are quite stable across diverse hyperparameters (\textbf{Supplementary Information~\ref{app:cognitive}}).
Furthermore, model comparisons are based on \textit{relative} performance between more and less constrained models within a given hyperparameterization, so it is not necessary to maximize absolute generalization performance in order to make comparisons.
Our software implementation of CDRNNs (\url{https://github.com/coryshain/cdr}) distributes with detailed documentation, provides default hyperparameters that are reasonable for many cases, and requires no familiarity with programming or deep learning.
Finally, in case of discrepant results between two hyperparameterizations, there is a simple model selection principle: prediction likelihood.
That is, the results from the hyperparameterization with the higher out-of-sample likelihood should be trusted more.
This principle also permits the use of model selection based on validation set performance in order to adapt CDRNNs to new domains, although such tuning may not be necessary in many cases, since the default parameters used in this study generalize well across diverse datasets (\textbf{Supplementary Information~\ref{app:cognitive}}).
Note that out-of-sample model selection obviates the need for the heuristic penalties on model complexity assumed in commonly-used information criteria \cite{akaike74,schwarz78}, since it directly quantifies the prediction likelihood that information criteria implicitly estimate.

In conclusion, we have demonstrated that CDRNNs obtain better estimates than standard regression analyses of observational time series and can directly capture complex nonlinear relationships between variables, permitting testing of fine-grained questions that are otherwise difficult to study.
CDRNNs therefore constitute an important advance for flexible and interpretable modeling of data from complex processes in nature.
\section{Materials and Methods}

\subsection{Datasets}

\subsubsection{Eye-Tracking (Dundee)}

Dundee \cite{kennedyetal03} is an eye-tracking dataset containing newspaper editorials read by 10 participants.
The dataset contains a total of 340,840 events (where one event is a single participant's eyes entering and then exiting a single word region).
Studies of language processing use measures derived from the eye-tracking record as indices of readers' comprehension difficulty, in order to test theories about the underlying comprehension processes.
A number of such measures exist in the literature \cite{rayner98}.
In this work, we use the following three measures:
\begin{itemize}
    \item \textit{Scan path duration}: time elapsed between entering a word region and entering a different word region.
    \item \textit{First pass duration}: time elapsed between entering a word region from the left and entering a different word region.
    \item \textit{Go-past duration}: time elapsed between entering a word region from the left and entering a different word region to the right.
\end{itemize}

Following \cite{shainschuler19}, unfixated items were excluded as well as (a) items following saccades longer than 4 words, (b) starts and ends of sentences, screens, documents, and lines, and (c) items whose duration included a blink \cite{schotteretal18}.

\subsubsection{Self-Paced Reading (Natural Stories)}

Natural Stories \cite{futrelletal20} is a crowd-sourced self-paced reading (SPR) dataset consisting of narratives and non-fiction passages read by 181 participants.
In a self-paced reading task, participants step through words in the passage on a screen by pressing a button, and the time between button presses is recorded.
The dataset contains a total of 1,013,290 events (where one event is a single participant viewing a single word token).
Following \cite{shainschuler19}, items were excluded if they have fixations shorter than 100ms or longer than 3000ms, if they start or end a sentence, if the participant missed 4 or more subsequent comprehension questions, or if the participant had fewer than 100 responses after application of the other filters.

\subsubsection{Functional Magnetic Resonance Imaging (Natural Stories)}

The Natural Stories fMRI dataset \cite{shainetal20} contains fMRI responses from 78 participants who listened to audio recordings of the \cite{futrelletal20} materials while in an MRI scanner.
Following \cite{shainschuler19}, we modeled mean activity in the core language network: six left-hemisphere fronto-temporal functional regions of interest (fROIs) that were functionally identified in each individual participant, treating \textit{fROI} as a random effect in addition to \textit{Participant}.

\subsubsection{Data Split}

Each dataset is partitioned into training (50\%), exploratory (25\%), and test (25\%) sets, using the same partitioning scheme as \cite{shainschuler19}.

\subsection{Predictors}
We use the same predictors as \cite{shainschuler19}, namely:
\begin{itemize}
    \item \textbf{Rate} (ET, SPR, fMRI): a ``deconvolutional intercept''; that is, a timestamped vector of 1's that is convolved by the model to yield an IRF representing the baseline response to an event, so named because variability in the response is driven by the rate of stimulus events in time.
    \item \textbf{Unigram surprisal} (ET, SPR, fMRI): the negative log probability of a word derived from a KenLM unigram model \cite{heafieldetal13} trained on the Gigaword 3 corpus \cite{graffetal07}.
    \item \textbf{5-gram surprisal} (ET, SPR, fMRI): the negative log probability of a word in context derived from a KenLM 5-gram model trained on the Gigaword 3 corpus.
    \item \textbf{Word length} (ET, SPR): word length in characters.
    \item \textbf{Saccade length} (ET): incoming saccade length in words.
    \item \textbf{In regression} (ET): whether a fixation is part of a regressive (backward) eye movement.
    \item \textbf{Previous was fixated} (ET): boolean indicator for whether the preceding word was fixated.
    \item \textbf{Sound power} (fMRI): Root mean squared signal power of the audio recording as computed by the Librosa software library \cite{mcfeeetal15}.
\end{itemize}

To account for the possibility of qualitatively different scan path responses to linguistic variables in regressive vs.\ non-regressive eye movements, in the Dundee scan path analyses we follow \cite{shainschuler19} in partitioning all variables in the scan path analyses into +reg and --reg variants as a function of whether the fixation occurred within a regression (+reg) or not (--reg).
Indexical predictors used by \cite{shainschuler19}, such as the position of the word within the experiment, are not needed in a CDRNN framework due to nonstationarity, and are therefore omitted.
For detailed motivation and interpretation of this set of language processing variables, see \cite{shainschuler19}.

\subsection{Model Design}
\label{sect:model_design}

We start from a ``base'' set of hyperparameters (see \textbf{Supplementary Information~\ref{sect:si-implementation}}) manually selected based on a combination of factors, including parsimony, training speed, exploratory set performance, and consistency of estimates/performance across replicates.
To explore the influence of these hyperparameter choices, we perform a limited grid search over models that deviate (up or down) from the base configuration in one of the following dimensions: number of hidden layers in the IRF, number of units per hidden layer of the IRF, L2 penalty strength on the IRF weights, L2 penalty strength on random IRF effects by participant, dropout level, learning rate, and batch size.

\subsection{Model Convergence}
\label{sect:model_convergence}

Convergence diagnosis follows the time-loss criterion of \cite{shainschuler19}.
In brief, the correlation of a performance metric with training time is tested statistically using $\alpha = 0.5$ until at least half of the most recent 100 training epochs have failed to reject the null hypothesis of no correlation, indicating that performance has stopped increasing.
For full details, see \cite{shainschuler19}.
The fMRI models used to exemplify exploratory and confirmatory CDRNN analysis used out-of-sample exploratory set likelihood (evaluated every 10 training epochs) as the diagnostic metric.
All other models used in-sample training set likelihood (evaluated every epoch) as the diagnostic metric.

\subsection{Model Comparison}
Performance of CDRNN models is statistically compared to that of LME \cite{batesetal15}, GAM \cite{wood06}, and GAMLSS \cite{rigbystasinopoulos05} models.
For Dundee and Natural Stories self-paced reading, we consider variants both with and without three additional lags per predictor to help capture delayed effects.
For Natural Stories fMRI, we pre-convolve the predictors with the canonical HRF, following evidence from \cite{shainschuler19} that this approach outperforms alternatives (linear interpolation, temporal binning, and Lanczos interpolation) that attempt to fit the HRF using discrete-time approaches.
Lagged regressors are therefore not included in the fMRI models, since the delays are already taken into account by the assumed HRF.
Performance gains in \textbf{Figure~\ref{fig:main}A} are relative to the least expressive model overall (LME with no lagged predictors).
For the Dundee and Natural Stories SPR datasets, LME and GAM baseline results in \textbf{Figure~\ref{fig:main}A} reflect the performance of models with additional lagged predictors (which is why LME performance differs from baseline in these cases).
For the fMRI dataset in which no lagged predictors were used, the LME performance gain is 0 because the LME model is identical to the baseline.
CDRNNs are also compared to (non-neural) CDR \cite{shainschuler19}.

Following \cite{shainschuler19}, LME and GAM models include by-subject random effects for every fixed effect in the model.
We attempted to follow this protocol with GAMLSS but found this to result in a range of numerical problems and fatal crashes.
GAMLSS models would only reliably run to completion when all random effects were removed except the by-subject intercept, which is the configuration used in all reported experiments.
All GAMLSS nonlinearities assume penalized B-splines with default parameters.

All baseline models except the GAM and GAMLSS models for the fMRI dataset are the same as those used in \cite{shainschuler19}.
All statistical comparisons use paired permutation tests \cite{demvsar06} of the conditional log likelihood assigned by each model to an out-of-sample test set.
To avoid unnecessary statistical comparisons, only the reference implementation of CDRNN (CDRNN base, see \textbf{Supplementary Information~\ref{sect:si-implementation}}) is evaluated on the test set.
For technical description of the permutation testing procedures, see \textbf{Supplementary Information~\ref{sect:si-nhst}}.
For reader-friendly versions of the R-style model formulae used to define each baseline, see \textbf{Supplementary Information~\ref{app:si-formulae}}.
Full implementation details necessary for reproducing both the baseline and CDRNN models are available in the public codebase: \url{https://github.com/coryshain/cdr}.

\subsection{Data Availability Statement}

All datasets analyzed in this study are publicly available and accessible online as described in \textbf{Materials and Methods: Datasets} above.
Code used to preprocess text and experiment data is available at \url{https://github.com/modelblocks/modelblocks-release}.
Code to reproduce all analyses reported in this study is available at \url{https://github.com/coryshain/cdr}.
\section*{Acknowledgments}

C.S.\ was supported by a postdoctoral fellowship from the Simons Center for the Social Brain at MIT (via the Simons Foundation).
W.S.\ was supported by the National Science Foundation grant \#1816891.
All views expressed are those of the authors and do not necessarily reflect the views of the National Science Foundation.

We would also like to thank Clara Meister and Tiago Pimentel for valuable discussion around bootstrap methods for hypothesis testing, and Ev Fedorenko for comments on the manuscript draft.

\bibliography{main_ms_arxiv}

\pagebreak

\resetlinenumber

\addappheadtotoc 

\begin{appendices}

\setcounter{table}{0}
\setcounter{figure}{0}
\renewcommand{\thetable}{A\arabic{table}}
\renewcommand{\thefigure}{A\arabic{figure}}

\section{Extended Background and Motivation}
\label{sect:si-motivation}

Here we briefly review existing approaches to analyzing observational time series and discuss key simplifying assumptions (which can be relaxed by CDRNNs) that are made in some combination by each of them: discrete time, linearity, stationarity, and homoscedasticity.
We exemplify potential problems associated with each assumption using evidence from studies of human language processing.

\subsection{Current Regression Analyses for Time Series}

In regression analyses of observational time series, linear models (LMs) are currently the dominant method.
Linear regression attempts to identify the vector of parameters $\mathbf{b}$ that models the expected value of response $y$ via linear combination with predictor vector $\mathbf{x}$:
\begin{equation}
    \text{E}(y) = \mathbf{x}^{\top}\mathbf{b}
\end{equation}
A common way of relaxing this linearity assumption is the \textit{generalized additive model} (GAM; \textit{\citen{hastietibshirani86,wood06}}), which permits arbitrary non-linear spline functions $f$ on the subsets of predictors $\mathbf{v}_1, \ldots, \mathbf{v}_k \in \mathcal{P}(\mathbf{x})$ (vectors corresponding to the powerset of elements in $\mathbf{x}$):
\begin{equation}
    \text{E}(y) = f_1(\mathbf{v_1}) + \ldots + f_k(\mathbf{v}_k)
\end{equation}
Both linear and GAM regression models can be augmented with random effects terms to capture hierarchical structure in the observations \cite{batesetal15}.
Errors in these models are assumed to be independent and identically distributed, and normal error is commonly assumed for continuous response variables.
This entails that variance is assumed constant (homoscedastic), since only the expectation $\text{E}(y)$ (and not any other distributional parameter) is modeled as a function of $\mathbf{x}$. 
This assumption can relaxed using generalized additive models for location, scale and shape (GAMLSS; \textit{\citen{rigbystasinopoulos05}}), a generalization of GAMs that admits additive nonlinear influences of predictors on up to four parameters of the predictive distribution $\mathcal{F}$ with parameter vector $\mathbf{s}$ over response $y$:
\begin{align}
    \mathbf{s} & = f_1(\mathbf{v_1}) + \ldots + f_k(\mathbf{v}_k)\\
    y & \sim \mathcal{F}(\mathbf{s})
\end{align}

When naively applied to time series, these approaches make strong temporal independence assumptions: the response $y_i$ depends solely on the corresponding predictors $\mathbf{x}_i$ and is independent of any predictor values that precede (or follow) $y_i$ in time.
This assumption can be relaxed in the design of $\mathbf{x}$ e.g., by including regressors from previous events---yielding a \textit{distributed lag} \cite{koyck54} or \textit{finite impulse response} (FIR; \textit{\citen{neuvoetal84}}) model (also called ``spillover'' in some research communities; \textit{\citen{mitchell84}})---or by including variables encoding the passage of time, which are especially useful in GAMs to permit modeling of nonstationarity \cite{baayenetal18}.
However, as argued at length in \cite{shainschuler19}, these approaches are limited in their ability to infer continuous dynamics from data with variable event durations, which are a ubiquitous feature of natural processes.\footnote{
One recent study used convolutional neural networks to parameterize discrete-time models like finite impulse response \cite{machadogivigi18}, rather than the standard approach of using fixed weights.
Although this shares our approach of using deep learning for time series analysis, it inherits the limitations of discrete-time models.
}

In response to this limitation, \cite{shainschuler19} proposed \textit{continuous-time deconvolutional regression} (CDR), a kernel-based variational Bayesian model that infers the parameterization of continuous-time impulse response functions (IRFs) from data.
In brief, in CDR, the expected value of $y_t$ at timestamp $t$ is a linear model on $\mathbf{x}'_t$, where $\mathbf{x}'_t$ is a convolution over time of preceding inputs $x(t)$ with convolution weights derived from the estimated IRF $g(t)$:
\begin{align}
    \label{eq:cdr_main}
    \text{E}(y_t) &= (\mathbf{x}'_t)^{\top}\mathbf{b}\\
    \label{eq:cdr_conv}
    \mathbf{x}'_t &= \int_0^t x(\tau)g(t - \tau) d\tau
\end{align}
CDR otherwise assumes homoscedasticity and stationarity (like LMs and GAMs) and linear/additive effects (like LMs).
The IRFs estimated by CDR describe diffusion of effects over time in continuous-time dynamical systems (like the human mind) in which previous events may continue to influence the response as the experiment unfolds.
CDR substantially improves fit to naturalistic human language processing data, while also shedding light on important aspects of processing dynamics that are otherwise difficult to obtain \cite{shainschuler19}.
For in-depth review of these and related approaches to time series analysis, especially under the possibility of delayed effects, see \cite{shainschuler19}.

CDRNNs relate the predictors $\mathbf{x}$ to the probability distribution over response $y$ using deep neural networks whose architecture ensures continuous-time deconvolution (\textbf{Introduction: The CDRNN Model} of the main article), thereby relaxing assumptions made in some combination by all methods reviewed above.
The consequences of model definitions for the kinds of information about the underlying process that can and cannot be captured by a given model type are summarized in \textbf{Table~\ref{tab:feature_comparison}} of the main article.
LMs, GAMs, and GAMLSS can only model discrete-time IRFs, whereas CDR and CDRNNs can additionally model continuous-time IRFs.
LMs and CDR can only model linear effects, whereas GAMs, GAMLSS, and CDRNNs can also model nonlinear effects.
LMs and CDR can only model linear effect interactions that are sparse (i.e., specified by the analyst), whereas GAMs and GAMLSS can also model nonlinear interactions through sparse tensor-product spline functions.
Only CDRNNs can model arbitrary nonlinear interactions over the full set of predictors, while also permitting explicit constraints on interactions and nonlinearity under an appropriate model definition.
LMs and CDR can only capture nonstationarity in the form of linear trends over time, whereas GAMs, GAMLSS, and CDRNNs can capture arbitrary nonstationarity via interactions with the time dimension.
Finally, only GAMLSS and CDRNNs can directly model influences of predictors on all parameters of the predictive distribution (distributional regression), and thereby capture heteroscedasticity in the modeled system (some Bayesian generalizations of LMs/GAMs also have this ability, see e.g., \textit{\citen{burkner18}}).
CDRNNs therefore merge the advantages of continuous-time deconvolutional modeling from CDR with the advantages of nonlinear modeling from GAM(LSS).

CDRNNs bear a close conceptual relationship to multiple recent toolkits that build either on the LM/GAM frameworks reviewed above or on deep learning.
For example, the multivariate temporal response function (mTRF) toolbox \cite{crosseetal16} supports regularized linear modeling for impulse response identification.
Nonlinear generalizations of this idea have been developed using generalized additive models \cite{ehingerdimigen19} and recurrent neural networks \cite{chehabetal22}.
These approaches are all underlyingly discrete-time in that they assume a regular sampling interval for the response variable to which the stimulus sequence must be aligned.
For high temporal resolution measures like EEG and MEG in which the sampling interval is both regular and fast compared to the stimulus stream, this assumption is appropriate.
However, it becomes problematic when events have variable duration and the sampling density of the response is low relative to the stimulus, as in language experiments using reading times or fMRI \cite{shainschuler19}.
Thus, in addition to their advantages for capturing complexities like nonlinearity, interactions, non-stationarity, and heteroscedasticity, CDRNNs can be applied to a broader range of domains than existing tools.

We now turn to the specific simplifying assumptions made by these models that underlie the constraints summarized in \textbf{Table~\ref{tab:feature_comparison}} of the main article and discuss evidence that each assumption is potentially problematic for the domain of language processing.

\subsection{Assumption: Discrete-Time Impulse Response}
\label{sect:discrete_time}

The impulse response functions within the solution spaces of LMs, GAMs, and GAMLSS are defined in discrete time.
Delayed effects must be captured by some fixed number of lagged regressors to preceding events.
There is a core difficulty in applying these models to time series generated by an underlyingly continuous system responding to variably spaced events: the discrete structure forces an indexical rather than continuous notion of time.
Methods of coercing the model and/or data are needed in order to align the lags with preceding events, which either destroys temporal information or compromises model identifiability (for further discussion, see \textit{\citen{shainschuler19}}).
CDR(NN) relaxes this assumption by permitting a continuous impulse response in the form of continuous kernels with trainable parameters.

Delayed effects are ubiquitous in human language comprehension \cite{kutashillyard80,mitchell84,vandyke07,smithlevy13,shainschuler19}
and prior evidence indicates that this discrete-time assumption may be ill-suited to capture them.
In particular, multiple lines of evidence indicate that, for diverse processing phenomena, the key determinant of delayed effects is how long ago the trigger word occurred \textit{in time}, rather than how many words back it occurred.
A large electrophysiological literature on human language processing investigates event-related potentials (ERPs, in electroencephalography, EEG) or event-related fields (ERFs, in magnetoencephalography, MEG), that is, IRFs that characterize the brain response to words in context.
ERPs are described by their average peak delay in ms, such as the N400 (a negative deflection occurring around 400ms after word onset) and the P600 (a positive deflection occurring around 600ms after word onset).
Studies consistently find effects consistent with well-known stereotyped continuous-time ERPs in response to phonological \cite{connollyphillips94,kaanetal07}, morphological \cite{osterhoutmobley95,allenetal03},  syntactic \cite{osterhoutholcomb92,ainsworthdarnelletal98}, and semantic \cite{kutashillyard80,vanberkumetal99} aspects of language, despite variable word presentation rates across experiments.
This suggests that the relevant cognitive processes unfold in continuous, rather than discrete, time.

Related work suggests that the human language processor may allow information processing to lag behind perception when processing load spikes \cite{boumadevoogd74,erlichrayner83,kliegletal06,mollicapiantadosi17}.
If these lags are driven by rate-limited processing \cite{mollicapiantadosi17}, this entails that the processing mechanisms that underlie them unfold in continuous time, rather than e.g., delaying processing until the next word is encountered, as implied by discrete-time models.

Evidence not only indicates that effect delays in human language comprehension are largely continuous-time rather than discrete-time, but also that discrete-time approximations to them are likely often poor quality due to extensive variability in word duration in natural language, whether spoken \cite{bakerbradlow09,dembergetal12} or read \cite{franketal13,futrelletal20}.
There is likely a substantial difference in the level of influence exerted by the preceding word depending on whether it occurred 100ms vs. 1000ms ago, a difference which is ignored by discrete-time models.
Relaxing the discrete-time assumption using CDR leads to substantial improvements to model fit in reading and neuroimaging measures of human language processing relative to comparable discrete-time controls \cite{shainschuler19}, suggesting that these controls lack access to critical information about underlyingly continuous comprehension processes.\footnote{Note that because CDR subsumes linear mixed-effects (LME) models (since any LME model can be expressed as a CDR model where $g$ of eq.~\ref{eq:cdr_conv} is fixed to be the Dirac $\delta$ function), discrete-time IRFs are still available when needed in a CDR framework, simply by including lagged regressors in the same way.}

\subsection{Assumption: Additive Linear Effects}

LMs, GAMs, GAMLSS, and CDR all model the response as a weighted sum of the predictors, and LMs and CDR additionally assume that these weights scale linearly on the predictors (GAMs relax the latter assumption by deriving weights through nonlinear spline functions).
The linearity assumptions of LMs and CDR can be problematic for model interpretation (the best-fit line may be a poor fit to a non-linear function), and may prevent the discovery of theoretically-relevant nonlinearities.

Predictor interactions are subject to the same constraints, and can only be modeled if explicitly included by the analyst.
As a result, each of these models has the following two properties:
\begin{enumerate}
    \item \textit{Correlated predictors are in zero-sum competition.}
    Increasing the effect of one covariate requires a corresponding decrease in the effect of the other.
    In cases of sufficiently high correlation, this can result in large-magnitude estimates of opposite sign \cite{wurmfisicaro14}, which are difficult to interpret.
    \item \textit{Interactions must be anticipated in advance.} This is of course a practical constraint: any LM or GAM could in principle include all possible interactions, and GAMs can further include multivariate spline functions of the full set of predictors.
    However, beyond a small handful of predictors, these approaches quickly produce problems for inference and computation due to combinatorial explosion.
    Analysts are therefore typically constrained by model identifiability considerations to a include small subset of interactions of interest based on prior evidence or other domain knowledge.
\end{enumerate}
These two properties can be problematic for natural processes, which often involve many correlated and potentially interacting variables.
They also prevent flexible inference of unanticipated interactions, which could serve as the basis for new discoveries.

The functional form and interaction structure of effects are of great interest to key questions in the study of human language processing.
For example, one prominent debate concerns the functional form of predictability effects in reading \cite{levyjaeger07,smithlevy13,brotherskuperberg21,wilcoxetal20}, which has implications for extant theories of human language processing \cite{smithlevy13}.
Another debate concerns the existence of an interaction between word frequency and word predictability effects on incremental language comprehension effort \cite{rayneretal04,ashbyetal05,kretzschmaretal15}, which also has implications for theories of human language processing \cite{reichleetal98,norris06,levy08,coltheartetal01}.
These debates concentrate on known theoretical implications for the functional form and interaction structure of language processing effects, but, given the complexity of the task of inferring meaning from language, it is likely that there exist other kinds of nonlinearities and effect interactions not yet covered by existing theory.
Discovering such patterns could advance the field, but this is not possible in standard analysis frameworks unless analysts deliberately look for them.

\subsection{Assumption: Stationarity (Time-Invariance)}

Naively implemented, LM, GAM, GAMLSS, and CDR models of time series assume a stationary (time-invariant) function mapping predictors to responses.
If the underlying response function is nonstationary (time-dependent), this can lead to poor fit and misleading estimates.
Some control of nonstationarity is nonetheless possible under these approaches by including autoregressive terms \cite{baayenetal17} or adding temporal features to the predictors \cite{baayenetal18}.
The kinds of nonstationarity that models can capture is thus determined by the kinds of effects they can capture: LMs and CDR can capture nonstationarity in the form of linear trends along some representation of the time dimension, whereas GAMs and GAMLSS can also capture nonlinear effects of time via spline functions.\footnote{Note that models thus defined are nonstationary only in that temporal features have been included in their inputs.
The mathematical function mapping inputs to outputs remains stationary; that function can simply condition on a representation of time.
}

Existing evidence indicates that responses in studies of human language processing are nonstationary, and in ways that arguably affect scientific inferences if not taken into account.
For example, participants are known to habituate strongly to tasks in language processing experiments, such that e.g., response times decrease dramatically and nonlinearly over the course of the experiment \cite{baayenetal18,prasadlinzen19}.
\cite{prasadlinzen19} have even argued that this task adaptation effect may have driven previous reports of ``syntactic priming'' \cite{fineetal13}, and that syntactic priming effects may only be detectable with much larger sample sizes.
\cite{baayenetal18} have likewise argued for an important influence of latent factors like attention and fatigue, which change over time, affect responses, and cannot be directly observed.
The full extent of the impact of these kinds of nonstationarities on estimates of cognitive effects of interest is not yet well understood.

\subsection{Assumption: Homoscedasticity (Constant Variance)}

LMs, GAMs, and CDR all assume a homoscedastic data-generating model: the predictors influence the mean response, but the variance (and/or any other distributional parameter) is treated as constant across time.
The many ways in which this assumption can be violated by time series is the subject of a vast statistical literature \cite{koyck54,coxisham80,sims80,engle82}, as are the implications of such violations for statistical inferences \cite{trenkler84,longervin00,youetal07,rosopaetal13,cattaneoetal18}.
These concerns take on special importance for (a) analyses in which the entire predictive distribution over the response (not just the expectation) is a quantity of interest, or (b) likelihood-based out-of-sample comparisons between hypotheses, where poor fit between the modeled and true predictive distribution can lead to failure to generalize.

Both of these concerns are pertinent to the study of the study of language processing.
For example, prior work has argued that cognitive variables like word frequency and predictability have differential effects on different parameters of the distribution of eye gaze during reading, and thus correspond to distinct cognitive mechanisms \cite{staubetal10,staub2011:exgauss}.
In addition, with growing interest in larger-scale naturalistic datasets for language processing research \cite{kennedyetal03,lukechristianson16,copetal17,futrelletal20,shainetal20} comes the growing possibility of drawing conclusions from overfitted statistical models of these data.
One approach to addressing this possibility is to perform statistical comparisons based on the likelihood assigned by models to out-of-sample data, ensuring that tests favor models with more generalizable descriptions of the modeled system \cite{shainetal20,shainschuler19}.
This approach crucially relies not just on an accurate model of the expected response, but on an accurate model of the \textit{distribution} of responses.
Models that fail to capture the structure of that distribution will struggle in the out-of-sample evaluation, with poor likelihood at points where the variance is over- or underestimated.

\subsection{Relaxing Assumptions with Deep Learning}

\textit{Deep learning} is the use of multilayer artificial neural networks for function approximation.
An artificial neural network is a supervised machine learning algorithm that transforms inputs into outputs via nonlinear transformations with learned parameters.
A \textit{deep} neural network (DNN) involves sequential transformations of the network's own hidden states, allowing the network to learn complex nonlinear interactions of the input features.
DNNs have been shown by mathematical analyses to be universal function approximators \cite{hornik91}, and thousands of practical applications have demonstrated their effectiveness for learning complex patterns in real data, to the point that DNNs now dominate engineering fields like natural language processing and computer vision \cite{lecunetal15}.
A DNN is trained by \textit{backpropagation} \cite{rumelhartetal86}, which involves (1) computing the partial derivatives of some objective function (e.g., negative log likelihood) with respect to each of the model's parameters and (2) changing those parameters via a deterministic function of the computed derivatives, seeking to minimize the objective.

The relevance of deep learning to the present goal of flexible time series analysis is its potential to relax all of the aforementioned simplifying assumptions by fitting continuous nonlinear multivariate transforms of the input.
The IRF can be represented using a deep neural transformation, and the inputs and outputs of that transformation determine which key assumptions it can relax.
If the input includes the scalar offset in time between a predictor and a target response, then the IRF is continuous- rather than discrete-time.
If the input includes the predictors themselves, then the IRF includes arbitrary nonlinear interactions between the full set of predictors.
If the input includes the timestamp of the predictor vector, then the IRF can represent nonstationarity of effects.
And if the output includes all parameters of the predictive distribution, then the IRF models stimulus-driven heteroscedasticity.
The design details that enable these abilities are described in \textbf{Introduction: The CDRNN Model} of the main article.

\section{Detailed Motivation for Proposed CDRNN Architecture}
\label{sect:si-buildup}

\input{fig2/buildup}

Our proposed CDRNN architecture (\textbf{Figure~\ref{fig:model}} of the main article) contains design elements whose motivation may not be immediately transparent to all readers.
Therefore, in this section, we build up to our proposed architecture in a step-by-step conceptual progression, starting from a maximally simple multiple regression model.
This progression is visualized in \textbf{Supplementary Figure~\ref{fig:buildup}}.

\textbf{Supplementary Figure~\ref{fig:buildup-a}} depicts a simple multiple regression model consisting of a mapping from the predictor vector to their evoked effects on the expected response, relative to an intercept or bias term.
If this mapping is linear, then \textbf{Supplementary Figure~\ref{fig:buildup-a}} depicts a linear regression model.
However, the mapping could be non-linear, as in a generalized additive model or a neural network.
Since (for reasons discussed in the main article) we seek to develop a neural network approach to time series regression, we allow this mapping to be instantiated as a neural network, and therefore to be potentially non-linear on the predictors.

However, even assuming a neural network mapping, the model in \textbf{Supplementary Figure~\ref{fig:buildup-a}} is highly constrained.
In particular, it assumes stationarity over time, homoscedasticity, and an absence of delayed effects.
To relax the stationarity assumption, we add the timestamp of the predictor vector to the inputs to the network, as shown in \textbf{Supplementary Figure~\ref{fig:buildup-b}}.
This allows the network to capture arbitrary non-linear effects of time on the response evoked by the predictor vector.
To relax the homoscedasticity assumption, we allow the model to generate a vector of parameters for the predictive distribution over the response, rather than generating the expected response directly, as shown in \textbf{Supplementary Figure~\ref{fig:buildup-c}}.
In this way, uncertainty over the response is also modeled as a function of the predictors, thereby allowing for non-constant variance across the time series.

However, a major shortcoming of this more flexible model is that its flexibility cannot be constrained in any way.
Predictors can be added to or removed from the model, but it would not be possible to e.g., force the model to respect a linear effect of a predictor.
This severely restricts the range of hypotheses that can be tested (by restricting the range of constrained null hypotheses that can be implemented and compared to the alternative hypothesis of e.g., a non-linear effect).
To address this concern, we instead use the neural network to generate a \textit{linear map} from predictor space to the parameter space of the predictive distribution (\textbf{Supplementary Figure~\ref{fig:buildup-d}}).
This permits greater control over the model by e.g., allowing a predictor to be removed from the input to the neural network while it is retained in the output, thereby enforcing a linear effect.

This model still lacks any capacity to estimate an IRF describing the change in evoked response over time.
To address this, we additionally include the time offset $d$ as an input to the network, and we add to the output of the linear map a bias term (\textit{Rate}, a vector of 1's) capturing overall effects of stimulus timing (\textbf{Supplementary Figure~\ref{fig:buildup-e}}).
These modifications allow us to apply the network simultaneously to multiple inputs within some time window of the target response, resulting in a deconvolutional model that implicitly estimates an IRF describing the influence of an impulse as a function of (i) its properties and (ii) its distance in time from the target.

However, a major shortcoming of this design is that nonlinearity in predictor space cannot be estimated independently of the impulse response over time.
This prevents independent statistical testing of these two aspects of the response.
To allow the model to decouple non-linearity in predictor space from the temporal dynamics of the response, we include an additional neural network component $\fin$ for ``input processing'' (\textbf{Supplementary Figure~\ref{fig:buildup-f}}).
Nonlinearities in predictor space can be estimated by $\fin$ while e.g., enforcing a Dirac $\delta$ IRF, allowing the model to approximate a GAM-like non-linear regression model that lacks continuous-time temporal dynamics.

Note that the full model in \textbf{Supplementary Figure~\ref{fig:buildup-f}} subsumes all submodels described in \textbf{Supplementary Figure~\ref{fig:buildup}}, and thus can be pared down to any of these submodels as motivated by the analyst's goals and domain knowledge.
By manipulating information flow through this final network as exemplified above, a wide range of constrained null models can be implemented and statistically tested against their relaxations.
Our proposed CDRNN architecture thus provides a powerful and flexible framework for scientific inference.

\section{Extended CDRNN Model}
\label{sect:si-extended}

The simple CDRNN model defined in the main article assumes independence between impulses in their effects on the response.
This assumption is helpful for model interpretation, since the causal effects of input features on estimated responses can be queried directly without reference to context.
However, this assumption can also be violated in practice: the response evoked by an impulse may depend in part on the nature and timing of prior impulses.
Relaxing this assumption requires a generalized definition of the regression model to allow $\fin$ and $\firf$ to condition on a \textit{sequence} of impulses, rather than on a single impulse, as follows:
\begin{align}
    \Xprime & \defeq \fin\left(\begin{bmatrix}\t & \X \end{bmatrix}; \uin \right)\\
    \label{eq:cdrnn-extended}
    \G_1, \ldots, \G_N & \defeq \firf \left(\begin{bmatrix} \d & \t & \Xprime \end{bmatrix}; \uirf\right)
\end{align}
This additional flexibility presents a danger to interpretability, since, by conditioning on the entire design matrix, it can leverage context in uninterpretable ways, potentially decoupling the IRF outputs $\G_n$ from their inputs $\x_n$.
This property would make it impractical or impossible to reliably query impulse response functions from the model's estimates.
There are thus many possible implementations of eq.~\ref{eq:cdrnn-extended} that would not constitute an interpretable deconvolutional model---e.g., a transformer \cite{vaswanietal17} implementation of $\firf$.

However, as explained below, at least one kind of neural network time series model can relax independence over time while largely retaining a capacity for interpretable IRF estimation: a recurrent neural network (\textit{RNN; \citen{elman91}}).
An RNN $\frnn$ with transition function $r$ and parameters $\w \in \R^{W}$ is a time series model that applies recursively to sequential inputs $\x_t, 1 \leq t \leq T$ as follows (where $\frnn(\x_0)$ is an ``initial state'', typically $\mathbf{0}$):
\begin{equation}
    \frnn(\x_t) \defeq r\left(\x_t, \frnn(\x_{t-1}); \w\right)
\end{equation}
Many possible definitions of $r$ have been proposed in the deep learning literature.
In this work, $r$ is assumed to be the transition function of a long short-term memory (LSTM) network \cite{Hochreiter1997}.

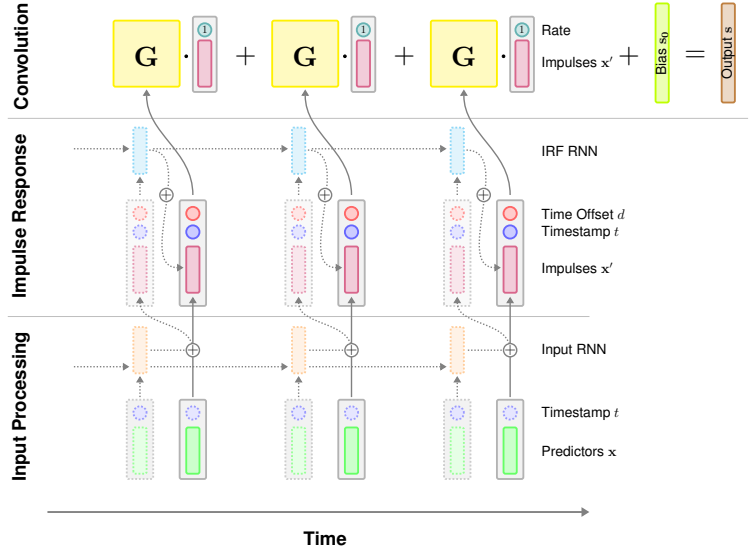
\begin{figure}
\centering
\sffamily
\resizebox{0.65\textwidth}{!}{
\begin{tikzpicture}
        
    \draw[draw=white,fill=white] (-10.5, 14.25) rectangle ++(1, 5.5);
    \draw[line width=1,color=gray!50] (-12, 13.9) to  (16, 13.9);
    \draw[line width=1,color=gray!50] (-12, 6.4) to  (10, 6.4);
    \draw[arrows={-Triangle},line width=2,color=gray] (-10.5, -1) to  (10, -1);
    \draw[draw=lime,fill=lime!30,rounded corners=2pt,line width=2] (12.5, 14.5) rectangle ++(0.5, 3.75);
    \draw[draw=brown,fill=brown!30,rounded corners=2pt,line width=2] (15, 14.5) rectangle ++(0.5, 3.75);
    \node[scale=2] at (0, -2) {\textbf{Time}};
    \node[scale=3] at (-3, 16.25) {$+$};
    \node[scale=3] at (3, 16.25) {$+$};
    \node[scale=3] at (11.5, 16.25) {$+$};
    \node[scale=3] at (14, 16.25) {$=$};

    \node[rotate=90,scale=2] at (-11.5, 16.25) {\textbf{Convolution}};
    \node[rotate=90,scale=2] at (-11.5, 10.2) {\textbf{Impulse Response}};
    \node[rotate=90,scale=2] at (-11.5, 3) {\textbf{Input Processing}};
    
    \node[anchor=west,scale=1.5] at (8, 17.25) {Rate};
    \node[anchor=west,scale=1.5] at (8, 15.975) {Impulses $\xprime$};
    \node[anchor=west,scale=1.5] at (8, 12.675) {IRF RNN};
    \node[anchor=west,scale=1.5] at (8, 10.3) {Time Offset $d$};
    \node[anchor=west,scale=1.5] at (8, 9.6) {Timestamp $t$};
    \node[anchor=west,scale=1.5] at (8, 8.175) {Impulses $\xprime$};
    \node[anchor=west,scale=1.5] at (8, 5.125) {Input RNN};
    \node[anchor=west,scale=1.5] at (8, 2.75) {Timestamp $t$};
    \node[anchor=west,scale=1.5] at (8, 1.325) {Predictors $\x$};
    \node[rotate=90,scale=1.5] at (12.75, 16.375) {Bias $\snull$};
    \node[rotate=90,scale=1.5] at (15.25, 16.375) {Output $\s$};

    \begin{scope}[shift={(-6, 0)}]
            
        \begin{scope}[shift={(-2,15)}]
            \draw[draw=yellow,fill=yellow!30,rounded corners=2pt,line width=2] (0, 0.1) rectangle ++(2.5, 2.5);
            \node[scale=3] at (1.25, 1.25) {$\G$};
            \filldraw[color=black] (2.75, 1.25) circle (0.05);
        \end{scope}
        \begin{scope}[shift={(1.2, 15)}]
            \draw[draw=gray!60,fill=gray!10,rounded corners=2pt,line width=2] (-0.2, -0.1) rectangle ++(0.9,2.85);
            \draw[draw=purple!60,fill=purple!20,rounded corners=2pt,line width=2] (0, 0.1) rectangle ++(0.5,1.75);
            \draw[draw=teal!60,fill=teal!20,line width=2] (0.25, 2.25) circle(0.25);
            \node at (0.25, 2.25) {\large$1$};
        \end{scope}
        
        \draw[arrows={-Triangle},line width=1.5,dotted,color=gray] (-3.5, 12.75) to  (-1.4, 12.75);
        
        
        \draw[draw=cyan!40,fill=cyan!10,dotted,rounded corners=2pt,line width=2] (-1.25,11.8) rectangle ++(0.5,1.75);
        
        \begin{scope}[shift={(-1, 7.1)}]
            \draw[draw=gray!40,fill=gray!5,dotted,rounded corners=2pt,line width=2] (-0.5,-0.3) rectangle ++(1,4);
            \draw[draw=blue!40,fill=blue!10,dotted,line width=2] (0,2.5) circle(0.25);
            \draw[draw=red!40,fill=red!10,dotted,line width=2] (0,3.2) circle(0.25);
            \draw[draw=purple!40,fill=purple!10,dotted,rounded corners=2pt,line width=2] (-0.25,0.2) rectangle ++(0.5,1.75);
        \end{scope}
        
        \begin{scope}[shift={(1, 7.1)}]
            \draw[draw=gray!60,fill=gray!10,rounded corners=2pt,line width=2] (-0.5,-0.3) rectangle ++(1,4);
            \draw[draw=red!60,fill=red!20,line width=2] (0,3.2) circle(0.25);
            \draw[draw=blue!60,fill=blue!20,line width=2] (0,2.5) circle(0.25);
            \draw[draw=purple!60,fill=purple!20,rounded corners=2pt,line width=2] (-0.25,0.2) rectangle ++(0.5,1.75);
        \end{scope}
        
        \draw[arrows={-Triangle},line width=1.5,dotted,color=gray] (-1, 11) to  (-1, 11.7);
        \draw[arrows={-Triangle},line width=1.5,dotted,color=gray] (-0.65, 12.7) to [out=0,in=180] (0.65, 8.25);
        \draw[arrows={-Triangle},line width=1.5,color=gray] (1, 11) to [out=90, in=270]  (-0.75, 15);
        \begin{scope}[shift={(0, 11)}]
            \draw[fill=white,line width=1.5,draw=gray] (0, 0) circle (.25);
            \node at (0, 0) {\LARGE$+$};
        \end{scope}

        
        
        \draw[draw=orange!40,fill=orange!10,dotted,rounded corners=2pt,line width=2] (-1.25,4.25) rectangle ++(0.5,1.75);
        
        \begin{scope}[shift={(-1, 0.25)}]
            \draw[draw=gray!40,fill=gray!10,dotted,rounded corners=2pt,line width=2] (-0.5,0) rectangle ++(1,3);
            \draw[draw=blue!40,fill=blue!10,dotted,line width=2] (0,2.5) circle(0.25);
            \draw[draw=green!40,fill=green!10,dotted,rounded corners=2pt,line width=2] (-0.25,0.2) rectangle ++(0.5,1.75);
        \end{scope}
        
        \begin{scope}[shift={(1, 0.25)}]
            \draw[draw=gray!60,fill=gray!10,rounded corners=2pt,line width=2] (-0.5,0) rectangle ++(1,3);
            \draw[draw=blue!40,fill=blue!10,dotted,line width=2] (0,2.5) circle(0.25);
            \draw[draw=green!60,fill=green!20,rounded corners=2pt,line width=2] (-0.25,0.2) rectangle ++(0.5,1.75);
        \end{scope}
        
        \draw[arrows={-Triangle},line width=1.5,dotted,color=gray] (-1, 3.35) to  (-1, 4.15);
        \draw[line width=1.5,dotted,color=gray] (-0.6, 5.125) to  (1, 5.125);
        \draw[arrows={-Triangle},line width=1.5,color=gray] (1, 3.35) to  (1, 7.15);
        \draw[arrows={-Triangle},line width=1.5,dotted,color=gray] (1, 5.125) to [out=90,in=270] (-1, 7.15);
        \draw[arrows={-Triangle},line width=1.5,dotted,color=gray] (-3.5, 4.5) to  (-1.4, 4.5);
        \begin{scope}[shift={(1, 5.125)}]
            \draw[fill=white,line width=1.5,draw=gray] (0, 0) circle (.25);
            \node at (0, 0) {\LARGE$+$};
        \end{scope}
    \end{scope}

    \begin{scope}[shift={(0, 0)}]
            
        \begin{scope}[shift={(-2,15)}]
            \draw[draw=yellow,fill=yellow!30,rounded corners=2pt,line width=2] (0, 0.1) rectangle ++(2.5, 2.5);
            \node[scale=3] at (1.25, 1.25) {$\G$};
            \filldraw[color=black] (2.75, 1.25) circle (0.05);
        \end{scope}
        \begin{scope}[shift={(1.2, 15)}]
            \draw[draw=gray!60,fill=gray!10,rounded corners=2pt,line width=2] (-0.2, -0.1) rectangle ++(0.9,2.85);
            \draw[draw=purple!60,fill=purple!20,rounded corners=2pt,line width=2] (0, 0.1) rectangle ++(0.5,1.75);
            \draw[draw=teal!60,fill=teal!20,line width=2] (0.25, 2.25) circle(0.25);
            \node at (0.25, 2.25) {\large$1$};
        \end{scope}
        
        \draw[arrows={-Triangle},line width=1.5,dotted,color=gray] (-6.5, 12.75) to  (-1.4, 12.75);

        
        \draw[draw=cyan!40,fill=cyan!10,dotted,rounded corners=2pt,line width=2] (-1.25,11.8) rectangle ++(0.5,1.75);
        
        \begin{scope}[shift={(-1, 7.1)}]
            \draw[draw=gray!40,fill=gray!5,dotted,rounded corners=2pt,line width=2] (-0.5,-0.3) rectangle ++(1,4);
            \draw[draw=blue!40,fill=blue!10,dotted,line width=2] (0,2.5) circle(0.25);
            \draw[draw=red!40,fill=red!10,dotted,line width=2] (0,3.2) circle(0.25);
            \draw[draw=purple!40,fill=purple!10,dotted,rounded corners=2pt,line width=2] (-0.25,0.2) rectangle ++(0.5,1.75);
        \end{scope}
        
        \begin{scope}[shift={(1, 7.1)}]
            \draw[draw=gray!60,fill=gray!10,rounded corners=2pt,line width=2] (-0.5,-0.3) rectangle ++(1,4);
            \draw[draw=red!60,fill=red!20,line width=2] (0,3.2) circle(0.25);
            \draw[draw=blue!60,fill=blue!20,line width=2] (0,2.5) circle(0.25);
            \draw[draw=purple!60,fill=purple!20,rounded corners=2pt,line width=2] (-0.25,0.2) rectangle ++(0.5,1.75);
        \end{scope}
        
        \draw[arrows={-Triangle},line width=1.5,dotted,color=gray] (-1, 11) to  (-1, 11.7);
        \draw[arrows={-Triangle},line width=1.5,dotted,color=gray] (-0.65, 12.7) to [out=0,in=180] (0.65, 8.25);
        \draw[arrows={-Triangle},line width=1.5,color=gray] (1, 11) to [out=90, in=270]  (-0.75, 15);
        \draw[arrows={-Triangle},line width=1.5,dotted,color=gray] (-6.6, 4.5) to  (-1.4, 4.5);
        \begin{scope}[shift={(0, 11)}]
            \draw[fill=white,line width=1.5,draw=gray] (0, 0) circle (.25);
            \node at (0, 0) {\LARGE$+$};
        \end{scope}

        
        
        \draw[draw=orange!40,fill=orange!10,dotted,rounded corners=2pt,line width=2] (-1.25,4.25) rectangle ++(0.5,1.75);
        
        \begin{scope}[shift={(-1, 0.25)}]
            \draw[draw=gray!40,fill=gray!10,dotted,rounded corners=2pt,line width=2] (-0.5,0) rectangle ++(1,3);
            \draw[draw=blue!40,fill=blue!10,dotted,line width=2] (0,2.5) circle(0.25);
            \draw[draw=green!40,fill=green!10,dotted,rounded corners=2pt,line width=2] (-0.25,0.2) rectangle ++(0.5,1.75);
        \end{scope}
        
        \begin{scope}[shift={(1, 0.25)}]
            \draw[draw=gray!60,fill=gray!10,rounded corners=2pt,line width=2] (-0.5,0) rectangle ++(1,3);
            \draw[draw=blue!40,fill=blue!10,dotted,line width=2] (0,2.5) circle(0.25);
            \draw[draw=green!60,fill=green!20,rounded corners=2pt,line width=2] (-0.25,0.2) rectangle ++(0.5,1.75);
        \end{scope}
        
        \draw[arrows={-Triangle},line width=1.5,dotted,color=gray] (-1, 3.35) to  (-1, 4.15);
        \draw[line width=1.5,dotted,color=gray] (-0.6, 5.125) to  (1, 5.125);
        \draw[arrows={-Triangle},line width=1.5,color=gray] (1, 3.35) to  (1, 7.15);
        \draw[arrows={-Triangle},line width=1.5,dotted,color=gray] (1, 5.125) to [out=90,in=270] (-1, 7.15);
        \begin{scope}[shift={(1, 5.125)}]
            \draw[fill=white,line width=1.5,draw=gray] (0, 0) circle (.25);
            \node at (0, 0) {\LARGE$+$};
        \end{scope}
    \end{scope}

    \begin{scope}[shift={(6, 0)}]
            
        \begin{scope}[shift={(-2,15)}]
            \draw[draw=yellow,fill=yellow!30,rounded corners=2pt,line width=2] (0, 0.1) rectangle ++(2.5, 2.5);
            \node[scale=3] at (1.25, 1.25) {$\G$};
            \filldraw[color=black] (2.75, 1.25) circle (0.05);
        \end{scope}
        \begin{scope}[shift={(1.2, 15)}]
            \draw[draw=gray!60,fill=gray!10,rounded corners=2pt,line width=2] (-0.2, -0.1) rectangle ++(0.9,2.85);
            \draw[draw=purple!60,fill=purple!20,rounded corners=2pt,line width=2] (0, 0.1) rectangle ++(0.5,1.75);
            \draw[draw=teal!60,fill=teal!20,line width=2] (0.25, 2.25) circle(0.25);
            \node at (0.25, 2.25) {\large$1$};
        \end{scope}
        
        \draw[arrows={-Triangle},line width=1.5,dotted,color=gray] (-6.5, 12.75) to  (-1.4, 12.75);
        
        
        \draw[draw=cyan!40,fill=cyan!10,dotted,rounded corners=2pt,line width=2] (-1.25,11.8) rectangle ++(0.5,1.75);
        
        \begin{scope}[shift={(-1, 7.1)}]
            \draw[draw=gray!40,fill=gray!5,dotted,rounded corners=2pt,line width=2] (-0.5,-0.3) rectangle ++(1,4);
            \draw[draw=blue!40,fill=blue!10,dotted,line width=2] (0,2.5) circle(0.25);
            \draw[draw=red!40,fill=red!10,dotted,line width=2] (0,3.2) circle(0.25);
            \draw[draw=purple!40,fill=purple!10,dotted,rounded corners=2pt,line width=2] (-0.25,0.2) rectangle ++(0.5,1.75);
        \end{scope}
        
        \begin{scope}[shift={(1, 7.1)}]
            \draw[draw=gray!60,fill=gray!10,rounded corners=2pt,line width=2] (-0.5,-0.3) rectangle ++(1,4);
            \draw[draw=red!60,fill=red!20,line width=2] (0,3.2) circle(0.25);
            \draw[draw=blue!60,fill=blue!20,line width=2] (0,2.5) circle(0.25);
            \draw[draw=purple!60,fill=purple!20,rounded corners=2pt,line width=2] (-0.25,0.2) rectangle ++(0.5,1.75);
        \end{scope}
        
        \draw[arrows={-Triangle},line width=1.5,dotted,color=gray] (-1, 11) to  (-1, 11.7);
        \draw[arrows={-Triangle},line width=1.5,dotted,color=gray] (-0.65, 12.7) to [out=0,in=180] (0.65, 8.25);
        \draw[arrows={-Triangle},line width=1.5,color=gray] (1, 11) to [out=90, in=270]  (-0.75, 15);
        \begin{scope}[shift={(0, 11)}]
            \draw[fill=white,line width=1.5,draw=gray] (0, 0) circle (.25);
            \node at (0, 0) {\LARGE$+$};
        \end{scope}

        
        
        \draw[draw=orange!40,fill=orange!10,dotted,rounded corners=2pt,line width=2] (-1.25,4.25) rectangle ++(0.5,1.75);
        
        \begin{scope}[shift={(-1, 0.25)}]
            \draw[draw=gray!40,fill=gray!10,dotted,rounded corners=2pt,line width=2] (-0.5,0) rectangle ++(1,3);
            \draw[draw=blue!40,fill=blue!10,dotted,line width=2] (0,2.5) circle(0.25);
            \draw[draw=green!40,fill=green!10,dotted,rounded corners=2pt,line width=2] (-0.25,0.2) rectangle ++(0.5,1.75);
        \end{scope}
        
        \begin{scope}[shift={(1, 0.25)}]
            \draw[draw=gray!60,fill=gray!10,rounded corners=2pt,line width=2] (-0.5,0) rectangle ++(1,3);
            \draw[draw=blue!40,fill=blue!10,dotted,line width=2] (0,2.5) circle(0.25);
            \draw[draw=green!60,fill=green!20,rounded corners=2pt,line width=2] (-0.25,0.2) rectangle ++(0.5,1.75);
        \end{scope}
        
        \draw[arrows={-Triangle},line width=1.5,dotted,color=gray] (-1, 3.35) to  (-1, 4.15);
        \draw[line width=1.5,dotted,color=gray] (-0.6, 5.125) to  (1, 5.125);
        \draw[arrows={-Triangle},line width=1.5,color=gray] (1, 3.35) to  (1, 7.15);
        \draw[arrows={-Triangle},line width=1.5,dotted,color=gray] (1, 5.125) to [out=90,in=270] (-1, 7.15);
        \draw[arrows={-Triangle},line width=1.5,dotted,color=gray] (-6.6, 4.5) to  (-1.4, 4.5);
        \begin{scope}[shift={(1, 5.125)}]
            \draw[fill=white,line width=1.5,draw=gray] (0, 0) circle (.25);
            \node at (0, 0) {\LARGE$+$};
        \end{scope}
    \end{scope}

\end{tikzpicture}}

\caption{\textbf{CDRNN (extended) architecture}. A graphical depiction of the extended CDRNN forward pass (including recurrent connections for context-dependence) for generating one prediction.
Scalars are shown as circles, vectors are shown as narrow boxes, matrices are shown as wider boxes, and deep neural network transformations are shown as arrows.
Computation proceeds in three stages (bottom-to-top): (i) processing the inputs, (ii) applying the impulse response, and (iii) convolving the impulses with the IRF (convolution weights) over time to generate a parameterization for the predictive distribution over the response.
At the convolution stage, the impulses are augmented with bias term (\textit{rate}) that allows the model to capture generalized effects of the rate of events in time.
Components shown with dotted lines are not used in the base CDRNN implementation in this study (although they are explored in the full set of analyses).
}
\label{fig:model-full}

\end{figure}

We incorporate RNNs into the CDRNN model design as schematized in \textbf{Supplementary Figure~\ref{fig:model-full}}.
As shown, RNNs can be used to introduce context-dependence into either $\fin$ or $\firf$.
In both cases, the RNN hidden states additively are mixed into the feedforward hidden states, allowing the input processing and/or impulse response functions to change for the same predictor vector as a function of context.

The advantage of an RNN for CDRNN modeling is that---unlike convolutional neural networks \cite{lecunetal89} or transformers \cite{vaswanietal17}---it is \textit{stateful}: all contextual influences on model behavior must be mediated through a fixed dimensional state vector that evolves over time.
The IRF can thus be queried from the model \textit{without supplying explicit context} simply by fixing the state at some value.
For example, our CDRNN implementation collects an exponential moving average of RNN hidden state values during training and uses this to parameterize the network for average-case IRF estimation.
This procedure results in qualitatively similar average-case IRF estimates and predictive performance to feedforward-only models (\textbf{Supplementary Information~\ref{app:si-synth} \& \ref{app:cognitive}}).
Thus, we consider our RNN-based approach to be an effective compromise between (i) flexibility in relaxing independence assumptions and (ii) capacity for interpretable IRF estimation.

\section{Implementation}
\label{sect:si-implementation}

These analyses use a publicly available implementation of CDR that has been extended to additionally support CDRNN (\url{https://github.com/coryshain/cdr}).
The system is written in Python TensorFlow \cite{abadietal15}, an open-source deep learning library.
All models reported here use $\fin = \text{identity}$, allowing effect interactions and nonlinearities to be handled by $\firf$.
The model of reference throughout this article uses a feedforward implementation of $\firf$ with the following key hyperparameters:
\begin{itemize}
    \item \textbf{Hidden layers:} 2
    \item \textbf{Units per hidden layer:} 32
    \item \textbf{L2 weight regularization penalty:} 5
    \item \textbf{L2 random effects regularization penalty:} 10
    \item \textbf{Dropout level:} 0.2
    \item \textbf{Learning rate:} 0.003
    \item \textbf{Batch size:} 1024
\end{itemize}
Thus, the inputs to the IRF are transformed through two internal feedforward layers (each with 32 artificial neurons) before being projected into the convolution weights $\G_n$.
This is a very small network by modern deep learning standards, which we take to be a strength, since it makes the model both parsimonious and fast to train and evaluate.
As we show below, the model is nonetheless capable of learning accurate and richly detailed IRFs, and increasing the size of the model provides little gain for the datasets analyzed here.
The network is penalized in proportion to the square of its weights of its internal layers, and the random offsets $\V$ are also subject to a shrinkage penalty to encourage valid population-level estimates (by discouraging the model from pushing population-level variation into the random effects).

We apply dropout \cite{srivastavaetal14} after the nonlinearity of each hidden layer.
Dropout is a form of regularization that randomly sets neurons of the layer to 0 during training with probability $p$ (per above, $p = 0.2$ in the reference model) and rescales the remaining units by $1 / (1 - p)$ to preserve the average magnitude.
Dropout approximates model averaging \cite{srivastavaetal14}, and it has become a widely-used regularization technique in deep learning.
In our case, the use of dropout critically underlies our ability to compute variational Bayesian estimates of uncertainty (see \textbf{Supplementary Information~\ref{sect:si-interpretation}}).

All models of human subjects data include by-subject random effects.
Although in principle each parameter of the neural IRF can have its own by-subject deviation, we found that it was simple, effective, and computationally efficient to restrict by-subject deviation to the bias terms of the internal IRF layers only, leaving all other neural network weights fixed across subjects.
The network is nonetheless able to capture substantial variation in IRF shape across subjects.\footnote{Following \cite{shainschuler19}, by-word random effects were not included to due their tendency to induce overfitting (poor generalization to unseen data) in the cognitive datasets evaluated in this study.}

Unless otherwise specified, models assume a univariate normal predictive distribution and are estimated with black box variational Bayes \cite{ranganathetal14}, using a variant of gradient descent (the Adam optimizer, \textit{\citen{kingmaba14}}) to maximize a regularized log likelihood.
We found that explicitly placing a variational prior on the neural network weights and biases led to poor performance (see also e.g., \textit{\citen{graves11,galghahramani16}}).
We therefore instead rely on dropout for uncertainty quantification in the deep neural component, since dropout has been shown to variationally approximate a Bayesian deep Gaussian process \cite{galghahramani16}.
Following \cite{shainschuler19}, we additionally place independent normal variational priors on the fixed effects components of coefficients $\b$ and predictive distribution parameter biases $\s_0$, with standard deviation equal to the standard deviation of the response in the training set.
Again following \cite{shainschuler19}, we also place independent normal variational priors on the random effects components of $\b$ and $\s_0$, with standard deviation equal to one tenth the standard deviation of the response in the training set, to encourage shrinkage.

Additional key implementation details common to all models are as follows:
\begin{itemize}
    \item Adam optimizer \cite{kingmaba14} with default Tensorflow parameters (aside from learning rate, which was directly investigated).
    
    \item RNN-internal activations follow LSTM defaults (tanh and sigmoid activations, see \textit{\citen{Hochreiter1997}}).
    
    \item All other nonlinearities are defined as a computationally efficient approximation to the Gaussian error linear unit (GELU, \textit{\citen{hendrycksgimpel16}}) used in current state of the art neural language models \cite{devlinetal19,radfordetal19}:
    \begin{align}
        \text{GELU}(v) & \defeq v\, \text{sigmoid}(1.702v)
    \end{align}

    \item Models use iterate averaging \cite{polyakjuditsky92} with exponential moving average decay rate 0.999 (updates after each minibatch).
    
    \item To aid training, all numeric predictors and responses are underlyingly rescaled by their training set standard deviations.
    These transforms are inverted for evaluation, visualization, and likelihood computation, allowing model estimates to be queried on the original scale.
    
    \item Stable training behavior is particularly important in this application, where convergence is diagnosed automatically based on the sequence of losses and users may not always visually inspect learning curves.
    To this end, these experiments use a number of safeguards against loss spikes, catastrophic forgetting, and numerical instability.
        \begin{itemize}
            \item The norm of the global gradient is clipped at 1.
            \item A constant $\epsilon$ = 1e-5 is added to bounded parameters (e.g., standard deviation).
            \item The outputs of the convolution $\s$ (eq.~\ref{eq:cdrnn_main}) are rescaled by $\frac{1}{TK}$ (i.e.\ divided by the number of timesteps times the number of features), since otherwise poor fit at initialization accumulates over both dimensions during convolution, leading to large early losses and training divergence.
            \item Large outlier losses are diagnosed based on de-biased exponential moving averages of the first and second moments of the loss by batch with decay rate 0.999.
            If the loss at any batch exceeds 1,000 moving standard deviations above the moving mean, training restarts from the last checkpoint.
        \end{itemize}
    
    \item 4 CPUs and 1 GPU per training run (specific hardware varied according to availability in our compute resource).
    
\end{itemize}

The hyperparameters enumerated above were selected based on exploratory analyses of synthetic and human data with respect to a combination of factors, including parsimony, training speed, exploratory set performance, and consistency of estimates/performance across replicates.
No models were evaluated on the test set of any dataset until all analyses were completed.

The same CDRNN implementation details used for eye-tracking and self-paced reading data work well on the noisier fMRI domain except that the random effects are underregularized, leading to poorer generalization error.
We find substantial improvement from increasing the regularizer strength on the random effects, and we therefore define CDRNN base to use a random effects regularization strength of 1000 rather than 10 in $\firf$ and a prior over the random effects components of $\snull$ and $\b$ of one hundredth rather than one tenth the standard deviation of the response in the training set.
The fact that different regularization levels are optimal for different datasets is an established pattern in deep learning research.
However, the degree of implementational overlap in CDRNN models across synthetic, reading, and fMRI data suggests that the base hyperparameters are a good starting point that may need little to no tuning when applied to novel domains.

The hyperparameters above define the ``base'' model in all analyses below.
To explore the influence of these hyperparameter choices, we additionally perform a limited grid search over models that deviate (up or down) from the base configuration in one of the following dimensions: number of hidden layers in the IRF, number of units per hidden layer of the IRF, L2 penalty strength on the IRF weights, L2 penalty strength on random IRF effects by participant, dropout level, learning rate, and batch size.
We show that results are relatively consistent across a range of parameterizations, suggesting that models are not deeply sensitive to particular choices for these values.

\section{Effect Estimation}
\label{sect:si-interpretation}

To query the model's estimates, an input configuration $\mathcal{C} = \left<\x, t_{\x}, d_{\x}\right>$ is constructed and fed to the model, which generates an expected response.
By systematically manipulating $\mathcal{C}$ and repeating this procedure, a wide range of estimates can be extracted \textit{post hoc} from the fitted model.
For example, to compute the model's estimated baseline IRF (the ``deconvolutional intercept'', also called \textit{rate} by \textit{\citen{shainschuler19}}), a reference stimulus $\mathcal{C}_{\text{ref}}$ can be constructed and fed to the model $f$ to yield an estimated response $f(\mathcal{C}_{\text{ref}})$.
Stimulus-driven deviation from this reference (e.g., the effect of increasing a predictor by $c$) can be measured by constructing an alternative configuration $\mathcal{C}_{\text{alt}}$ and computing the difference $f(\mathcal{C}_{\text{alt}}) - f(\mathcal{C}_{\text{ref}})$.
Note that $f$ can return any statistic of the predictive distribution $\F(\s)$, including each element of parameter vector $\s$, which the IRF directly models, but also e.g., moments or quantiles.
This design therefore provides a highly general procedure for effect estimation.
It can be used to run any model query that can be instantiated as a set of input configurations paired with a response statistic of interest, regardless of the internal structure of the model.

Given the nonlinear, interactive, and (in some cases) context-dependent nature of CDRNN-estimated IRFs, the choice of $\mathcal{C}_{\text{ref}}$ plays an important role in effect estimation: different effects might be obtained relative to different reference inputs.
We consider the mean predictor vector (in the training set) to be an appropriate default choice for $\mathcal{C}_{\text{ref}}$, since it constitutes an unbiased estimate of the expected values of the predictors, and we therefore use this approach in all analyses presented here.
However, other research questions might motivate a different choice for $\mathcal{C}_{\text{ref}}$.
Note that, as in linear models, estimates for nonsensical or unattested predictor values should be treated with caution: the model continuously interpolates/extrapolates as a result of its design, but it is the analyst's responsibility to avoid over-interpreting effects derived from inappropriate regions of predictor space (e.g., estimates for non-integer values of a boolean indicator variable, or estimates outside the range of values attested in the training data).
For these reasons, $\mathbf{0}$ is not an appropriate choice for $\mathcal{C}_{\text{ref}}$ in the general case, since depending on the modeling problem it may never be attested in training.

Approximate uncertainty intervals around any effect estimate can be obtained by resampling the model many times from its variational posterior and performing perturbation analysis on each resampled model.
This is straightforward for parameters with parametric variational posteriors (e.g., $\snull$), but it can still be done for the parameters of $\fin$ and $\firf$ as long as the model is fitted using dropout, thanks to \cite{galghahramani16}, who showed that resampling models by resampling their dropout masks provides a variational approximation to the posterior of a deep Gaussian process \cite{damianoulawrence13}.
We therefore resample $\firf$ by continuing to apply dropout at evaluation time for \textit{post hoc} effect estimation/visualization, which provides a range of possible responses to $\mathcal{C}$ under different dropout masks.\footnote{Note that some queries involve responses at many different values of $\mathcal{C}$, e.g., estimating the IRF shape by evaluating it at many values of $d_{\x}$.
In these cases, the entire query should be run over each resampled model (rather than resampling the model for each point in the query, which will mix multiple models in computing a single sample).
Our software implementation enforces this behavior by resampling the model once before each full query.
}
Because the intervals thus computed derive from a variational approximating distribution rather than the true posterior, they should not be used for scientific hypothesis testing unless no alternative is feasible.
\section{Significance Testing through Model Comparison}
\label{sect:si-nhst}

\subsection{General Approach}

All statistical comparisons use paired permutation tests of out-of-sample conditional likelihood.
The logic of the test is as follows: if null hypothesis $H_0$ is true, then a model instantiating alternative hypothesis $H_1$ should predict unseen data no better than a model instantiating $H_0$.
The difference between $H_1$ and $H_0$ is cached out as a \textit{constraint} that is enforced in $H_0$ but not in $H_1$, and the test asks whether relaxing this constraint leads to significantly improved fit, and thus, a basis for rejecting the null.
The advantages of this approach are that (1) it requires no assumptions about the sampling distribution of effects in CDRNNs, (2) it can be applied to any null hypothesis that can be instantiated as a model constraint, and (3) it is based directly on generalization performance, thereby potentially improving replicability of results.

As in CDR \cite{shainschuler19}, in-sample tests of CDRNN models are discouraged: intervals-based tests are not necessarily valid because the intervals are variationally approximated, and in-sample tests like the likelihood ratio test are not necessarily valid because likelihood in a non-convex model like CDRNN cannot be guaranteed to be maximized.
Even if it could, the model is so expressive that in-sample likelihood gains can simply be due to overfitting.

However, as discussed in \textbf{Supplementary Information~\ref{app:cognitive}}, variability in CDRNNs' out-of-sample generalization performance across random seeds also presents a problem for null hypothesis significance testing via pairwise model comparison: performance differences can be driven by noise in the optimization process, in addition to any differences due to the constraint that distinguishes $H_0$ from $H_1$.
This problem can be mitigated by comparing \textit{ensembles} of models, where an ensemble is a set of $E$ replicates of a given model design.
To perform such a comparison, we generalize the \cite{shainschuler19} paired permutation test to ensembles of models.
Note that, given $N$ evaluation items and corresponding sets of log-likelihoods $\mathcal{E}_0$ and $\mathcal{E}_1$ respectively derived from the $E$ component models of the ensembles representing $H_0$ and $H_1$, the $2E$ likelihoods assigned to each datum in the evaluation set are exchangeable under the null hypothesis of no performance difference between models.
We therefore compute a bootstrap sample in a way that respects these exchangeability criteria (see also e.g., \textit{\citen{winkleretal14,winkleretal15}}) by randomly resampling ensembles by item, using the following procedure: 
\begin{enumerate}
    \item For each of the $N$ evaluation items $1 \leq n \leq N$, repartition the $2E$ log-likelihood statistics into two random sets of likelihoods $\hat{\mathcal{E}}_{1,n}$, $\hat{\mathcal{E}}_{2,n}$, each with $E$ elements.
    \item Compute the resampled dataset likelihood as the sum of averages within the resampled partition:
    
    $\hat{\mathcal{L}_i} = \sum_{n=1}^N \frac{1}{E} \sum_{e=1}^E \hat{\mathcal{E}}_{i,n,e}$.
    \item Compute and store the absolute difference $|\hat{\mathcal{L}_1} - \hat{\mathcal{L}_2}|$.
\end{enumerate}
This process is repeated many times  to construct an empirical null distribution over the likelihood differences between ensembles, which is then compared to the observed difference in mean likelihood between ensembles in order to compute a $p$ value.
The number of resampling iterations determines the minimum obtainable $p$ value.
Here, we use 10,000 resampling iterations, and thus the minimum detectable $p$ value is $0.0001$.
In cases where none of the resampled likelihood differences exceed the empirical one, this value serves as an upper bound on $p$.

Estimates from an entire ensemble are visualized by first sampling uniformly over the $E$ fitted ensemble components, then sampling a model from that component's variational posterior, then querying the sampled model.
Repeated many times, this procedure defines an empirical distribution over model estimates that takes into account uncertainty across the ensemble.
All plots derived from ensembles were computed in this way.
Unless otherwise stated, plots show the effect on the mean of the predictive distribution.
All ensembles in this study used $E = 10$.
Comparisons of CDRNNs to baseline models do not involve ensembling, since most baselines are deterministically optimized.
This is equivalent to using the procedure above with $E = 1$.

\subsection{Null Models Used in this Study}

Here we present the implementation details for the null models used to test effects in Examples A--E.

\subsubsection{Example A: The Existence of Effects}

To test for the presence of overall effects of a predictor, it is necessary to construct a null model in which the predictor is not present.
To do so, we simply remove the predictor from the model altogether, holding everything else constant.

\subsubsection{Example B: Linearity of Effects}

To test nonlinearity of effects, it is necessary to construct a null model model in which the critical effect is constrained to be linear.
To do so, we remove the predictor of interest from the inputs to the deep neural IRF (but retain it in the outputs), thereby enforcing a linear response (since the IRF cannot condition on the value of the predictor and therefore can only define a linear coefficient).

\subsubsection{Example C: Effect Interactions}

To test an interaction of predictors A and B, it is necessary to construct a null model in which both A and B are present but cannot interact.
To do so, we split the IRF into two distinct neural networks, one containing A along with all other predictors but B, and another containing B along with all other predictors but A.
To ensure a minimal comparison, we reimplement the full model using this two-network design without ablating any interactions.
In this way, the interaction between A and B is removed while retaining all remaining interactions.

\subsubsection{Example D: Distributional Regression}

To test an effect of a predictor on a specific distributional parameter, it is necessary to construct a null model in which only that effect is removed.
To do so, we split the IRF into two distinct neural networks, one that generates $\mu$, and another that generates $\sigma$.
Doing so permits selective removal of effects in one but not the other distributional parameter.
To ensure a minimal comparison, we reimplement the full model using this two-network design without ablating any effects on any distributional parameters.

\subsubsection{Example E: Nonstationarity}

To test nonstationarity of an effect, it is necessary to construct a null model in which the timestamp (which licenses nonstationarity) is removed from the input for the IRF to that effect only, retaining it for all others.
To do so, we split the IRF into two distinct neural networks, one that retains the timestamp in the input and convolves all predictors but the critical one, and one that omits the timestamp but convolves only the critical predictor, while additionally including all remaining predictors as network \textit{inputs} to ensure retention of interactions.
This guarantees that the response to the critical predictor is stationary.
To ensure a minimal comparison, we reimplement the full model using this two-network design without ablating the timestamp in either network.
\section{Relationship to Shain (2021)}

We proposed the central idea of this work---using a deep neural IRF kernel for deconvolutional regression---in a previous conference paper \cite{shain21}.
However, the current study goes substantially beyond the previous one in several ways.
First, unlike this work, \cite{shain21} did not distinguish between the inputs and outputs to the deep neural IRF: every predictor was used as an input to the IRF and convolved using an output of the IRF.
This distinction critically underlies our current CDRNN definition's support for flexible hypothesis testing, since the IRF can now condition on predictors that it does not convolve or convolve predictors that it does not condition on.
For example, to test linearity of effect A, a null model can now be built in which the IRF convolves but does not condition on A, thus enforcing linearity.
The lack of this ability in \cite{shain21} limited the scope of testable hypotheses relative to the current definition.

Second, we have improved the model design by factoring input processing from IRF computation.
This permits dissociation of the IRF shape (over time) from the functional form of effects, allowing researchers to target these components of the response separately if desired.
Such a factorization was not possible under the \cite{shain21} design.

Third, we have broadened the scope of mixed effects modeling to encompass all trainable model parameters.
By contrast, in \cite{shain21}, mixed effects were arbitrarily restricted to specific model components.

Fourth, we have expanded the feature set of the software implementation.
For example, it is now possible to use a composite IRF consisting of multiple distinct deep neural transforms, each with their own parameterization.
This feature is critical for testing hypotheses about whether a particular effect is e.g., stationary (by using two neural IRFs, a stationary one that convolves the critical predictor and a nonstationary one that convolves the other predictors) or specific to a certain distributional parameter (by using different neural IRFs for each distributional parameter).
See \textbf{Supplementary Information~\ref{sect:si-nhst}} for additional examples.

Fifth, our mathematical definition (\textbf{Introduction: The CDRNN Model} of the main article) has been simplified and clarified.
It also now generalizes a range of related regression methods for time series, including linear models, generalized additive models, and kernel-based continuous-time deconvolutional regression models.

Finally, our empirical evaluation goes well beyond the cursory evaluation in \cite{shain21}, which only evaluated a handful of model designs, used a small subset of the full synthetic dataset, and did no hypothesis testing.
By contrast, our study provides extensive evaluations on every dataset covered by \cite{shainschuler19}, permitting detailed comparison of the effects of different hyperparameterizations.
In addition, we propose methods for hypothesis testing and demonstrate these methods for a diverse set of questions.

\section{Full Results: Synthetic Experiments}
\label{app:si-synth}

\begin{figure}

    \footnotesize
    \sffamily
    \centering
    
    \textbf{\Large Synth: Noise, $\sigma_{\epsilon}=0$}
    
    \vspace{1em}
    
    \begin{subfigure}[t]{0.49\textwidth}
        \centering
        \makebox[0.49\textwidth]{\centering \textbf{True}}
        
        \includegraphics[width=0.49\textwidth]{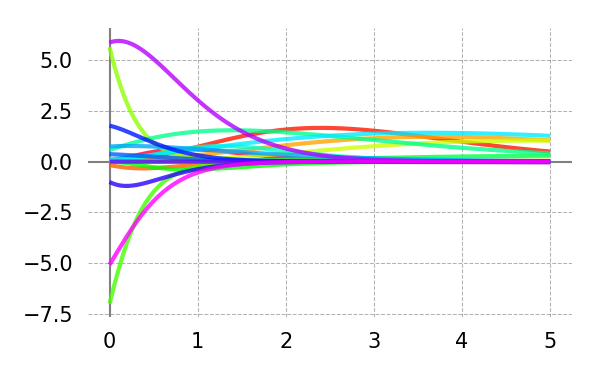}
    \end{subfigure}
    
    \begin{subfigure}[t]{0.49\textwidth}
        \centering
        \makebox[0.49\textwidth]{\centering Base}%
        \makebox[0.49\textwidth]{\centering + RNN}
        \begin{overpic}[width=0.49\textwidth]{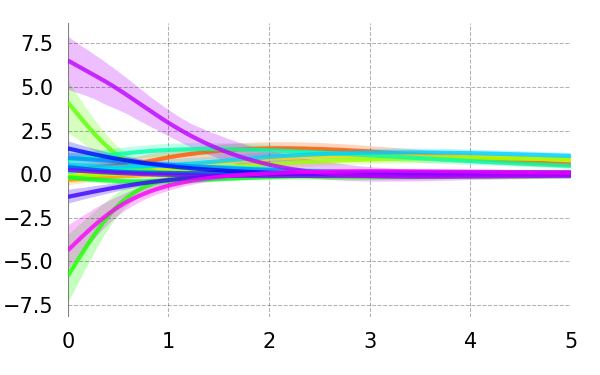}
        \end{overpic}%
        \includegraphics[width=0.49\textwidth]{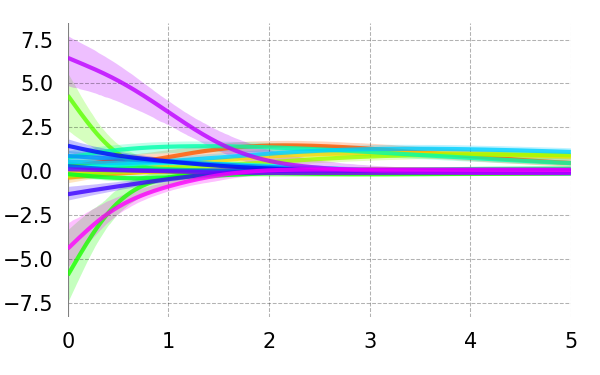}
    \end{subfigure}
    \begin{subfigure}[t]{0.49\textwidth}
        \centering
        \makebox[0.49\textwidth]{\centering Hidden Units $\div$ 2 (16)}%
        \makebox[0.49\textwidth]{\centering Hidden Units $\times$ 2 (64)}
        \includegraphics[width=0.49\textwidth]{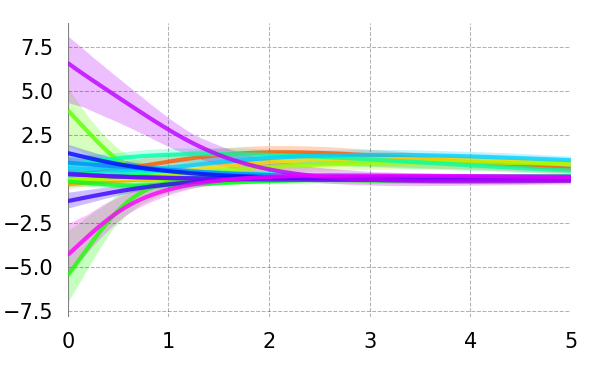}
        \includegraphics[width=0.49\textwidth]{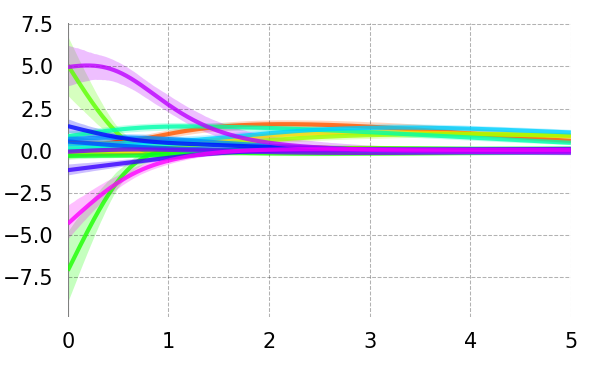}
    \end{subfigure}
    
    \begin{subfigure}[t]{0.49\textwidth}
        \centering
        \makebox[0.49\textwidth]{\centering Hidden Layers - 1 (1)}%
        \makebox[0.49\textwidth]{\centering Hidden Layers + 1 (3)}
        \includegraphics[width=0.49\textwidth]{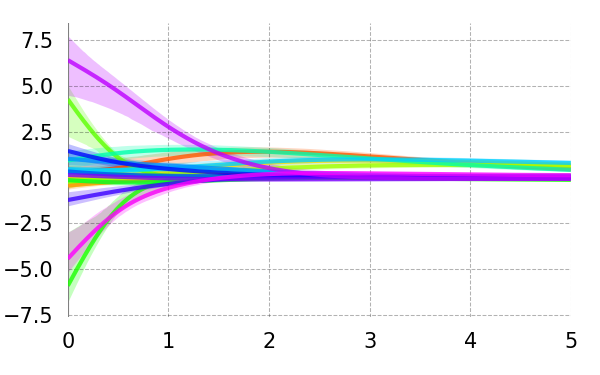}
        \includegraphics[width=0.49\textwidth]{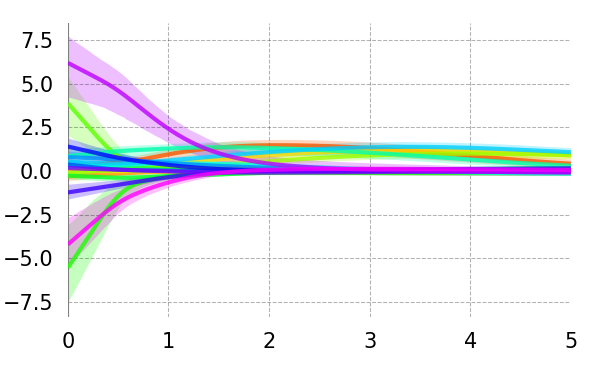}
    \end{subfigure}
    \begin{subfigure}[t]{0.49\textwidth}
        \centering
        \makebox[0.49\textwidth]{\centering Weight Reg $\div$ 5 (1)}%
        \makebox[0.49\textwidth]{\centering Weight Reg $\times$ 5 (5)}
        \includegraphics[width=0.49\textwidth]{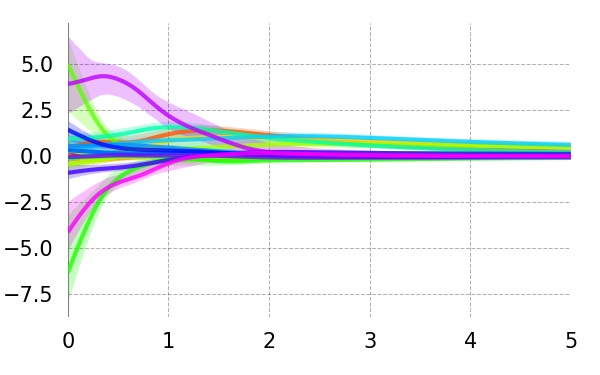}
        \includegraphics[width=0.49\textwidth]{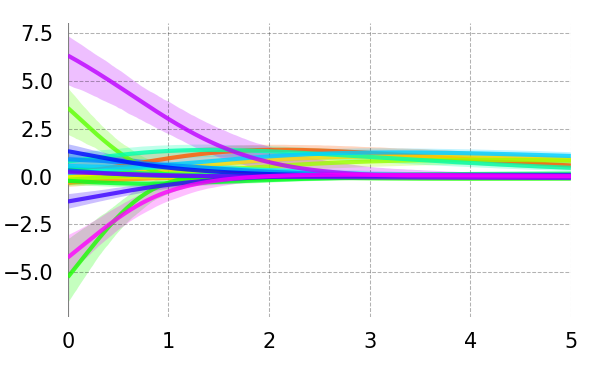}
    \end{subfigure}
    
    \begin{subfigure}[t]{0.49\textwidth}
        \centering
        \makebox[0.49\textwidth]{\centering Dropout $\div$ 2 (0.05)}%
        \makebox[0.49\textwidth]{\centering Dropout $\times$ 2 (0.2)}
        \includegraphics[width=0.49\textwidth]{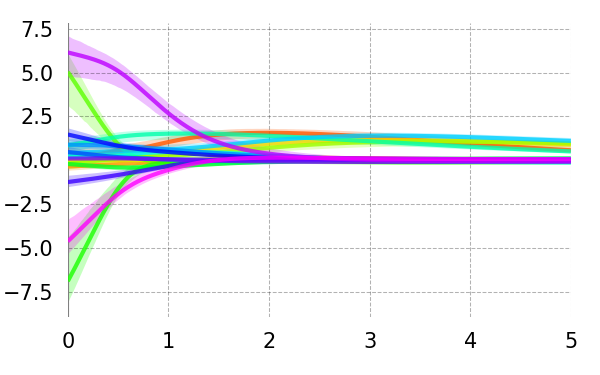}
        \includegraphics[width=0.49\textwidth]{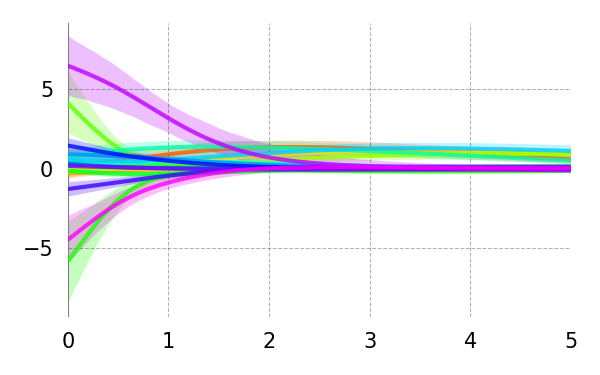}
    \end{subfigure}
    \begin{subfigure}[t]{0.49\textwidth}
        \centering
        \makebox[0.49\textwidth]{\centering Learning Rate $\div$ 3 (0.001)}%
        \makebox[0.49\textwidth]{\centering Learning Rate $\times$ 3 (0.009)}
        \includegraphics[width=0.49\textwidth]{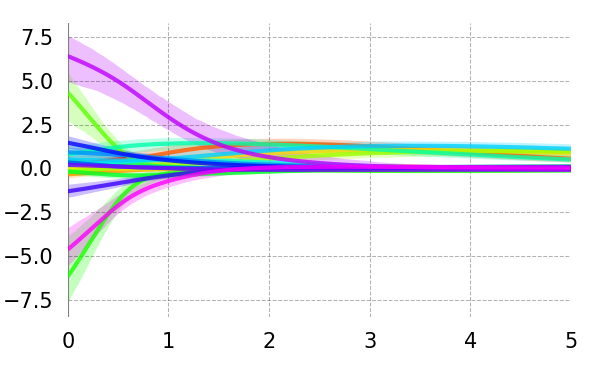}
        \includegraphics[width=0.49\textwidth]{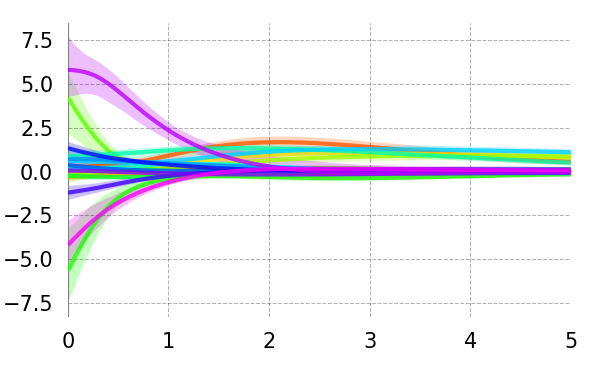}
    \end{subfigure}
    \begin{subfigure}[t]{0.49\textwidth}
        \centering
        \makebox[0.49\textwidth]{\centering Batch Size $\div$ 2 (512)}%
        \makebox[0.49\textwidth]{\centering Batch Size $\times$ 2 (2048)}
        \includegraphics[width=0.49\textwidth]{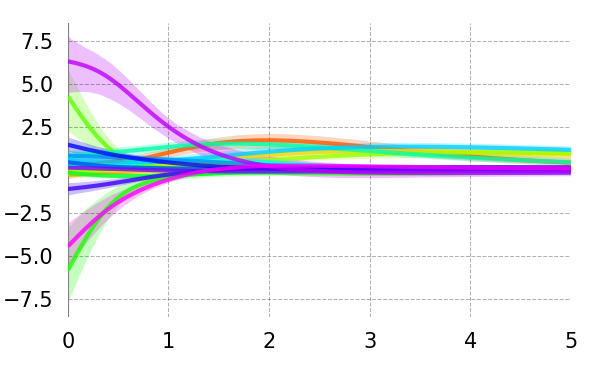}
        \includegraphics[width=0.49\textwidth]{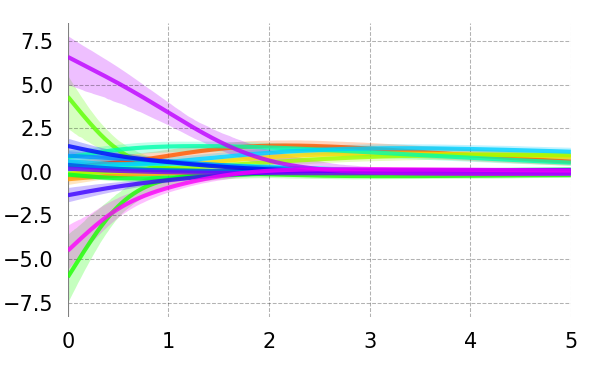}
    \end{subfigure}
    
    \vspace{1em}    
    
    \textbf{\large Consistency}
    
    \vspace{0.5em}
    
    \begin{subfigure}[t]{0.19\textwidth}
        \centering
        \makebox[0.9\textwidth]{\centering Rep 1}
        \includegraphics[width=\textwidth]{{results_cdrnn_journal_synth_noise_e0_CDR_main_irf_univariate_y_mc}.png}
    \end{subfigure}
    \begin{subfigure}[t]{0.19\textwidth}
        \centering
        \makebox[0.9\textwidth]{\centering Rep 2}
        \includegraphics[width=\textwidth]{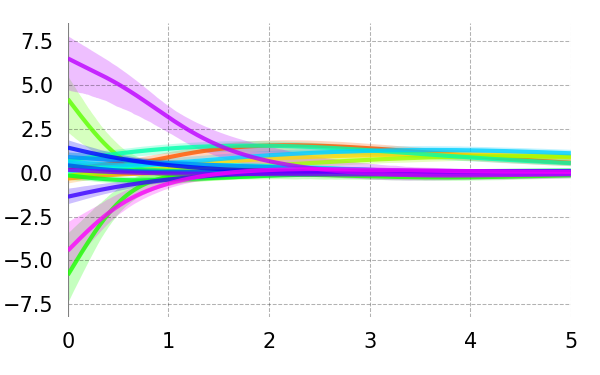}
    \end{subfigure}
    \begin{subfigure}[t]{0.19\textwidth}
        \centering
        \makebox[0.9\textwidth]{\centering Rep 3}
        \includegraphics[width=\textwidth]{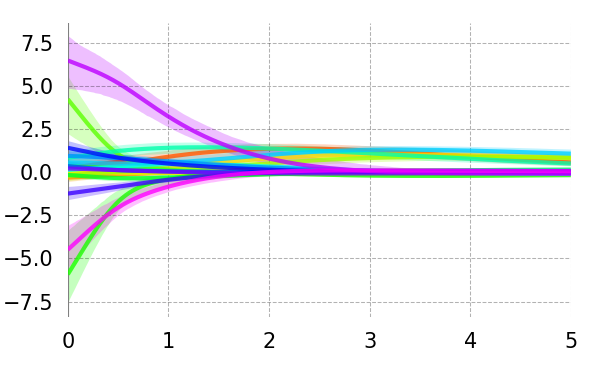}
    \end{subfigure}
    \begin{subfigure}[t]{0.19\textwidth}
        \centering
        \makebox[0.9\textwidth]{\centering Rep 4}
        \includegraphics[width=\textwidth]{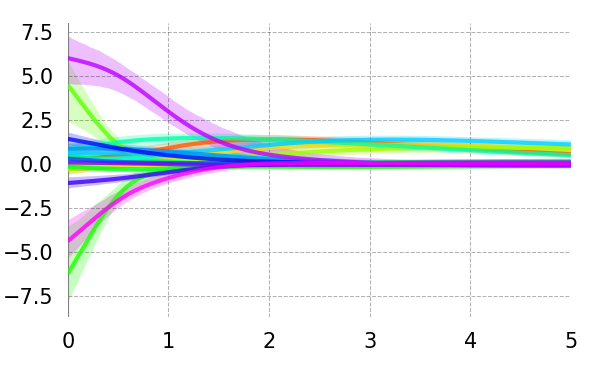}
    \end{subfigure}
    \begin{subfigure}[t]{0.19\textwidth}
        \centering
        \makebox[0.9\textwidth]{\centering Rep 5}
        \includegraphics[width=\textwidth]{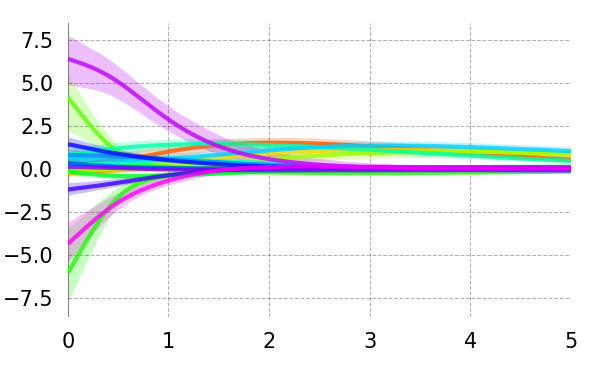}
    \end{subfigure}
    
    {\large Delay (s)}
    
    \vspace{0.5em}
    
    \caption{CDRNN estimated responses to synthetic data with \textbf{noise standard deviation $\sigma_{\epsilon} = 0$} (i.e. noise free). Estimates using base hyperparameters are compared to estimates from models that deviate from the base in some dimension. Plots under ``Consistency'' show estimates from five replicates of the ``base'' configuration, where ``Rep 1'' is the same model as ``base'' above, replotted for ease of comparison.}
    \label{fig:app-synth-noise-e0}
    
\end{figure}

\begin{figure}

    \footnotesize
    \sffamily
    \centering
    
    \textbf{\Large Synth: Noise, $\sigma_{\epsilon}=1$}
    
    \vspace{1em}
    
    \begin{subfigure}[t]{0.49\textwidth}
        \centering
        \makebox[0.49\textwidth]{\centering \textbf{True}}
        
        \includegraphics[width=0.49\textwidth]{{results_cl_synth_noise_e0_synthetic_true}.png}
    \end{subfigure}
    
    \begin{subfigure}[t]{0.49\textwidth}
        \centering
        \makebox[0.49\textwidth]{\centering Base}%
        \makebox[0.49\textwidth]{\centering + RNN}
        \begin{overpic}[width=0.49\textwidth]{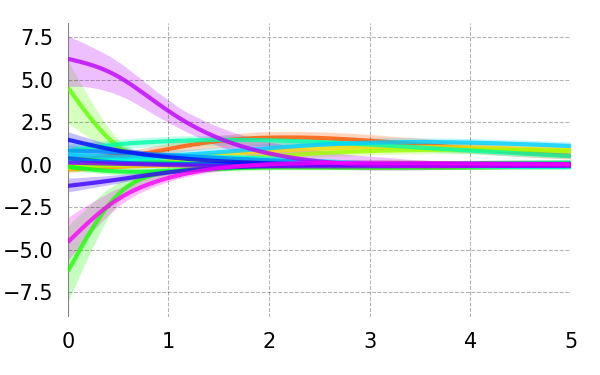}
        \end{overpic}%
        \includegraphics[width=0.49\textwidth]{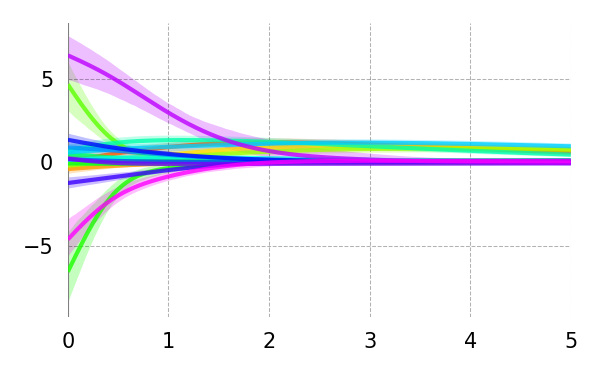}
    \end{subfigure}
    \begin{subfigure}[t]{0.49\textwidth}
        \centering
        \makebox[0.49\textwidth]{\centering Hidden Units $\div$ 2 (16)}%
        \makebox[0.49\textwidth]{\centering Hidden Units $\times$ 2 (64)}
        \includegraphics[width=0.49\textwidth]{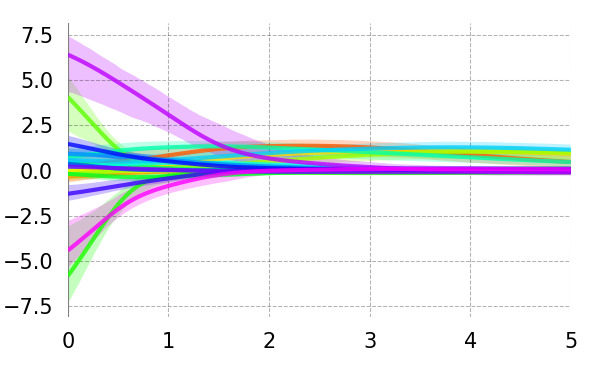}
        \includegraphics[width=0.49\textwidth]{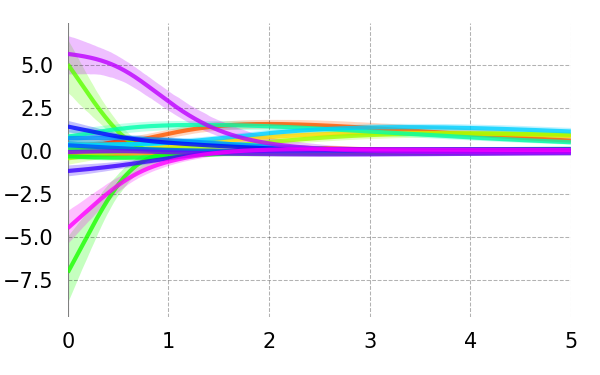}
    \end{subfigure}
    
    \begin{subfigure}[t]{0.49\textwidth}
        \centering
        \makebox[0.49\textwidth]{\centering Hidden Layers - 1 (1)}%
        \makebox[0.49\textwidth]{\centering Hidden Layers + 1 (3)}
        \includegraphics[width=0.49\textwidth]{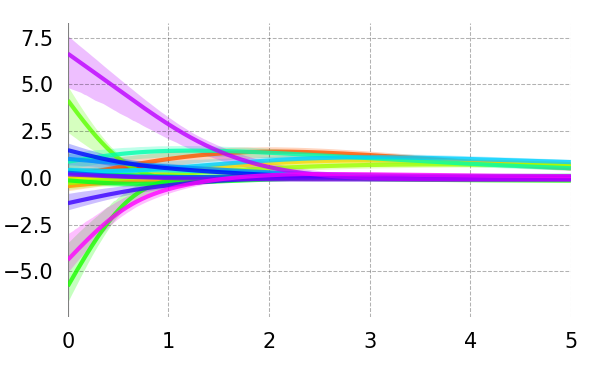}
        \includegraphics[width=0.49\textwidth]{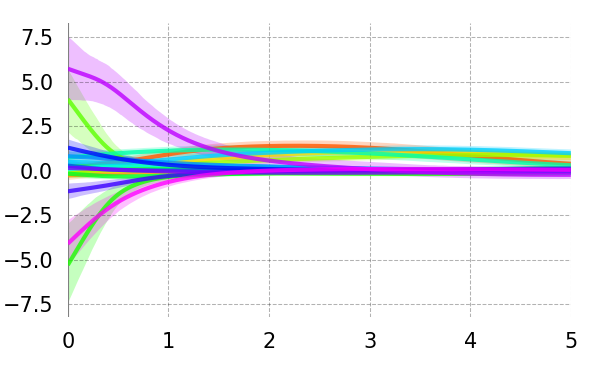}
    \end{subfigure}
    \begin{subfigure}[t]{0.49\textwidth}
        \centering
        \makebox[0.49\textwidth]{\centering Weight Reg $\div$ 5 (1)}%
        \makebox[0.49\textwidth]{\centering Weight Reg $\times$ 5 (5)}
        \includegraphics[width=0.49\textwidth]{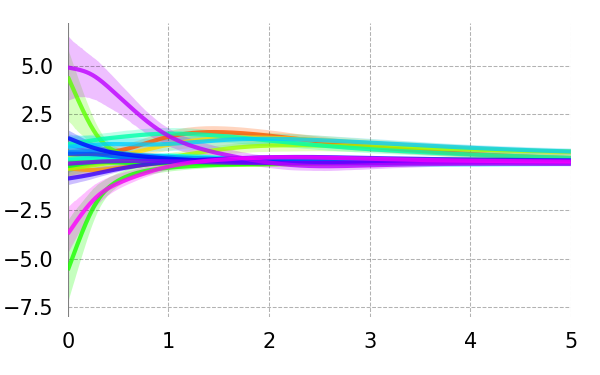}
        \includegraphics[width=0.49\textwidth]{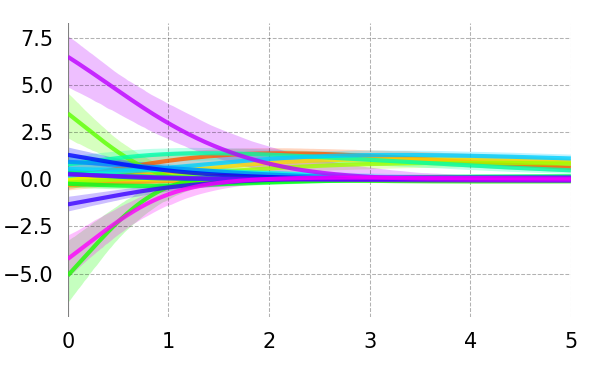}
    \end{subfigure}
    
    \begin{subfigure}[t]{0.49\textwidth}
        \centering
        \makebox[0.49\textwidth]{\centering Dropout $\div$ 2 (0.05)}%
        \makebox[0.49\textwidth]{\centering Dropout $\times$ 2 (0.2)}
        \includegraphics[width=0.49\textwidth]{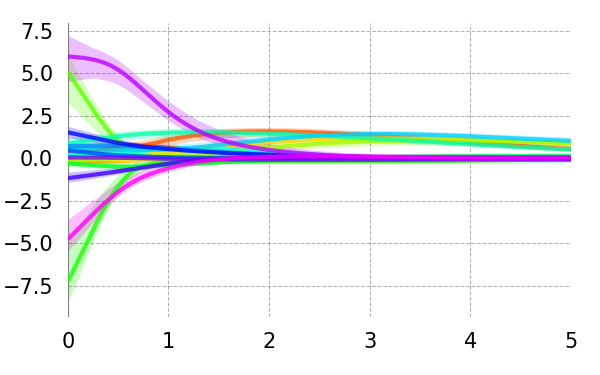}
        \includegraphics[width=0.49\textwidth]{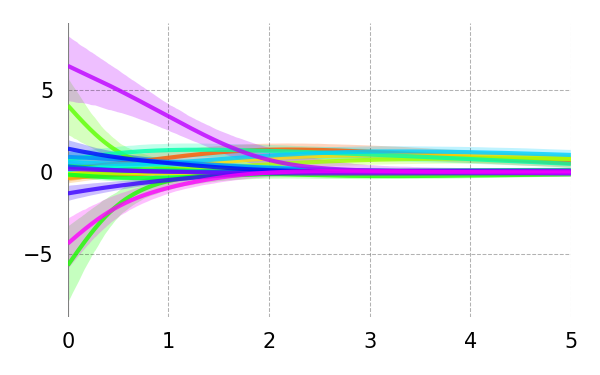}
    \end{subfigure}
    \begin{subfigure}[t]{0.49\textwidth}
        \centering
        \makebox[0.49\textwidth]{\centering Learning Rate $\div$ 3 (0.001)}%
        \makebox[0.49\textwidth]{\centering Learning Rate $\times$ 3 (0.009)}
        \includegraphics[width=0.49\textwidth]{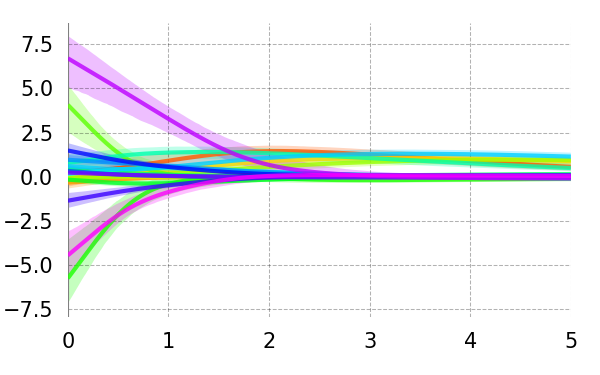}
        \includegraphics[width=0.49\textwidth]{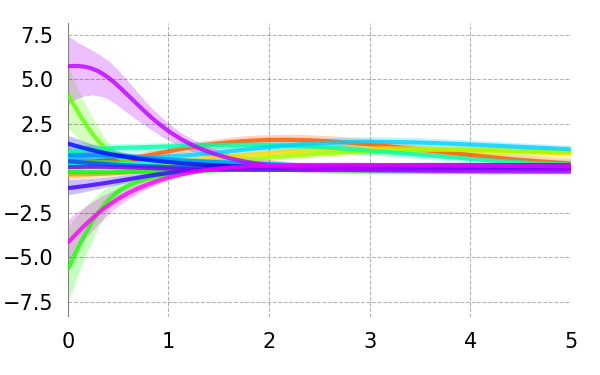}
    \end{subfigure}
    \begin{subfigure}[t]{0.49\textwidth}
        \centering
        \makebox[0.49\textwidth]{\centering Batch Size $\div$ 2 (512)}%
        \makebox[0.49\textwidth]{\centering Batch Size $\times$ 2 (2048)}
        \includegraphics[width=0.49\textwidth]{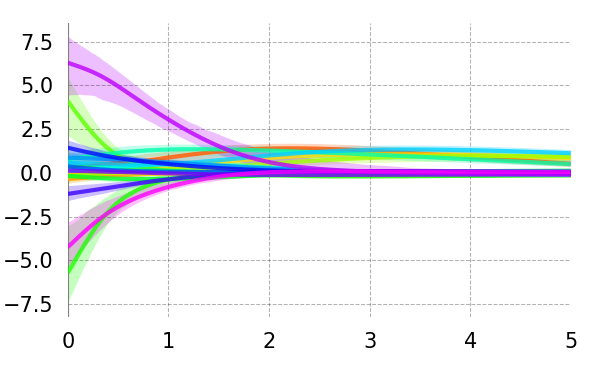}
        \includegraphics[width=0.49\textwidth]{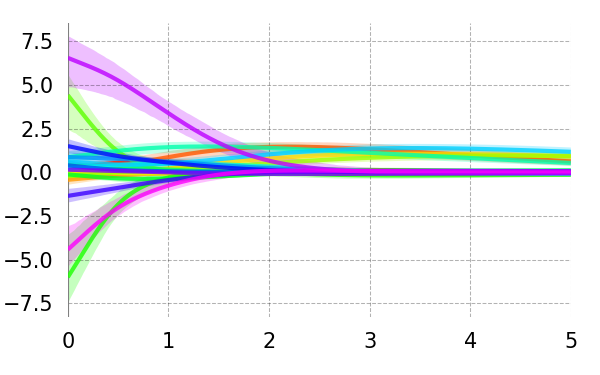}
    \end{subfigure}
    
    \vspace{1em}    
    
    \textbf{\large Consistency}
    
    \vspace{0.5em}
    
    \begin{subfigure}[t]{0.19\textwidth}
        \centering
        \makebox[0.9\textwidth]{\centering Rep 1}
        \includegraphics[width=\textwidth]{{results_cdrnn_journal_synth_noise_e1_CDR_main_irf_univariate_y_mc}.png}
    \end{subfigure}
    \begin{subfigure}[t]{0.19\textwidth}
        \centering
        \makebox[0.9\textwidth]{\centering Rep 2}
        \includegraphics[width=\textwidth]{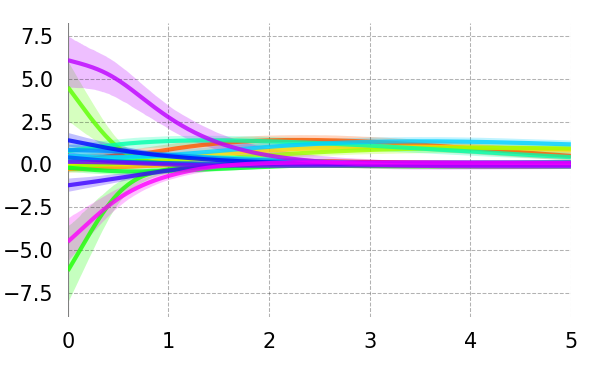}
    \end{subfigure}
    \begin{subfigure}[t]{0.19\textwidth}
        \centering
        \makebox[0.9\textwidth]{\centering Rep 3}
        \includegraphics[width=\textwidth]{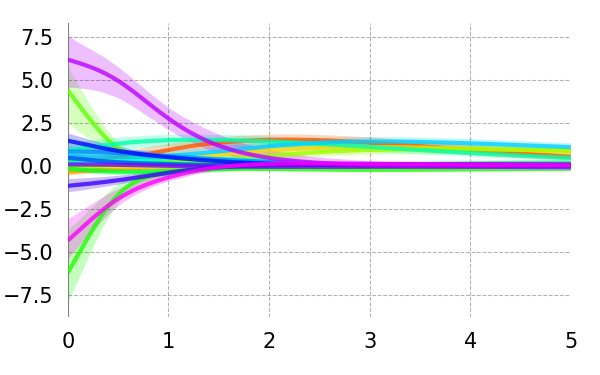}
    \end{subfigure}
    \begin{subfigure}[t]{0.19\textwidth}
        \centering
        \makebox[0.9\textwidth]{\centering Rep 4}
        \includegraphics[width=\textwidth]{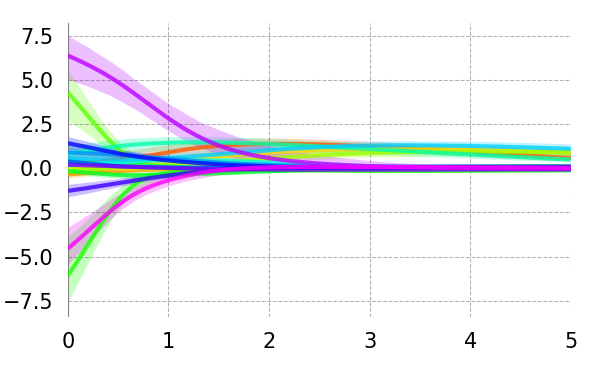}
    \end{subfigure}
    \begin{subfigure}[t]{0.19\textwidth}
        \centering
        \makebox[0.9\textwidth]{\centering Rep 5}
        \includegraphics[width=\textwidth]{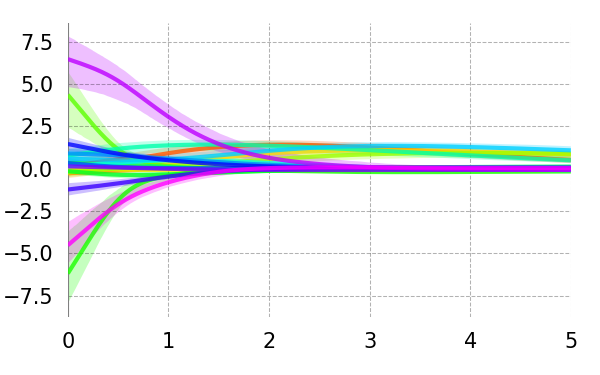}
    \end{subfigure}
    
    {\large Delay (s)}
    
    \vspace{0.5em}
    
    \caption{CDRNN estimated responses to synthetic data with \textbf{noise standard deviation $\sigma_{\epsilon} = 1$}. Estimates using base hyperparameters are compared to estimates from models that deviate from the base in some dimension. Plots under ``Consistency'' show estimates from five replicates of the ``base'' configuration, where ``Rep 1'' is the same model as ``base'' above, replotted for ease of comparison.}
    \label{fig:app-synth-noise-e1}
    
\end{figure}

\begin{figure}

    \footnotesize
    \sffamily
    \centering
    
    \textbf{\Large Synth: Noise, $\sigma_{\epsilon}=10$}
    
    \vspace{1em}
    
    \begin{subfigure}[t]{0.49\textwidth}
        \centering
        \makebox[0.49\textwidth]{\centering \textbf{True}}
        
        \includegraphics[width=0.49\textwidth]{{results_cl_synth_noise_e0_synthetic_true}.png}
    \end{subfigure}
    
    \begin{subfigure}[t]{0.49\textwidth}
        \centering
        \makebox[0.49\textwidth]{\centering Base}%
        \makebox[0.49\textwidth]{\centering + RNN}
        \begin{overpic}[width=0.49\textwidth]{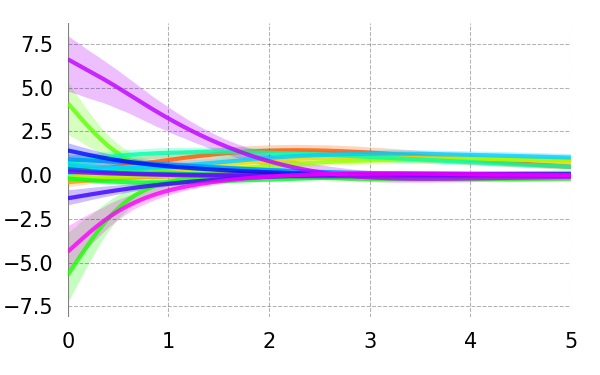}
        \end{overpic}%
        \includegraphics[width=0.49\textwidth]{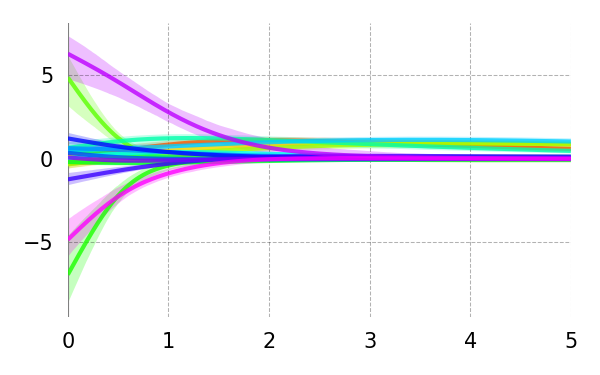}
    \end{subfigure}
    \begin{subfigure}[t]{0.49\textwidth}
        \centering
        \makebox[0.49\textwidth]{\centering Hidden Units $\div$ 2 (16)}%
        \makebox[0.49\textwidth]{\centering Hidden Units $\times$ 2 (64)}
        \includegraphics[width=0.49\textwidth]{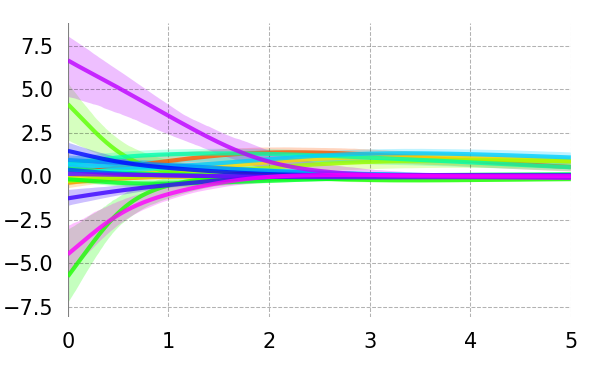}
        \includegraphics[width=0.49\textwidth]{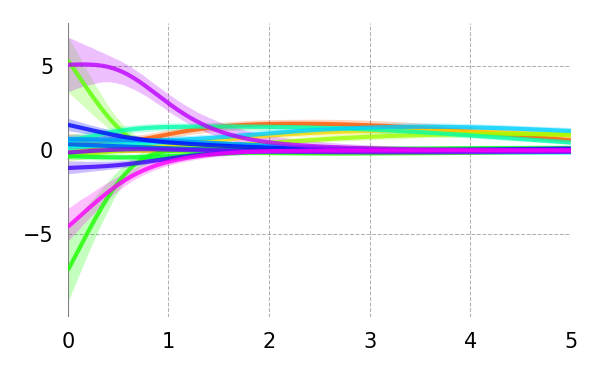}
    \end{subfigure}
    
    \begin{subfigure}[t]{0.49\textwidth}
        \centering
        \makebox[0.49\textwidth]{\centering Hidden Layers - 1 (1)}%
        \makebox[0.49\textwidth]{\centering Hidden Layers + 1 (3)}
        \includegraphics[width=0.49\textwidth]{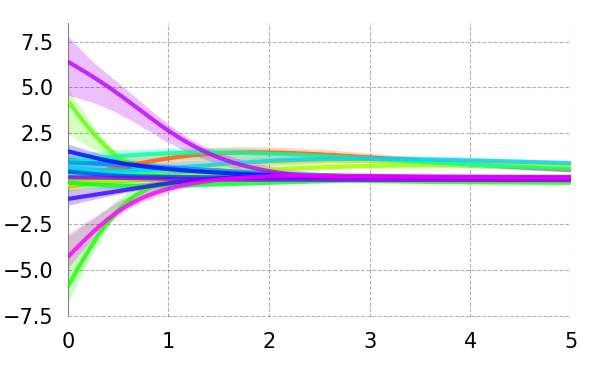}
        \includegraphics[width=0.49\textwidth]{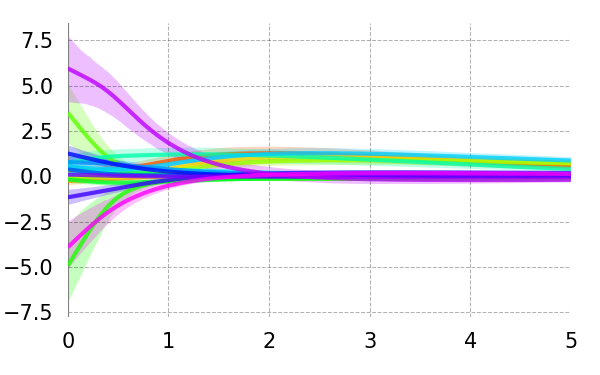}
    \end{subfigure}
    \begin{subfigure}[t]{0.49\textwidth}
        \centering
        \makebox[0.49\textwidth]{\centering Weight Reg $\div$ 5 (1)}%
        \makebox[0.49\textwidth]{\centering Weight Reg $\times$ 5 (5)}
        \includegraphics[width=0.49\textwidth]{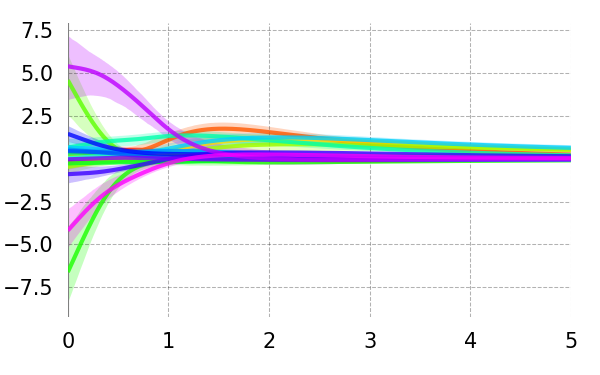}
        \includegraphics[width=0.49\textwidth]{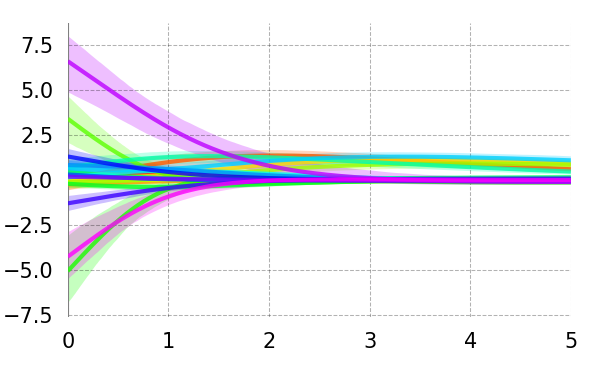}
    \end{subfigure}
    
    \begin{subfigure}[t]{0.49\textwidth}
        \centering
        \makebox[0.49\textwidth]{\centering Dropout $\div$ 2 (0.05)}%
        \makebox[0.49\textwidth]{\centering Dropout $\times$ 2 (0.2)}
        \includegraphics[width=0.49\textwidth]{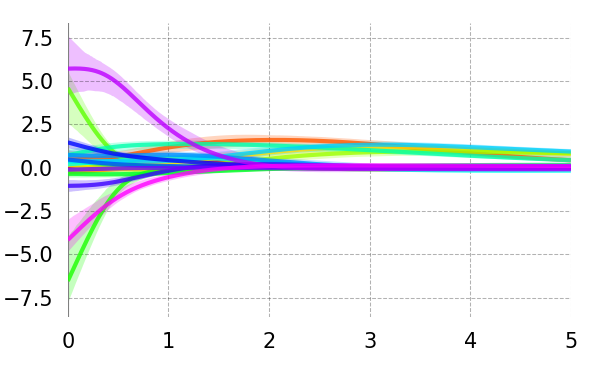}
        \includegraphics[width=0.49\textwidth]{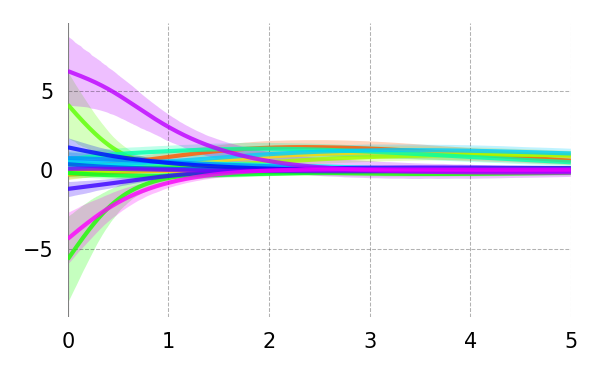}
    \end{subfigure}
    \begin{subfigure}[t]{0.49\textwidth}
        \centering
        \makebox[0.49\textwidth]{\centering Learning Rate $\div$ 3 (0.001)}%
        \makebox[0.49\textwidth]{\centering Learning Rate $\times$ 3 (0.009)}
        \includegraphics[width=0.49\textwidth]{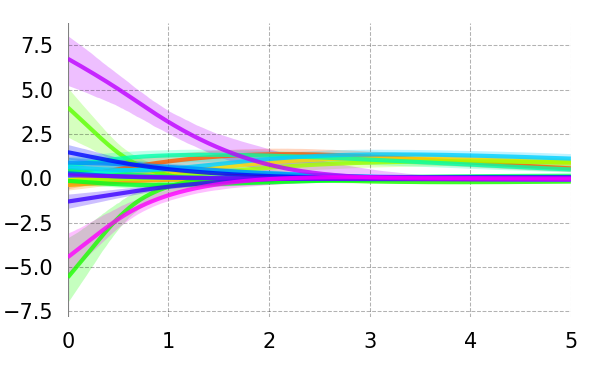}
        \includegraphics[width=0.49\textwidth]{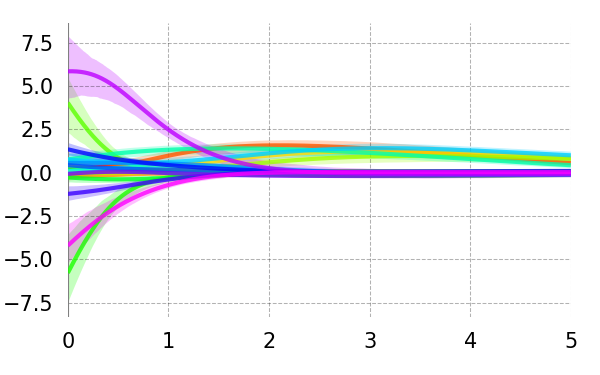}
    \end{subfigure}
    \begin{subfigure}[t]{0.49\textwidth}
        \centering
        \makebox[0.49\textwidth]{\centering Batch Size $\div$ 2 (512)}%
        \makebox[0.49\textwidth]{\centering Batch Size $\times$ 2 (2048)}
        \includegraphics[width=0.49\textwidth]{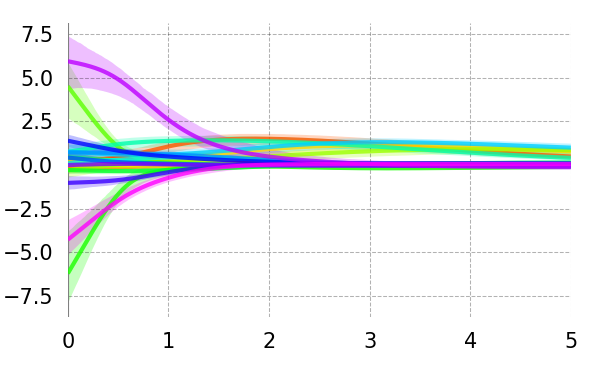}
        \includegraphics[width=0.49\textwidth]{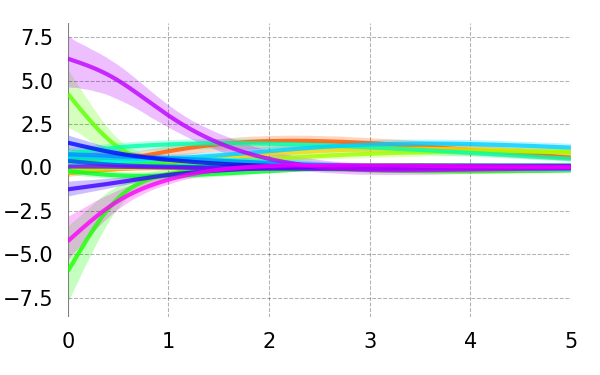}
    \end{subfigure}
    
    \vspace{1em}    
    
    \textbf{\large Consistency}
    
    \vspace{0.5em}
    
    \begin{subfigure}[t]{0.19\textwidth}
        \centering
        \makebox[0.9\textwidth]{\centering Rep 1}
        \includegraphics[width=\textwidth]{{results_cdrnn_journal_synth_noise_e10_CDR_main_irf_univariate_y_mc}.png}
    \end{subfigure}
    \begin{subfigure}[t]{0.19\textwidth}
        \centering
        \makebox[0.9\textwidth]{\centering Rep 2}
        \includegraphics[width=\textwidth]{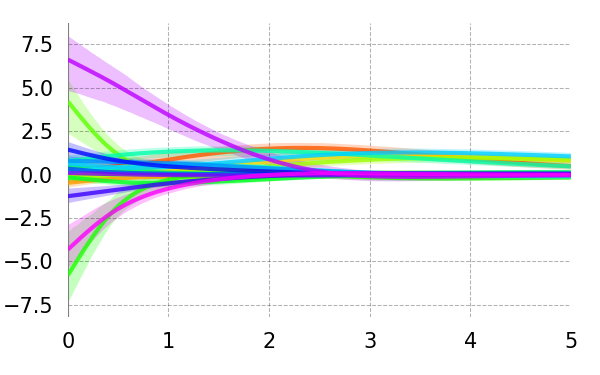}
    \end{subfigure}
    \begin{subfigure}[t]{0.19\textwidth}
        \centering
        \makebox[0.9\textwidth]{\centering Rep 3}
        \includegraphics[width=\textwidth]{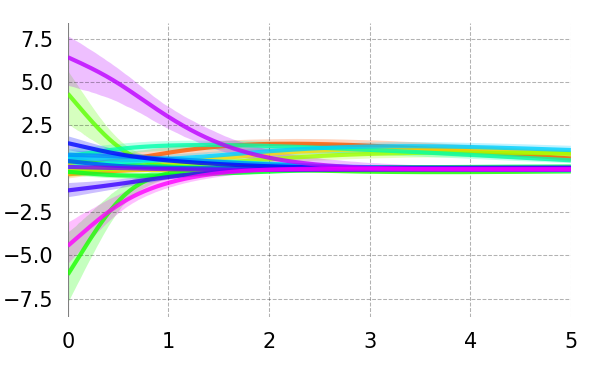}
    \end{subfigure}
    \begin{subfigure}[t]{0.19\textwidth}
        \centering
        \makebox[0.9\textwidth]{\centering Rep 4}
        \includegraphics[width=\textwidth]{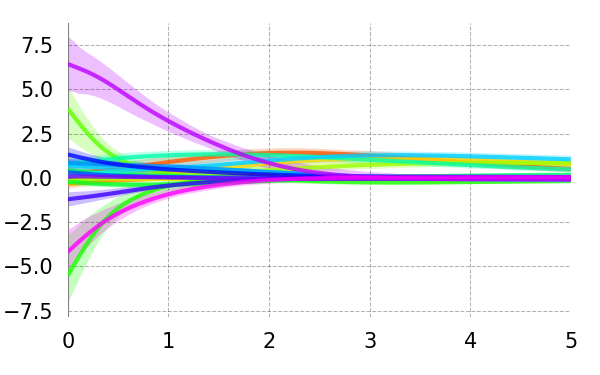}
    \end{subfigure}
    \begin{subfigure}[t]{0.19\textwidth}
        \centering
        \makebox[0.9\textwidth]{\centering Rep 5}
        \includegraphics[width=\textwidth]{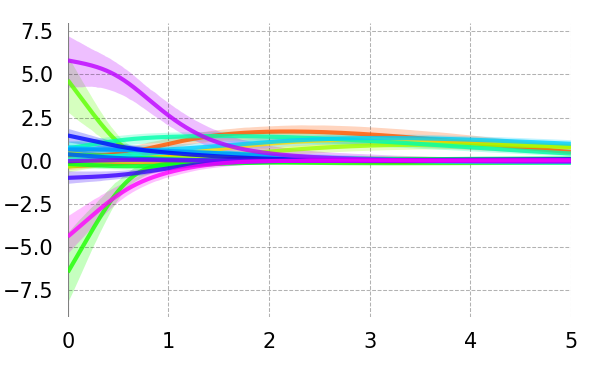}
    \end{subfigure}
    
    {\large Delay (s)}
    
    \vspace{0.5em}
    
    \caption{CDRNN estimated responses to synthetic data with \textbf{noise standard deviation $\sigma_{\epsilon} = 10$}. Estimates using base hyperparameters are compared to estimates from models that deviate from the base in some dimension. Plots under ``Consistency'' show estimates from five replicates of the ``base'' configuration, where ``Rep 1'' is the same model as ``base'' above, replotted for ease of comparison.}
    \label{fig:app-synth-noise-e10}
    
\end{figure}

\begin{figure}

    \footnotesize
    \sffamily
    \centering
    
    \textbf{\Large Synth: Noise, $\sigma_{\epsilon}=100$}
    
    \vspace{1em}
    
    \begin{subfigure}[t]{0.49\textwidth}
        \centering
        \makebox[0.49\textwidth]{\centering \textbf{True}}
        
        \includegraphics[width=0.49\textwidth]{{results_cl_synth_noise_e0_synthetic_true}.png}
    \end{subfigure}
    
    \begin{subfigure}[t]{0.49\textwidth}
        \centering
        \makebox[0.49\textwidth]{\centering Base}%
        \makebox[0.49\textwidth]{\centering + RNN}
        \begin{overpic}[width=0.49\textwidth]{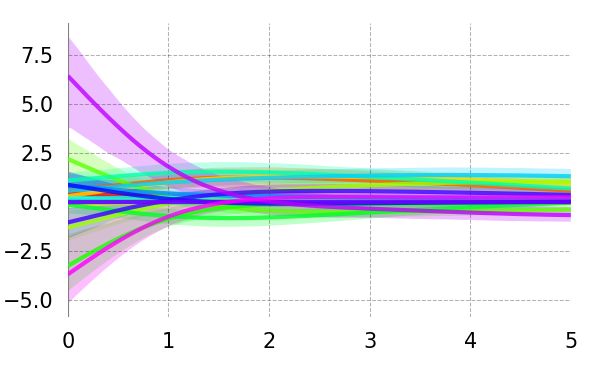}
        \end{overpic}%
        \includegraphics[width=0.49\textwidth]{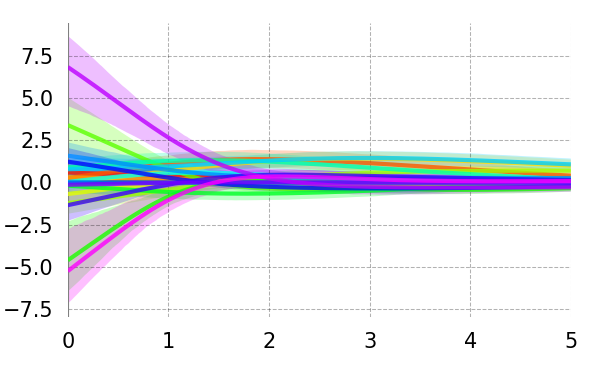}
    \end{subfigure}
    \begin{subfigure}[t]{0.49\textwidth}
        \centering
        \makebox[0.49\textwidth]{\centering Hidden Units $\div$ 2 (16)}%
        \makebox[0.49\textwidth]{\centering Hidden Units $\times$ 2 (64)}
        \includegraphics[width=0.49\textwidth]{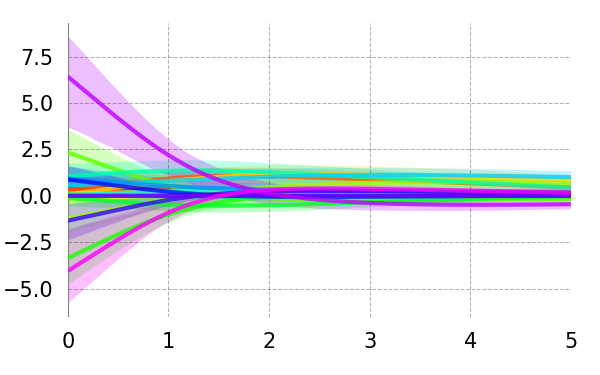}
        \includegraphics[width=0.49\textwidth]{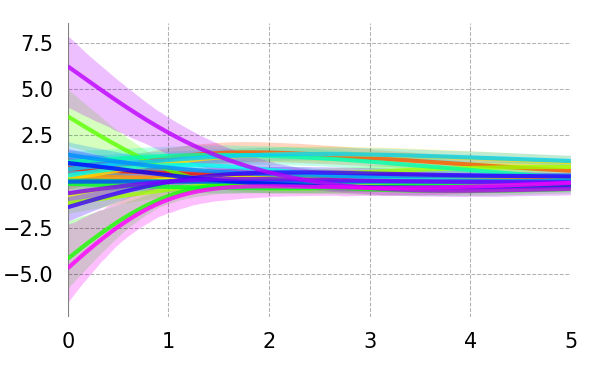}
    \end{subfigure}
    
    \begin{subfigure}[t]{0.49\textwidth}
        \centering
        \makebox[0.49\textwidth]{\centering Hidden Layers - 1 (1)}%
        \makebox[0.49\textwidth]{\centering Hidden Layers + 1 (3)}
        \includegraphics[width=0.49\textwidth]{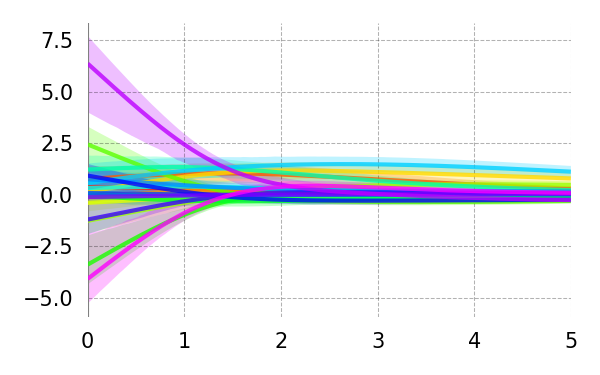}
        \includegraphics[width=0.49\textwidth]{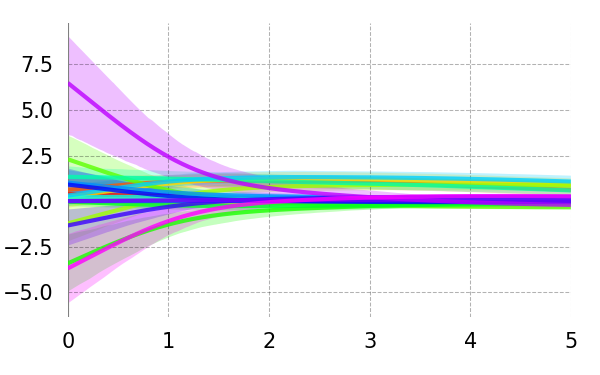}
    \end{subfigure}
    \begin{subfigure}[t]{0.49\textwidth}
        \centering
        \makebox[0.49\textwidth]{\centering Weight Reg $\div$ 5 (1)}%
        \makebox[0.49\textwidth]{\centering Weight Reg $\times$ 5 (5)}
        \includegraphics[width=0.49\textwidth]{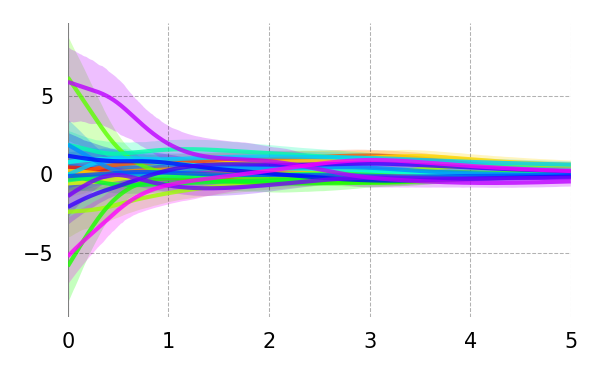}
        \includegraphics[width=0.49\textwidth]{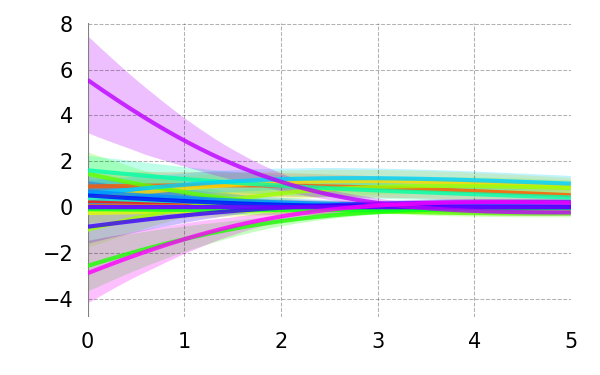}
    \end{subfigure}
    
    \begin{subfigure}[t]{0.49\textwidth}
        \centering
        \makebox[0.49\textwidth]{\centering Dropout $\div$ 2 (0.05)}%
        \makebox[0.49\textwidth]{\centering Dropout $\times$ 2 (0.2)}
        \includegraphics[width=0.49\textwidth]{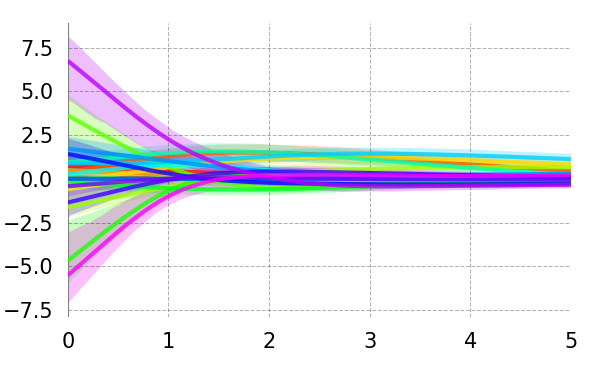}
        \includegraphics[width=0.49\textwidth]{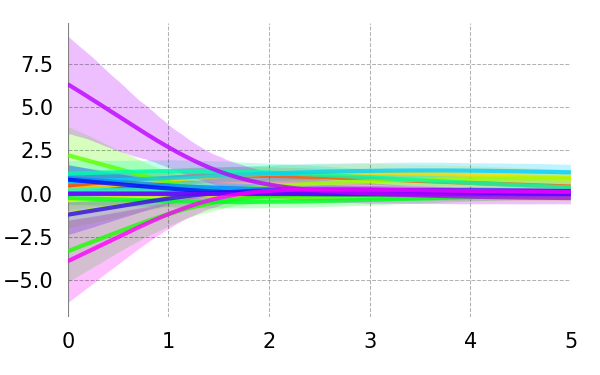}
    \end{subfigure}
    \begin{subfigure}[t]{0.49\textwidth}
        \centering
        \makebox[0.49\textwidth]{\centering Learning Rate $\div$ 3 (0.001)}%
        \makebox[0.49\textwidth]{\centering Learning Rate $\times$ 3 (0.009)}
        \includegraphics[width=0.49\textwidth]{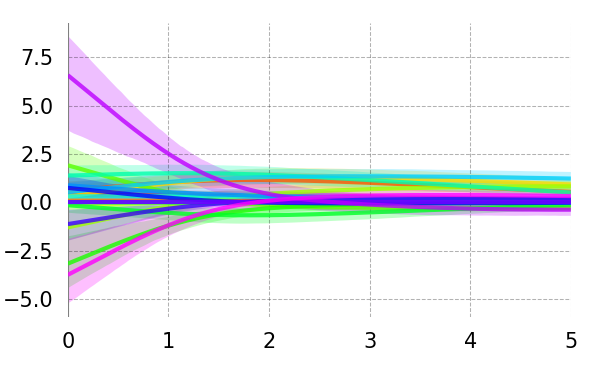}
        \includegraphics[width=0.49\textwidth]{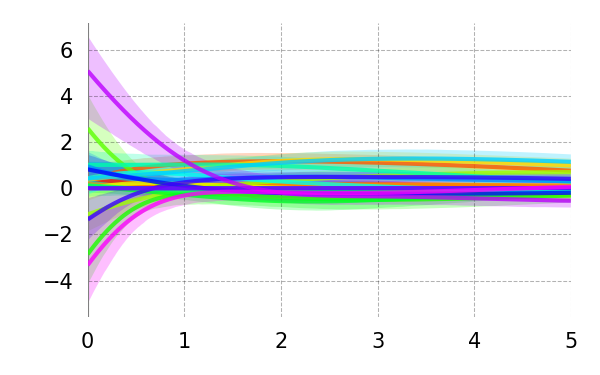}
    \end{subfigure}
    \begin{subfigure}[t]{0.49\textwidth}
        \centering
        \makebox[0.49\textwidth]{\centering Batch Size $\div$ 2 (512)}%
        \makebox[0.49\textwidth]{\centering Batch Size $\times$ 2 (2048)}
        \includegraphics[width=0.49\textwidth]{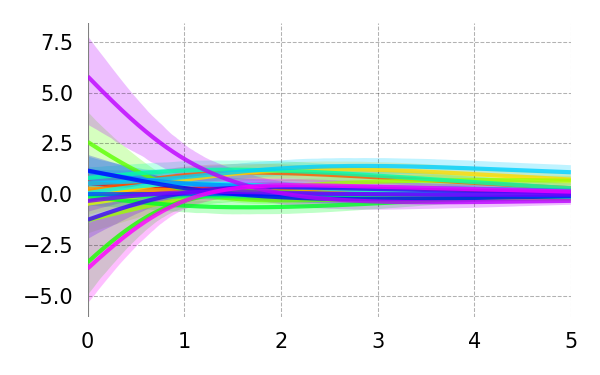}
        \includegraphics[width=0.49\textwidth]{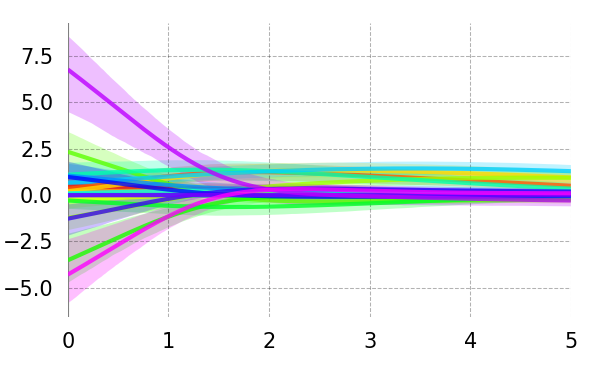}
    \end{subfigure}
    
    \vspace{1em}    
    
    \textbf{\large Consistency}
    
    \vspace{0.5em}
    
    \begin{subfigure}[t]{0.19\textwidth}
        \centering
        \makebox[0.9\textwidth]{\centering Rep 1}
        \includegraphics[width=\textwidth]{{results_cdrnn_journal_synth_noise_e100_CDR_main_irf_univariate_y_mc}.png}
    \end{subfigure}
    \begin{subfigure}[t]{0.19\textwidth}
        \centering
        \makebox[0.9\textwidth]{\centering Rep 2}
        \includegraphics[width=\textwidth]{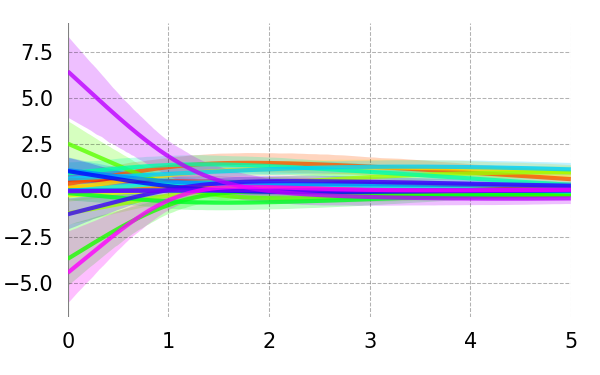}
    \end{subfigure}
    \begin{subfigure}[t]{0.19\textwidth}
        \centering
        \makebox[0.9\textwidth]{\centering Rep 3}
        \includegraphics[width=\textwidth]{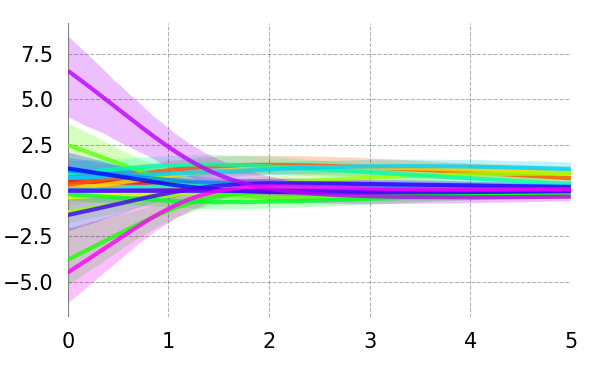}
    \end{subfigure}
    \begin{subfigure}[t]{0.19\textwidth}
        \centering
        \makebox[0.9\textwidth]{\centering Rep 4}
        \includegraphics[width=\textwidth]{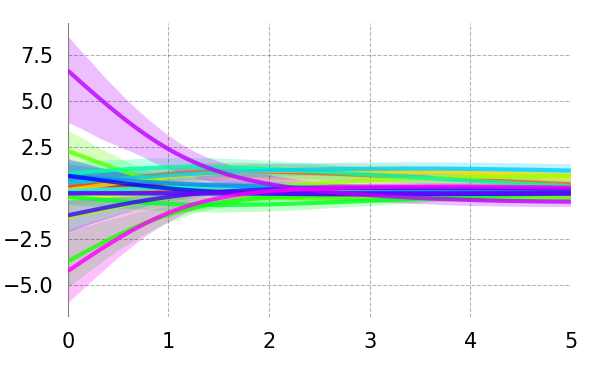}
    \end{subfigure}
    \begin{subfigure}[t]{0.19\textwidth}
        \centering
        \makebox[0.9\textwidth]{\centering Rep 5}
        \includegraphics[width=\textwidth]{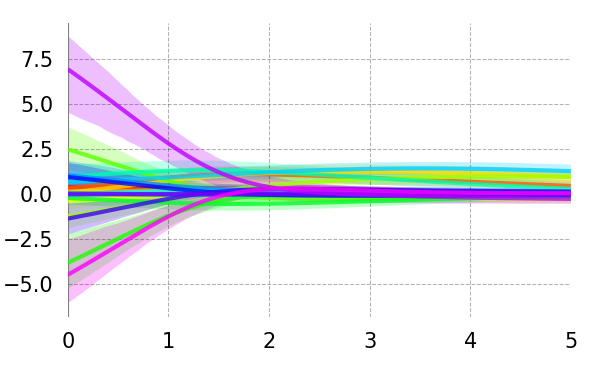}
    \end{subfigure}
    
    {\large Delay (s)}
    
    \vspace{0.5em}
    
    \caption{CDRNN estimated responses to synthetic data with \textbf{noise standard deviation $\sigma_{\epsilon} = 100$}. Estimates using base hyperparameters are compared to estimates from models that deviate from the base in some dimension. Plots under ``Consistency'' show estimates from five replicates of the ``base'' configuration, where ``Rep 1'' is the same model as ``base'' above, replotted for ease of comparison.}
    \label{fig:app-synth-noise-e100}
    
\end{figure}

\begin{figure}

    \footnotesize
    \sffamily
    \centering
    
    \textbf{\Large Synth: Time, Fixed Synchronous Short Interval (100ms)}
    \vspace{1em}
    
    \begin{subfigure}[t]{0.49\textwidth}
        \centering
        \makebox[0.49\textwidth]{\centering \textbf{True}}
        
        \includegraphics[width=0.49\textwidth]{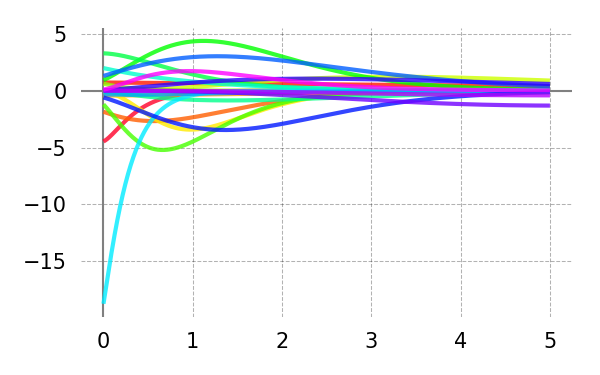}
    \end{subfigure}
    
    \begin{subfigure}[t]{0.49\textwidth}
        \centering
        \makebox[0.49\textwidth]{\centering Base}%
        \makebox[0.49\textwidth]{\centering + RNN}
        \begin{overpic}[width=0.49\textwidth]{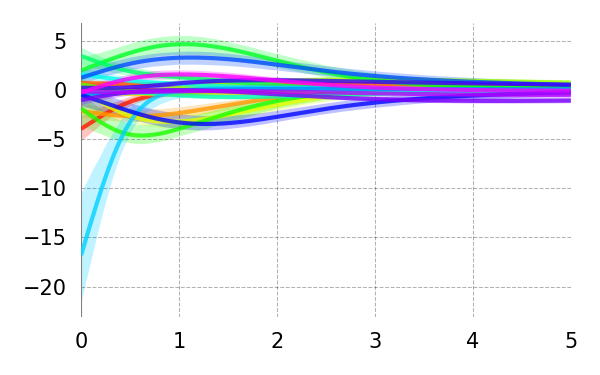}
        \end{overpic}%
        \includegraphics[width=0.49\textwidth]{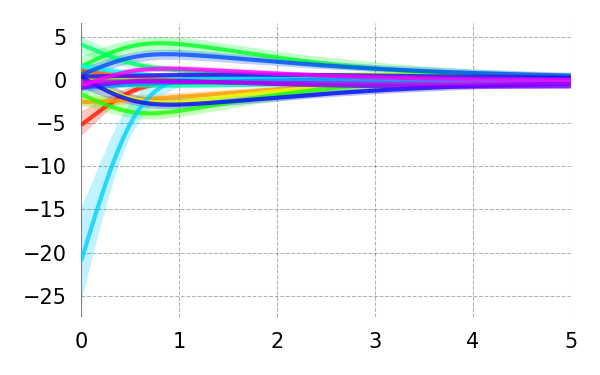}
    \end{subfigure}
    \begin{subfigure}[t]{0.49\textwidth}
        \centering
        \makebox[0.49\textwidth]{\centering Hidden Units $\div$ 2 (16)}%
        \makebox[0.49\textwidth]{\centering Hidden Units $\times$ 2 (64)}
        \includegraphics[width=0.49\textwidth]{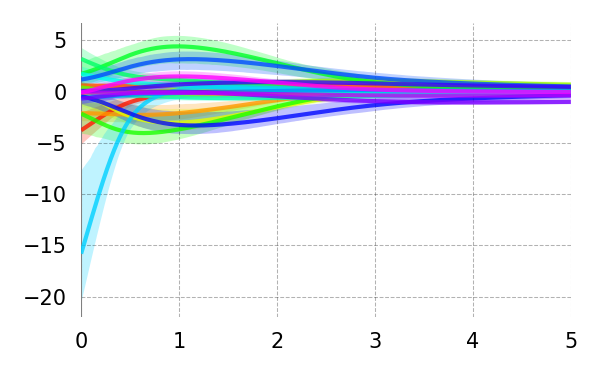}
        \includegraphics[width=0.49\textwidth]{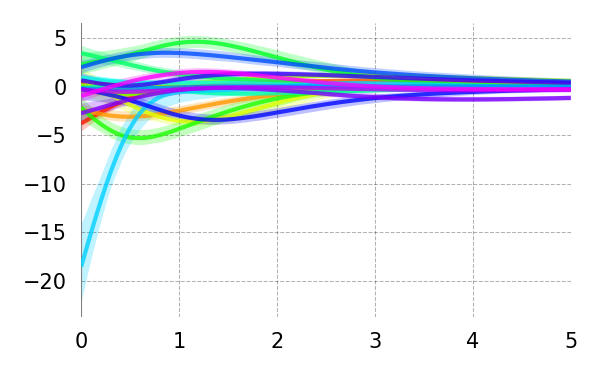}
    \end{subfigure}
    
    \begin{subfigure}[t]{0.49\textwidth}
        \centering
        \makebox[0.49\textwidth]{\centering Hidden Layers - 1 (1)}%
        \makebox[0.49\textwidth]{\centering Hidden Layers + 1 (3)}
        \includegraphics[width=0.49\textwidth]{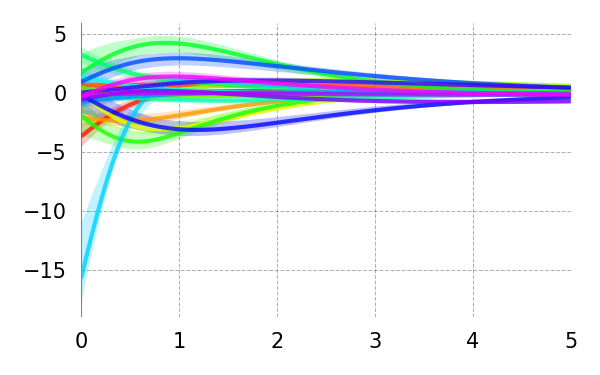}
        \includegraphics[width=0.49\textwidth]{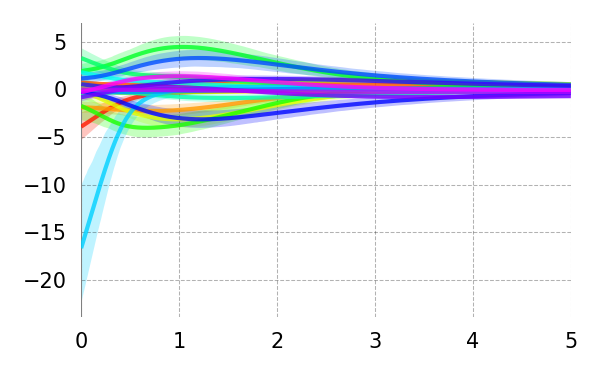}
    \end{subfigure}
    \begin{subfigure}[t]{0.49\textwidth}
        \centering
        \makebox[0.49\textwidth]{\centering Weight Reg $\div$ 5 (1)}%
        \makebox[0.49\textwidth]{\centering Weight Reg $\times$ 5 (5)}
        \includegraphics[width=0.49\textwidth]{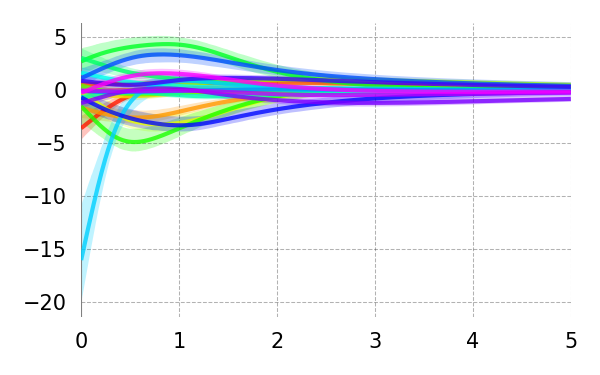}
        \includegraphics[width=0.49\textwidth]{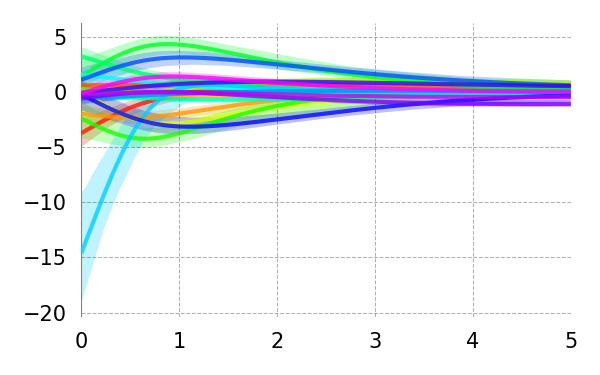}
    \end{subfigure}
    
    \begin{subfigure}[t]{0.49\textwidth}
        \centering
        \makebox[0.49\textwidth]{\centering Dropout $\div$ 2 (0.05)}%
        \makebox[0.49\textwidth]{\centering Dropout $\times$ 2 (0.2)}
        \includegraphics[width=0.49\textwidth]{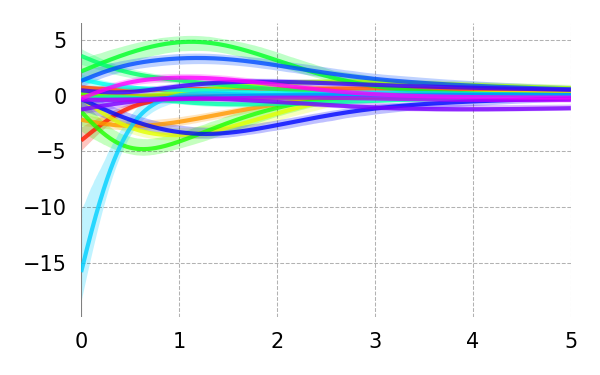}
        \includegraphics[width=0.49\textwidth]{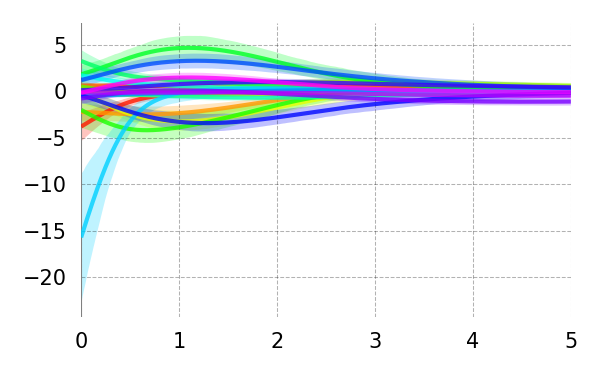}
    \end{subfigure}
    \begin{subfigure}[t]{0.49\textwidth}
        \centering
        \makebox[0.49\textwidth]{\centering Learning Rate $\div$ 3 (0.001)}%
        \makebox[0.49\textwidth]{\centering Learning Rate $\times$ 3 (0.009)}
        \includegraphics[width=0.49\textwidth]{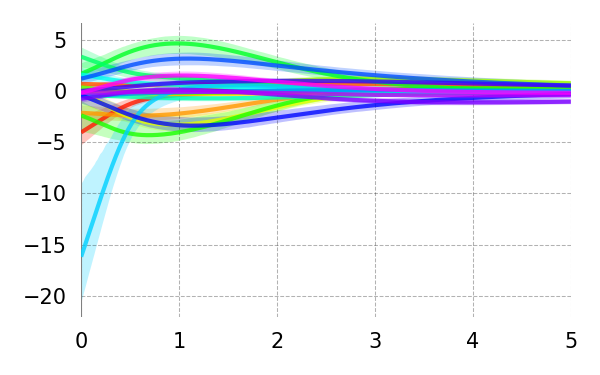}
        \includegraphics[width=0.49\textwidth]{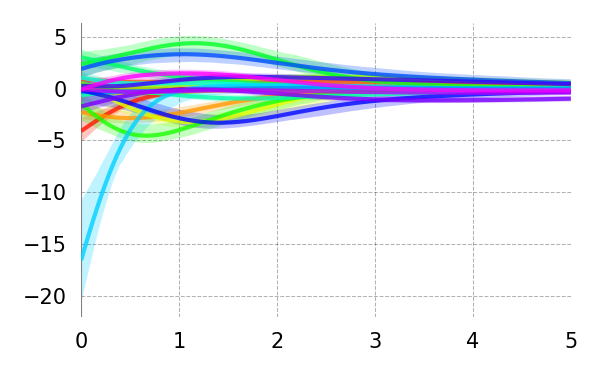}
    \end{subfigure}
    \begin{subfigure}[t]{0.49\textwidth}
        \centering
        \makebox[0.49\textwidth]{\centering Batch Size $\div$ 2 (512)}%
        \makebox[0.49\textwidth]{\centering Batch Size $\times$ 2 (2048)}
        \includegraphics[width=0.49\textwidth]{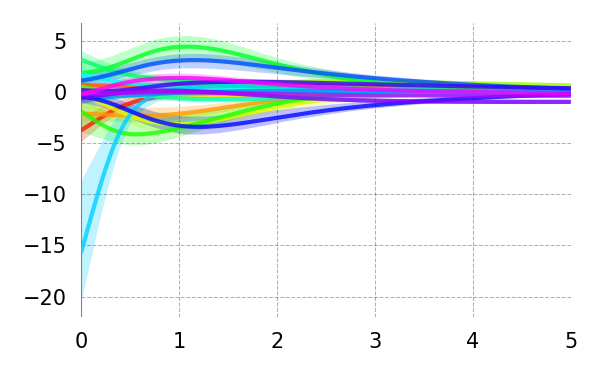}
        \includegraphics[width=0.49\textwidth]{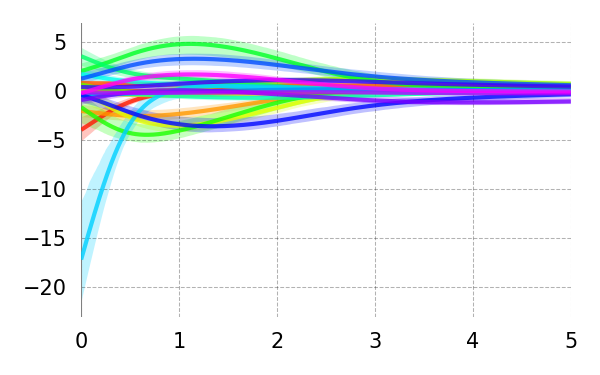}
    \end{subfigure}
    
    \vspace{1em}    
    
    \textbf{\large Consistency}
    
    \vspace{0.5em}
    
    \begin{subfigure}[t]{0.19\textwidth}
        \centering
        \makebox[0.9\textwidth]{\centering Rep 1}
        \includegraphics[width=\textwidth]{{results_cdrnn_journal_synth_time_fixed1_CDR_main_irf_univariate_y_mc}.png}
    \end{subfigure}
    \begin{subfigure}[t]{0.19\textwidth}
        \centering
        \makebox[0.9\textwidth]{\centering Rep 2}
        \includegraphics[width=\textwidth]{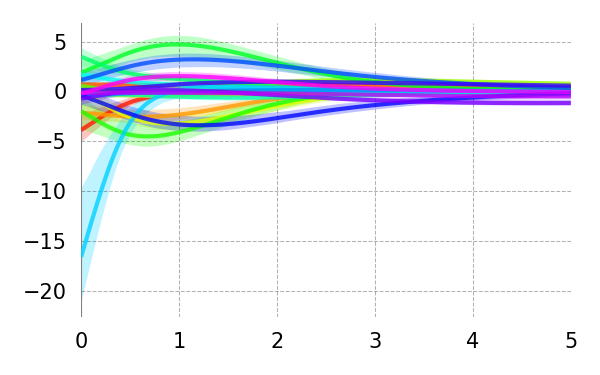}
    \end{subfigure}
    \begin{subfigure}[t]{0.19\textwidth}
        \centering
        \makebox[0.9\textwidth]{\centering Rep 3}
        \includegraphics[width=\textwidth]{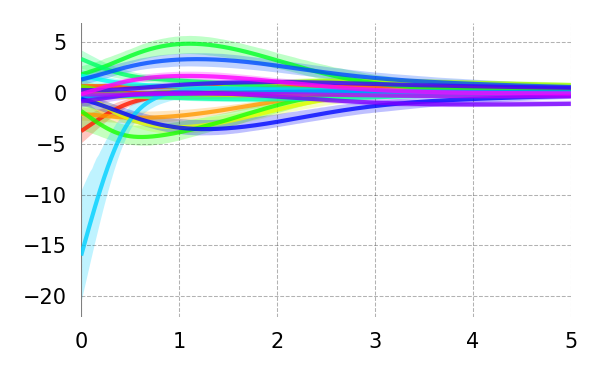}
    \end{subfigure}
    \begin{subfigure}[t]{0.19\textwidth}
        \centering
        \makebox[0.9\textwidth]{\centering Rep 4}
        \includegraphics[width=\textwidth]{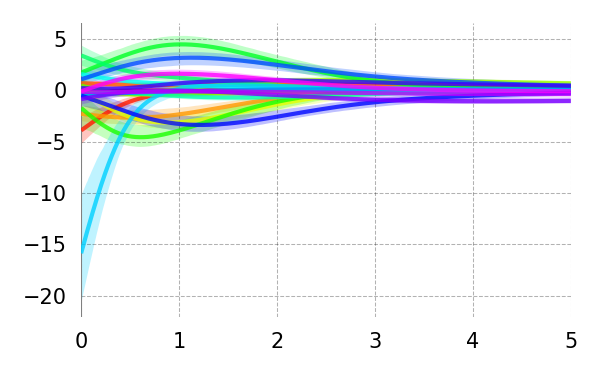}
    \end{subfigure}
    \begin{subfigure}[t]{0.19\textwidth}
        \centering
        \makebox[0.9\textwidth]{\centering Rep 5}
        \includegraphics[width=\textwidth]{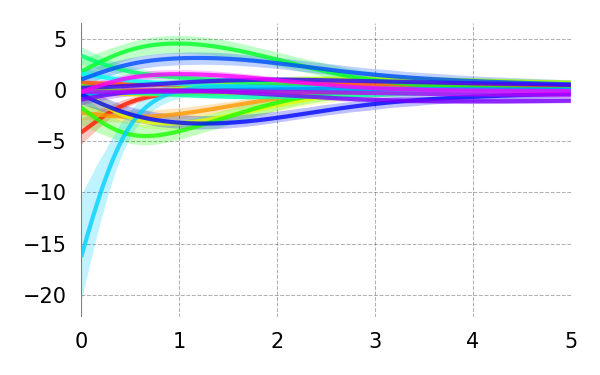}
    \end{subfigure}
    
    {\large Delay (s)}
    
    \vspace{0.5em}
    
    \caption{CDRNN estimated responses to synthetic data with a \textbf{synchronously measured predictors and responses with a fixed interval of 100ms between them}. Estimates using base hyperparameters are compared to estimates from models that deviate from the base in some dimension. Plots under ``Consistency'' show estimates from five replicates of the ``base'' configuration, where ``Rep 1'' is the same model as ``base'' above, replotted for ease of comparison.}
    \label{fig:app-synth-time-fixed1}
    
\end{figure}

\begin{figure}

    \footnotesize
    \sffamily
    \centering
    
    \textbf{\Large Synth: Time, Fixed Synchronous Long Interval (500ms)}
    
    \vspace{1em}
    
    \begin{subfigure}[t]{0.49\textwidth}
        \centering
        \makebox[0.49\textwidth]{\centering \textbf{True}}
        
        \includegraphics[width=0.49\textwidth]{{results_cl_synth_time_async1_synthetic_true}.png}
    \end{subfigure}
    
    \begin{subfigure}[t]{0.49\textwidth}
        \centering
        \makebox[0.49\textwidth]{\centering Base}%
        \makebox[0.49\textwidth]{\centering + RNN}
        \begin{overpic}[width=0.49\textwidth]{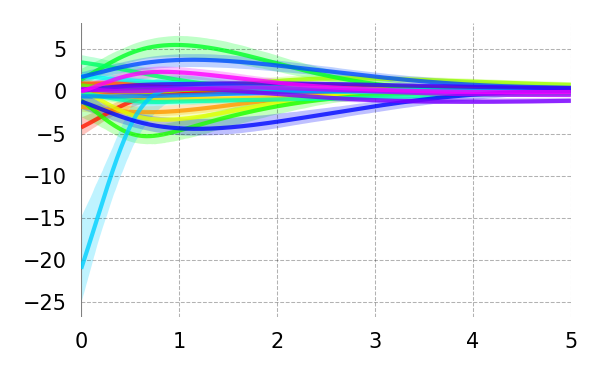}
        \end{overpic}%
        \includegraphics[width=0.49\textwidth]{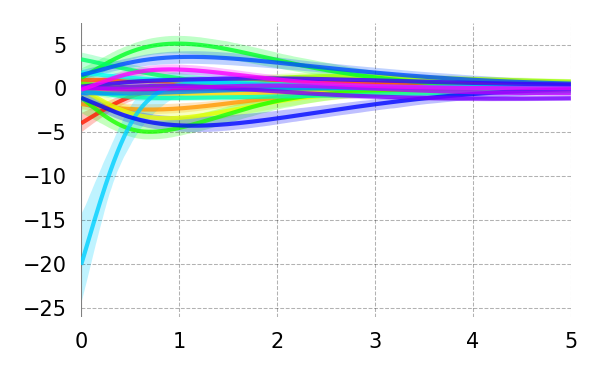}
    \end{subfigure}
    \begin{subfigure}[t]{0.49\textwidth}
        \centering
        \makebox[0.49\textwidth]{\centering Hidden Units $\div$ 2 (16)}%
        \makebox[0.49\textwidth]{\centering Hidden Units $\times$ 2 (64)}
        \includegraphics[width=0.49\textwidth]{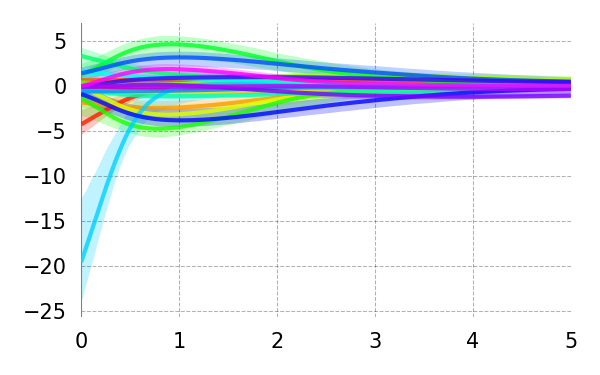}
        \includegraphics[width=0.49\textwidth]{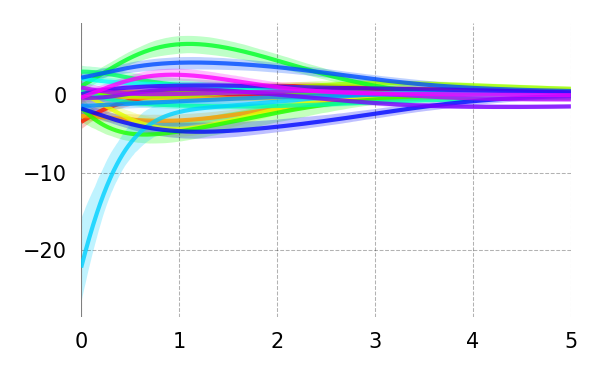}
    \end{subfigure}
    
    \begin{subfigure}[t]{0.49\textwidth}
        \centering
        \makebox[0.49\textwidth]{\centering Hidden Layers - 1 (1)}%
        \makebox[0.49\textwidth]{\centering Hidden Layers + 1 (3)}
        \includegraphics[width=0.49\textwidth]{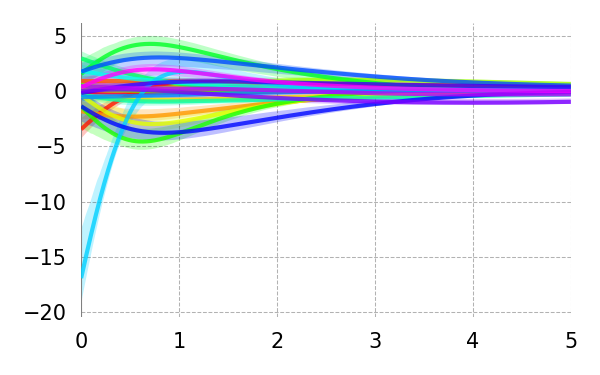}
        \includegraphics[width=0.49\textwidth]{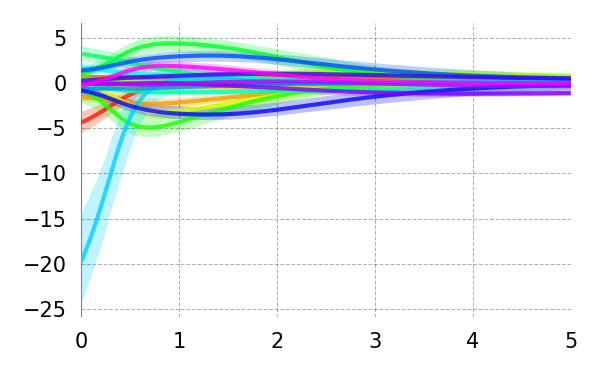}
    \end{subfigure}
    \begin{subfigure}[t]{0.49\textwidth}
        \centering
        \makebox[0.49\textwidth]{\centering Weight Reg $\div$ 5 (1)}%
        \makebox[0.49\textwidth]{\centering Weight Reg $\times$ 5 (5)}
        \includegraphics[width=0.49\textwidth]{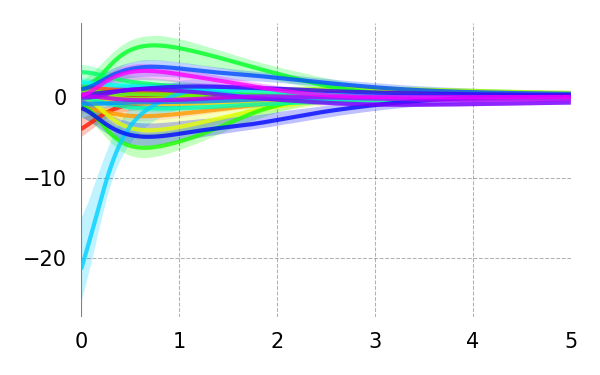}
        \includegraphics[width=0.49\textwidth]{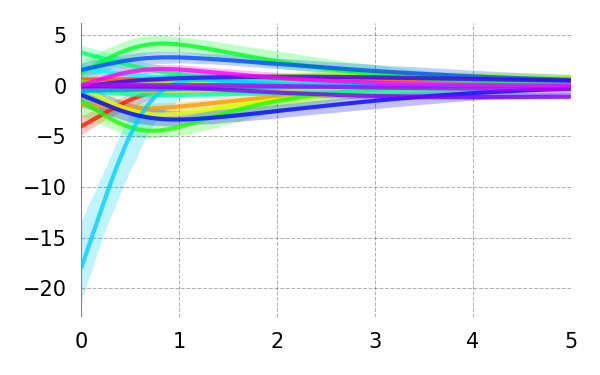}
    \end{subfigure}
    
    \begin{subfigure}[t]{0.49\textwidth}
        \centering
        \makebox[0.49\textwidth]{\centering Dropout $\div$ 2 (0.05)}%
        \makebox[0.49\textwidth]{\centering Dropout $\times$ 2 (0.2)}
        \includegraphics[width=0.49\textwidth]{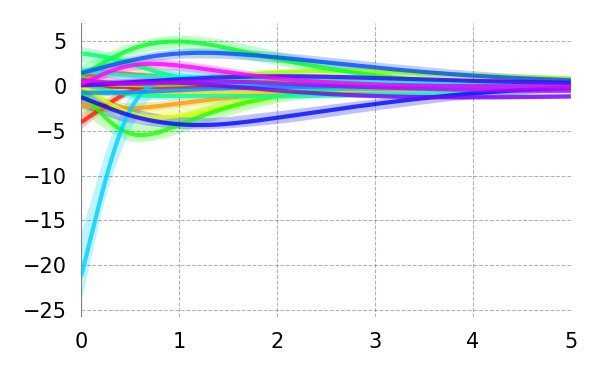}
        \includegraphics[width=0.49\textwidth]{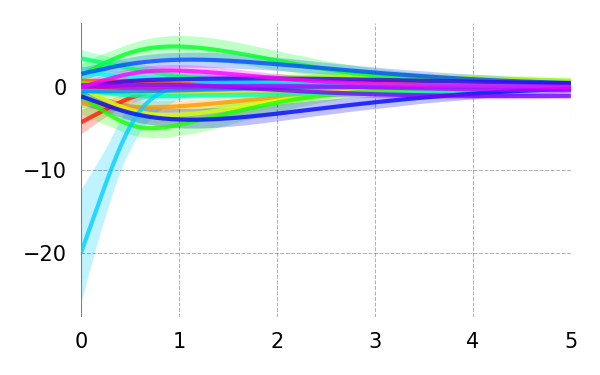}
    \end{subfigure}
    \begin{subfigure}[t]{0.49\textwidth}
        \centering
        \makebox[0.49\textwidth]{\centering Learning Rate $\div$ 3 (0.001)}%
        \makebox[0.49\textwidth]{\centering Learning Rate $\times$ 3 (0.009)}
        \includegraphics[width=0.49\textwidth]{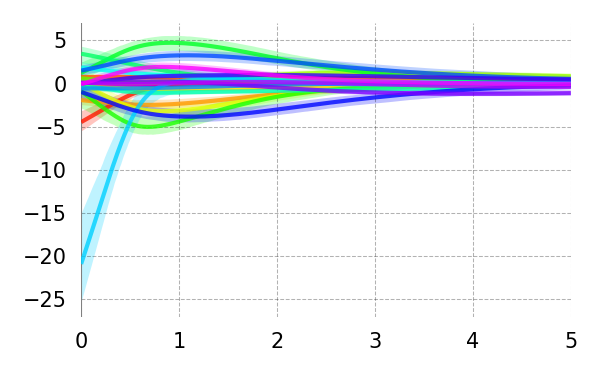}
        \includegraphics[width=0.49\textwidth]{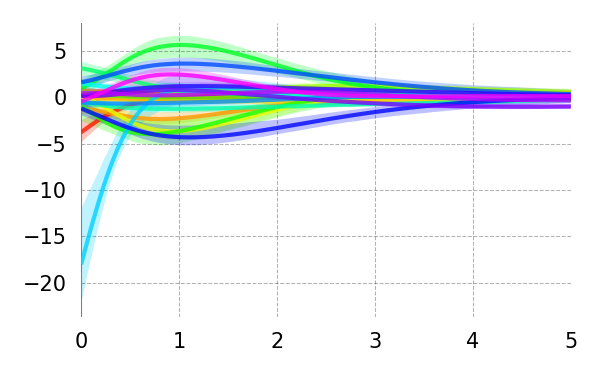}
    \end{subfigure}
    \begin{subfigure}[t]{0.49\textwidth}
        \centering
        \makebox[0.49\textwidth]{\centering Batch Size $\div$ 2 (512)}%
        \makebox[0.49\textwidth]{\centering Batch Size $\times$ 2 (2048)}
        \includegraphics[width=0.49\textwidth]{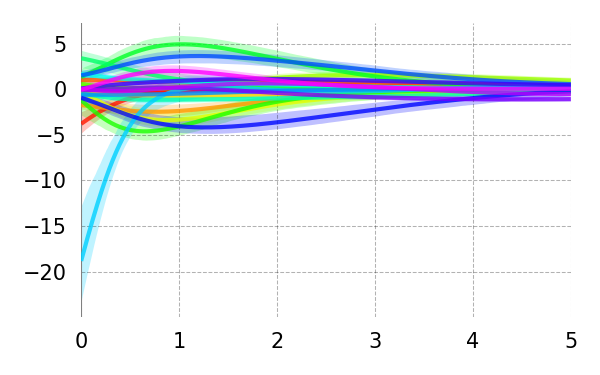}
        \includegraphics[width=0.49\textwidth]{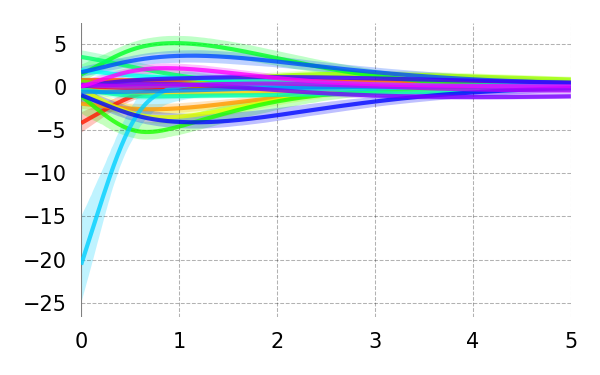}
    \end{subfigure}
    
    \vspace{1em}    
    
    \textbf{\large Consistency}
    
    \vspace{0.5em}
    
    \begin{subfigure}[t]{0.19\textwidth}
        \centering
        \makebox[0.9\textwidth]{\centering Rep 1}
        \includegraphics[width=\textwidth]{{results_cdrnn_journal_synth_time_fixed5_CDR_main_irf_univariate_y_mc}.png}
    \end{subfigure}
    \begin{subfigure}[t]{0.19\textwidth}
        \centering
        \makebox[0.9\textwidth]{\centering Rep 2}
        \includegraphics[width=\textwidth]{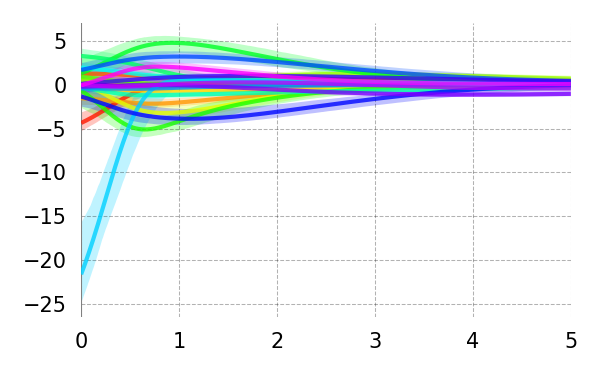}
    \end{subfigure}
    \begin{subfigure}[t]{0.19\textwidth}
        \centering
        \makebox[0.9\textwidth]{\centering Rep 3}
        \includegraphics[width=\textwidth]{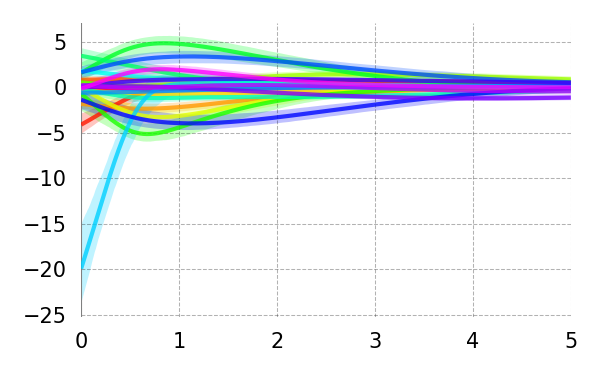}
    \end{subfigure}
    \begin{subfigure}[t]{0.19\textwidth}
        \centering
        \makebox[0.9\textwidth]{\centering Rep 4}
        \includegraphics[width=\textwidth]{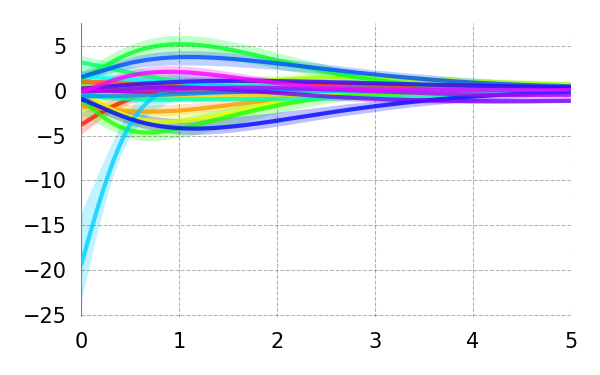}
    \end{subfigure}
    \begin{subfigure}[t]{0.19\textwidth}
        \centering
        \makebox[0.9\textwidth]{\centering Rep 5}
        \includegraphics[width=\textwidth]{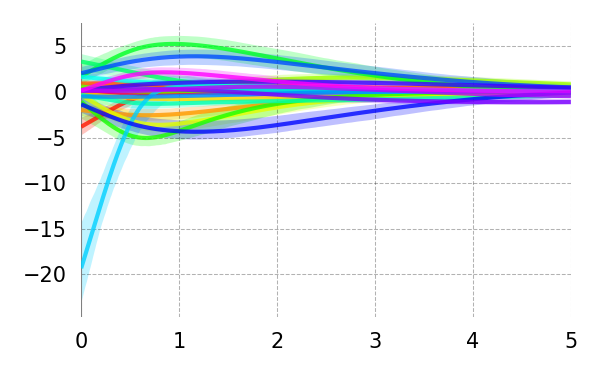}
    \end{subfigure}
    
    {\large Delay (s)}
    
    \vspace{0.5em}
    
    \caption{CDRNN estimated responses to synthetic data with a \textbf{synchronously measured predictors and responses with a fixed interval of 500ms between them}. Estimates using base hyperparameters are compared to estimates from models that deviate from the base in some dimension. Plots under ``Consistency'' show estimates from five replicates of the ``base'' configuration, where ``Rep 1'' is the same model as ``base'' above, replotted for ease of comparison.}
    \label{fig:app-synth-time-fixed5}
    
\end{figure}

\begin{figure}

    \footnotesize
    \sffamily
    \centering
    
    \textbf{\Large Synth: Time, Random Synchronous Short Interval (100ms)}
    
    \vspace{1em}
    
    \begin{subfigure}[t]{0.49\textwidth}
        \centering
        \makebox[0.49\textwidth]{\centering \textbf{True}}
        
        \includegraphics[width=0.49\textwidth]{{results_cl_synth_time_async1_synthetic_true}.png}
    \end{subfigure}
    
    \begin{subfigure}[t]{0.49\textwidth}
        \centering
        \makebox[0.49\textwidth]{\centering Base}%
        \makebox[0.49\textwidth]{\centering + RNN}
        \begin{overpic}[width=0.49\textwidth]{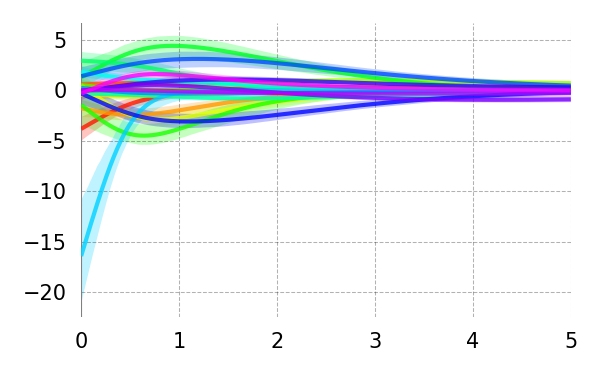}
        \end{overpic}%
        \includegraphics[width=0.49\textwidth]{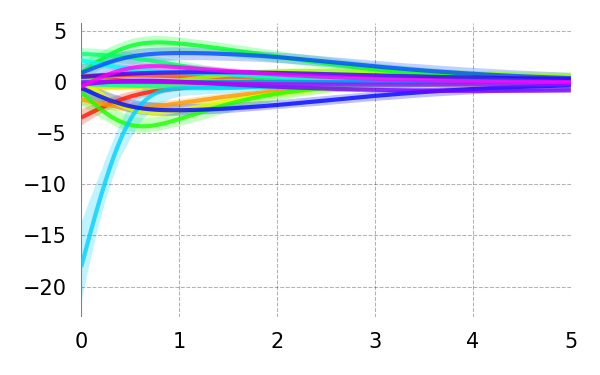}
    \end{subfigure}
    \begin{subfigure}[t]{0.49\textwidth}
        \centering
        \makebox[0.49\textwidth]{\centering Hidden Units $\div$ 2 (16)}%
        \makebox[0.49\textwidth]{\centering Hidden Units $\times$ 2 (64)}
        \includegraphics[width=0.49\textwidth]{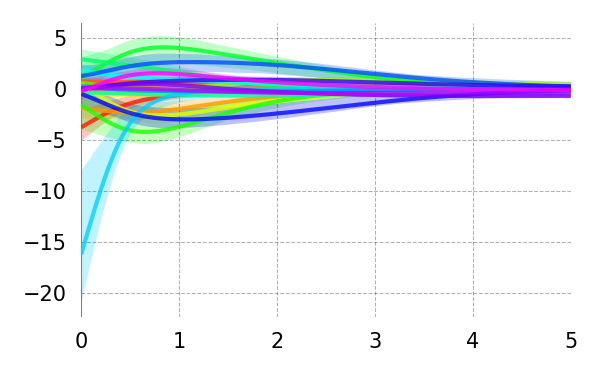}
        \includegraphics[width=0.49\textwidth]{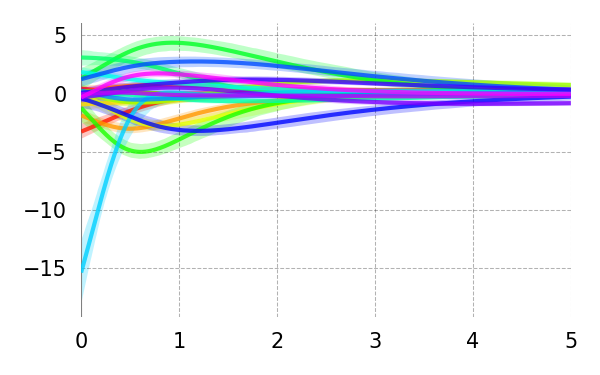}
    \end{subfigure}
    
    \begin{subfigure}[t]{0.49\textwidth}
        \centering
        \makebox[0.49\textwidth]{\centering Hidden Layers - 1 (1)}%
        \makebox[0.49\textwidth]{\centering Hidden Layers + 1 (3)}
        \includegraphics[width=0.49\textwidth]{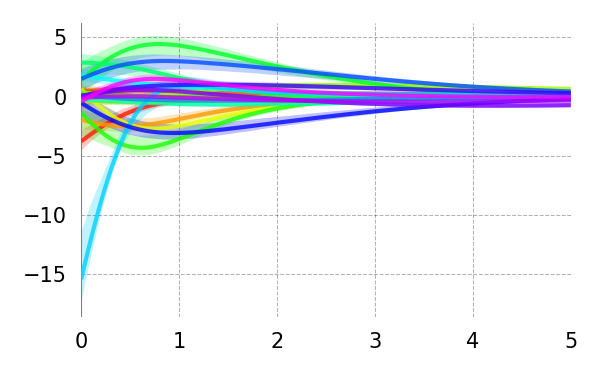}
        \includegraphics[width=0.49\textwidth]{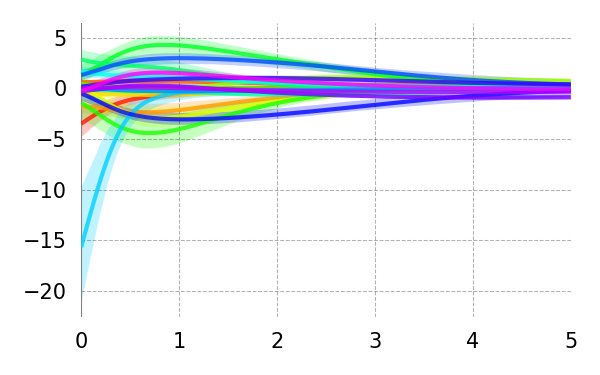}
    \end{subfigure}
    \begin{subfigure}[t]{0.49\textwidth}
        \centering
        \makebox[0.49\textwidth]{\centering Weight Reg $\div$ 5 (1)}%
        \makebox[0.49\textwidth]{\centering Weight Reg $\times$ 5 (5)}
        \includegraphics[width=0.49\textwidth]{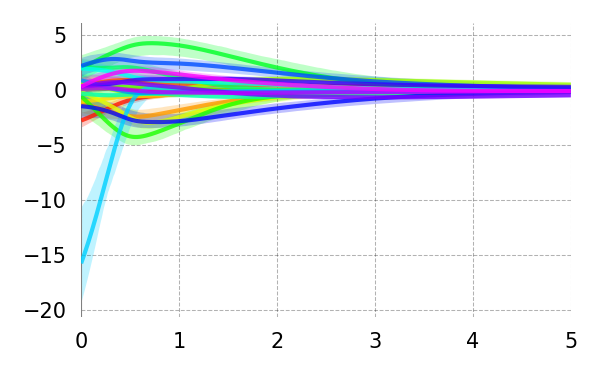}
        \includegraphics[width=0.49\textwidth]{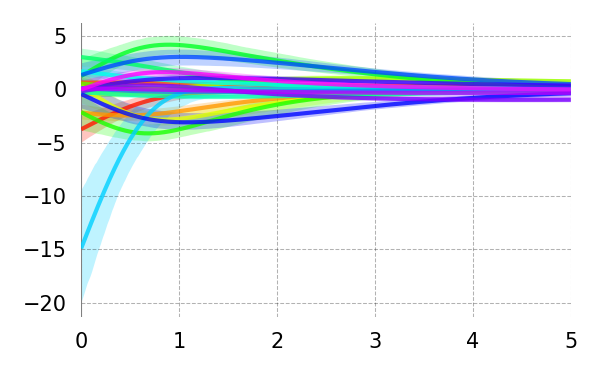}
    \end{subfigure}
    
    \begin{subfigure}[t]{0.49\textwidth}
        \centering
        \makebox[0.49\textwidth]{\centering Dropout $\div$ 2 (0.05)}%
        \makebox[0.49\textwidth]{\centering Dropout $\times$ 2 (0.2)}
        \includegraphics[width=0.49\textwidth]{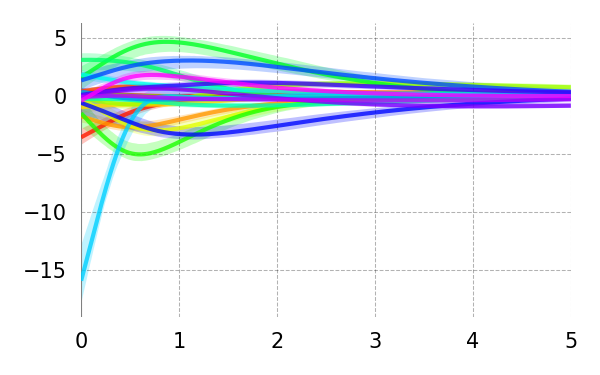}
        \includegraphics[width=0.49\textwidth]{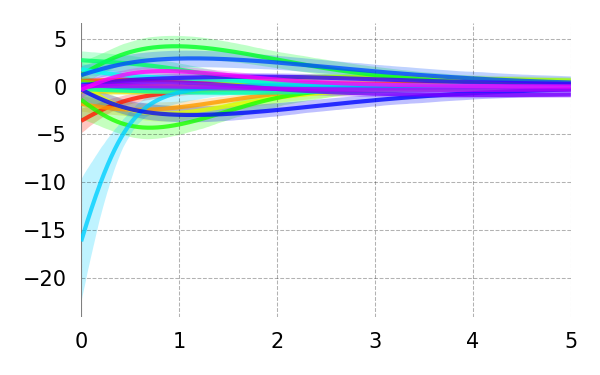}
    \end{subfigure}
    \begin{subfigure}[t]{0.49\textwidth}
        \centering
        \makebox[0.49\textwidth]{\centering Learning Rate $\div$ 3 (0.001)}%
        \makebox[0.49\textwidth]{\centering Learning Rate $\times$ 3 (0.009)}
        \includegraphics[width=0.49\textwidth]{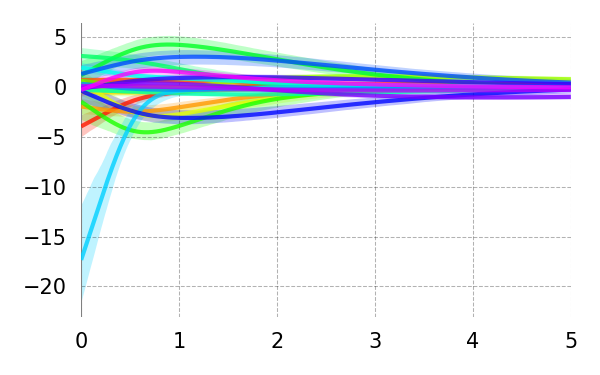}
        \includegraphics[width=0.49\textwidth]{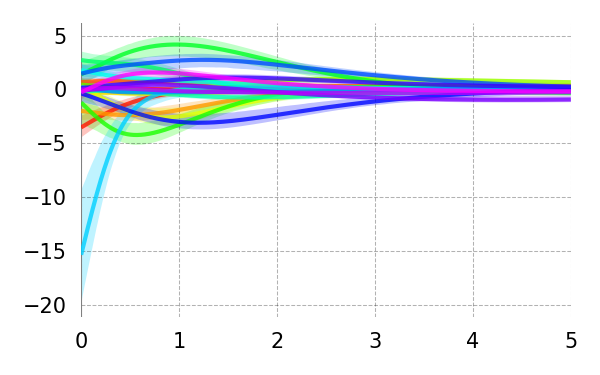}
    \end{subfigure}
    \begin{subfigure}[t]{0.49\textwidth}
        \centering
        \makebox[0.49\textwidth]{\centering Batch Size $\div$ 2 (512)}%
        \makebox[0.49\textwidth]{\centering Batch Size $\times$ 2 (2048)}
        \includegraphics[width=0.49\textwidth]{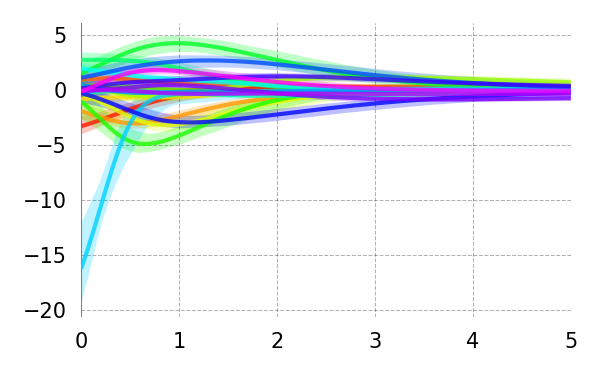}
        \includegraphics[width=0.49\textwidth]{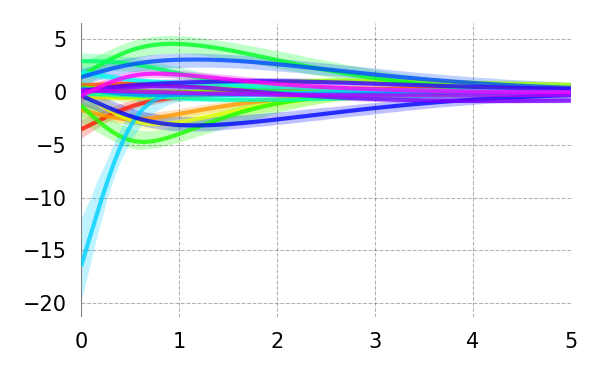}
    \end{subfigure}
    
    \vspace{1em}    
    
    \textbf{\large Consistency}
    
    \vspace{0.5em}
    
    \begin{subfigure}[t]{0.19\textwidth}
        \centering
        \makebox[0.9\textwidth]{\centering Rep 1}
        \includegraphics[width=\textwidth]{{results_cdrnn_journal_synth_time_rnd1_CDR_main_irf_univariate_y_mc}.png}
    \end{subfigure}
    \begin{subfigure}[t]{0.19\textwidth}
        \centering
        \makebox[0.9\textwidth]{\centering Rep 2}
        \includegraphics[width=\textwidth]{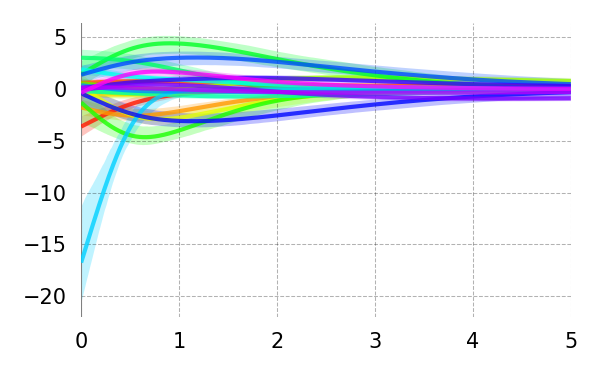}
    \end{subfigure}
    \begin{subfigure}[t]{0.19\textwidth}
        \centering
        \makebox[0.9\textwidth]{\centering Rep 3}
        \includegraphics[width=\textwidth]{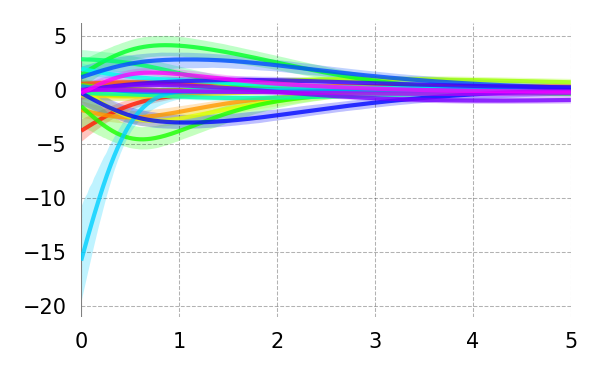}
    \end{subfigure}
    \begin{subfigure}[t]{0.19\textwidth}
        \centering
        \makebox[0.9\textwidth]{\centering Rep 4}
        \includegraphics[width=\textwidth]{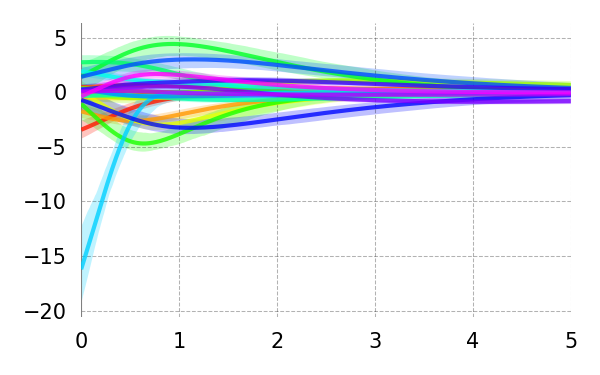}
    \end{subfigure}
    \begin{subfigure}[t]{0.19\textwidth}
        \centering
        \makebox[0.9\textwidth]{\centering Rep 5}
        \includegraphics[width=\textwidth]{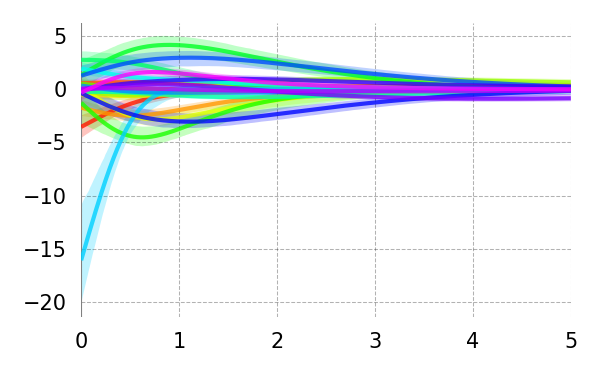}
    \end{subfigure}
    
    {\large Delay (s)}
    
    \vspace{0.5em}
    
    \caption{CDRNN estimated responses to synthetic data with a \textbf{synchronously measured predictors and responses with a variable intervals between them (mean interval 100ms)}. Estimates using base hyperparameters are compared to estimates from models that deviate from the base in some dimension. Plots under ``Consistency'' show estimates from five replicates of the ``base'' configuration, where ``Rep 1'' is the same model as ``base'' above, replotted for ease of comparison.}
    \label{fig:app-synth-time-rnd1}
    
\end{figure}

\begin{figure}

    \footnotesize
    \sffamily
    \centering
    
    \textbf{\Large Synth: Time, Random Synchronous Long Interval (500ms)}
    
    \vspace{1em}
    
    \begin{subfigure}[t]{0.49\textwidth}
        \centering
        \makebox[0.49\textwidth]{\centering \textbf{True}}
        
        \includegraphics[width=0.49\textwidth]{{results_cl_synth_time_async1_synthetic_true}.png}
    \end{subfigure}
    
    \begin{subfigure}[t]{0.49\textwidth}
        \centering
        \makebox[0.49\textwidth]{\centering Base}%
        \makebox[0.49\textwidth]{\centering + RNN}
        \begin{overpic}[width=0.49\textwidth]{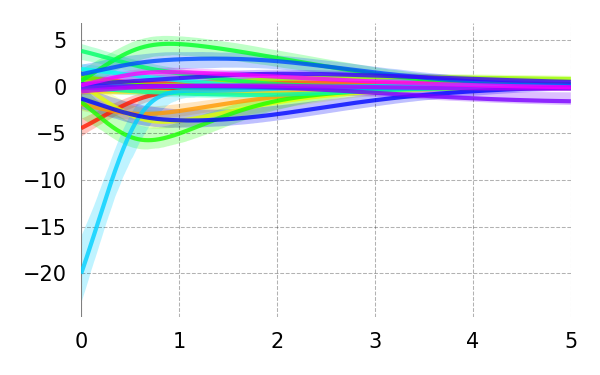}
        \end{overpic}%
        \includegraphics[width=0.49\textwidth]{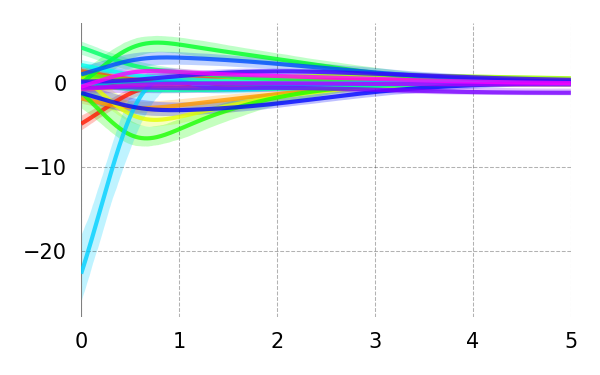}
    \end{subfigure}
    \begin{subfigure}[t]{0.49\textwidth}
        \centering
        \makebox[0.49\textwidth]{\centering Hidden Units $\div$ 2 (16)}%
        \makebox[0.49\textwidth]{\centering Hidden Units $\times$ 2 (64)}
        \includegraphics[width=0.49\textwidth]{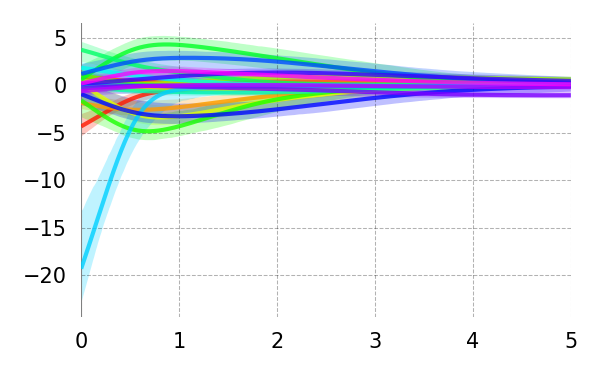}
        \includegraphics[width=0.49\textwidth]{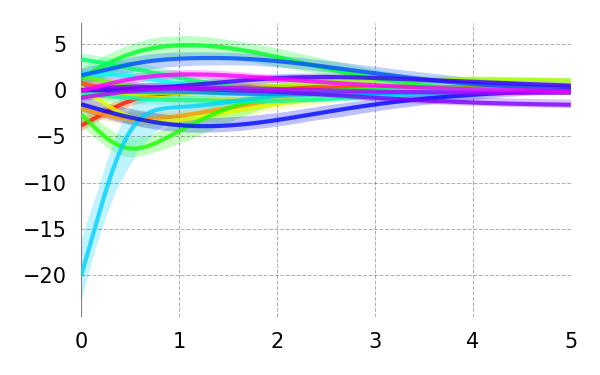}
    \end{subfigure}
    
    \begin{subfigure}[t]{0.49\textwidth}
        \centering
        \makebox[0.49\textwidth]{\centering Hidden Layers - 1 (1)}%
        \makebox[0.49\textwidth]{\centering Hidden Layers + 1 (3)}
        \includegraphics[width=0.49\textwidth]{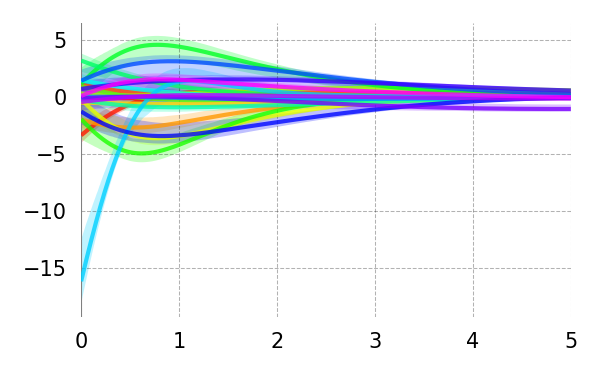}
        \includegraphics[width=0.49\textwidth]{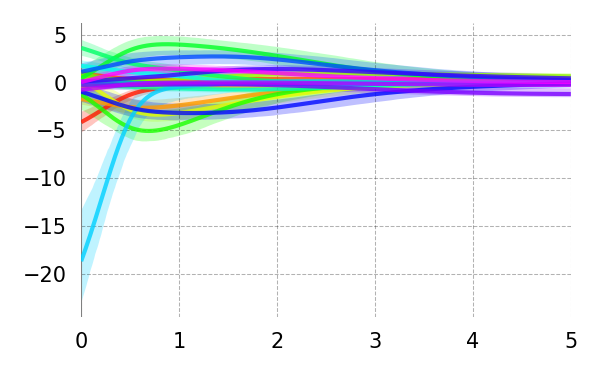}
    \end{subfigure}
    \begin{subfigure}[t]{0.49\textwidth}
        \centering
        \makebox[0.49\textwidth]{\centering Weight Reg $\div$ 5 (1)}%
        \makebox[0.49\textwidth]{\centering Weight Reg $\times$ 5 (5)}
        \includegraphics[width=0.49\textwidth]{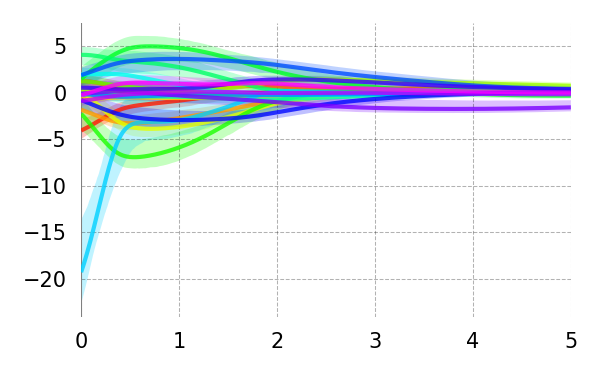}
        \includegraphics[width=0.49\textwidth]{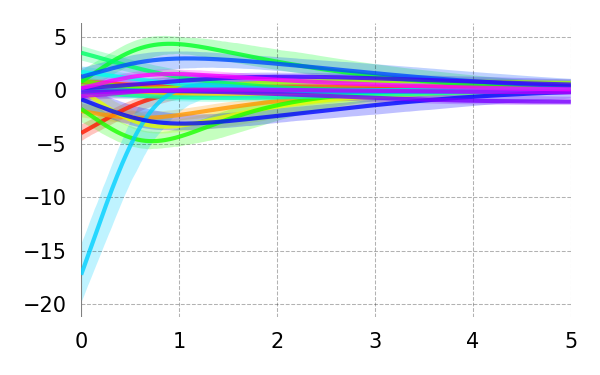}
    \end{subfigure}
    
    \begin{subfigure}[t]{0.49\textwidth}
        \centering
        \makebox[0.49\textwidth]{\centering Dropout $\div$ 2 (0.05)}%
        \makebox[0.49\textwidth]{\centering Dropout $\times$ 2 (0.2)}
        \includegraphics[width=0.49\textwidth]{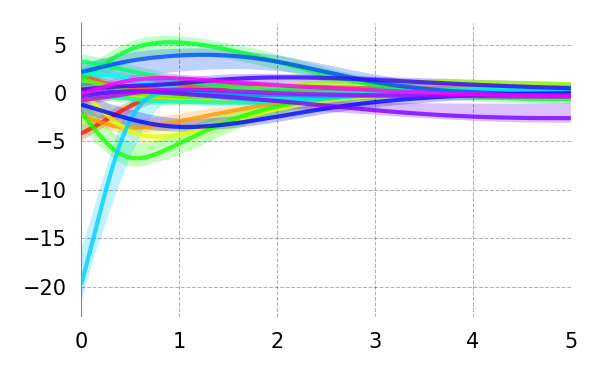}
        \includegraphics[width=0.49\textwidth]{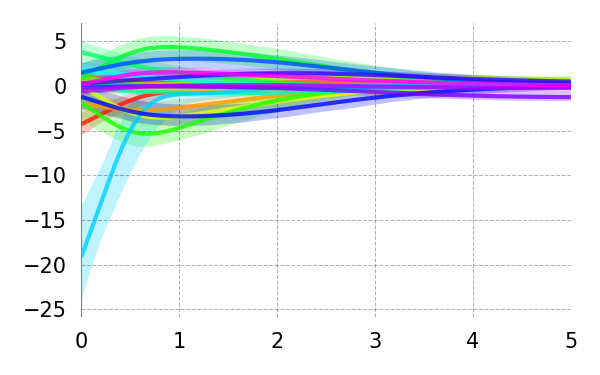}
    \end{subfigure}
    \begin{subfigure}[t]{0.49\textwidth}
        \centering
        \makebox[0.49\textwidth]{\centering Learning Rate $\div$ 3 (0.001)}%
        \makebox[0.49\textwidth]{\centering Learning Rate $\times$ 3 (0.009)}
        \includegraphics[width=0.49\textwidth]{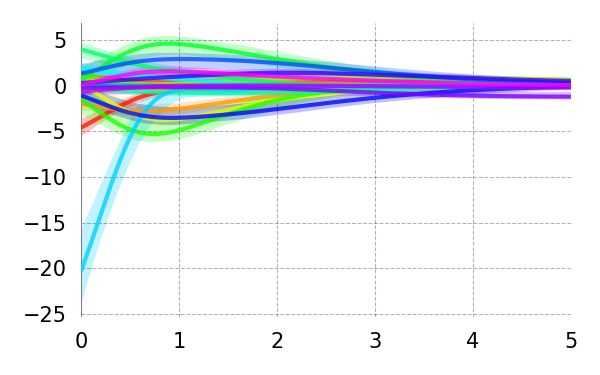}
        \includegraphics[width=0.49\textwidth]{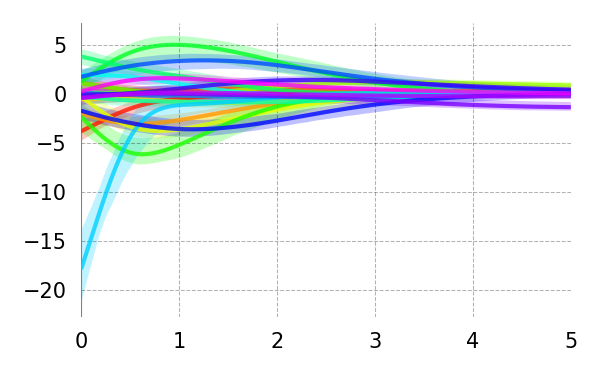}
    \end{subfigure}
    \begin{subfigure}[t]{0.49\textwidth}
        \centering
        \makebox[0.49\textwidth]{\centering Batch Size $\div$ 2 (512)}%
        \makebox[0.49\textwidth]{\centering Batch Size $\times$ 2 (2048)}
        \includegraphics[width=0.49\textwidth]{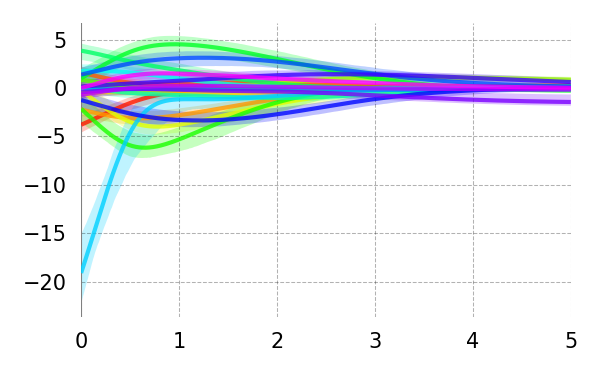}
        \includegraphics[width=0.49\textwidth]{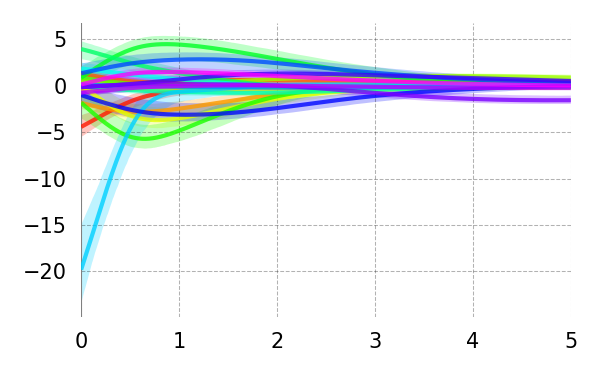}
    \end{subfigure}
    
    \vspace{1em}    
    
    \textbf{\large Consistency}
    
    \vspace{0.5em}
    
    \begin{subfigure}[t]{0.19\textwidth}
        \centering
        \makebox[0.9\textwidth]{\centering Rep 1}
        \includegraphics[width=\textwidth]{{results_cdrnn_journal_synth_time_rnd5_CDR_main_irf_univariate_y_mc}.png}
    \end{subfigure}
    \begin{subfigure}[t]{0.19\textwidth}
        \centering
        \makebox[0.9\textwidth]{\centering Rep 2}
        \includegraphics[width=\textwidth]{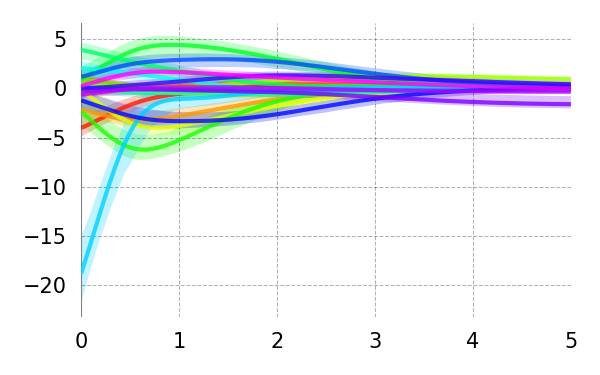}
    \end{subfigure}
    \begin{subfigure}[t]{0.19\textwidth}
        \centering
        \makebox[0.9\textwidth]{\centering Rep 3}
        \includegraphics[width=\textwidth]{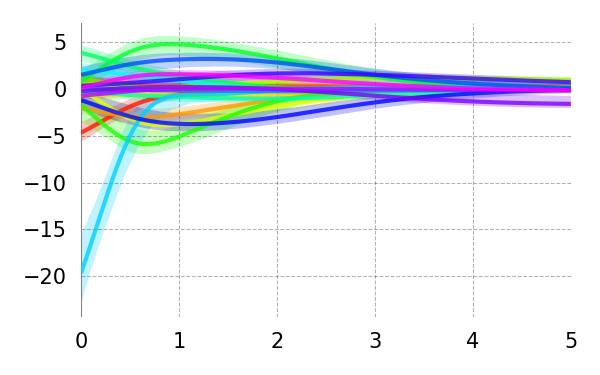}
    \end{subfigure}
    \begin{subfigure}[t]{0.19\textwidth}
        \centering
        \makebox[0.9\textwidth]{\centering Rep 4}
        \includegraphics[width=\textwidth]{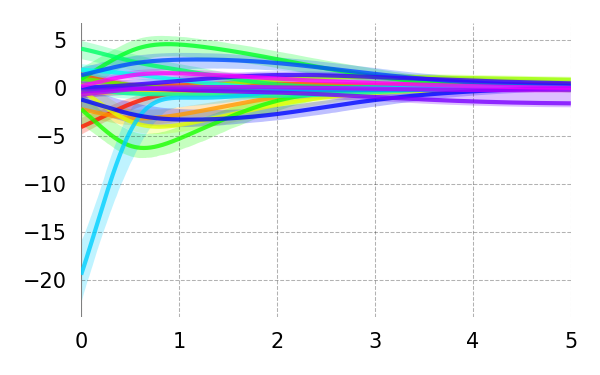}
    \end{subfigure}
    \begin{subfigure}[t]{0.19\textwidth}
        \centering
        \makebox[0.9\textwidth]{\centering Rep 5}
        \includegraphics[width=\textwidth]{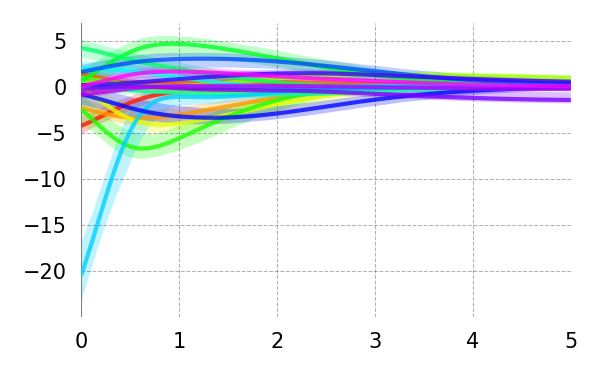}
    \end{subfigure}
    
    {\large Delay (s)}
    
    \vspace{0.5em}
    
    \caption{CDRNN estimated responses to synthetic data with a \textbf{synchronously measured predictors and responses with a variable intervals between them (mean interval 500ms)}. Estimates using base hyperparameters are compared to estimates from models that deviate from the base in some dimension. Plots under ``Consistency'' show estimates from five replicates of the ``base'' configuration, where ``Rep 1'' is the same model as ``base'' above, replotted for ease of comparison.}
    \label{fig:app-synth-time-rnd5}
    
\end{figure}

\begin{figure}

    \footnotesize
    \sffamily
    \centering
    
    \textbf{\Large Synth: Time, Random Asynchronous Short Interval (mean 100ms)}
    
    \vspace{1em}
    
    \begin{subfigure}[t]{0.49\textwidth}
        \centering
        \makebox[0.49\textwidth]{\centering \textbf{True}}
        
        \includegraphics[width=0.49\textwidth]{{results_cl_synth_time_async1_synthetic_true}.png}
    \end{subfigure}
    
    \begin{subfigure}[t]{0.49\textwidth}
        \centering
        \makebox[0.49\textwidth]{\centering Base}%
        \makebox[0.49\textwidth]{\centering + RNN}
        \begin{overpic}[width=0.49\textwidth]{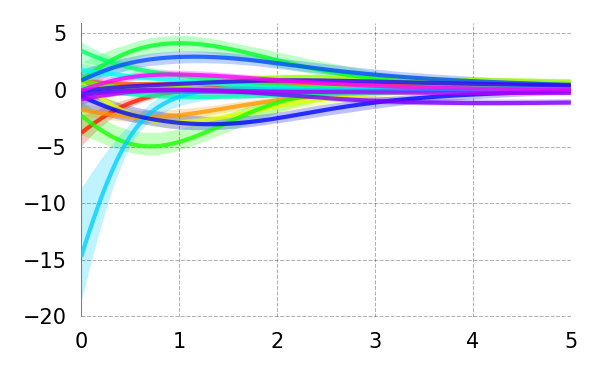}
        \end{overpic}%
        \includegraphics[width=0.49\textwidth]{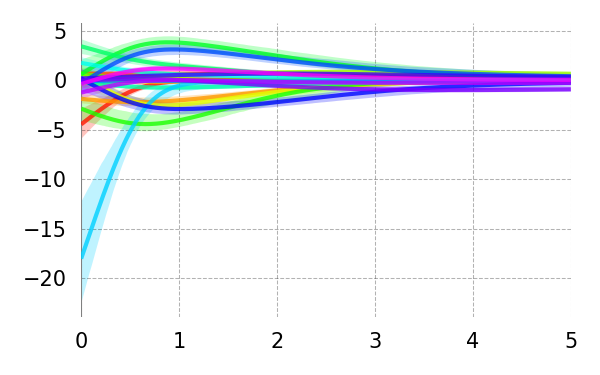}
    \end{subfigure}
    \begin{subfigure}[t]{0.49\textwidth}
        \centering
        \makebox[0.49\textwidth]{\centering Hidden Units $\div$ 2 (16)}%
        \makebox[0.49\textwidth]{\centering Hidden Units $\times$ 2 (64)}
        \includegraphics[width=0.49\textwidth]{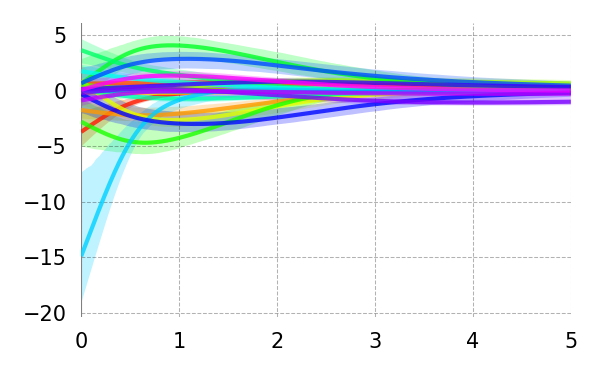}
        \includegraphics[width=0.49\textwidth]{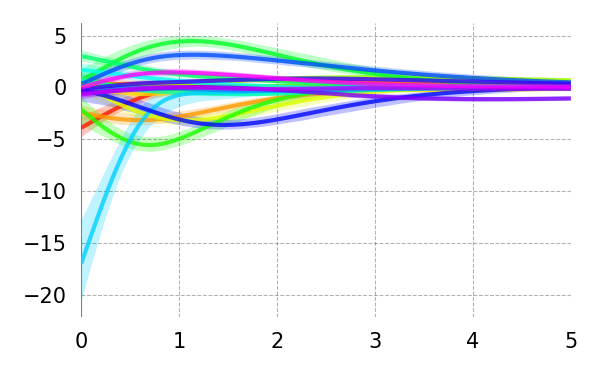}
    \end{subfigure}
    
    \begin{subfigure}[t]{0.49\textwidth}
        \centering
        \makebox[0.49\textwidth]{\centering Hidden Layers - 1 (1)}%
        \makebox[0.49\textwidth]{\centering Hidden Layers + 1 (3)}
        \includegraphics[width=0.49\textwidth]{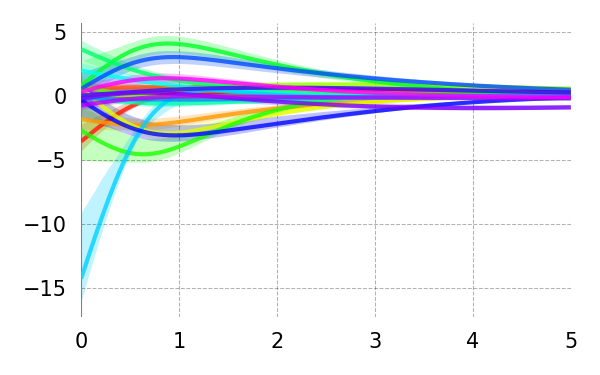}
        \includegraphics[width=0.49\textwidth]{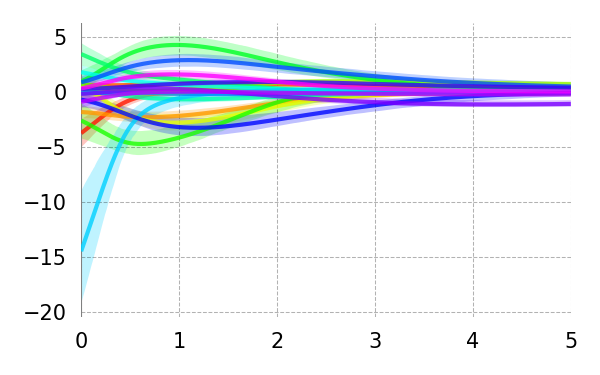}
    \end{subfigure}
    \begin{subfigure}[t]{0.49\textwidth}
        \centering
        \makebox[0.49\textwidth]{\centering Weight Reg $\div$ 5 (1)}%
        \makebox[0.49\textwidth]{\centering Weight Reg $\times$ 5 (5)}
        \includegraphics[width=0.49\textwidth]{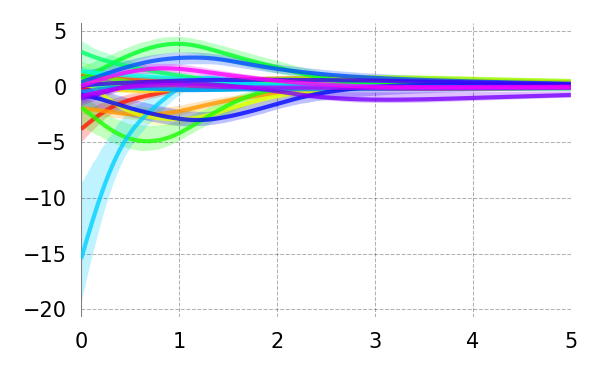}
        \includegraphics[width=0.49\textwidth]{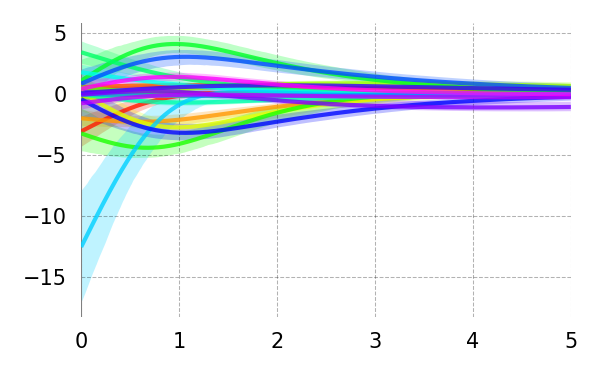}
    \end{subfigure}
    
    \begin{subfigure}[t]{0.49\textwidth}
        \centering
        \makebox[0.49\textwidth]{\centering Dropout $\div$ 2 (0.05)}%
        \makebox[0.49\textwidth]{\centering Dropout $\times$ 2 (0.2)}
        \includegraphics[width=0.49\textwidth]{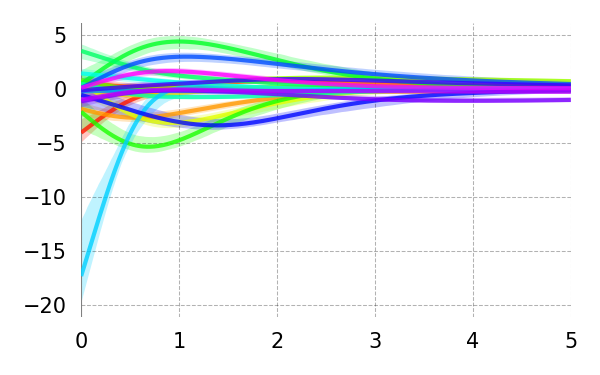}
        \includegraphics[width=0.49\textwidth]{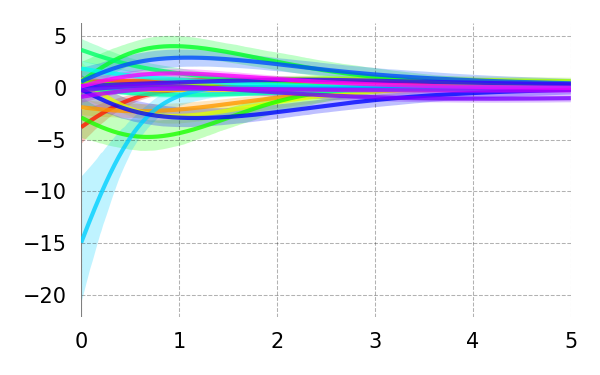}
    \end{subfigure}
    \begin{subfigure}[t]{0.49\textwidth}
        \centering
        \makebox[0.49\textwidth]{\centering Learning Rate $\div$ 3 (0.001)}%
        \makebox[0.49\textwidth]{\centering Learning Rate $\times$ 3 (0.009)}
        \includegraphics[width=0.49\textwidth]{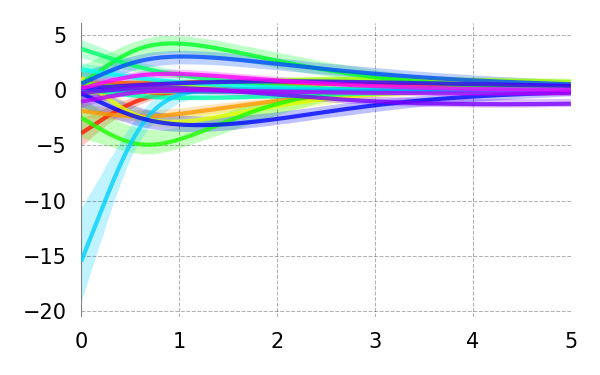}
        \includegraphics[width=0.49\textwidth]{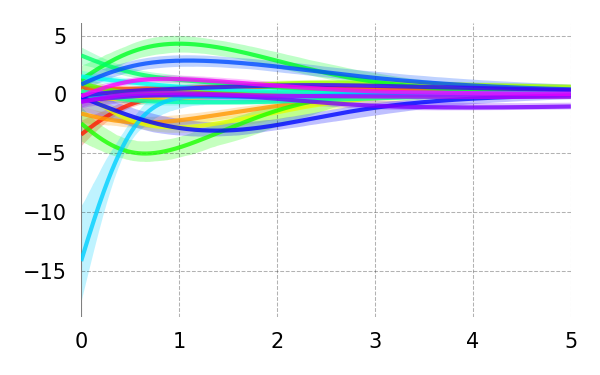}
    \end{subfigure}
    \begin{subfigure}[t]{0.49\textwidth}
        \centering
        \makebox[0.49\textwidth]{\centering Batch Size $\div$ 2 (512)}%
        \makebox[0.49\textwidth]{\centering Batch Size $\times$ 2 (2048)}
        \includegraphics[width=0.49\textwidth]{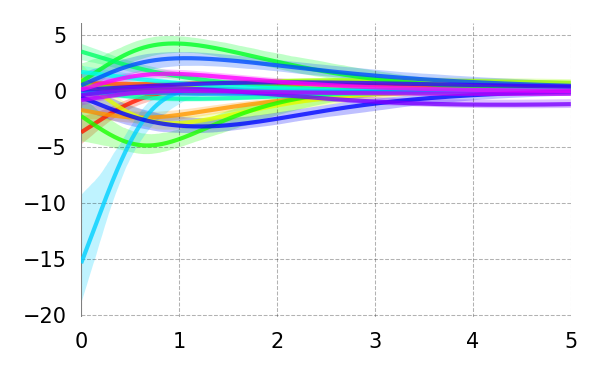}
        \includegraphics[width=0.49\textwidth]{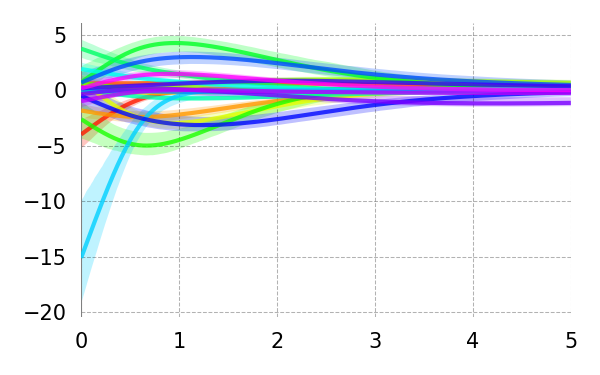}
    \end{subfigure}
    
    \vspace{1em}    
    
    \textbf{\large Consistency}
    
    \vspace{0.5em}
    
    \begin{subfigure}[t]{0.19\textwidth}
        \centering
        \makebox[0.9\textwidth]{\centering Rep 1}
        \includegraphics[width=\textwidth]{{results_cdrnn_journal_synth_time_async1_CDR_main_irf_univariate_y_mc}.png}
    \end{subfigure}
    \begin{subfigure}[t]{0.19\textwidth}
        \centering
        \makebox[0.9\textwidth]{\centering Rep 2}
        \includegraphics[width=\textwidth]{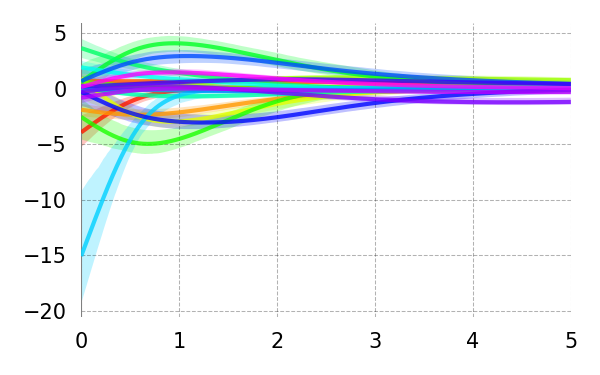}
    \end{subfigure}
    \begin{subfigure}[t]{0.19\textwidth}
        \centering
        \makebox[0.9\textwidth]{\centering Rep 3}
        \includegraphics[width=\textwidth]{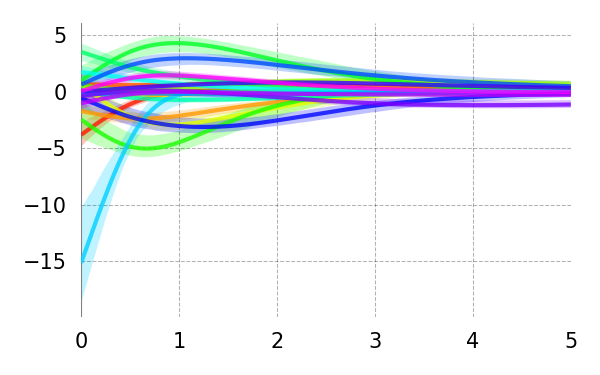}
    \end{subfigure}
    \begin{subfigure}[t]{0.19\textwidth}
        \centering
        \makebox[0.9\textwidth]{\centering Rep 4}
        \includegraphics[width=\textwidth]{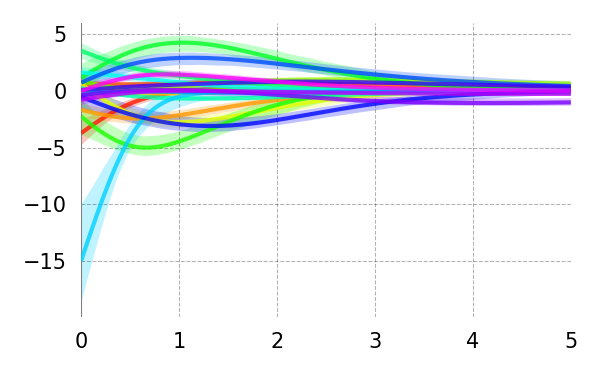}
    \end{subfigure}
    \begin{subfigure}[t]{0.19\textwidth}
        \centering
        \makebox[0.9\textwidth]{\centering Rep 5}
        \includegraphics[width=\textwidth]{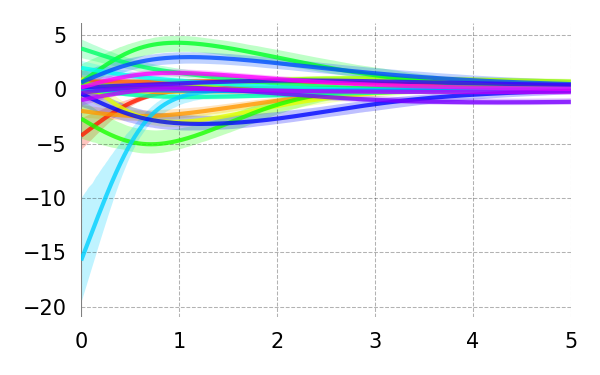}
    \end{subfigure}
    
    {\large Delay (s)}
    
    \vspace{0.5em}
    
    \caption{CDRNN estimated responses to synthetic data with a \textbf{asynchronously measured predictors and responses with variable intervals between them (mean interval 100ms)}. Estimates using base hyperparameters are compared to estimates from models that deviate from the base in some dimension. Plots under ``Consistency'' show estimates from five replicates of the ``base'' configuration, where ``Rep 1'' is the same model as ``base'' above, replotted for ease of comparison.}
    \label{fig:app-synth-time-async1}
    
\end{figure}

\begin{figure}

    \footnotesize
    \sffamily
    \centering
    
    \textbf{\Large Synth: Time, Random Asynchronous Long Interval (mean 500ms)}
    
    \vspace{1em}
    
    \begin{subfigure}[t]{0.49\textwidth}
        \centering
        \makebox[0.49\textwidth]{\centering \textbf{True}}
        
        \includegraphics[width=0.49\textwidth]{{results_cl_synth_time_async1_synthetic_true}.png}
    \end{subfigure}
    
    \begin{subfigure}[t]{0.49\textwidth}
        \centering
        \makebox[0.49\textwidth]{\centering Base}%
        \makebox[0.49\textwidth]{\centering + RNN}
        \begin{overpic}[width=0.49\textwidth]{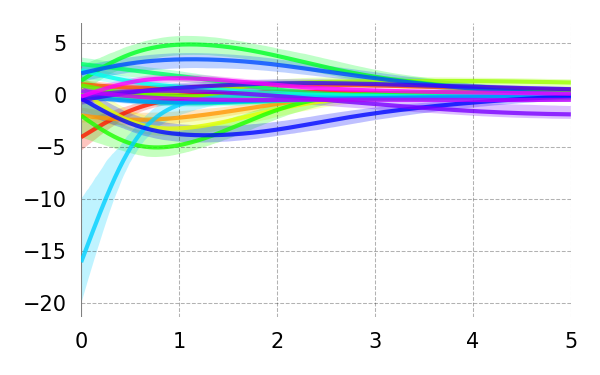}
        \end{overpic}%
        \includegraphics[width=0.49\textwidth]{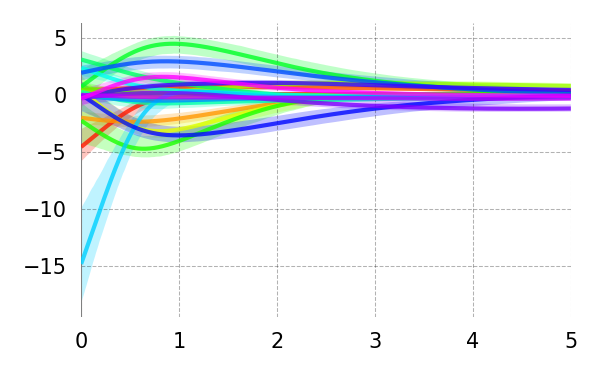}
    \end{subfigure}
    \begin{subfigure}[t]{0.49\textwidth}
        \centering
        \makebox[0.49\textwidth]{\centering Hidden Units $\div$ 2 (16)}%
        \makebox[0.49\textwidth]{\centering Hidden Units $\times$ 2 (64)}
        \includegraphics[width=0.49\textwidth]{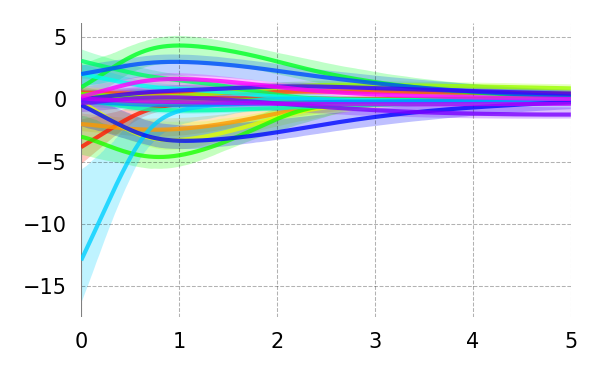}
        \includegraphics[width=0.49\textwidth]{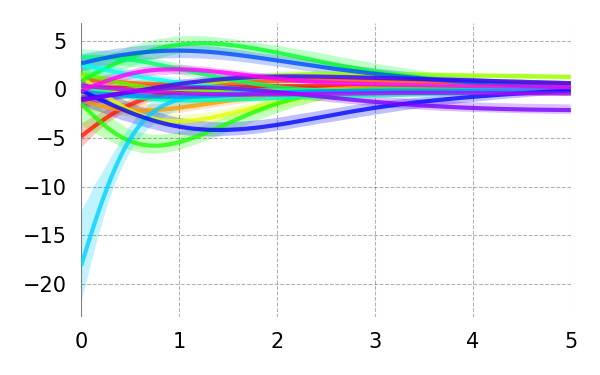}
    \end{subfigure}
    
    \begin{subfigure}[t]{0.49\textwidth}
        \centering
        \makebox[0.49\textwidth]{\centering Hidden Layers - 1 (1)}%
        \makebox[0.49\textwidth]{\centering Hidden Layers + 1 (3)}
        \includegraphics[width=0.49\textwidth]{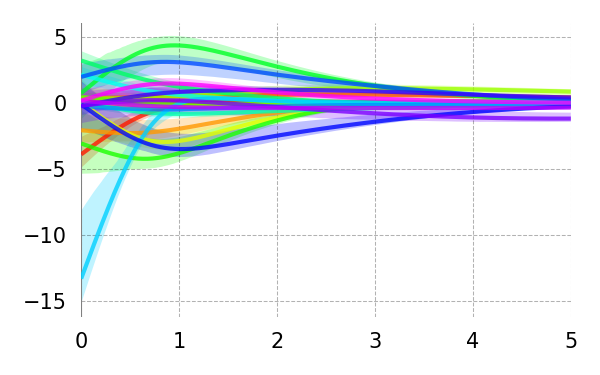}
        \includegraphics[width=0.49\textwidth]{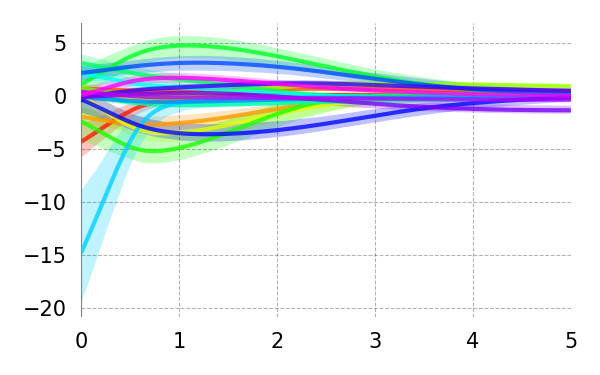}
    \end{subfigure}
    \begin{subfigure}[t]{0.49\textwidth}
        \centering
        \makebox[0.49\textwidth]{\centering Weight Reg $\div$ 5 (1)}%
        \makebox[0.49\textwidth]{\centering Weight Reg $\times$ 5 (5)}
        \includegraphics[width=0.49\textwidth]{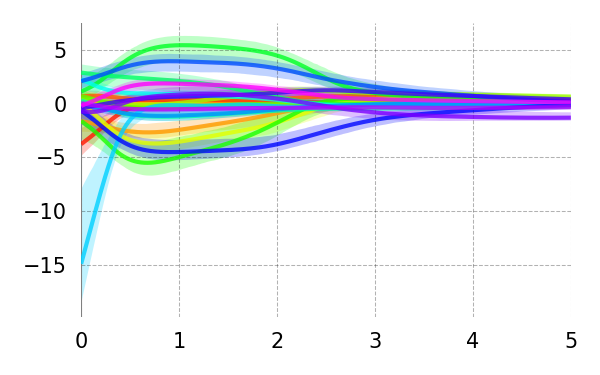}
        \includegraphics[width=0.49\textwidth]{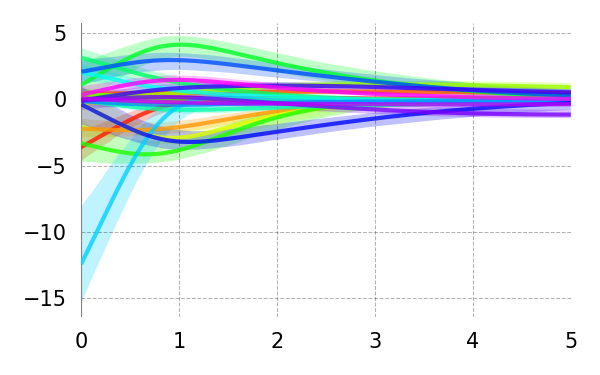}
    \end{subfigure}
    
    \begin{subfigure}[t]{0.49\textwidth}
        \centering
        \makebox[0.49\textwidth]{\centering Dropout $\div$ 2 (0.05)}%
        \makebox[0.49\textwidth]{\centering Dropout $\times$ 2 (0.2)}
        \includegraphics[width=0.49\textwidth]{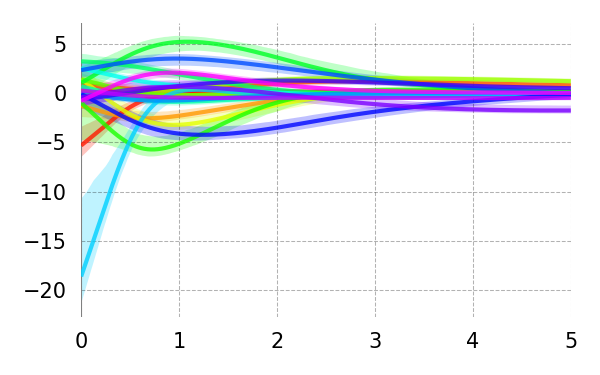}
        \includegraphics[width=0.49\textwidth]{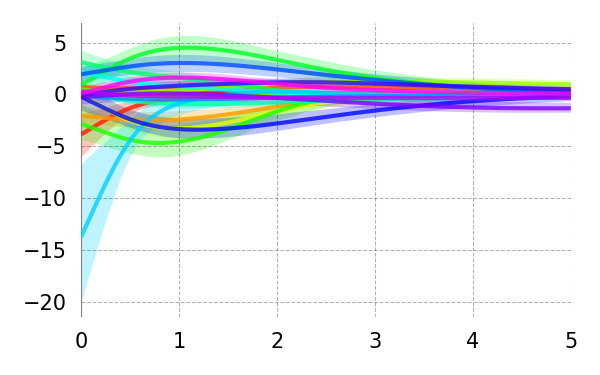}
    \end{subfigure}
    \begin{subfigure}[t]{0.49\textwidth}
        \centering
        \makebox[0.49\textwidth]{\centering Learning Rate $\div$ 3 (0.001)}%
        \makebox[0.49\textwidth]{\centering Learning Rate $\times$ 3 (0.009)}
        \includegraphics[width=0.49\textwidth]{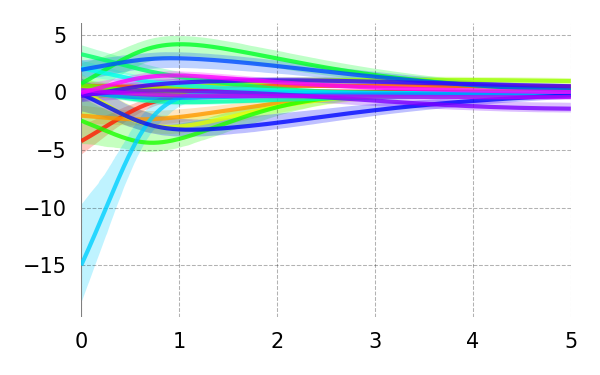}
        \includegraphics[width=0.49\textwidth]{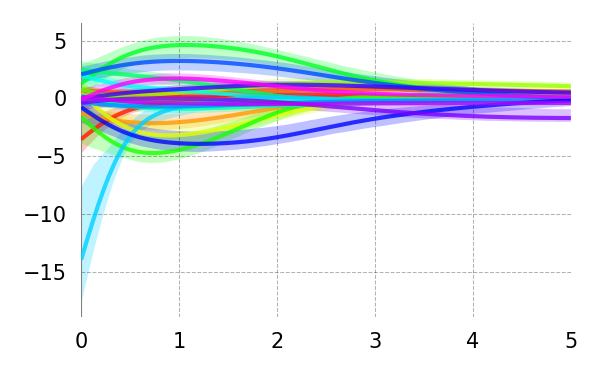}
    \end{subfigure}
    \begin{subfigure}[t]{0.49\textwidth}
        \centering
        \makebox[0.49\textwidth]{\centering Batch Size $\div$ 2 (512)}%
        \makebox[0.49\textwidth]{\centering Batch Size $\times$ 2 (2048)}
        \includegraphics[width=0.49\textwidth]{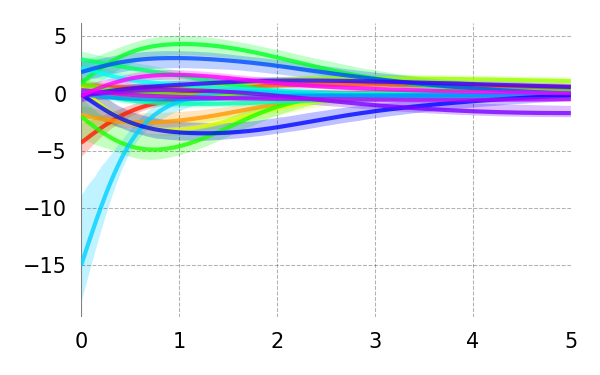}
        \includegraphics[width=0.49\textwidth]{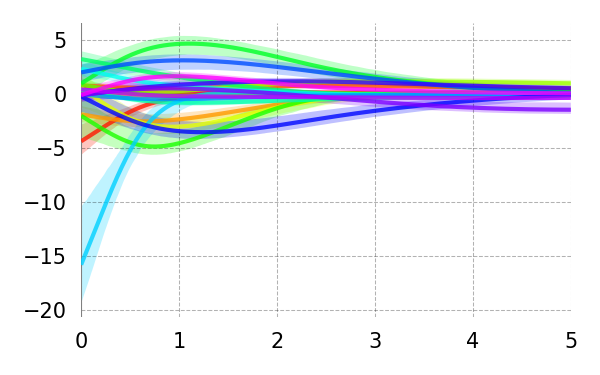}
    \end{subfigure}
    
    \vspace{1em}    
    
    \textbf{\large Consistency}
    
    \vspace{0.5em}
    
    \begin{subfigure}[t]{0.19\textwidth}
        \centering
        \makebox[0.9\textwidth]{\centering Rep 1}
        \includegraphics[width=\textwidth]{{results_cdrnn_journal_synth_time_async5_CDR_main_irf_univariate_y_mc}.png}
    \end{subfigure}
    \begin{subfigure}[t]{0.19\textwidth}
        \centering
        \makebox[0.9\textwidth]{\centering Rep 2}
        \includegraphics[width=\textwidth]{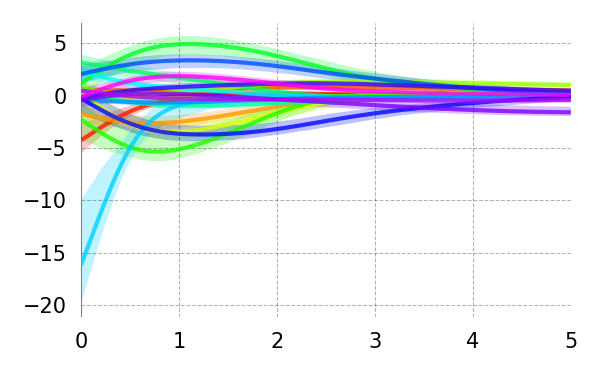}
    \end{subfigure}
    \begin{subfigure}[t]{0.19\textwidth}
        \centering
        \makebox[0.9\textwidth]{\centering Rep 3}
        \includegraphics[width=\textwidth]{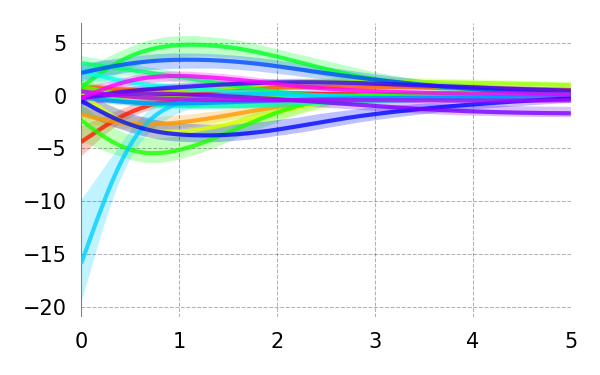}
    \end{subfigure}
    \begin{subfigure}[t]{0.19\textwidth}
        \centering
        \makebox[0.9\textwidth]{\centering Rep 4}
        \includegraphics[width=\textwidth]{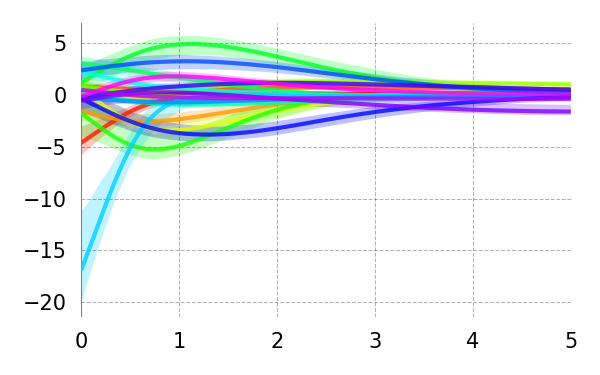}
    \end{subfigure}
    \begin{subfigure}[t]{0.19\textwidth}
        \centering
        \makebox[0.9\textwidth]{\centering Rep 5}
        \includegraphics[width=\textwidth]{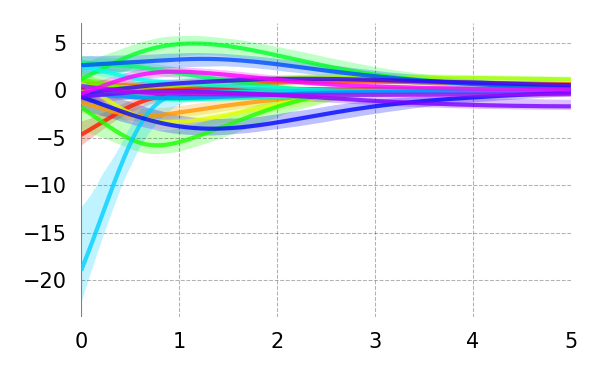}
    \end{subfigure}
    
    {\large Delay (s)}
    
    \vspace{0.5em}
    
    \caption{CDRNN estimated responses to synthetic data with a \textbf{asynchronously measured predictors and responses with variable intervals between them (mean interval 500ms)}. Estimates using base hyperparameters are compared to estimates from models that deviate from the base in some dimension. Plots under ``Consistency'' show estimates from five replicates of the ``base'' configuration, where ``Rep 1'' is the same model as ``base'' above, replotted for ease of comparison.}
    \label{fig:app-synth-time-async5}
    
\end{figure}

\begin{figure}

    \footnotesize
    \sffamily
    \centering
    
    \textbf{\Large Synth: Multicollinearity, $r=0$}
    
    \vspace{1em}
    
    \begin{subfigure}[t]{0.49\textwidth}
        \centering
        \makebox[0.49\textwidth]{\centering \textbf{True}}
        
        \includegraphics[width=0.49\textwidth]{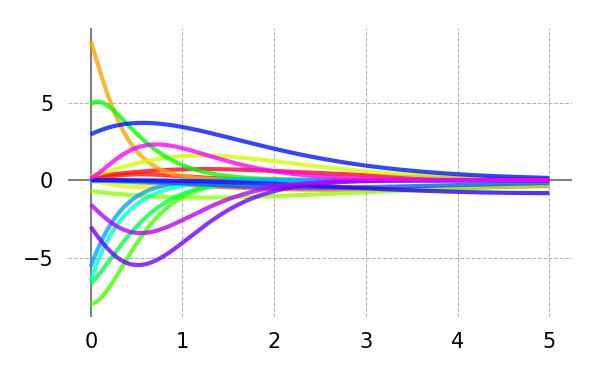}
    \end{subfigure}
    
    \begin{subfigure}[t]{0.49\textwidth}
        \centering
        \makebox[0.49\textwidth]{\centering Base}%
        \makebox[0.49\textwidth]{\centering + RNN}
        \begin{overpic}[width=0.49\textwidth]{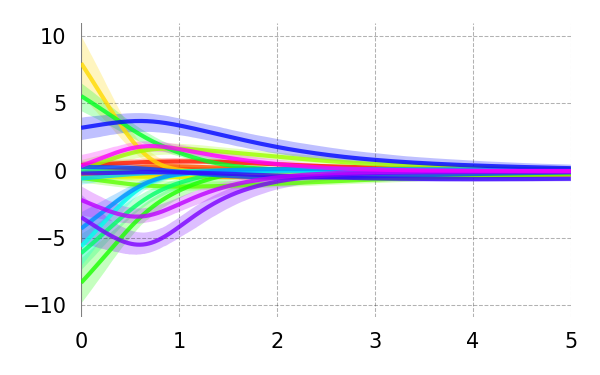}
        \end{overpic}%
        \includegraphics[width=0.49\textwidth]{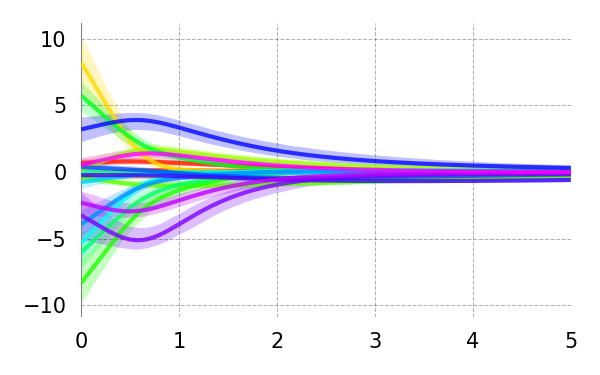}
    \end{subfigure}
    \begin{subfigure}[t]{0.49\textwidth}
        \centering
        \makebox[0.49\textwidth]{\centering Hidden Units $\div$ 2 (16)}%
        \makebox[0.49\textwidth]{\centering Hidden Units $\times$ 2 (64)}
        \includegraphics[width=0.49\textwidth]{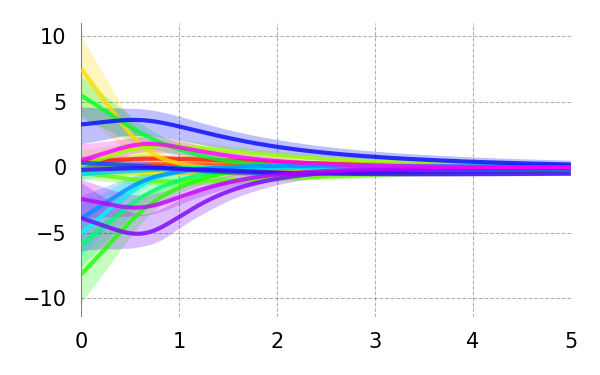}
        \includegraphics[width=0.49\textwidth]{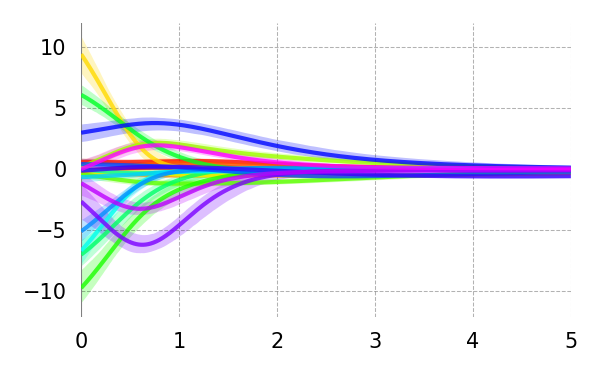}
    \end{subfigure}
    
    \begin{subfigure}[t]{0.49\textwidth}
        \centering
        \makebox[0.49\textwidth]{\centering Hidden Layers - 1 (1)}%
        \makebox[0.49\textwidth]{\centering Hidden Layers + 1 (3)}
        \includegraphics[width=0.49\textwidth]{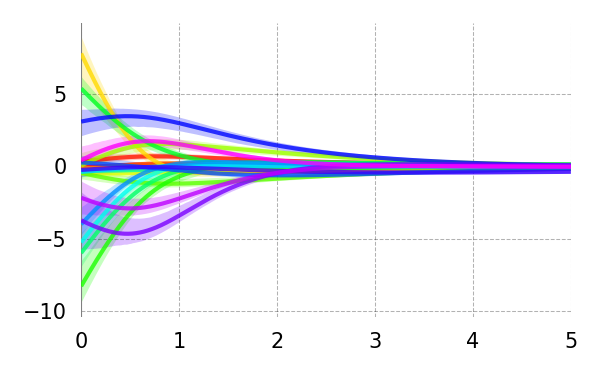}
        \includegraphics[width=0.49\textwidth]{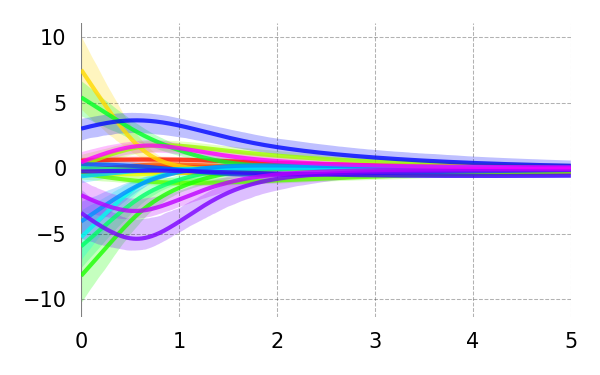}
    \end{subfigure}
    \begin{subfigure}[t]{0.49\textwidth}
        \centering
        \makebox[0.49\textwidth]{\centering Weight Reg $\div$ 5 (1)}%
        \makebox[0.49\textwidth]{\centering Weight Reg $\times$ 5 (5)}
        \includegraphics[width=0.49\textwidth]{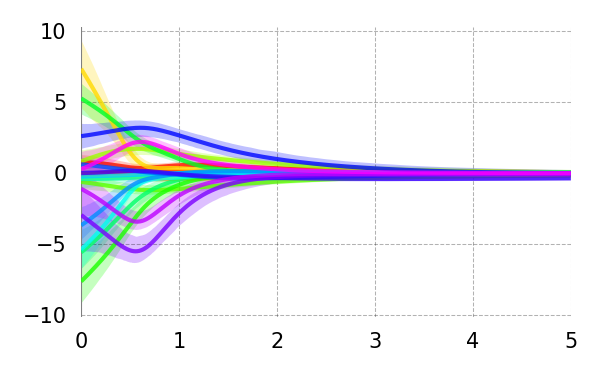}
        \includegraphics[width=0.49\textwidth]{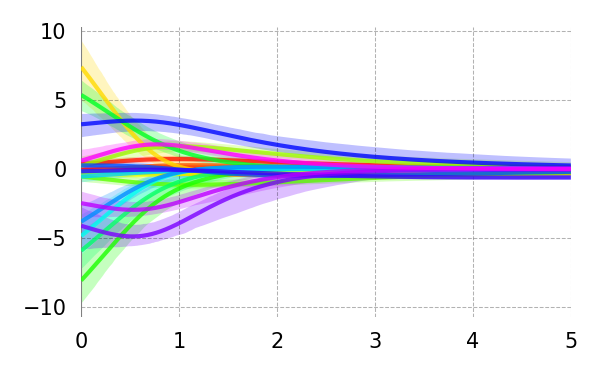}
    \end{subfigure}
    
    \begin{subfigure}[t]{0.49\textwidth}
        \centering
        \makebox[0.49\textwidth]{\centering Dropout $\div$ 2 (0.05)}%
        \makebox[0.49\textwidth]{\centering Dropout $\times$ 2 (0.2)}
        \includegraphics[width=0.49\textwidth]{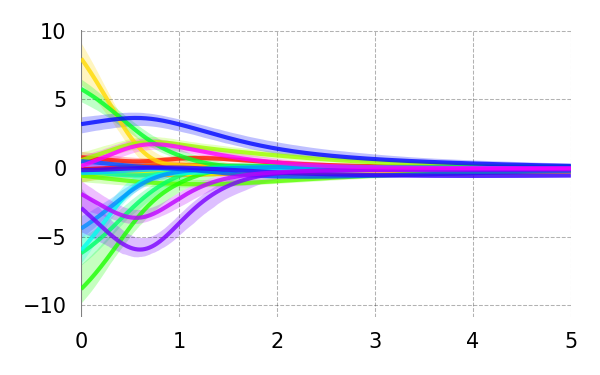}
        \includegraphics[width=0.49\textwidth]{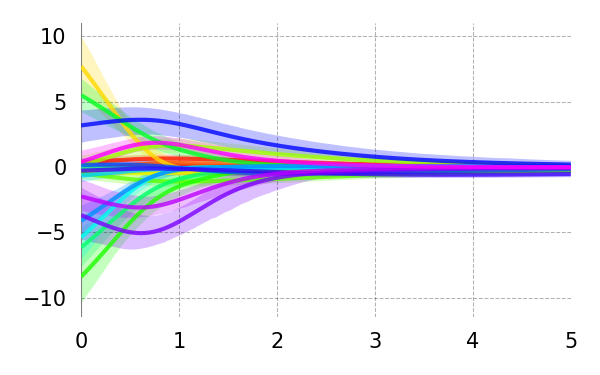}
    \end{subfigure}
    \begin{subfigure}[t]{0.49\textwidth}
        \centering
        \makebox[0.49\textwidth]{\centering Learning Rate $\div$ 3 (0.001)}%
        \makebox[0.49\textwidth]{\centering Learning Rate $\times$ 3 (0.009)}
        \includegraphics[width=0.49\textwidth]{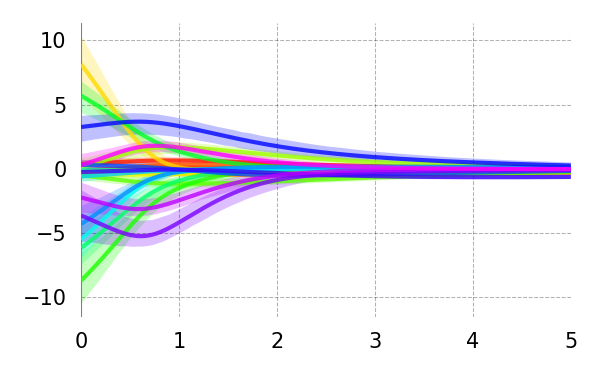}
        \includegraphics[width=0.49\textwidth]{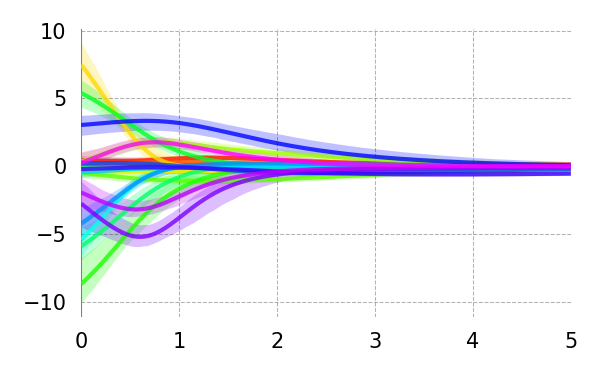}
    \end{subfigure}
    \begin{subfigure}[t]{0.49\textwidth}
        \centering
        \makebox[0.49\textwidth]{\centering Batch Size $\div$ 2 (512)}%
        \makebox[0.49\textwidth]{\centering Batch Size $\times$ 2 (2048)}
        \includegraphics[width=0.49\textwidth]{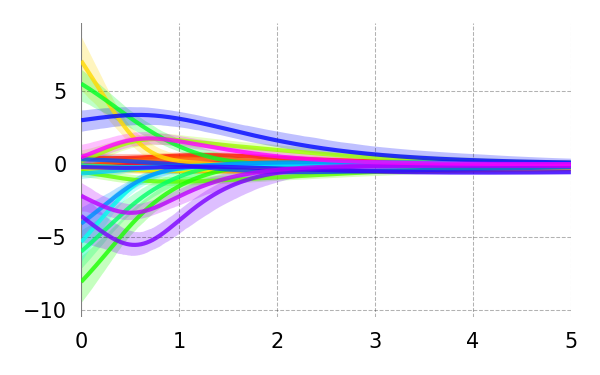}
        \includegraphics[width=0.49\textwidth]{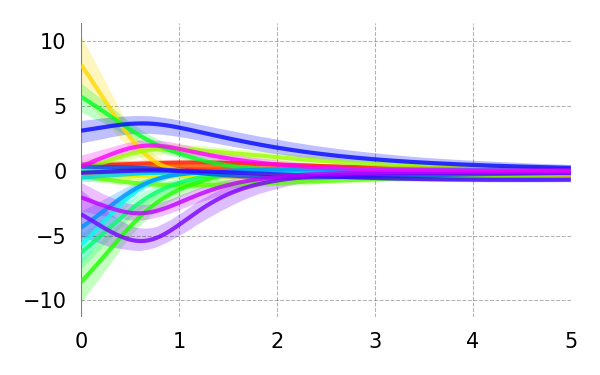}
    \end{subfigure}
    
    \vspace{1em}    
    
    \textbf{\large Consistency}
    
    \vspace{0.5em}
    
    \begin{subfigure}[t]{0.19\textwidth}
        \centering
        \makebox[0.9\textwidth]{\centering Rep 1}
        \includegraphics[width=\textwidth]{{results_cdrnn_journal_synth_multicollinearity_r0.00_CDR_main_irf_univariate_y_mc}.png}
    \end{subfigure}
    \begin{subfigure}[t]{0.19\textwidth}
        \centering
        \makebox[0.9\textwidth]{\centering Rep 2}
        \includegraphics[width=\textwidth]{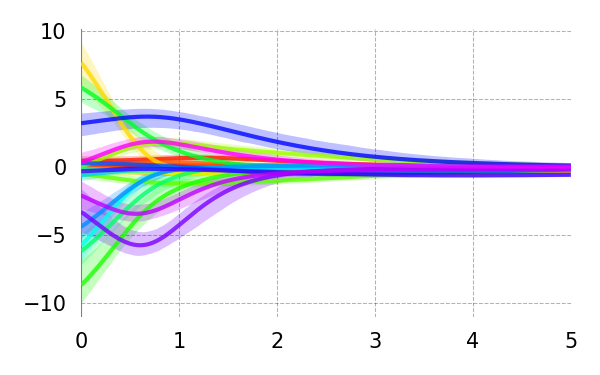}
    \end{subfigure}
    \begin{subfigure}[t]{0.19\textwidth}
        \centering
        \makebox[0.9\textwidth]{\centering Rep 3}
        \includegraphics[width=\textwidth]{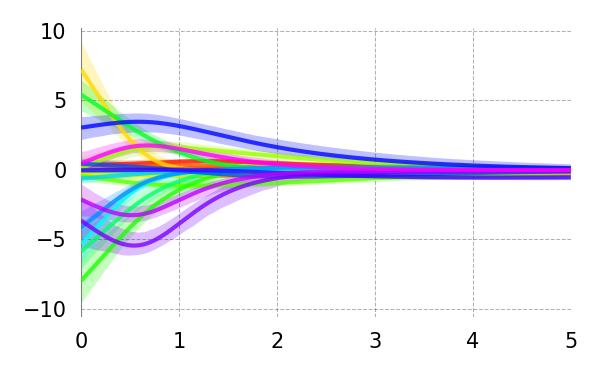}
    \end{subfigure}
    \begin{subfigure}[t]{0.19\textwidth}
        \centering
        \makebox[0.9\textwidth]{\centering Rep 4}
        \includegraphics[width=\textwidth]{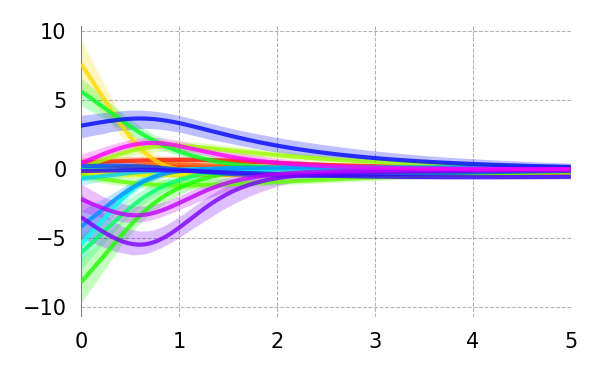}
    \end{subfigure}
    \begin{subfigure}[t]{0.19\textwidth}
        \centering
        \makebox[0.9\textwidth]{\centering Rep 5}
        \includegraphics[width=\textwidth]{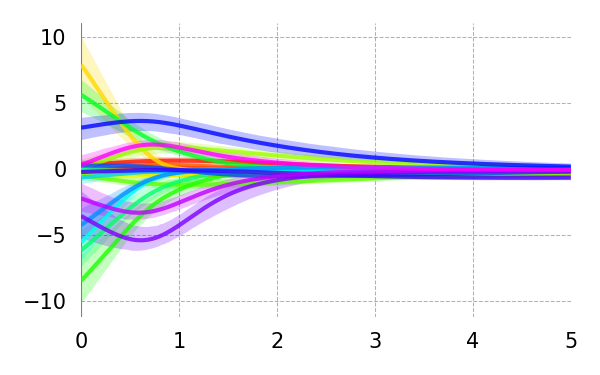}
    \end{subfigure}
    
    {\large Delay (s)}
    
    \vspace{0.5em}
    
    \caption{CDRNN estimated responses to synthetic data with \textbf{pairwise predictor multicollinearity of $r = 0$}. Estimates using base hyperparameters are compared to estimates from models that deviate from the base in some dimension. Plots under ``Consistency'' show estimates from five replicates of the ``base'' configuration, where ``Rep 1'' is the same model as ``base'' above, replotted for ease of comparison.}
    \label{fig:app-synth-multicollinearity-r00}
    
\end{figure}

\begin{figure}

    \footnotesize
    \sffamily
    \centering
    
    \textbf{\Large Synth: Multicollinearity, $r=0.25$}
    
    \vspace{1em}
    
    \begin{subfigure}[t]{0.49\textwidth}
        \centering
        \makebox[0.49\textwidth]{\centering \textbf{True}}
        
        \includegraphics[width=0.49\textwidth]{{results_cl_synth_multicollinearity_r0.00_synthetic_true}.png}
    \end{subfigure}
    
    \begin{subfigure}[t]{0.49\textwidth}
        \centering
        \makebox[0.49\textwidth]{\centering Base}%
        \makebox[0.49\textwidth]{\centering + RNN}
        \begin{overpic}[width=0.49\textwidth]{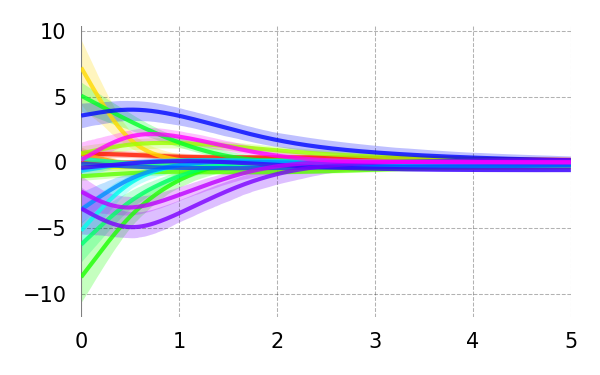}
        \end{overpic}%
        \includegraphics[width=0.49\textwidth]{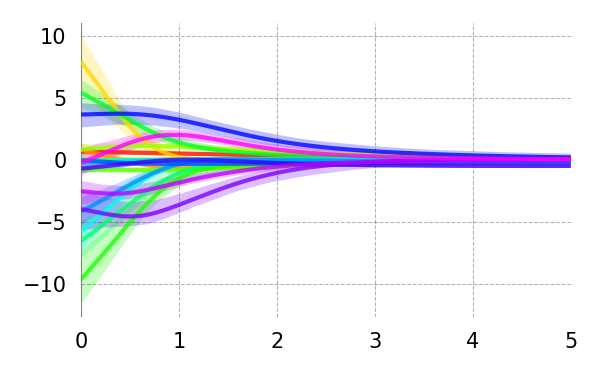}
    \end{subfigure}
    \begin{subfigure}[t]{0.49\textwidth}
        \centering
        \makebox[0.49\textwidth]{\centering Hidden Units $\div$ 2 (16)}%
        \makebox[0.49\textwidth]{\centering Hidden Units $\times$ 2 (64)}
        \includegraphics[width=0.49\textwidth]{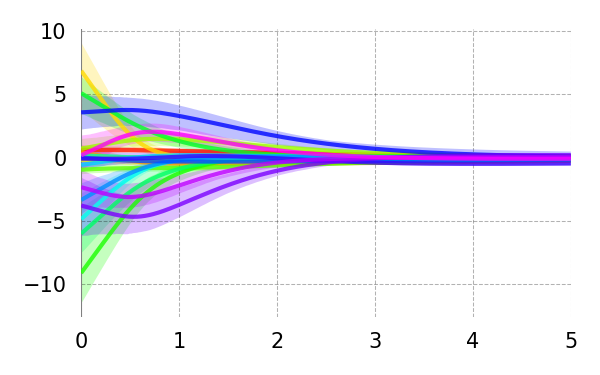}
        \includegraphics[width=0.49\textwidth]{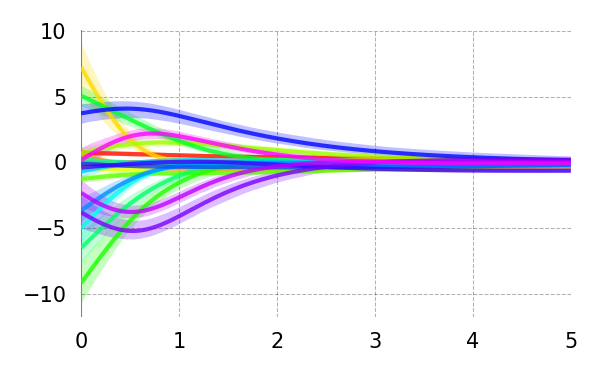}
    \end{subfigure}
    
    \begin{subfigure}[t]{0.49\textwidth}
        \centering
        \makebox[0.49\textwidth]{\centering Hidden Layers - 1 (1)}%
        \makebox[0.49\textwidth]{\centering Hidden Layers + 1 (3)}
        \includegraphics[width=0.49\textwidth]{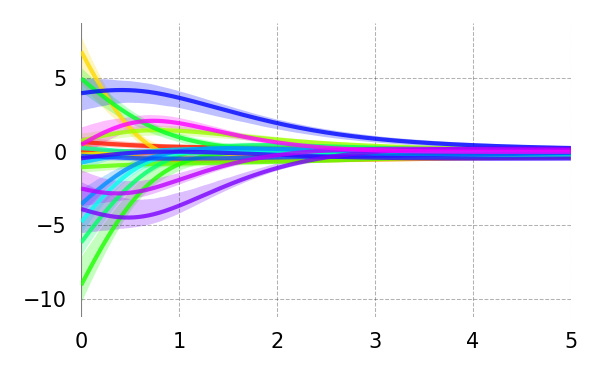}
        \includegraphics[width=0.49\textwidth]{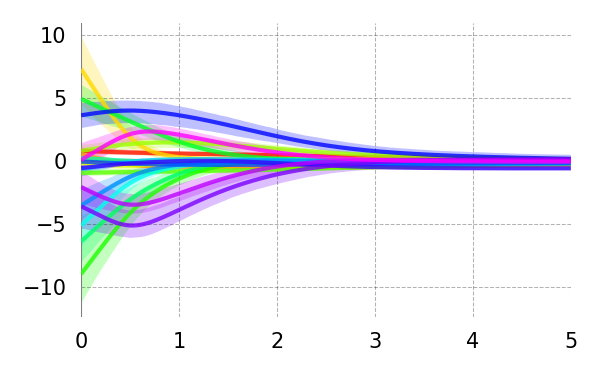}
    \end{subfigure}
    \begin{subfigure}[t]{0.49\textwidth}
        \centering
        \makebox[0.49\textwidth]{\centering Weight Reg $\div$ 5 (1)}%
        \makebox[0.49\textwidth]{\centering Weight Reg $\times$ 5 (5)}
        \includegraphics[width=0.49\textwidth]{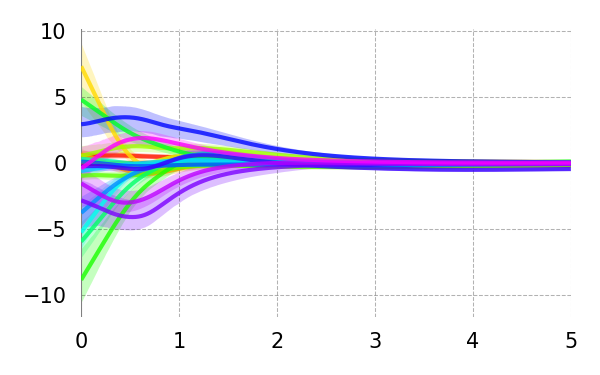}
        \includegraphics[width=0.49\textwidth]{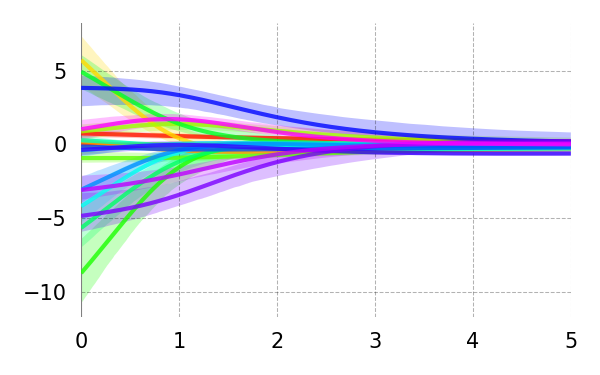}
    \end{subfigure}
    
    \begin{subfigure}[t]{0.49\textwidth}
        \centering
        \makebox[0.49\textwidth]{\centering Dropout $\div$ 2 (0.05)}%
        \makebox[0.49\textwidth]{\centering Dropout $\times$ 2 (0.2)}
        \includegraphics[width=0.49\textwidth]{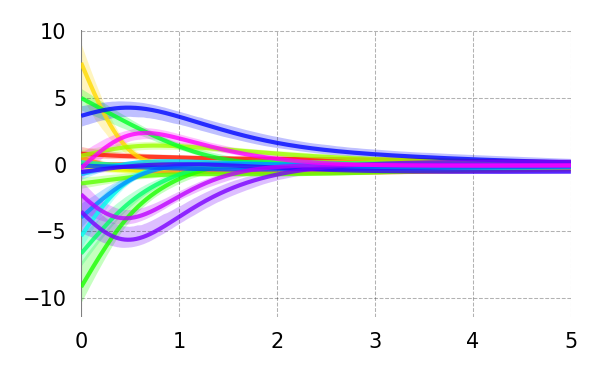}
        \includegraphics[width=0.49\textwidth]{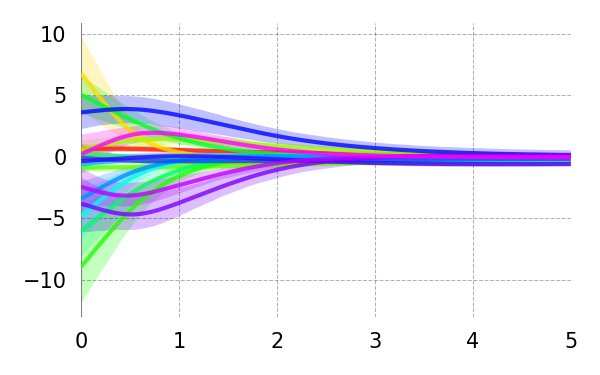}
    \end{subfigure}
    \begin{subfigure}[t]{0.49\textwidth}
        \centering
        \makebox[0.49\textwidth]{\centering Learning Rate $\div$ 3 (0.001)}%
        \makebox[0.49\textwidth]{\centering Learning Rate $\times$ 3 (0.009)}
        \includegraphics[width=0.49\textwidth]{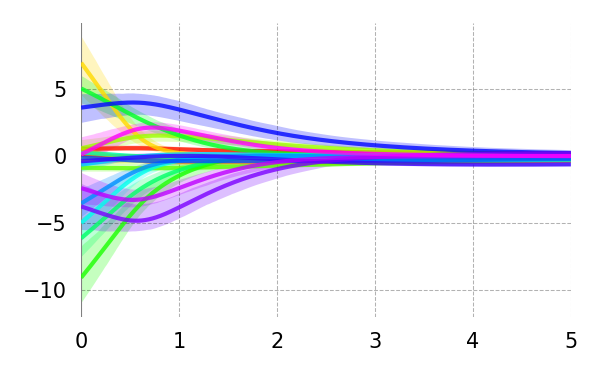}
        \includegraphics[width=0.49\textwidth]{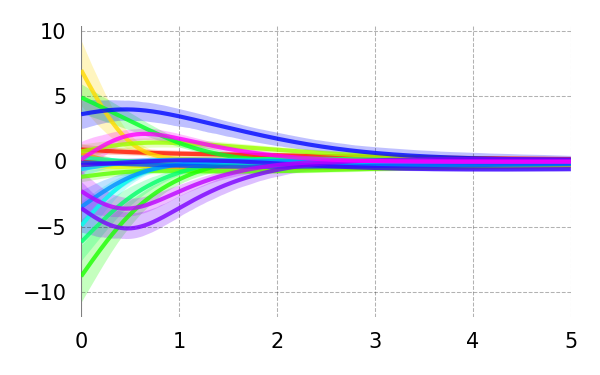}
    \end{subfigure}
    \begin{subfigure}[t]{0.49\textwidth}
        \centering
        \makebox[0.49\textwidth]{\centering Batch Size $\div$ 2 (512)}%
        \makebox[0.49\textwidth]{\centering Batch Size $\times$ 2 (2048)}
        \includegraphics[width=0.49\textwidth]{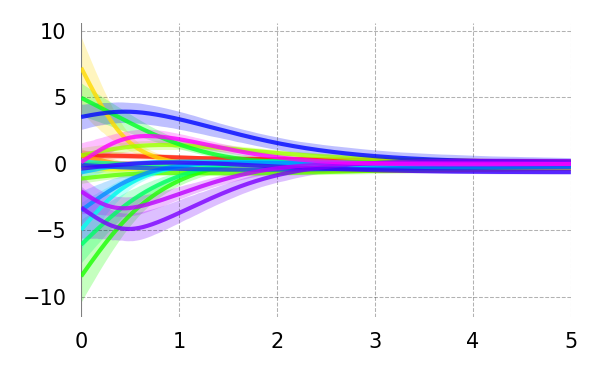}
        \includegraphics[width=0.49\textwidth]{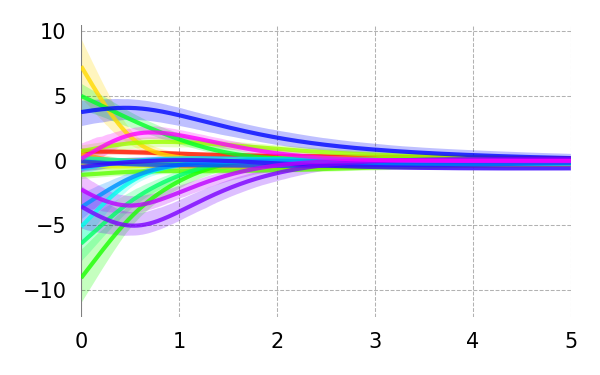}
    \end{subfigure}
    
    \vspace{1em}    
    
    \textbf{\large Consistency}
    
    \vspace{0.5em}
    
    \begin{subfigure}[t]{0.19\textwidth}
        \centering
        \makebox[0.9\textwidth]{\centering Rep 1}
        \includegraphics[width=\textwidth]{{results_cdrnn_journal_synth_multicollinearity_r0.25_CDR_main_irf_univariate_y_mc}.png}
    \end{subfigure}
    \begin{subfigure}[t]{0.19\textwidth}
        \centering
        \makebox[0.9\textwidth]{\centering Rep 2}
        \includegraphics[width=\textwidth]{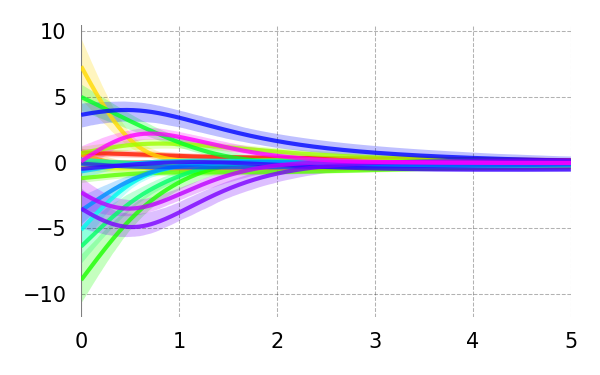}
    \end{subfigure}
    \begin{subfigure}[t]{0.19\textwidth}
        \centering
        \makebox[0.9\textwidth]{\centering Rep 3}
        \includegraphics[width=\textwidth]{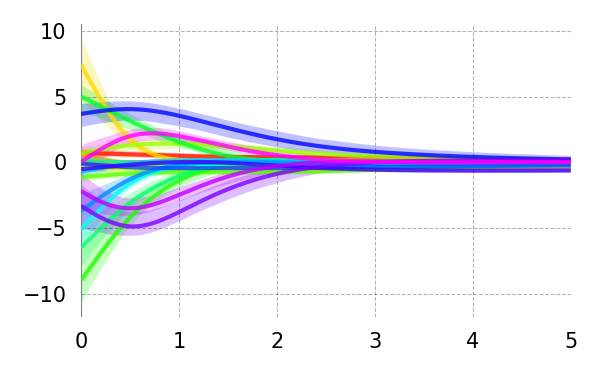}
    \end{subfigure}
    \begin{subfigure}[t]{0.19\textwidth}
        \centering
        \makebox[0.9\textwidth]{\centering Rep 4}
        \includegraphics[width=\textwidth]{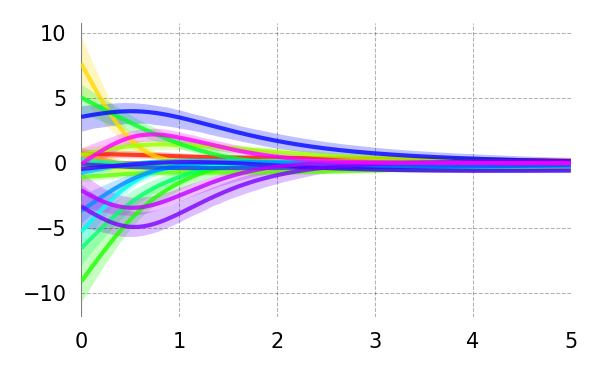}
    \end{subfigure}
    \begin{subfigure}[t]{0.19\textwidth}
        \centering
        \makebox[0.9\textwidth]{\centering Rep 5}
        \includegraphics[width=\textwidth]{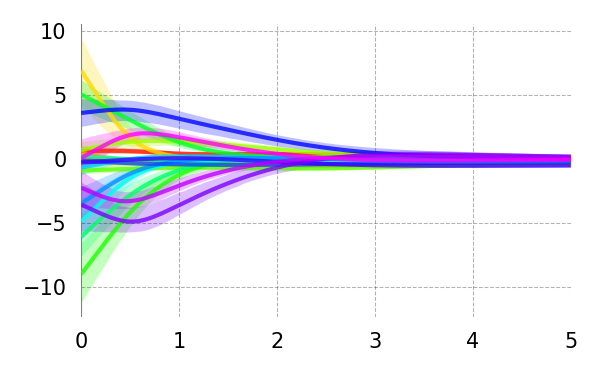}
    \end{subfigure}
    
    {\large Delay (s)}
    
    \vspace{0.5em}
    
    \caption{CDRNN estimated responses to synthetic data with \textbf{pairwise predictor multicollinearity of $r = 0.25$}. Estimates using base hyperparameters are compared to estimates from models that deviate from the base in some dimension. Plots under ``Consistency'' show estimates from five replicates of the ``base'' configuration, where ``Rep 1'' is the same model as ``base'' above, replotted for ease of comparison.}
    \label{fig:app-synth-multicollinearity-r25}
    
\end{figure}

\begin{figure}

    \footnotesize
    \sffamily
    \centering
    
    \textbf{\Large Synth: Multicollinearity, $r=0.5$}
    
    \vspace{1em}
    
    \begin{subfigure}[t]{0.49\textwidth}
        \centering
        \makebox[0.49\textwidth]{\centering \textbf{True}}
        
        \includegraphics[width=0.49\textwidth]{{results_cl_synth_multicollinearity_r0.00_synthetic_true}.png}
    \end{subfigure}
    
    \begin{subfigure}[t]{0.49\textwidth}
        \centering
        \makebox[0.49\textwidth]{\centering Base}%
        \makebox[0.49\textwidth]{\centering + RNN}
        \begin{overpic}[width=0.49\textwidth]{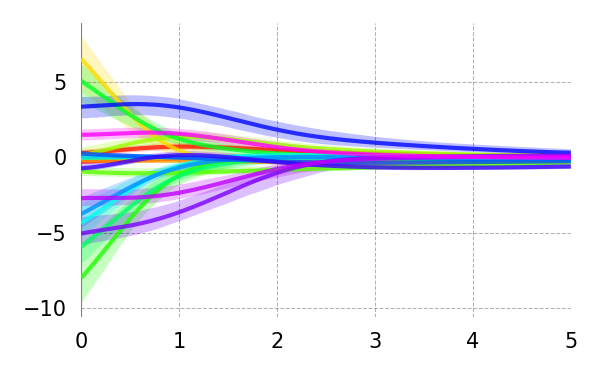}
        \end{overpic}%
        \includegraphics[width=0.49\textwidth]{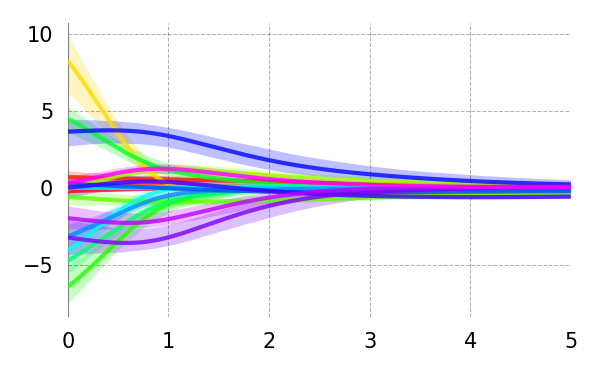}
    \end{subfigure}
    \begin{subfigure}[t]{0.49\textwidth}
        \centering
        \makebox[0.49\textwidth]{\centering Hidden Units $\div$ 2 (16)}%
        \makebox[0.49\textwidth]{\centering Hidden Units $\times$ 2 (64)}
        \includegraphics[width=0.49\textwidth]{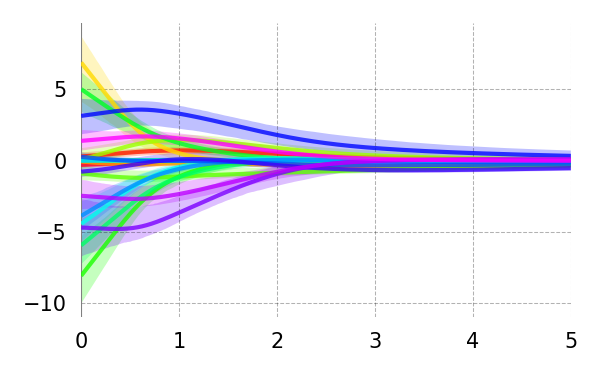}
        \includegraphics[width=0.49\textwidth]{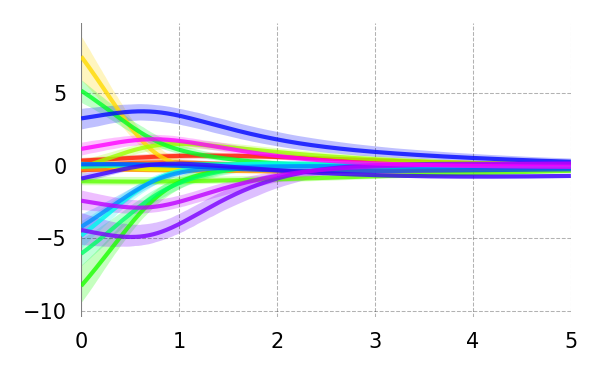}
    \end{subfigure}
    
    \begin{subfigure}[t]{0.49\textwidth}
        \centering
        \makebox[0.49\textwidth]{\centering Hidden Layers - 1 (1)}%
        \makebox[0.49\textwidth]{\centering Hidden Layers + 1 (3)}
        \includegraphics[width=0.49\textwidth]{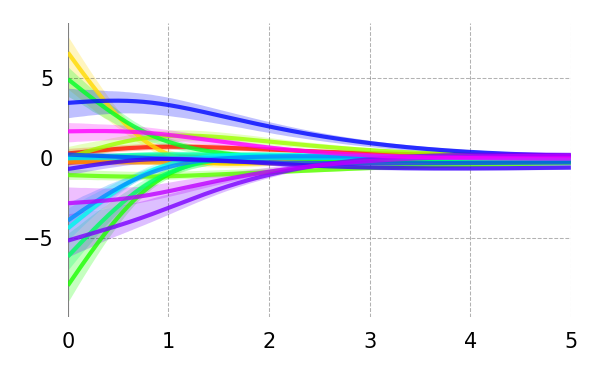}
        \includegraphics[width=0.49\textwidth]{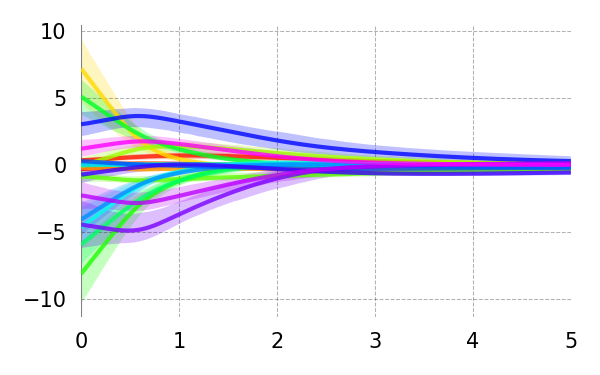}
    \end{subfigure}
    \begin{subfigure}[t]{0.49\textwidth}
        \centering
        \makebox[0.49\textwidth]{\centering Weight Reg $\div$ 5 (1)}%
        \makebox[0.49\textwidth]{\centering Weight Reg $\times$ 5 (5)}
        \includegraphics[width=0.49\textwidth]{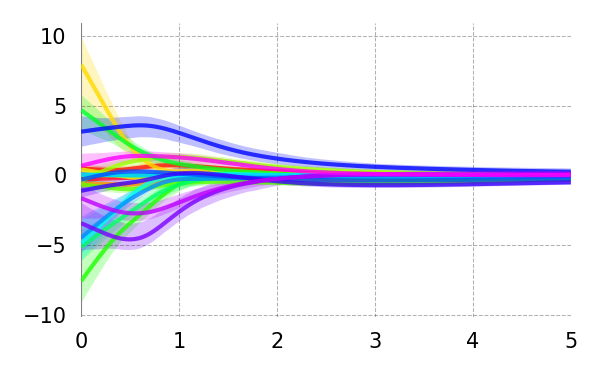}
        \includegraphics[width=0.49\textwidth]{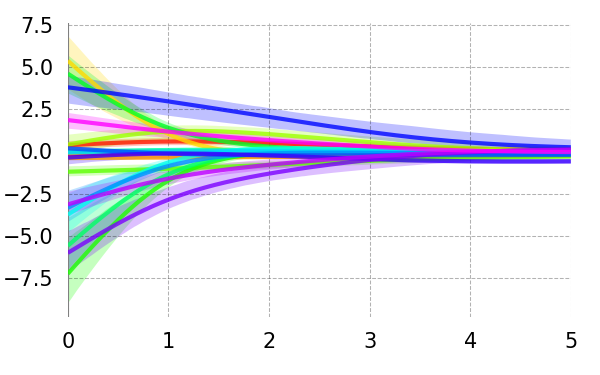}
    \end{subfigure}
    
    \begin{subfigure}[t]{0.49\textwidth}
        \centering
        \makebox[0.49\textwidth]{\centering Dropout $\div$ 2 (0.05)}%
        \makebox[0.49\textwidth]{\centering Dropout $\times$ 2 (0.2)}
        \includegraphics[width=0.49\textwidth]{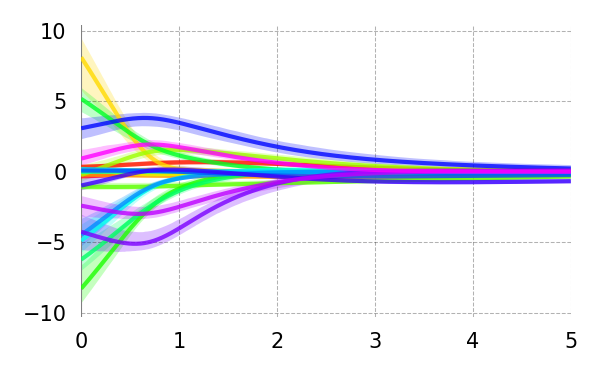}
        \includegraphics[width=0.49\textwidth]{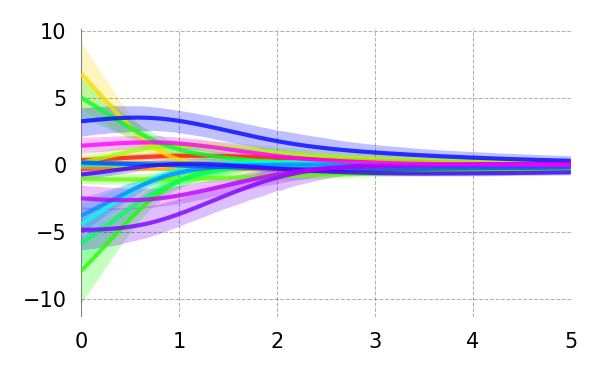}
    \end{subfigure}
    \begin{subfigure}[t]{0.49\textwidth}
        \centering
        \makebox[0.49\textwidth]{\centering Learning Rate $\div$ 3 (0.001)}%
        \makebox[0.49\textwidth]{\centering Learning Rate $\times$ 3 (0.009)}
        \includegraphics[width=0.49\textwidth]{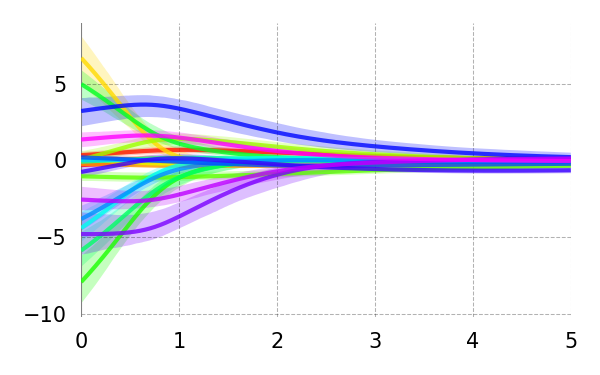}
        \includegraphics[width=0.49\textwidth]{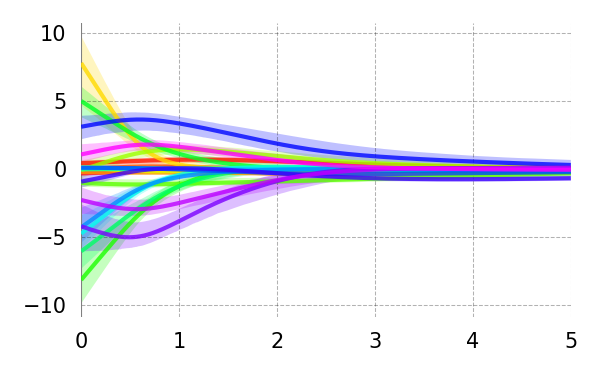}
    \end{subfigure}
    \begin{subfigure}[t]{0.49\textwidth}
        \centering
        \makebox[0.49\textwidth]{\centering Batch Size $\div$ 2 (512)}%
        \makebox[0.49\textwidth]{\centering Batch Size $\times$ 2 (2048)}
        \includegraphics[width=0.49\textwidth]{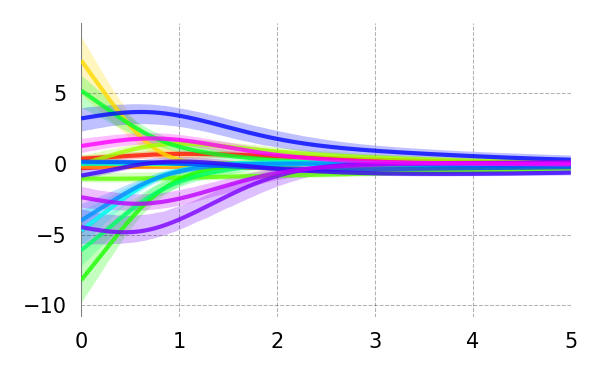}
        \includegraphics[width=0.49\textwidth]{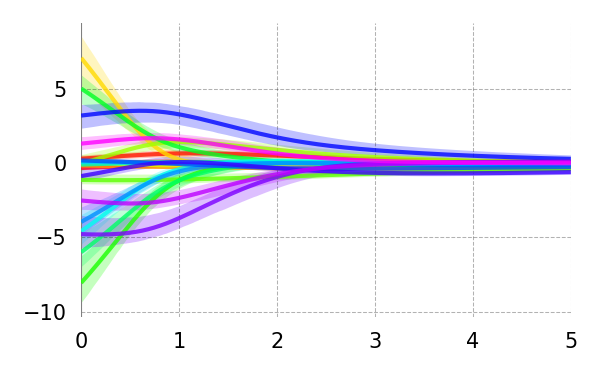}
    \end{subfigure}
    
    \vspace{1em}    
    
    \textbf{\large Consistency}
    
    \vspace{0.5em}
    
    \begin{subfigure}[t]{0.19\textwidth}
        \centering
        \makebox[0.9\textwidth]{\centering Rep 1}
        \includegraphics[width=\textwidth]{{results_cdrnn_journal_synth_multicollinearity_r0.50_CDR_main_irf_univariate_y_mc}.png}
    \end{subfigure}
    \begin{subfigure}[t]{0.19\textwidth}
        \centering
        \makebox[0.9\textwidth]{\centering Rep 2}
        \includegraphics[width=\textwidth]{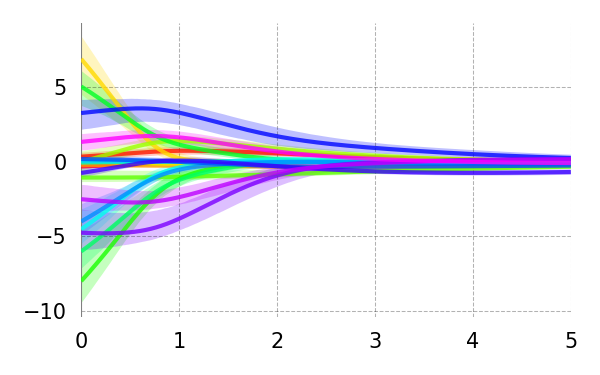}
    \end{subfigure}
    \begin{subfigure}[t]{0.19\textwidth}
        \centering
        \makebox[0.9\textwidth]{\centering Rep 3}
        \includegraphics[width=\textwidth]{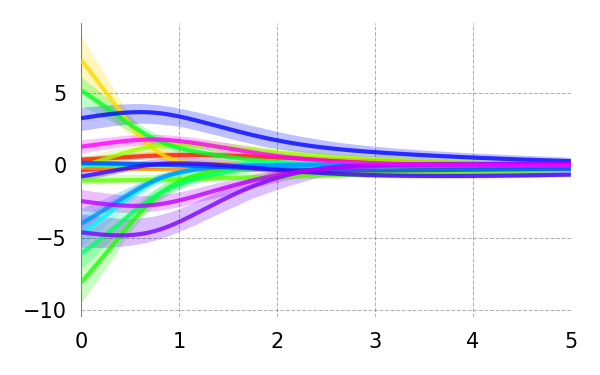}
    \end{subfigure}
    \begin{subfigure}[t]{0.19\textwidth}
        \centering
        \makebox[0.9\textwidth]{\centering Rep 4}
        \includegraphics[width=\textwidth]{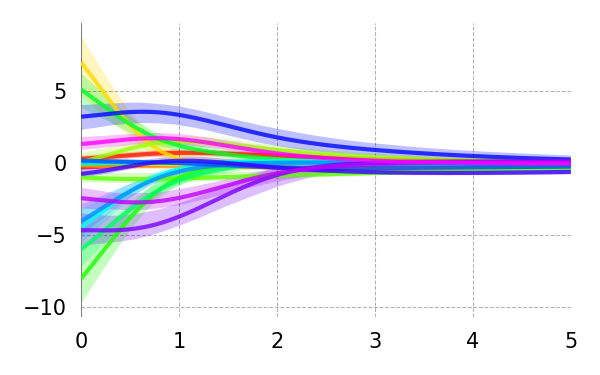}
    \end{subfigure}
    \begin{subfigure}[t]{0.19\textwidth}
        \centering
        \makebox[0.9\textwidth]{\centering Rep 5}
        \includegraphics[width=\textwidth]{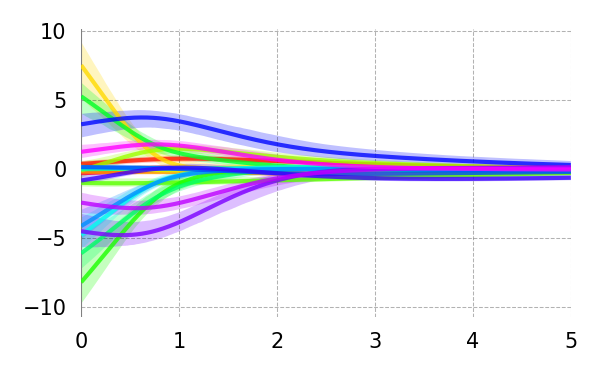}
    \end{subfigure}
    
    {\large Delay (s)}
    
    \vspace{0.5em}
    
    \caption{CDRNN estimated responses to synthetic data with \textbf{pairwise predictor multicollinearity of $r = 0.5$}. Estimates using base hyperparameters are compared to estimates from models that deviate from the base in some dimension. Plots under ``Consistency'' show estimates from five replicates of the ``base'' configuration, where ``Rep 1'' is the same model as ``base'' above, replotted for ease of comparison.}
    \label{fig:app-synth-multicollinearity-r50}
    
\end{figure}

\begin{figure}

    \footnotesize
    \sffamily
    \centering
    
    \textbf{\Large Synth: Multicollinearity, $r=0.75$}
    
    \vspace{1em}
    
    \begin{subfigure}[t]{0.49\textwidth}
        \centering
        \makebox[0.49\textwidth]{\centering \textbf{True}}
        
        \includegraphics[width=0.49\textwidth]{{results_cl_synth_multicollinearity_r0.00_synthetic_true}.png}
    \end{subfigure}
    
    \begin{subfigure}[t]{0.49\textwidth}
        \centering
        \makebox[0.49\textwidth]{\centering Base}%
        \makebox[0.49\textwidth]{\centering + RNN}
        \begin{overpic}[width=0.49\textwidth]{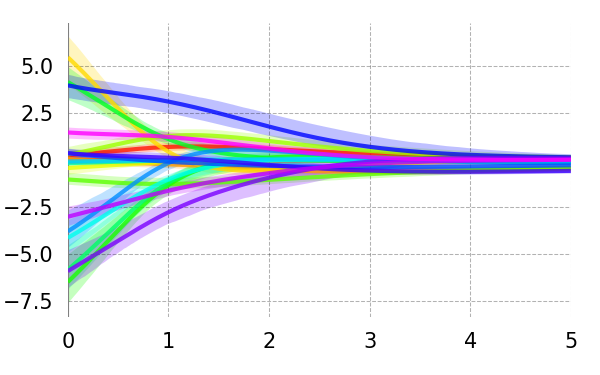}
        \end{overpic}%
        \includegraphics[width=0.49\textwidth]{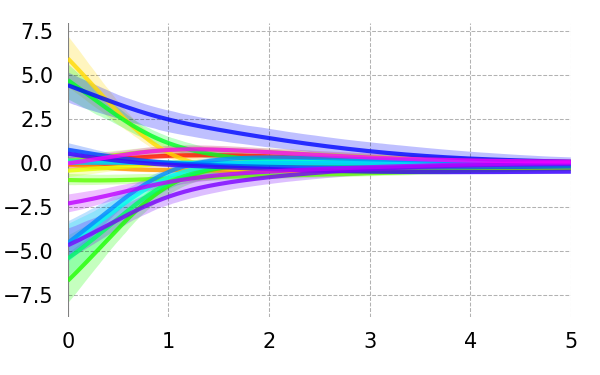}
    \end{subfigure}
    \begin{subfigure}[t]{0.49\textwidth}
        \centering
        \makebox[0.49\textwidth]{\centering Hidden Units $\div$ 2 (16)}%
        \makebox[0.49\textwidth]{\centering Hidden Units $\times$ 2 (64)}
        \includegraphics[width=0.49\textwidth]{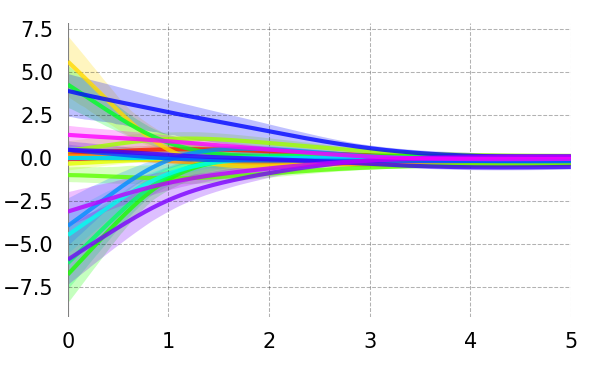}
        \includegraphics[width=0.49\textwidth]{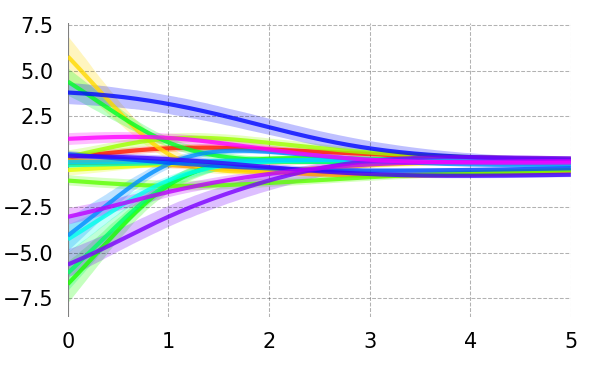}
    \end{subfigure}
    
    \begin{subfigure}[t]{0.49\textwidth}
        \centering
        \makebox[0.49\textwidth]{\centering Hidden Layers - 1 (1)}%
        \makebox[0.49\textwidth]{\centering Hidden Layers + 1 (3)}
        \includegraphics[width=0.49\textwidth]{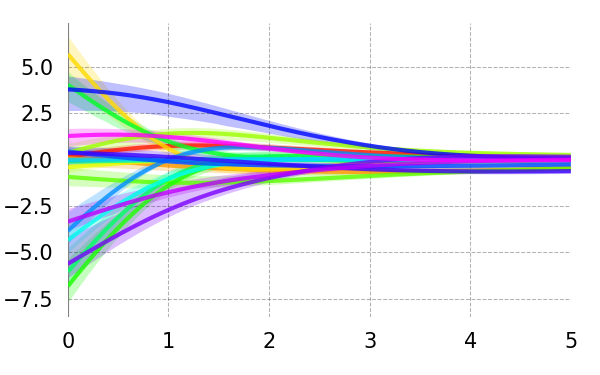}
        \includegraphics[width=0.49\textwidth]{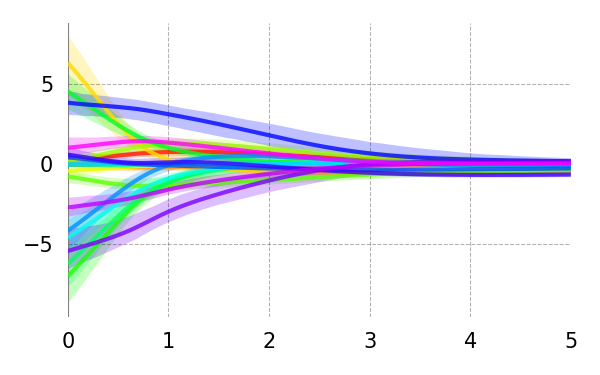}
    \end{subfigure}
    \begin{subfigure}[t]{0.49\textwidth}
        \centering
        \makebox[0.49\textwidth]{\centering Weight Reg $\div$ 5 (1)}%
        \makebox[0.49\textwidth]{\centering Weight Reg $\times$ 5 (5)}
        \includegraphics[width=0.49\textwidth]{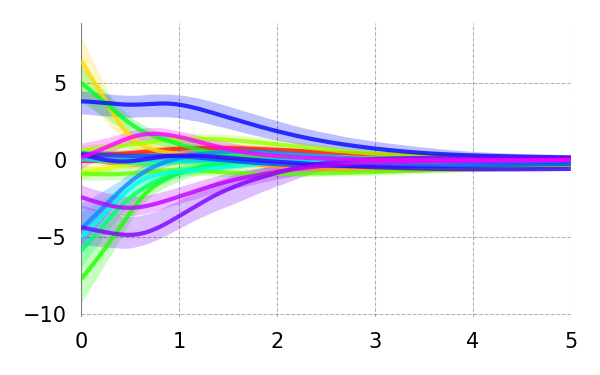}
        \includegraphics[width=0.49\textwidth]{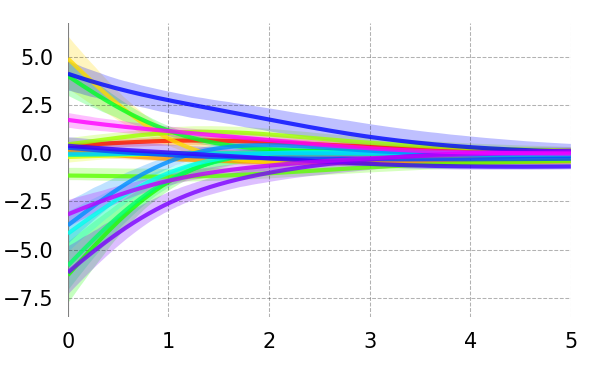}
    \end{subfigure}
    
    \begin{subfigure}[t]{0.49\textwidth}
        \centering
        \makebox[0.49\textwidth]{\centering Dropout $\div$ 2 (0.05)}%
        \makebox[0.49\textwidth]{\centering Dropout $\times$ 2 (0.2)}
        \includegraphics[width=0.49\textwidth]{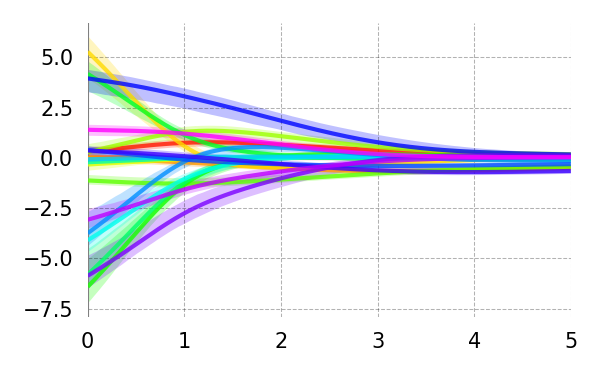}
        \includegraphics[width=0.49\textwidth]{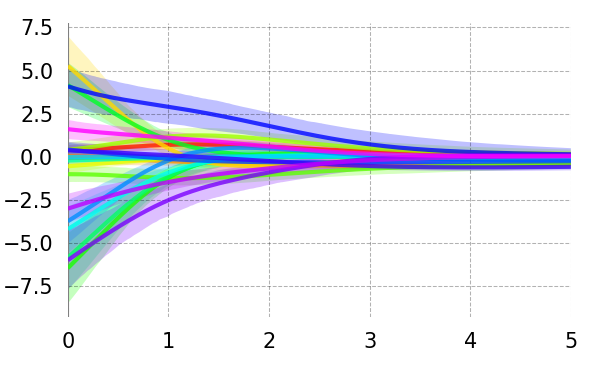}
    \end{subfigure}
    \begin{subfigure}[t]{0.49\textwidth}
        \centering
        \makebox[0.49\textwidth]{\centering Learning Rate $\div$ 3 (0.001)}%
        \makebox[0.49\textwidth]{\centering Learning Rate $\times$ 3 (0.009)}
        \includegraphics[width=0.49\textwidth]{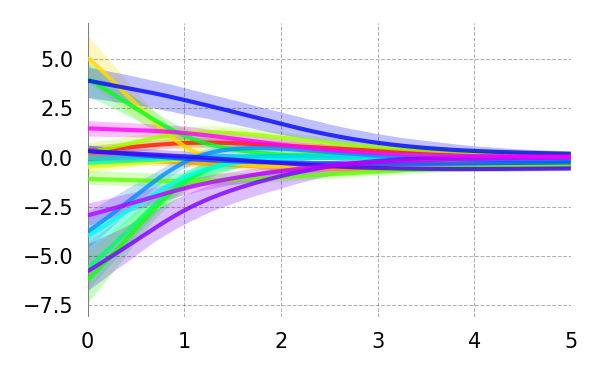}
        \includegraphics[width=0.49\textwidth]{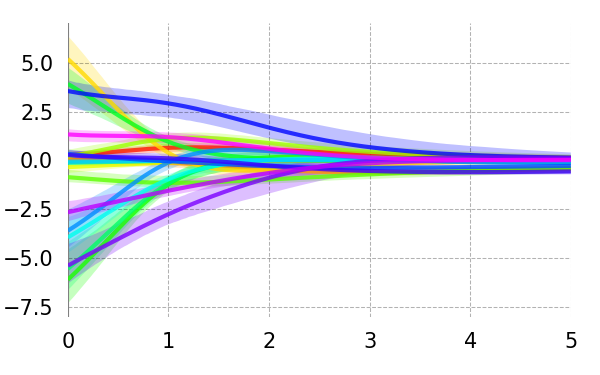}
    \end{subfigure}
    \begin{subfigure}[t]{0.49\textwidth}
        \centering
        \makebox[0.49\textwidth]{\centering Batch Size $\div$ 2 (512)}%
        \makebox[0.49\textwidth]{\centering Batch Size $\times$ 2 (2048)}
        \includegraphics[width=0.49\textwidth]{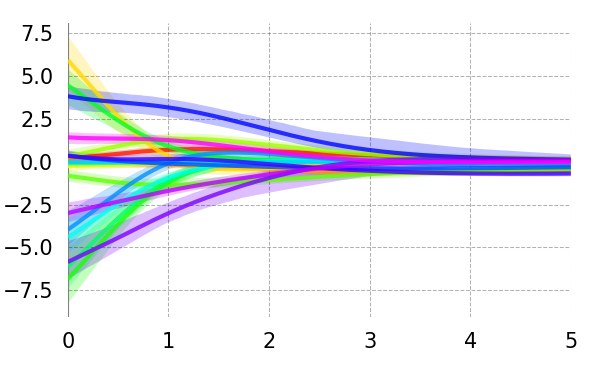}
        \includegraphics[width=0.49\textwidth]{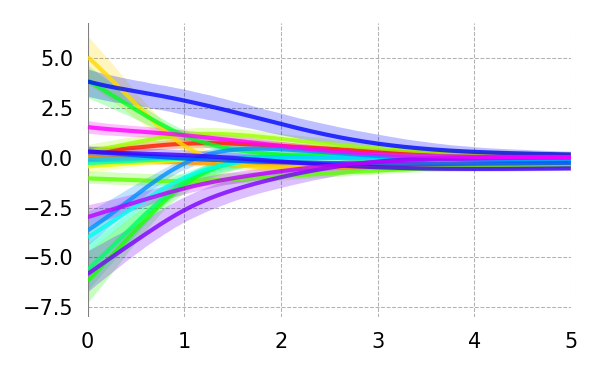}
    \end{subfigure}
    
    \vspace{1em}    
    
    \textbf{\large Consistency}
    
    \vspace{0.5em}
    
    \begin{subfigure}[t]{0.19\textwidth}
        \centering
        \makebox[0.9\textwidth]{\centering Rep 1}
        \includegraphics[width=\textwidth]{{results_cdrnn_journal_synth_multicollinearity_r0.75_CDR_main_irf_univariate_y_mc}.png}
    \end{subfigure}
    \begin{subfigure}[t]{0.19\textwidth}
        \centering
        \makebox[0.9\textwidth]{\centering Rep 2}
        \includegraphics[width=\textwidth]{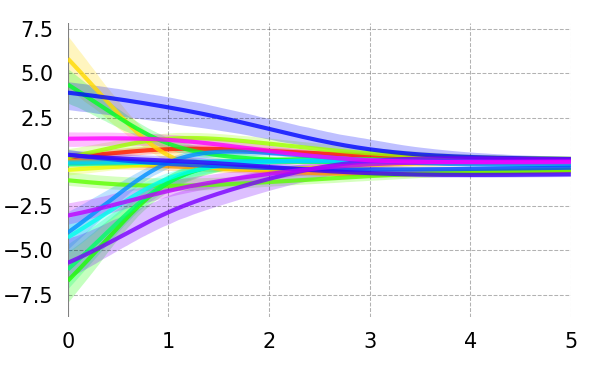}
    \end{subfigure}
    \begin{subfigure}[t]{0.19\textwidth}
        \centering
        \makebox[0.9\textwidth]{\centering Rep 3}
        \includegraphics[width=\textwidth]{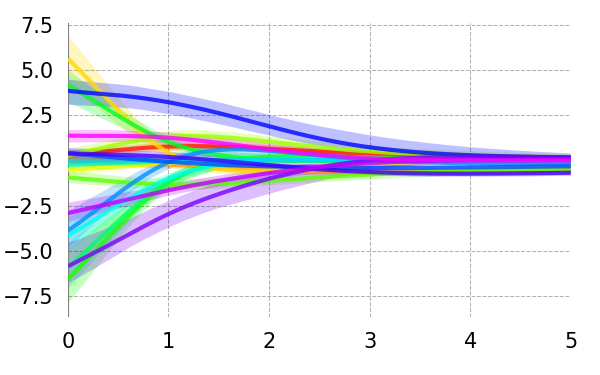}
    \end{subfigure}
    \begin{subfigure}[t]{0.19\textwidth}
        \centering
        \makebox[0.9\textwidth]{\centering Rep 4}
        \includegraphics[width=\textwidth]{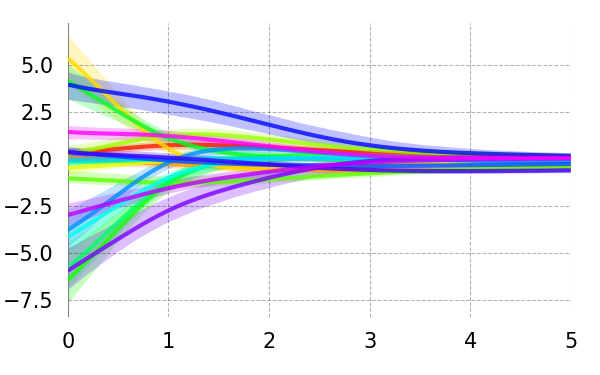}
    \end{subfigure}
    \begin{subfigure}[t]{0.19\textwidth}
        \centering
        \makebox[0.9\textwidth]{\centering Rep 5}
        \includegraphics[width=\textwidth]{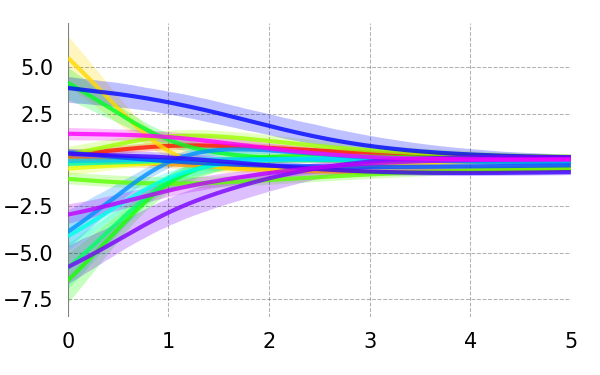}
    \end{subfigure}
    
    {\large Delay (s)}
    
    \vspace{0.5em}
    
    \caption{CDRNN estimated responses to synthetic data with \textbf{pairwise predictor multicollinearity of $r = 0.75$}. Estimates using base hyperparameters are compared to estimates from models that deviate from the base in some dimension. Plots under ``Consistency'' show estimates from five replicates of the ``base'' configuration, where ``Rep 1'' is the same model as ``base'' above, replotted for ease of comparison.}
    \label{fig:app-synth-multicollinearity-r75}
    
\end{figure}

\begin{figure}

    \footnotesize
    \sffamily
    \centering
    
    \textbf{\Large Synth: Multicollinearity, $r=0.9$}
    
    \vspace{1em}
    
    \begin{subfigure}[t]{0.49\textwidth}
        \centering
        \makebox[0.49\textwidth]{\centering \textbf{True}}
        
        \includegraphics[width=0.49\textwidth]{{results_cl_synth_multicollinearity_r0.00_synthetic_true}.png}
    \end{subfigure}
    
    \begin{subfigure}[t]{0.49\textwidth}
        \centering
        \makebox[0.49\textwidth]{\centering Base}%
        \makebox[0.49\textwidth]{\centering + RNN}
        \begin{overpic}[width=0.49\textwidth]{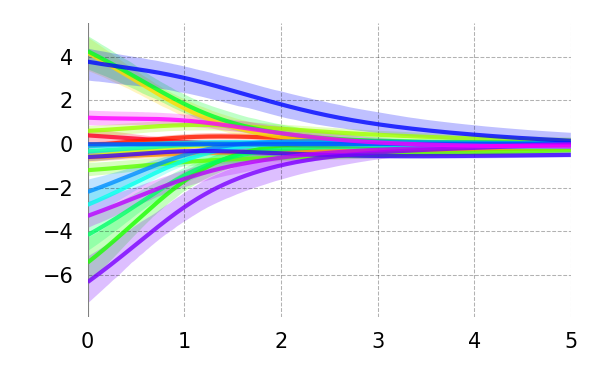}
        \end{overpic}%
        \includegraphics[width=0.49\textwidth]{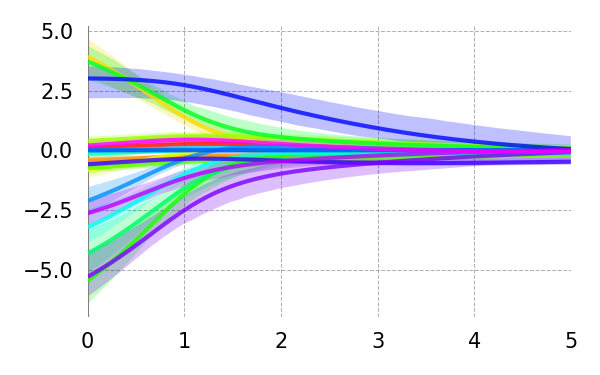}
    \end{subfigure}
    \begin{subfigure}[t]{0.49\textwidth}
        \centering
        \makebox[0.49\textwidth]{\centering Hidden Units $\div$ 2 (16)}%
        \makebox[0.49\textwidth]{\centering Hidden Units $\times$ 2 (64)}
        \includegraphics[width=0.49\textwidth]{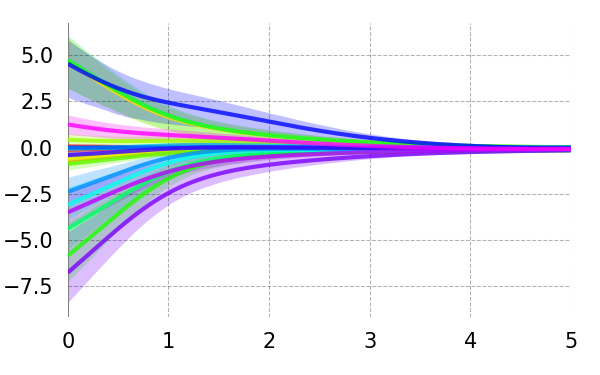}
        \includegraphics[width=0.49\textwidth]{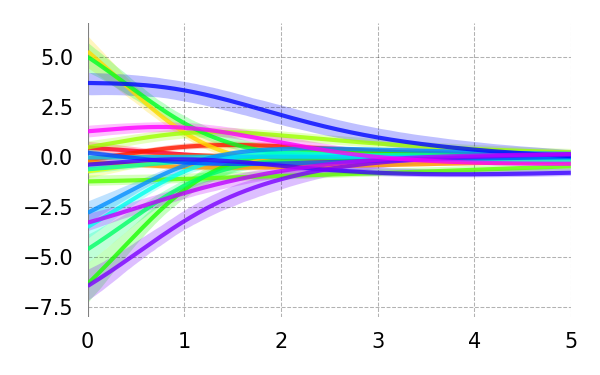}
    \end{subfigure}
    
    \begin{subfigure}[t]{0.49\textwidth}
        \centering
        \makebox[0.49\textwidth]{\centering Hidden Layers - 1 (1)}%
        \makebox[0.49\textwidth]{\centering Hidden Layers + 1 (3)}
        \includegraphics[width=0.49\textwidth]{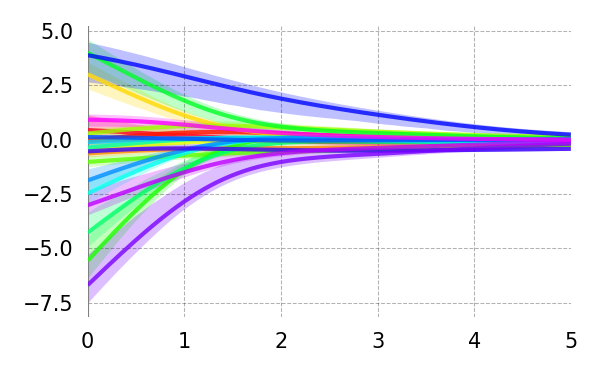}
        \includegraphics[width=0.49\textwidth]{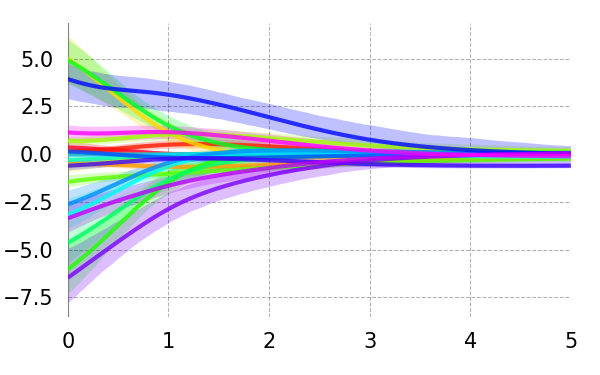}
    \end{subfigure}
    \begin{subfigure}[t]{0.49\textwidth}
        \centering
        \makebox[0.49\textwidth]{\centering Weight Reg $\div$ 5 (1)}%
        \makebox[0.49\textwidth]{\centering Weight Reg $\times$ 5 (5)}
        \includegraphics[width=0.49\textwidth]{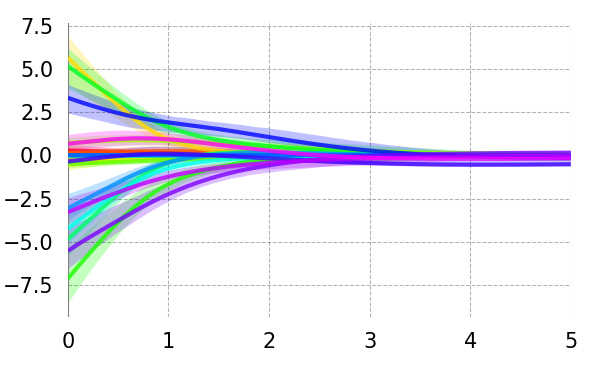}
        \includegraphics[width=0.49\textwidth]{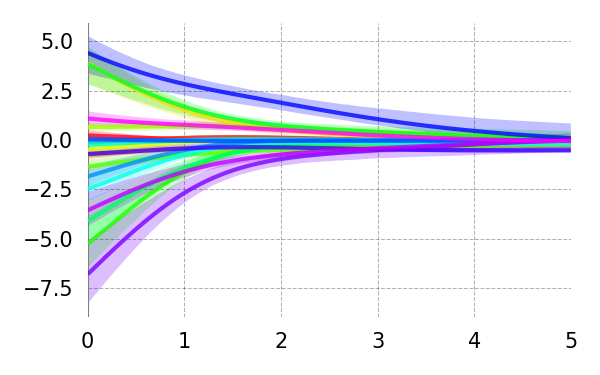}
    \end{subfigure}
    
    \begin{subfigure}[t]{0.49\textwidth}
        \centering
        \makebox[0.49\textwidth]{\centering Dropout $\div$ 2 (0.05)}%
        \makebox[0.49\textwidth]{\centering Dropout $\times$ 2 (0.2)}
        \includegraphics[width=0.49\textwidth]{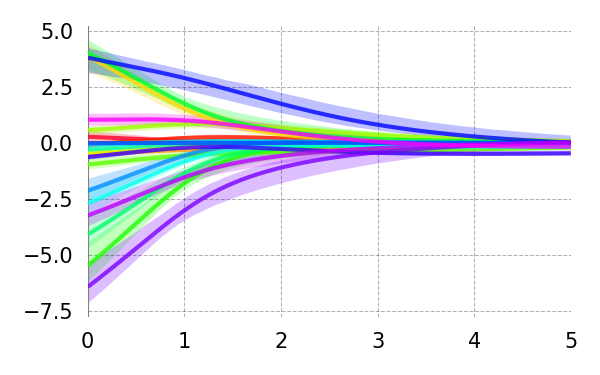}
        \includegraphics[width=0.49\textwidth]{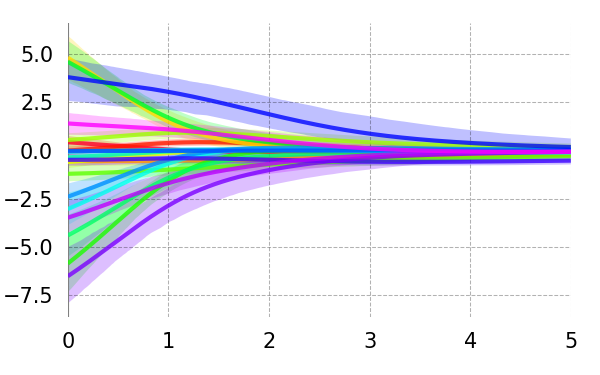}
    \end{subfigure}
    \begin{subfigure}[t]{0.49\textwidth}
        \centering
        \makebox[0.49\textwidth]{\centering Learning Rate $\div$ 3 (0.001)}%
        \makebox[0.49\textwidth]{\centering Learning Rate $\times$ 3 (0.009)}
        \includegraphics[width=0.49\textwidth]{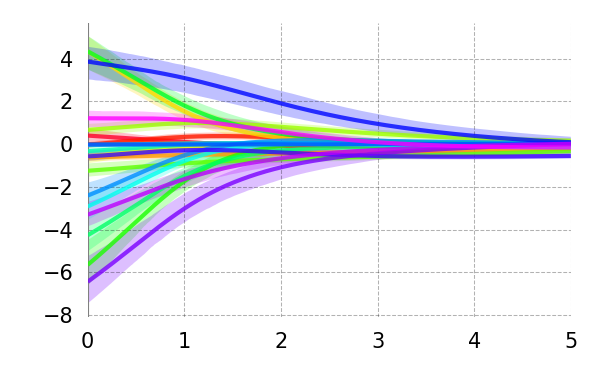}
        \includegraphics[width=0.49\textwidth]{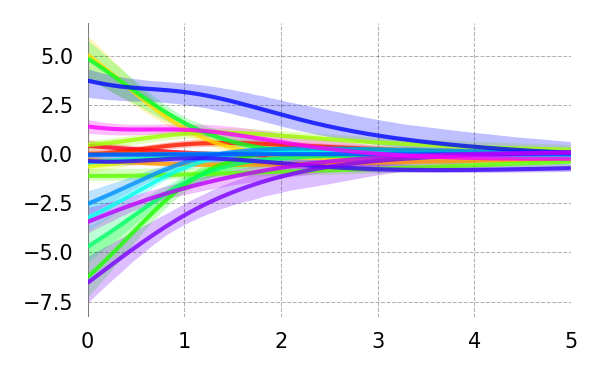}
    \end{subfigure}
    \begin{subfigure}[t]{0.49\textwidth}
        \centering
        \makebox[0.49\textwidth]{\centering Batch Size $\div$ 2 (512)}%
        \makebox[0.49\textwidth]{\centering Batch Size $\times$ 2 (2048)}
        \includegraphics[width=0.49\textwidth]{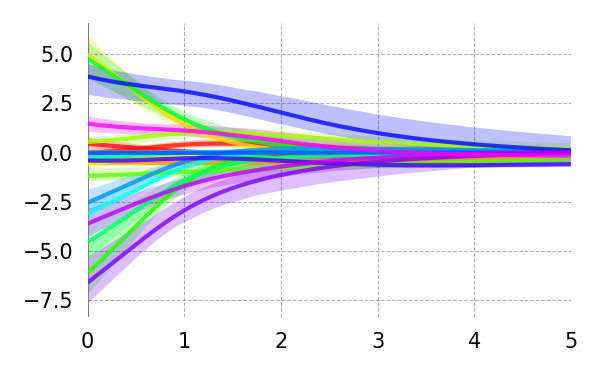}
        \includegraphics[width=0.49\textwidth]{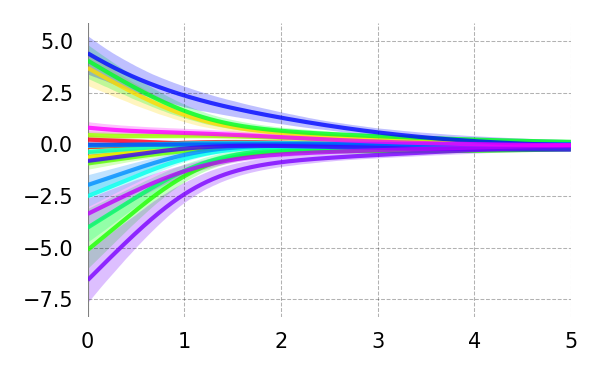}
    \end{subfigure}
    
    \vspace{1em}    
    
    \textbf{\large Consistency}
    
    \vspace{0.5em}
    
    \begin{subfigure}[t]{0.19\textwidth}
        \centering
        \makebox[0.9\textwidth]{\centering Rep 1}
        \includegraphics[width=\textwidth]{{results_cdrnn_journal_synth_multicollinearity_r0.90_CDR_main_irf_univariate_y_mc}.png}
    \end{subfigure}
    \begin{subfigure}[t]{0.19\textwidth}
        \centering
        \makebox[0.9\textwidth]{\centering Rep 2}
        \includegraphics[width=\textwidth]{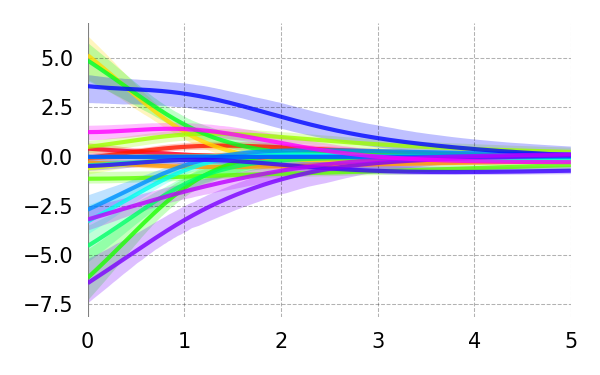}
    \end{subfigure}
    \begin{subfigure}[t]{0.19\textwidth}
        \centering
        \makebox[0.9\textwidth]{\centering Rep 3}
        \includegraphics[width=\textwidth]{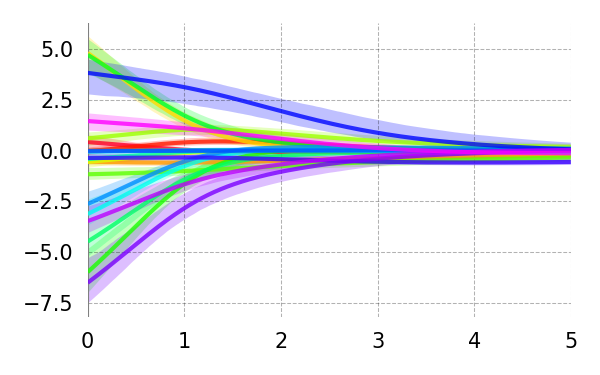}
    \end{subfigure}
    \begin{subfigure}[t]{0.19\textwidth}
        \centering
        \makebox[0.9\textwidth]{\centering Rep 4}
        \includegraphics[width=\textwidth]{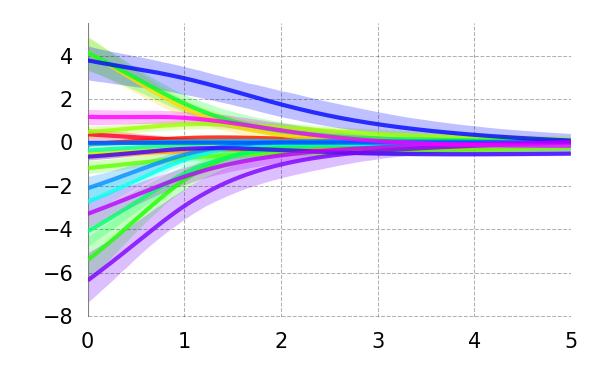}
    \end{subfigure}
    \begin{subfigure}[t]{0.19\textwidth}
        \centering
        \makebox[0.9\textwidth]{\centering Rep 5}
        \includegraphics[width=\textwidth]{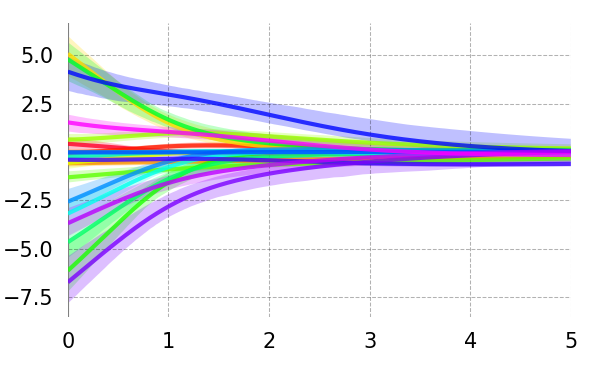}
    \end{subfigure}
    
    {\large Delay (s)}
    
    \vspace{0.5em}
    
    \caption{CDRNN estimated responses to synthetic data with \textbf{pairwise predictor multicollinearity of $r = 0.9$}. Estimates using base hyperparameters are compared to estimates from models that deviate from the base in some dimension. Plots under ``Consistency'' show estimates from five replicates of the ``base'' configuration, where ``Rep 1'' is the same model as ``base'' above, replotted for ease of comparison.}
    \label{fig:app-synth-multicollinearity-r90}
    
\end{figure}

\begin{figure}

    \footnotesize
    \sffamily
    \centering
    
    \textbf{\Large Synth: Multicollinearity, $r=0.95$}
    
    \vspace{1em}
    
    \begin{subfigure}[t]{0.49\textwidth}
        \centering
        \makebox[0.49\textwidth]{\centering \textbf{True}}
        
        \includegraphics[width=0.49\textwidth]{{results_cl_synth_multicollinearity_r0.00_synthetic_true}.png}
    \end{subfigure}
    
    \begin{subfigure}[t]{0.49\textwidth}
        \centering
        \makebox[0.49\textwidth]{\centering Base}%
        \makebox[0.49\textwidth]{\centering + RNN}
        \begin{overpic}[width=0.49\textwidth]{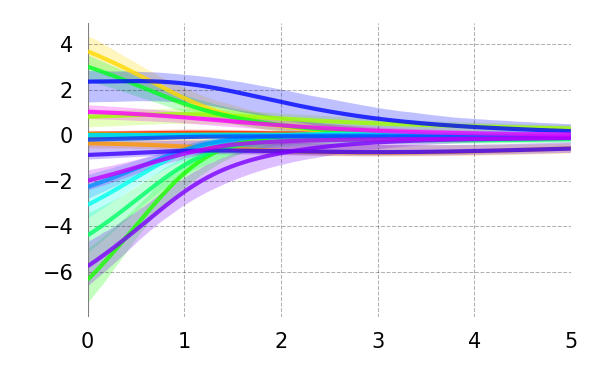}
        \end{overpic}%
        \includegraphics[width=0.49\textwidth]{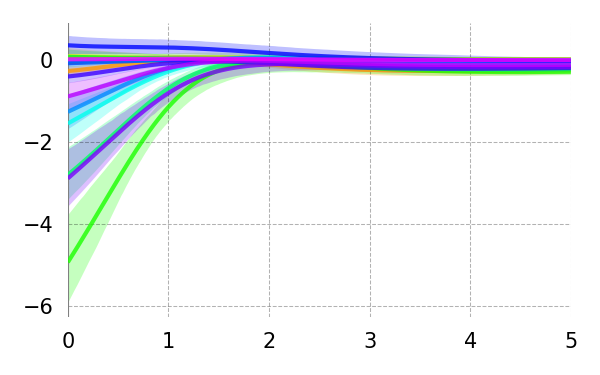}
    \end{subfigure}
    \begin{subfigure}[t]{0.49\textwidth}
        \centering
        \makebox[0.49\textwidth]{\centering Hidden Units $\div$ 2 (16)}%
        \makebox[0.49\textwidth]{\centering Hidden Units $\times$ 2 (64)}
        \includegraphics[width=0.49\textwidth]{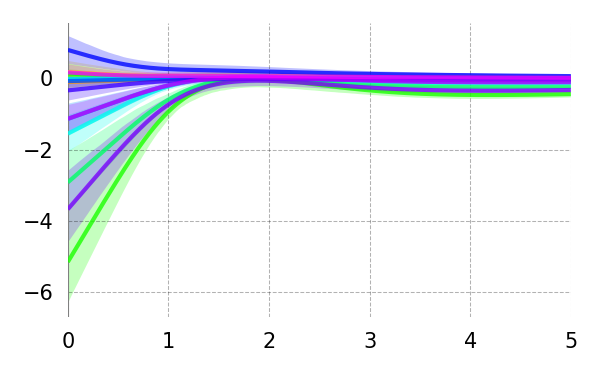}
        \includegraphics[width=0.49\textwidth]{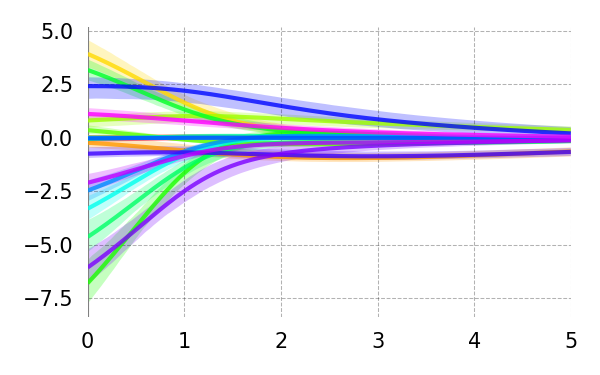}
    \end{subfigure}
    
    \begin{subfigure}[t]{0.49\textwidth}
        \centering
        \makebox[0.49\textwidth]{\centering Hidden Layers - 1 (1)}%
        \makebox[0.49\textwidth]{\centering Hidden Layers + 1 (3)}
        \includegraphics[width=0.49\textwidth]{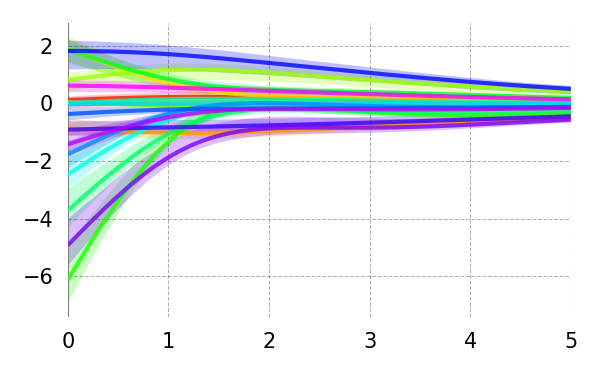}
        \includegraphics[width=0.49\textwidth]{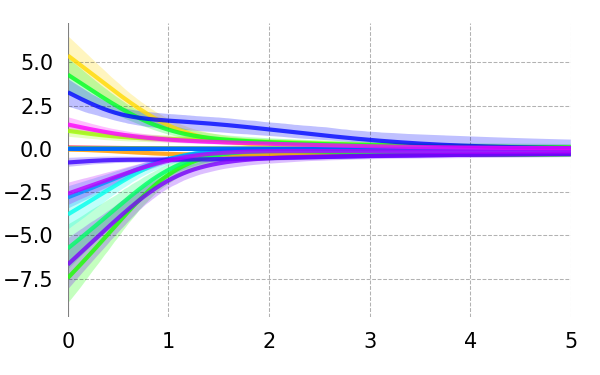}
    \end{subfigure}
    \begin{subfigure}[t]{0.49\textwidth}
        \centering
        \makebox[0.49\textwidth]{\centering Weight Reg $\div$ 5 (1)}%
        \makebox[0.49\textwidth]{\centering Weight Reg $\times$ 5 (5)}
        \includegraphics[width=0.49\textwidth]{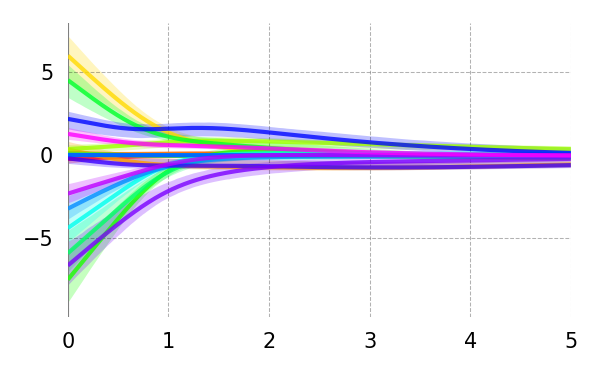}
        \includegraphics[width=0.49\textwidth]{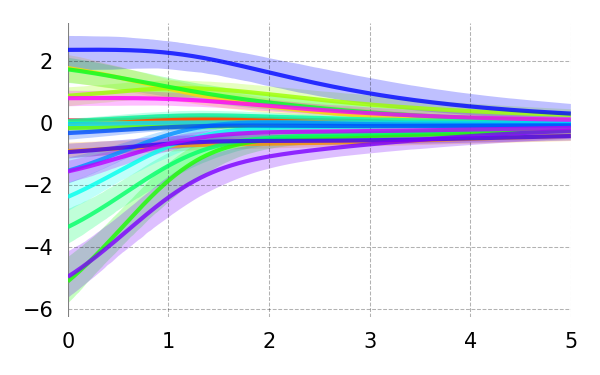}
    \end{subfigure}
    
    \begin{subfigure}[t]{0.49\textwidth}
        \centering
        \makebox[0.49\textwidth]{\centering Dropout $\div$ 2 (0.05)}%
        \makebox[0.49\textwidth]{\centering Dropout $\times$ 2 (0.2)}
        \includegraphics[width=0.49\textwidth]{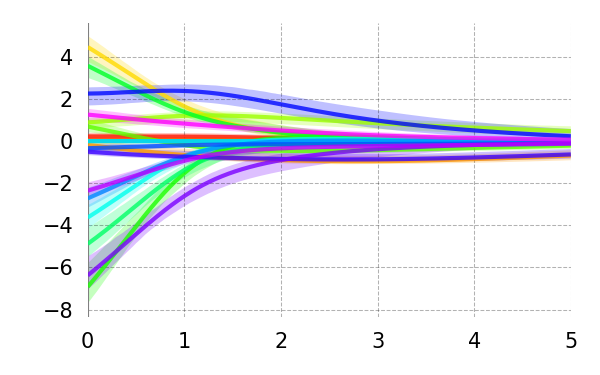}
        \includegraphics[width=0.49\textwidth]{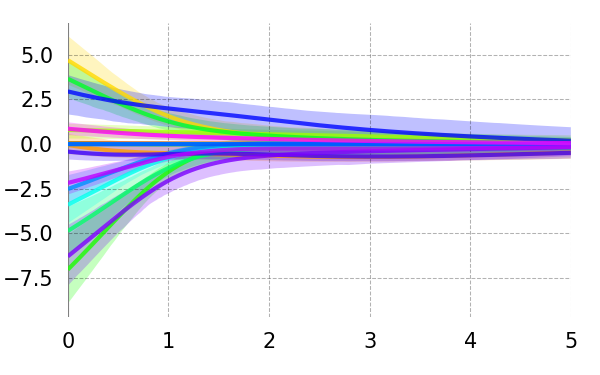}
    \end{subfigure}
    \begin{subfigure}[t]{0.49\textwidth}
        \centering
        \makebox[0.49\textwidth]{\centering Learning Rate $\div$ 3 (0.001)}%
        \makebox[0.49\textwidth]{\centering Learning Rate $\times$ 3 (0.009)}
        \includegraphics[width=0.49\textwidth]{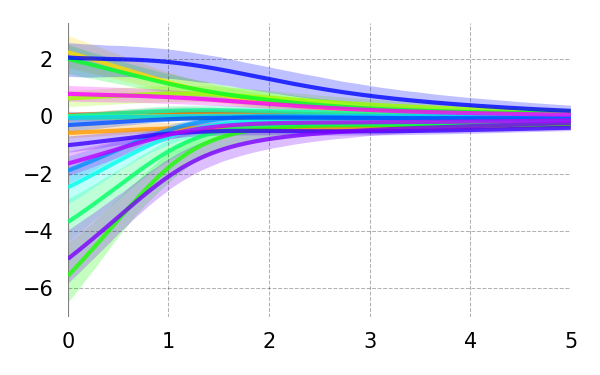}
        \includegraphics[width=0.49\textwidth]{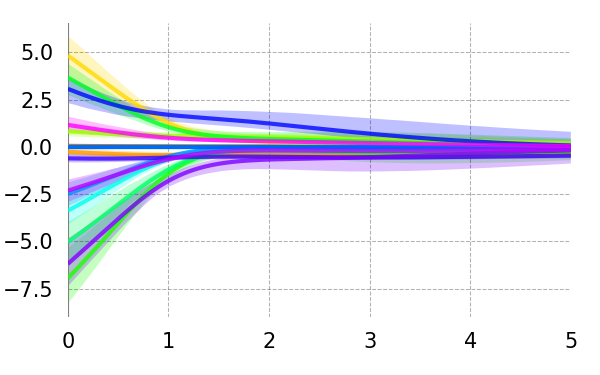}
    \end{subfigure}
    \begin{subfigure}[t]{0.49\textwidth}
        \centering
        \makebox[0.49\textwidth]{\centering Batch Size $\div$ 2 (512)}%
        \makebox[0.49\textwidth]{\centering Batch Size $\times$ 2 (2048)}
        \includegraphics[width=0.49\textwidth]{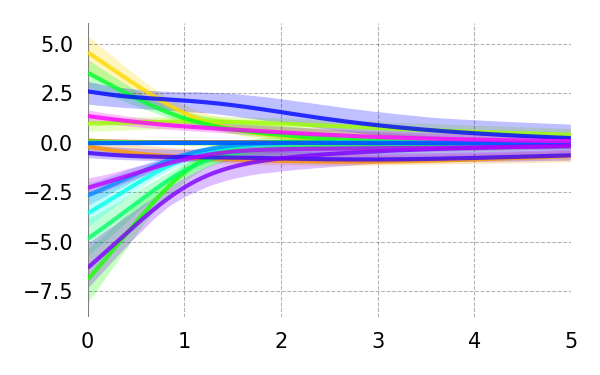}
        \includegraphics[width=0.49\textwidth]{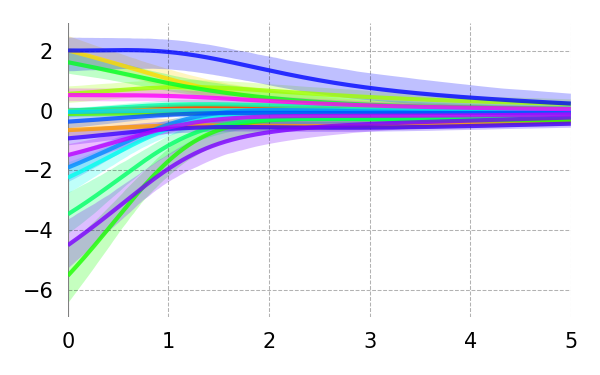}
    \end{subfigure}
    
    \vspace{1em}    
    
    \textbf{\large Consistency}
    
    \vspace{0.5em}
    
    \begin{subfigure}[t]{0.19\textwidth}
        \centering
        \makebox[0.9\textwidth]{\centering Rep 1}
        \includegraphics[width=\textwidth]{{results_cdrnn_journal_synth_multicollinearity_r0.95_CDR_main_irf_univariate_y_mc}.png}
    \end{subfigure}
    \begin{subfigure}[t]{0.19\textwidth}
        \centering
        \makebox[0.9\textwidth]{\centering Rep 2}
        \includegraphics[width=\textwidth]{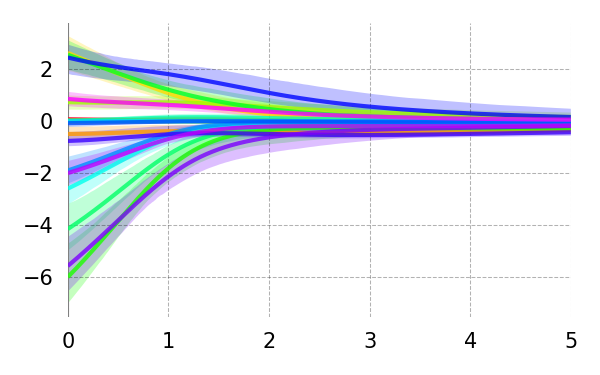}
    \end{subfigure}
    \begin{subfigure}[t]{0.19\textwidth}
        \centering
        \makebox[0.9\textwidth]{\centering Rep 3}
        \includegraphics[width=\textwidth]{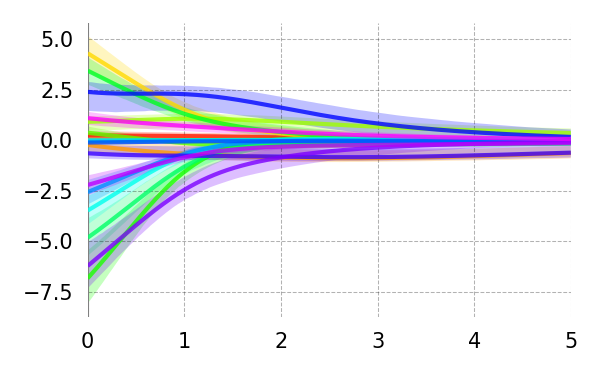}
    \end{subfigure}
    \begin{subfigure}[t]{0.19\textwidth}
        \centering
        \makebox[0.9\textwidth]{\centering Rep 4}
        \includegraphics[width=\textwidth]{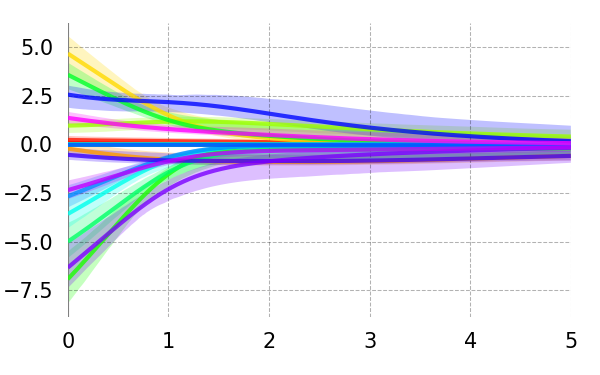}
    \end{subfigure}
    \begin{subfigure}[t]{0.19\textwidth}
        \centering
        \makebox[0.9\textwidth]{\centering Rep 5}
        \includegraphics[width=\textwidth]{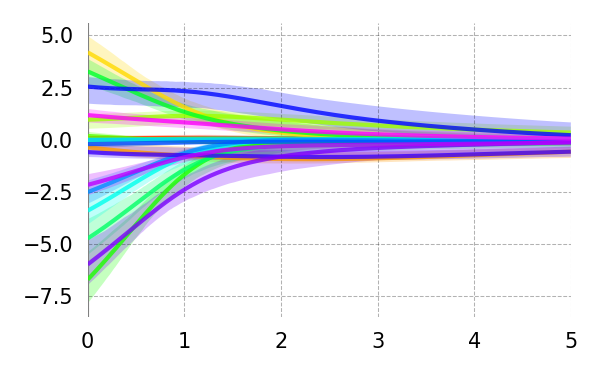}
    \end{subfigure}
    
    {\large Delay (s)}
    
    \vspace{0.5em}
    
    \caption{CDRNN estimated responses to synthetic data with \textbf{pairwise predictor multicollinearity of $r = 0.95$}. Estimates using base hyperparameters are compared to estimates from models that deviate from the base in some dimension. Plots under ``Consistency'' show estimates from five replicates of the ``base'' configuration, where ``Rep 1'' is the same model as ``base'' above, replotted for ease of comparison.}
    \label{fig:app-synth-multicollinearity-r95}
    
\end{figure}

\begin{figure}

    \footnotesize
    \sffamily
    \centering
    
    \textbf{\Large Synth: Exponential Ground Truth IRF}
    
    \vspace{1em}
    
    \begin{subfigure}[t]{0.49\textwidth}
        \centering
        \makebox[0.49\textwidth]{\centering \textbf{True}}
        
        \includegraphics[width=0.49\textwidth]{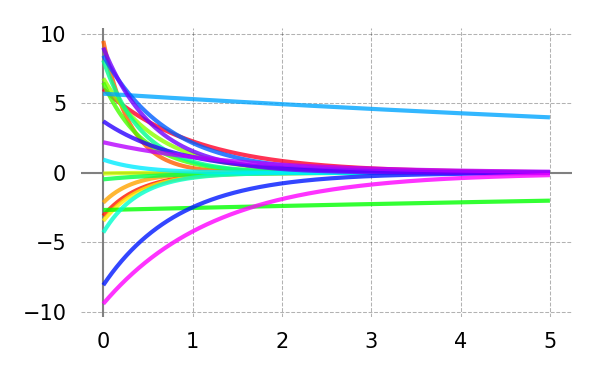}
    \end{subfigure}
    
    \begin{subfigure}[t]{0.49\textwidth}
        \centering
        \makebox[0.49\textwidth]{\centering Base}%
        \makebox[0.49\textwidth]{\centering + RNN}
        \begin{overpic}[width=0.49\textwidth]{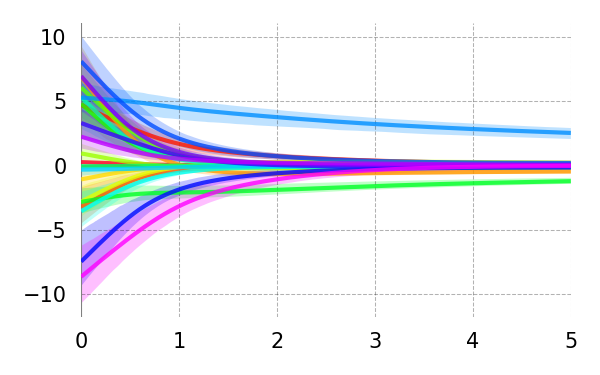}
        \end{overpic}%
        \includegraphics[width=0.49\textwidth]{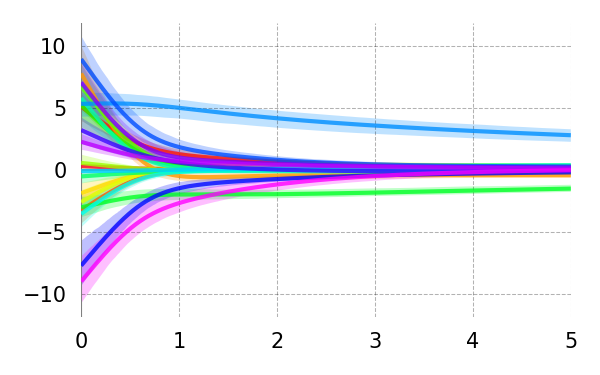}
    \end{subfigure}
    \begin{subfigure}[t]{0.49\textwidth}
        \centering
        \makebox[0.49\textwidth]{\centering Hidden Units $\div$ 2 (16)}%
        \makebox[0.49\textwidth]{\centering Hidden Units $\times$ 2 (64)}
        \includegraphics[width=0.49\textwidth]{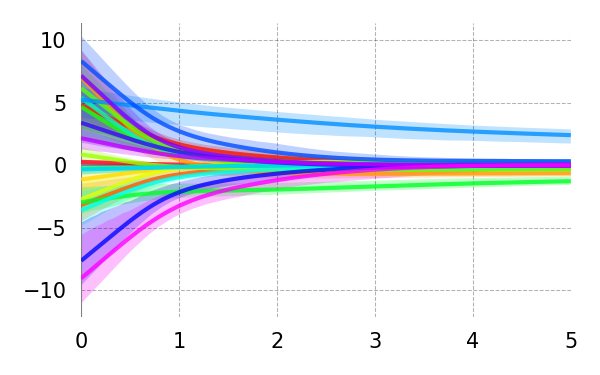}
        \includegraphics[width=0.49\textwidth]{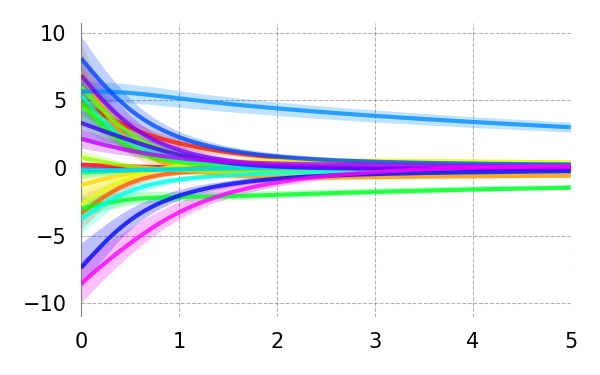}
    \end{subfigure}
    
    \begin{subfigure}[t]{0.49\textwidth}
        \centering
        \makebox[0.49\textwidth]{\centering Hidden Layers - 1 (1)}%
        \makebox[0.49\textwidth]{\centering Hidden Layers + 1 (3)}
        \includegraphics[width=0.49\textwidth]{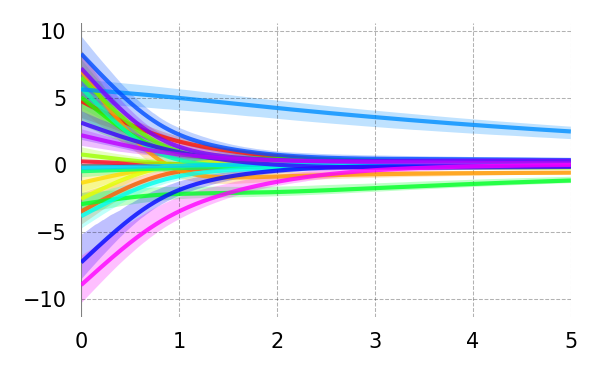}
        \includegraphics[width=0.49\textwidth]{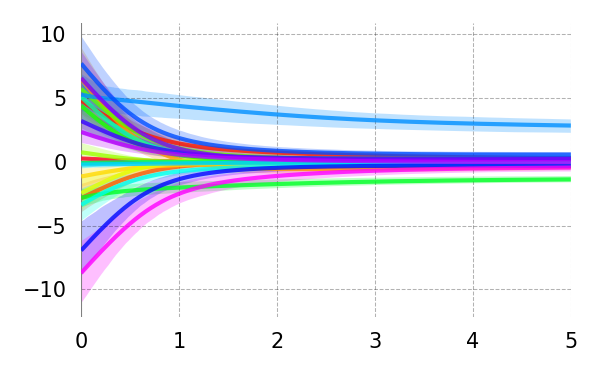}
    \end{subfigure}
    \begin{subfigure}[t]{0.49\textwidth}
        \centering
        \makebox[0.49\textwidth]{\centering Weight Reg $\div$ 5 (1)}%
        \makebox[0.49\textwidth]{\centering Weight Reg $\times$ 5 (5)}
        \includegraphics[width=0.49\textwidth]{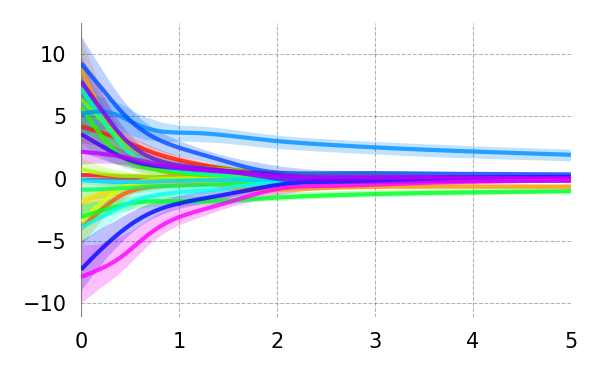}
        \includegraphics[width=0.49\textwidth]{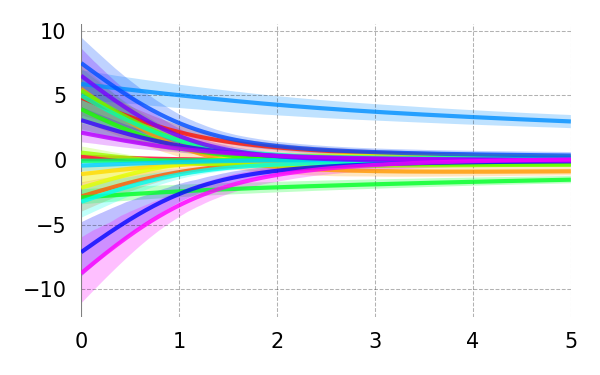}
    \end{subfigure}
    
    \begin{subfigure}[t]{0.49\textwidth}
        \centering
        \makebox[0.49\textwidth]{\centering Dropout $\div$ 2 (0.05)}%
        \makebox[0.49\textwidth]{\centering Dropout $\times$ 2 (0.2)}
        \includegraphics[width=0.49\textwidth]{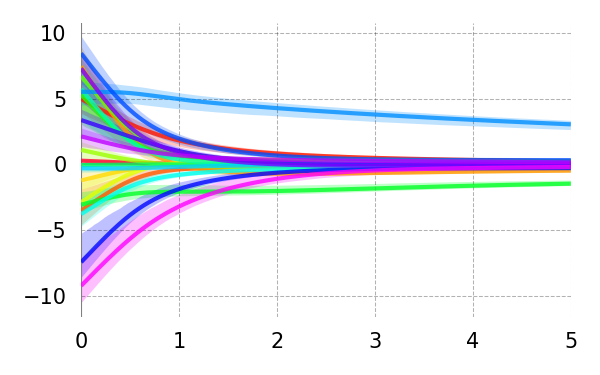}
        \includegraphics[width=0.49\textwidth]{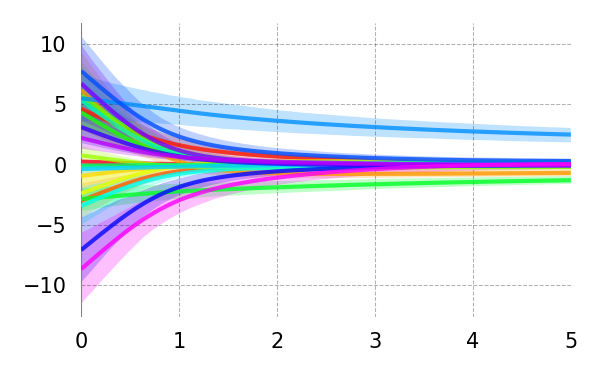}
    \end{subfigure}
    \begin{subfigure}[t]{0.49\textwidth}
        \centering
        \makebox[0.49\textwidth]{\centering Learning Rate $\div$ 3 (0.001)}%
        \makebox[0.49\textwidth]{\centering Learning Rate $\times$ 3 (0.009)}
        \includegraphics[width=0.49\textwidth]{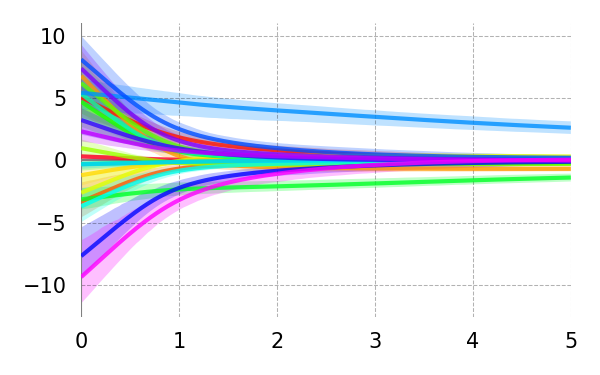}
        \includegraphics[width=0.49\textwidth]{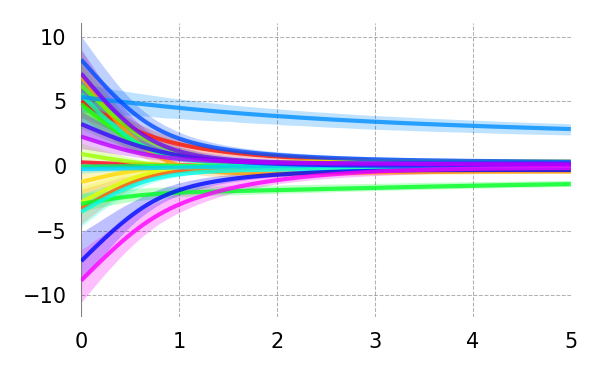}
    \end{subfigure}
    \begin{subfigure}[t]{0.49\textwidth}
        \centering
        \makebox[0.49\textwidth]{\centering Batch Size $\div$ 2 (512)}%
        \makebox[0.49\textwidth]{\centering Batch Size $\times$ 2 (2048)}
        \includegraphics[width=0.49\textwidth]{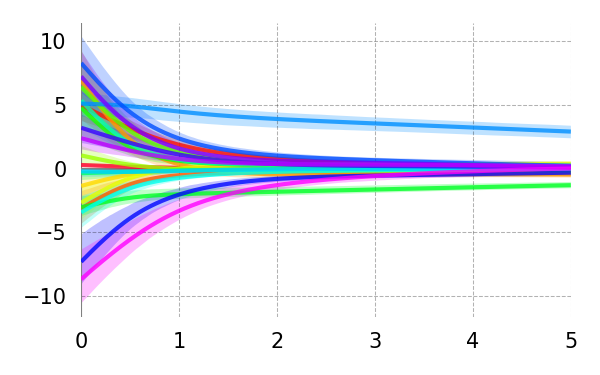}
        \includegraphics[width=0.49\textwidth]{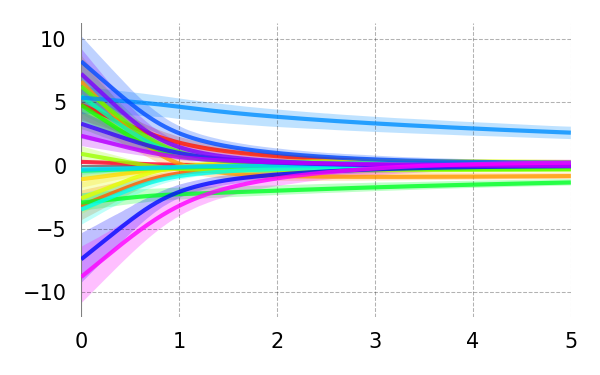}
    \end{subfigure}
    
    \vspace{1em}    
    
    \textbf{\large Consistency}
    
    \vspace{0.5em}
    
    \begin{subfigure}[t]{0.19\textwidth}
        \centering
        \makebox[0.9\textwidth]{\centering Rep 1}
        \includegraphics[width=\textwidth]{{results_cdrnn_journal_synth_misspecification_e_CDR_main_irf_univariate_y_mc}.png}
    \end{subfigure}
    \begin{subfigure}[t]{0.19\textwidth}
        \centering
        \makebox[0.9\textwidth]{\centering Rep 2}
        \includegraphics[width=\textwidth]{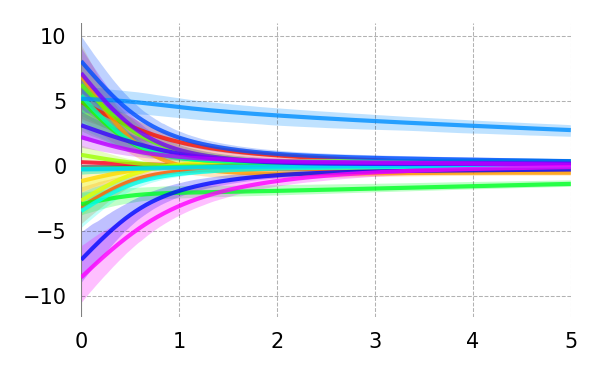}
    \end{subfigure}
    \begin{subfigure}[t]{0.19\textwidth}
        \centering
        \makebox[0.9\textwidth]{\centering Rep 3}
        \includegraphics[width=\textwidth]{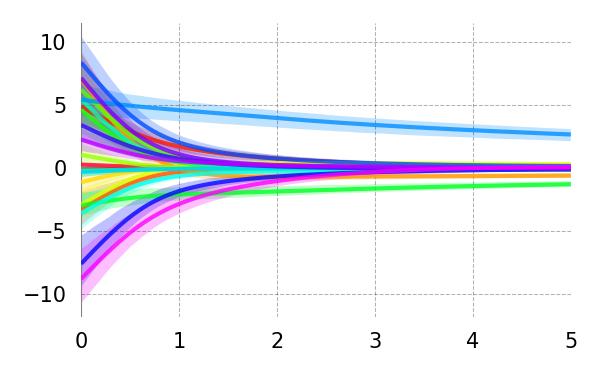}
    \end{subfigure}
    \begin{subfigure}[t]{0.19\textwidth}
        \centering
        \makebox[0.9\textwidth]{\centering Rep 4}
        \includegraphics[width=\textwidth]{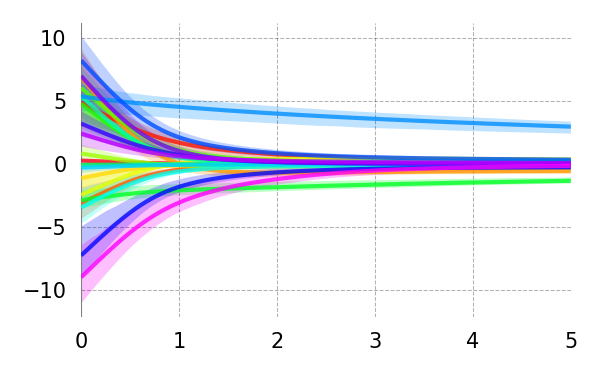}
    \end{subfigure}
    \begin{subfigure}[t]{0.19\textwidth}
        \centering
        \makebox[0.9\textwidth]{\centering Rep 5}
        \includegraphics[width=\textwidth]{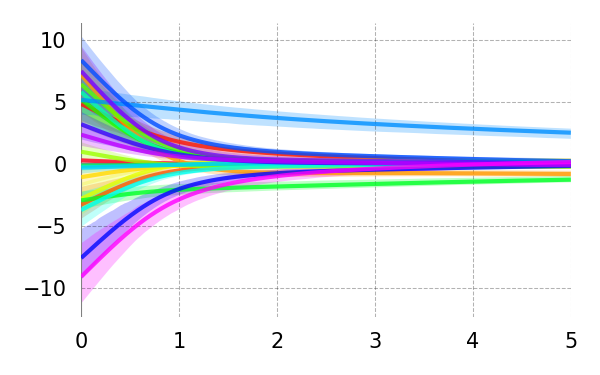}
    \end{subfigure}
    
    {\large Delay (s)}
    
    \vspace{0.5em}
    
    \caption{CDRNN estimated responses to synthetic data convolved using an \textbf{exponential IRF}. Estimates using base hyperparameters are compared to estimates from models that deviate from the base in some dimension. Plots under ``Consistency'' show estimates from five replicates of the ``base'' configuration, where ``Rep 1'' is the same model as ``base'' above, replotted for ease of comparison.}
    \label{fig:app-synth-misspecification-e}
    
\end{figure}

\begin{figure}

    \footnotesize
    \sffamily
    \centering
    
    \textbf{\Large Synth: Normal (Gaussian) Ground Truth IRF}
    
    \vspace{1em}
    
    \begin{subfigure}[t]{0.49\textwidth}
        \centering
        \makebox[0.49\textwidth]{\centering \textbf{True}}
        
        \includegraphics[width=0.49\textwidth]{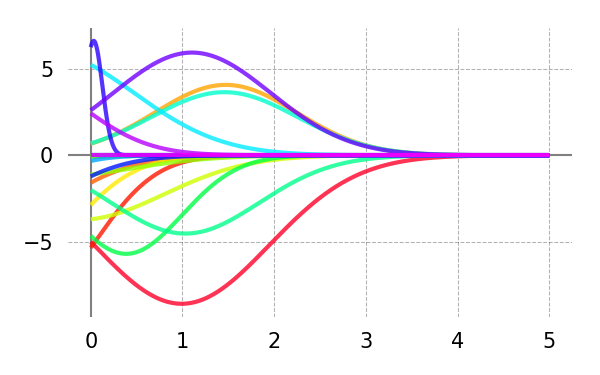}
    \end{subfigure}
    
    \begin{subfigure}[t]{0.49\textwidth}
        \centering
        \makebox[0.49\textwidth]{\centering Base}%
        \makebox[0.49\textwidth]{\centering + RNN}
        \begin{overpic}[width=0.49\textwidth]{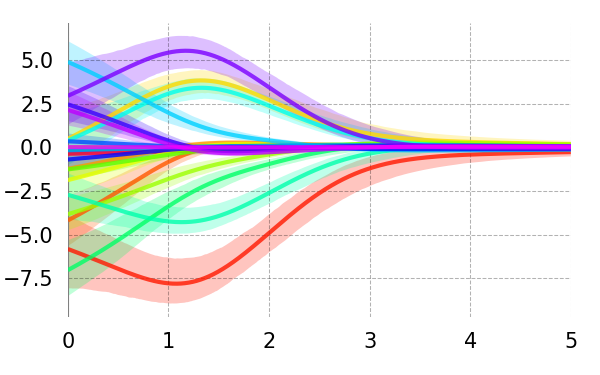}
        \end{overpic}%
        \includegraphics[width=0.49\textwidth]{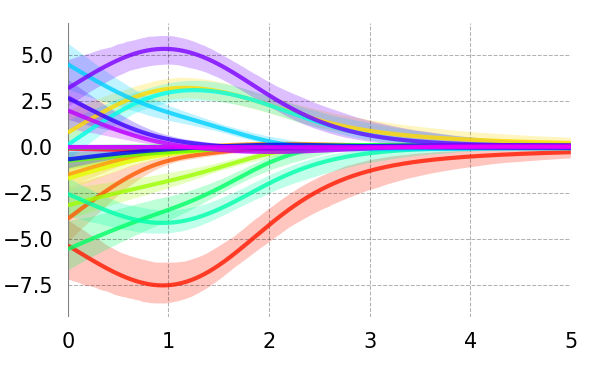}
    \end{subfigure}
    \begin{subfigure}[t]{0.49\textwidth}
        \centering
        \makebox[0.49\textwidth]{\centering Hidden Units $\div$ 2 (16)}%
        \makebox[0.49\textwidth]{\centering Hidden Units $\times$ 2 (64)}
        \includegraphics[width=0.49\textwidth]{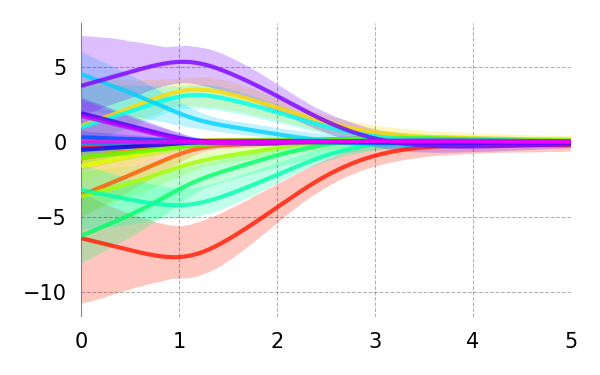}
        \includegraphics[width=0.49\textwidth]{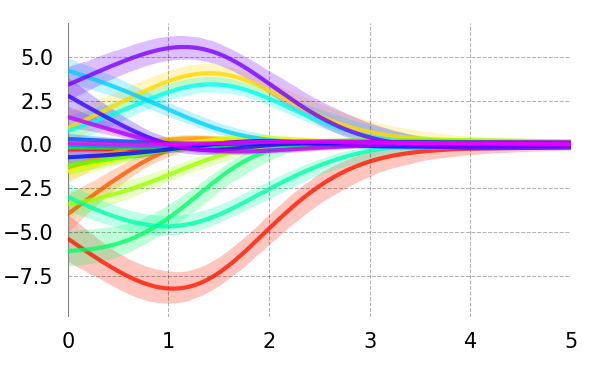}
    \end{subfigure}
    
    \begin{subfigure}[t]{0.49\textwidth}
        \centering
        \makebox[0.49\textwidth]{\centering Hidden Layers - 1 (1)}%
        \makebox[0.49\textwidth]{\centering Hidden Layers + 1 (3)}
        \includegraphics[width=0.49\textwidth]{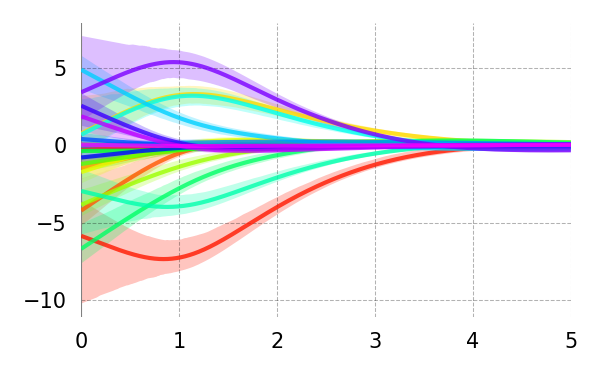}
        \includegraphics[width=0.49\textwidth]{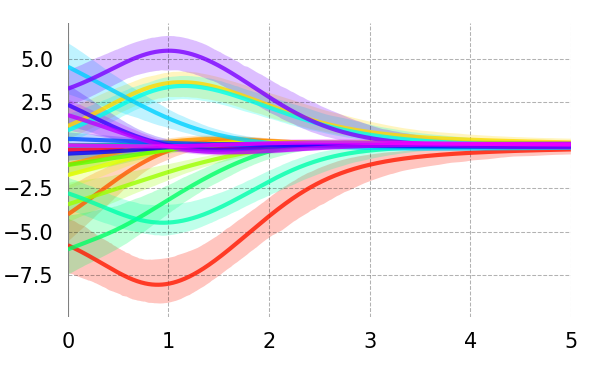}
    \end{subfigure}
    \begin{subfigure}[t]{0.49\textwidth}
        \centering
        \makebox[0.49\textwidth]{\centering Weight Reg $\div$ 5 (1)}%
        \makebox[0.49\textwidth]{\centering Weight Reg $\times$ 5 (5)}
        \includegraphics[width=0.49\textwidth]{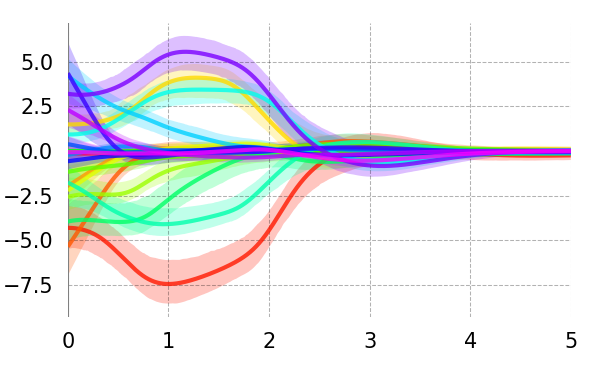}
        \includegraphics[width=0.49\textwidth]{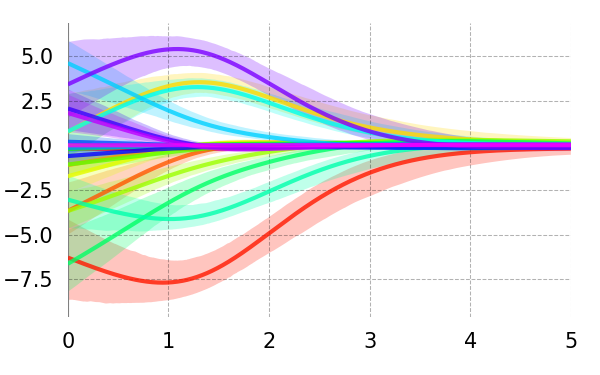}
    \end{subfigure}
    
    \begin{subfigure}[t]{0.49\textwidth}
        \centering
        \makebox[0.49\textwidth]{\centering Dropout $\div$ 2 (0.05)}%
        \makebox[0.49\textwidth]{\centering Dropout $\times$ 2 (0.2)}
        \includegraphics[width=0.49\textwidth]{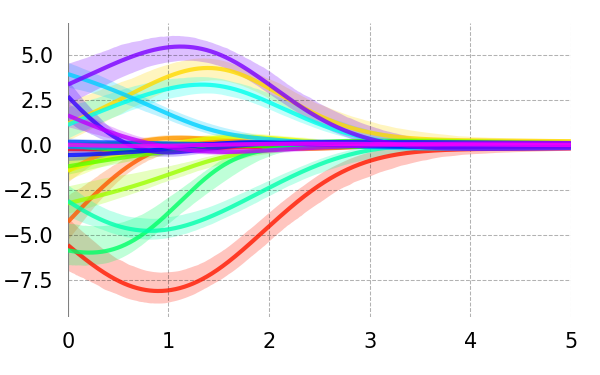}
        \includegraphics[width=0.49\textwidth]{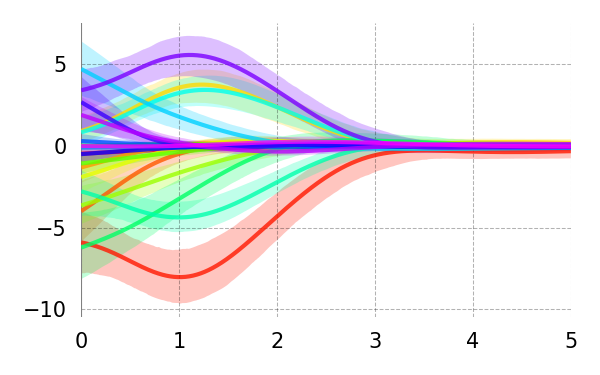}
    \end{subfigure}
    \begin{subfigure}[t]{0.49\textwidth}
        \centering
        \makebox[0.49\textwidth]{\centering Learning Rate $\div$ 3 (0.001)}%
        \makebox[0.49\textwidth]{\centering Learning Rate $\times$ 3 (0.009)}
        \includegraphics[width=0.49\textwidth]{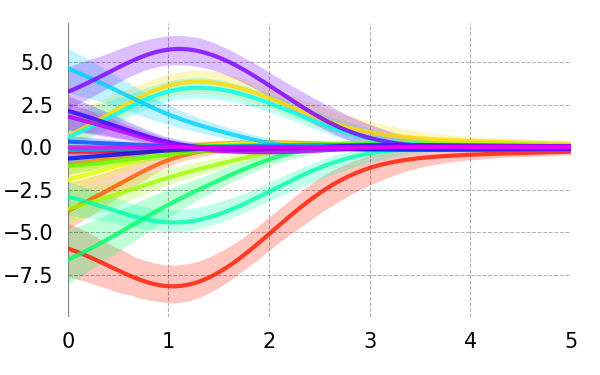}
        \includegraphics[width=0.49\textwidth]{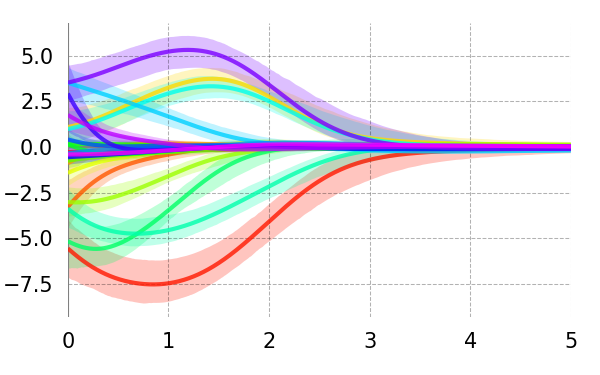}
    \end{subfigure}
    \begin{subfigure}[t]{0.49\textwidth}
        \centering
        \makebox[0.49\textwidth]{\centering Batch Size $\div$ 2 (512)}%
        \makebox[0.49\textwidth]{\centering Batch Size $\times$ 2 (2048)}
        \includegraphics[width=0.49\textwidth]{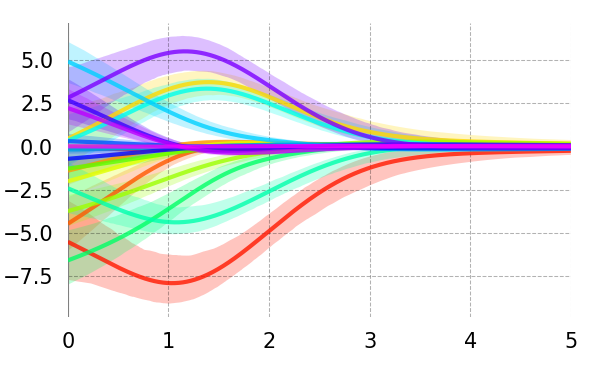}
        \includegraphics[width=0.49\textwidth]{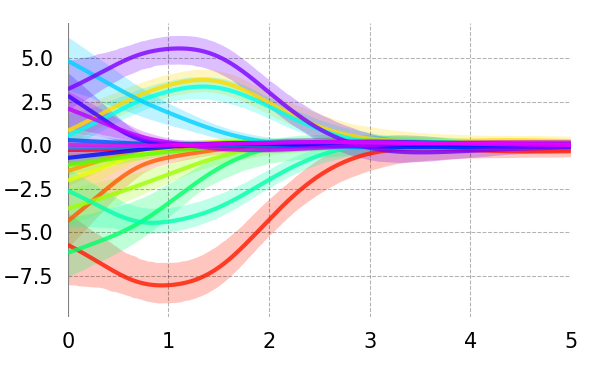}
    \end{subfigure}
    
    \vspace{1em}    
    
    \textbf{\large Consistency}
    
    \vspace{0.5em}
    
    \begin{subfigure}[t]{0.19\textwidth}
        \centering
        \makebox[0.9\textwidth]{\centering Rep 1}
        \includegraphics[width=\textwidth]{{results_cdrnn_journal_synth_misspecification_n_CDR_main_irf_univariate_y_mc}.png}
    \end{subfigure}
    \begin{subfigure}[t]{0.19\textwidth}
        \centering
        \makebox[0.9\textwidth]{\centering Rep 2}
        \includegraphics[width=\textwidth]{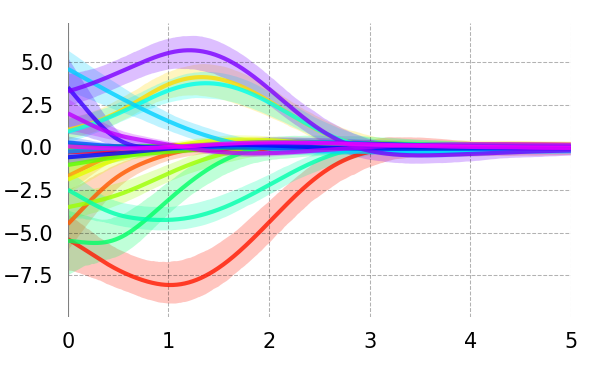}
    \end{subfigure}
    \begin{subfigure}[t]{0.19\textwidth}
        \centering
        \makebox[0.9\textwidth]{\centering Rep 3}
        \includegraphics[width=\textwidth]{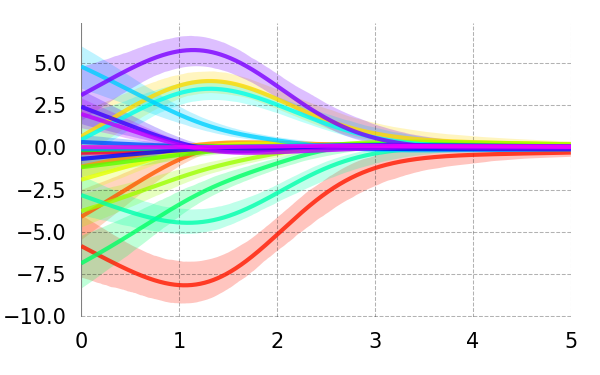}
    \end{subfigure}
    \begin{subfigure}[t]{0.19\textwidth}
        \centering
        \makebox[0.9\textwidth]{\centering Rep 4}
        \includegraphics[width=\textwidth]{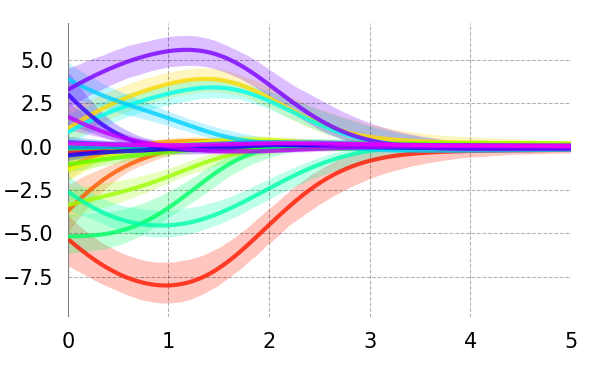}
    \end{subfigure}
    \begin{subfigure}[t]{0.19\textwidth}
        \centering
        \makebox[0.9\textwidth]{\centering Rep 5}
        \includegraphics[width=\textwidth]{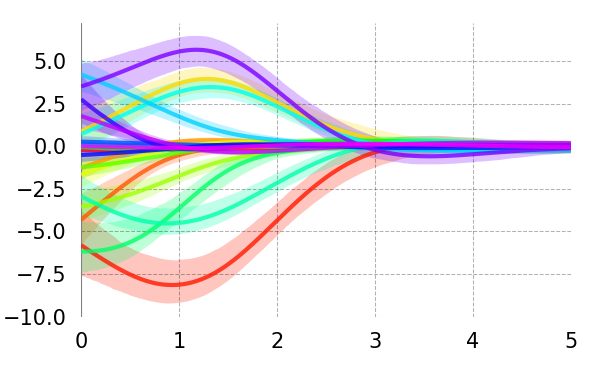}
    \end{subfigure}
    
    {\large Delay (s)}
    
    \vspace{0.5em}
    
    \caption{CDRNN estimated responses to synthetic data convolved using a \textbf{Normal (Gaussian) IRF}. Estimates using base hyperparameters are compared to estimates from models that deviate from the base in some dimension. Plots under ``Consistency'' show estimates from five replicates of the ``base'' configuration, where ``Rep 1'' is the same model as ``base'' above, replotted for ease of comparison.}
    \label{fig:app-synth-misspecification-n}
    
\end{figure}

\begin{figure}

    \footnotesize
    \sffamily
    \centering
    
    \textbf{\Large Synth: Shifted Gamma Ground Truth IRF}
    
    \vspace{1em}
    
    \begin{subfigure}[t]{0.49\textwidth}
        \centering
        \makebox[0.49\textwidth]{\centering \textbf{True}}
        
        \includegraphics[width=0.49\textwidth]{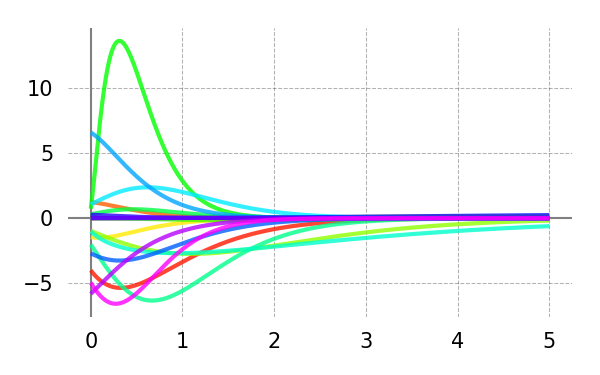}
    \end{subfigure}
    
    \begin{subfigure}[t]{0.49\textwidth}
        \centering
        \makebox[0.49\textwidth]{\centering Base}%
        \makebox[0.49\textwidth]{\centering + RNN}
        \begin{overpic}[width=0.49\textwidth]{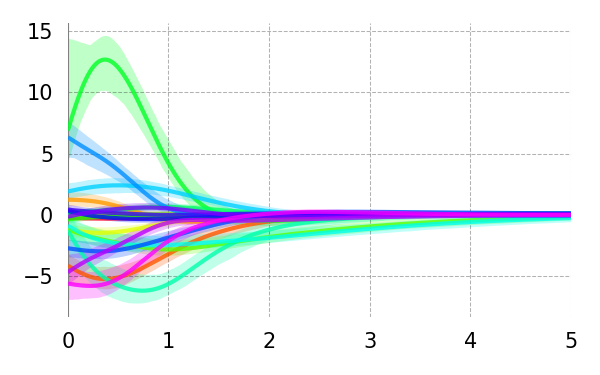}
        \end{overpic}%
        \includegraphics[width=0.49\textwidth]{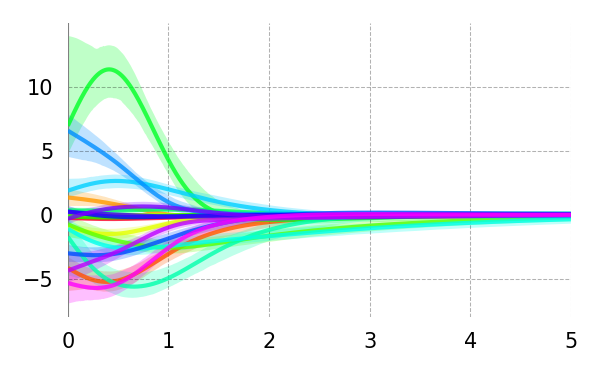}
    \end{subfigure}
    \begin{subfigure}[t]{0.49\textwidth}
        \centering
        \makebox[0.49\textwidth]{\centering Hidden Units $\div$ 2 (16)}%
        \makebox[0.49\textwidth]{\centering Hidden Units $\times$ 2 (64)}
        \includegraphics[width=0.49\textwidth]{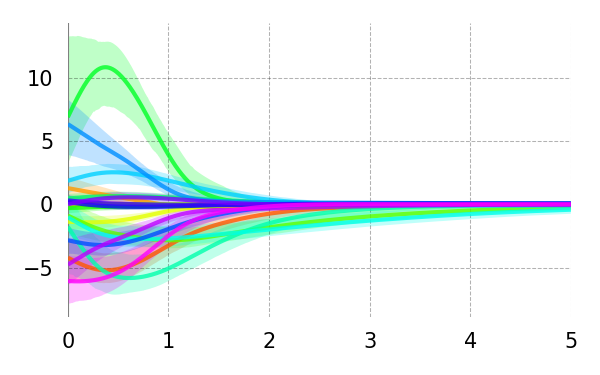}
        \includegraphics[width=0.49\textwidth]{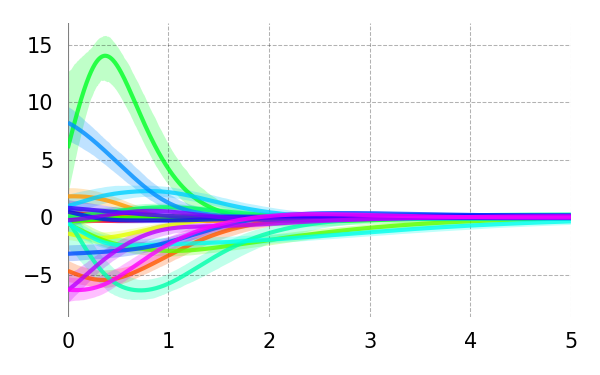}
    \end{subfigure}
    
    \begin{subfigure}[t]{0.49\textwidth}
        \centering
        \makebox[0.49\textwidth]{\centering Hidden Layers - 1 (1)}%
        \makebox[0.49\textwidth]{\centering Hidden Layers + 1 (3)}
        \includegraphics[width=0.49\textwidth]{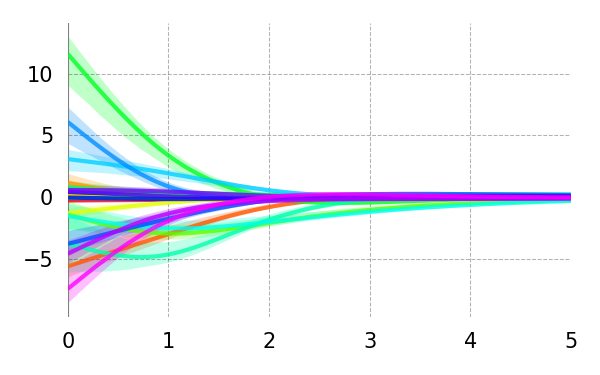}
        \includegraphics[width=0.49\textwidth]{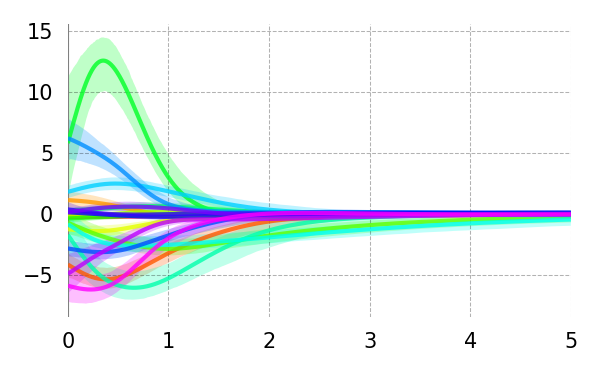}
    \end{subfigure}
    \begin{subfigure}[t]{0.49\textwidth}
        \centering
        \makebox[0.49\textwidth]{\centering Weight Reg $\div$ 5 (1)}%
        \makebox[0.49\textwidth]{\centering Weight Reg $\times$ 5 (5)}
        \includegraphics[width=0.49\textwidth]{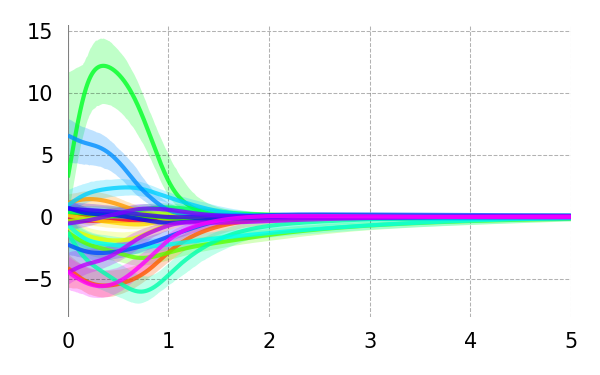}
        \includegraphics[width=0.49\textwidth]{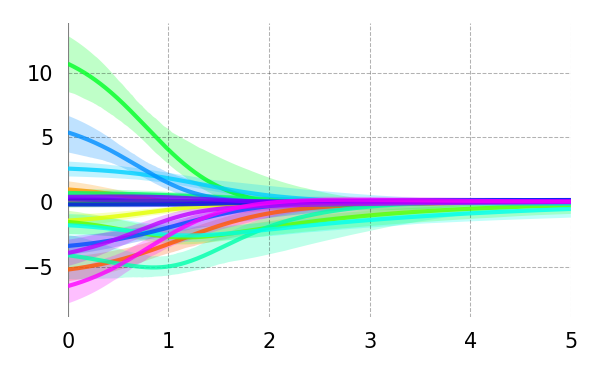}
    \end{subfigure}
    
    \begin{subfigure}[t]{0.49\textwidth}
        \centering
        \makebox[0.49\textwidth]{\centering Dropout $\div$ 2 (0.05)}%
        \makebox[0.49\textwidth]{\centering Dropout $\times$ 2 (0.2)}
        \includegraphics[width=0.49\textwidth]{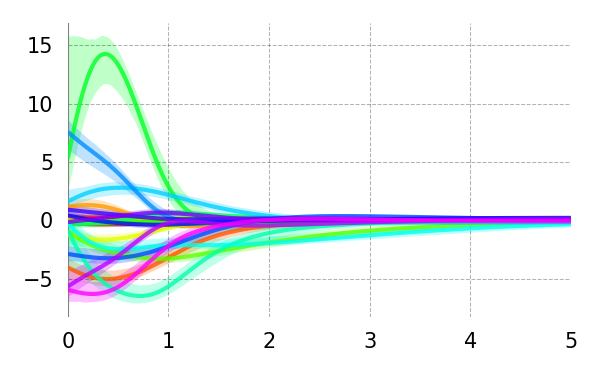}
        \includegraphics[width=0.49\textwidth]{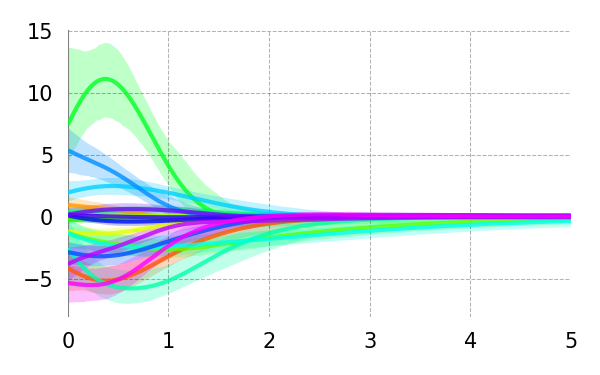}
    \end{subfigure}
    \begin{subfigure}[t]{0.49\textwidth}
        \centering
        \makebox[0.49\textwidth]{\centering Learning Rate $\div$ 3 (0.001)}%
        \makebox[0.49\textwidth]{\centering Learning Rate $\times$ 3 (0.009)}
        \includegraphics[width=0.49\textwidth]{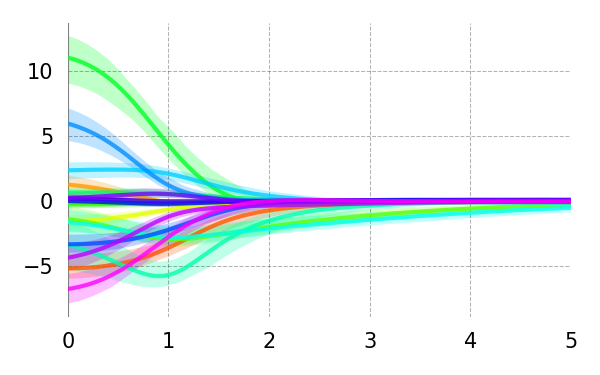}
        \includegraphics[width=0.49\textwidth]{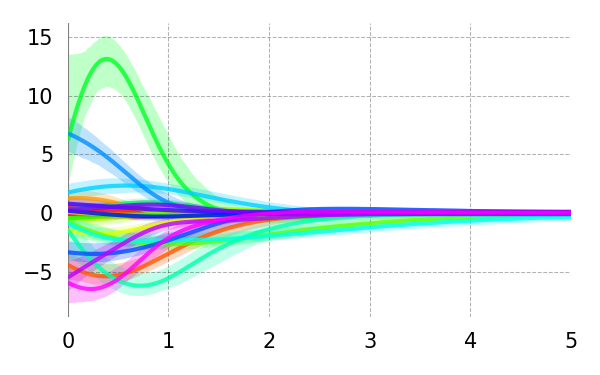}
    \end{subfigure}
    \begin{subfigure}[t]{0.49\textwidth}
        \centering
        \makebox[0.49\textwidth]{\centering Batch Size $\div$ 2 (512)}%
        \makebox[0.49\textwidth]{\centering Batch Size $\times$ 2 (2048)}
        \includegraphics[width=0.49\textwidth]{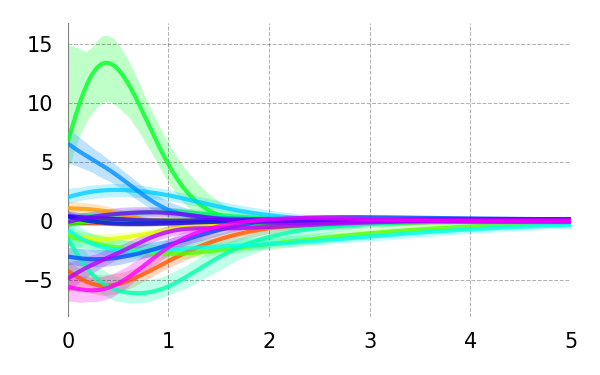}
        \includegraphics[width=0.49\textwidth]{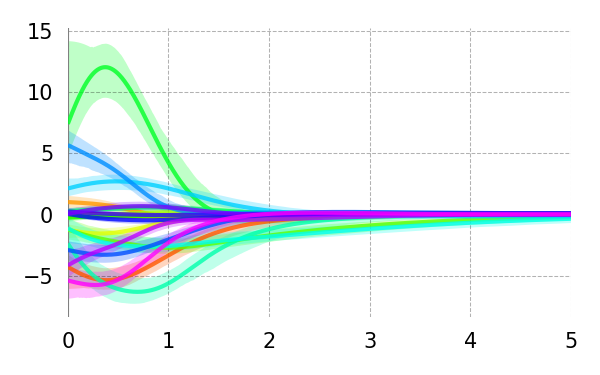}
    \end{subfigure}
    
    \vspace{1em}    
    
    \textbf{\large Consistency}
    
    \vspace{0.5em}
    
    \begin{subfigure}[t]{0.19\textwidth}
        \centering
        \makebox[0.9\textwidth]{\centering Rep 1}
        \includegraphics[width=\textwidth]{{results_cdrnn_journal_synth_misspecification_g_CDR_main_irf_univariate_y_mc}.png}
    \end{subfigure}
    \begin{subfigure}[t]{0.19\textwidth}
        \centering
        \makebox[0.9\textwidth]{\centering Rep 2}
        \includegraphics[width=\textwidth]{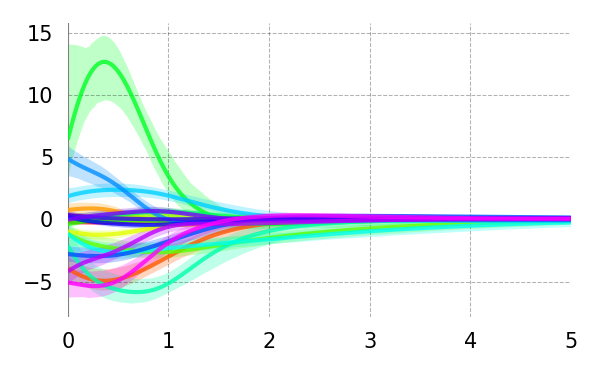}
    \end{subfigure}
    \begin{subfigure}[t]{0.19\textwidth}
        \centering
        \makebox[0.9\textwidth]{\centering Rep 3}
        \includegraphics[width=\textwidth]{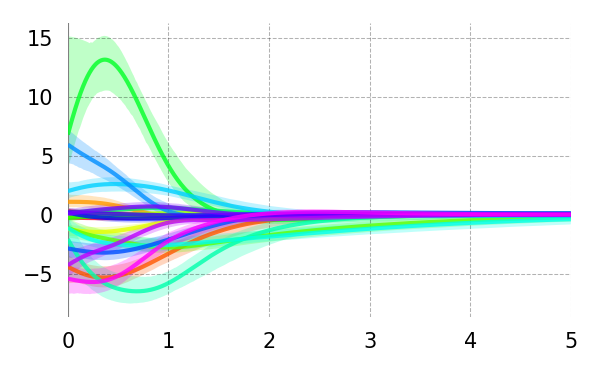}
    \end{subfigure}
    \begin{subfigure}[t]{0.19\textwidth}
        \centering
        \makebox[0.9\textwidth]{\centering Rep 4}
        \includegraphics[width=\textwidth]{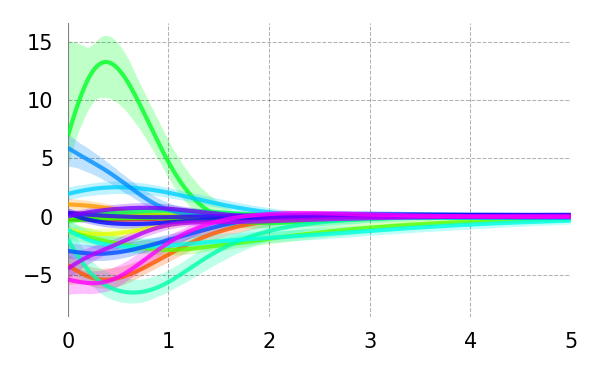}
    \end{subfigure}
    \begin{subfigure}[t]{0.19\textwidth}
        \centering
        \makebox[0.9\textwidth]{\centering Rep 5}
        \includegraphics[width=\textwidth]{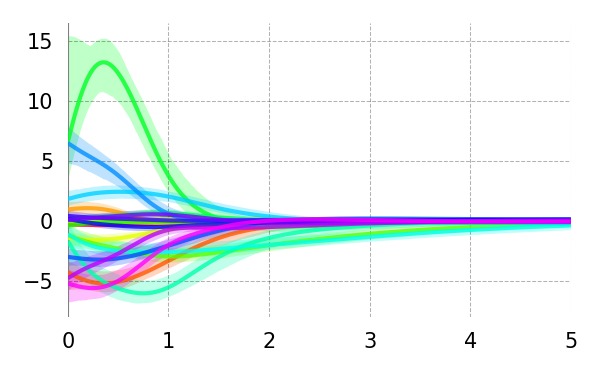}
    \end{subfigure}
    
    {\large Delay (s)}
    
    \vspace{0.5em}
    
    \caption{CDRNN estimated responses to synthetic data convolved using a \textbf{Shifted Gamma IRF}. Estimates using base hyperparameters are compared to estimates from models that deviate from the base in some dimension. Plots under ``Consistency'' show estimates from five replicates of the ``base'' configuration, where ``Rep 1'' is the same model as ``base'' above, replotted for ease of comparison.}
    \label{fig:app-synth-misspecification-g}
    
\end{figure}

Here we present results from the synthetic datasets in \cite{shainschuler19}.
\textbf{Supplementary Figures~\ref{fig:app-synth-noise-e0}--\ref{fig:app-synth-noise-e100}} vary the signal-to-noise ratio (injected noise with standard deviation 0, 0.1, 1, and 10).
\textbf{Supplementary Figures~\ref{fig:app-synth-time-fixed1}--\ref{fig:app-synth-time-async5}} vary temporal structure in the data (fixed vs.\ random intervals between events, short vs.\ long intervals between events, and synchronous vs.\ asynchronous predictors and responses).
\textbf{Supplementary Figures~\ref{fig:app-synth-multicollinearity-r00}--\ref{fig:app-synth-multicollinearity-r95}} vary pairwise predictor multicollinearity (pairwise correlation level $r$ = 0, 0.25, 0.5, 0.75, 0.9, and 0.95).
\textbf{Supplementary Figures~\ref{fig:app-synth-misspecification-e}--\ref{fig:app-synth-misspecification-g}} vary the underlying shape of the true response function (the true kernel is the probability density function of an exponential, normal, or shifted gamma distribution).
See \cite{shainschuler19} for full details about the synthetic data.

Across datasets and model configurations, CDRNNs accurately recover the ground truth IRFs.
Unsurprisingly, certain adverse conditions harm the accuracy of estimates, especially high levels of noise (e.g., error SD = 100) and high levels of predictor multicollinearity (e.g., all pairs of predictors correlated at $r = 0.9$).
Unlike the CDR models of \cite{shainschuler19}, which benefited in these evaluations from valid assumptions of IRF kernel family, linearity, stationarity, and homoscedasticity, the CDRNN models evaluated here had no such foreknowledge, and instead had to learn these features of the response from data.
The fact that they do so successfully supports the general applicability of the CDRNN approach to analyzing continuous-time dynamical systems, even if the full power of a CDRNN may not be needed for a given modeling task.

Because CDRNNs are neural networks that stochastically optimize a highly non-convex objective, we additionally use the synthetic datasets to investigate the degree to which estimates differ across replicates of the CDRNN base model with different random seeds.
As shown, estimates are almost indistinguishable in each synthetic task, supporting a high degree of consistency across replicates.

\FloatBarrier

\section{Full Results: Cognitive Experiments}
\label{app:cognitive}

Here we present the full results from all cognitive datasets evaluated in this study: Dundee (eye-tracking, ET), Natural Stories (self-paced reading, SPR), and Natural Stories (fMRI).


\subsection{Dundee (Eye-Tracking)}

In-sample and out-of-sample predictive performance on scan path, first pass, and go-past durations in Dundee are given in \textbf{Supplementary Tables~\ref{tab:app-dundee-sp-ll-bakeoff}--\ref{tab:app-dundee-gp-ll-bakeoff}}.
Effect estimates in Dundee are plotted in \textbf{Supplementary Figures~\ref{fig:app-dundee-raw-sp}--\ref{fig:app-dundee-log-gp}}.
Estimates are plausible and highly consistent across model variants.

\begin{table}
    \footnotesize
    \sffamily
    \centering
    \begin{tabular}{r|ccc|ccc}
        & \multicolumn{3}{|c}{Dundee Scan Path Duration (ms)} & \multicolumn{3}{|c}{Dundee Scan Path Duration (log ms)}\\
        Model & Train & Expl & Test & Train & Expl & Test\\
        \hline
        LME & -779299\textsuperscript{\textdagger} & -391437\textsuperscript{\textdagger} & -391996\textsuperscript{\textdagger} & -70203\textsuperscript{\textdagger} & -35380\textsuperscript{\textdagger} & -35331\textsuperscript{\textdagger} \\
        LME-S & --- & --- & --- & --- & --- & --- \\
        GAM & -778411 & -391105 & -391653 & -69660 & -35169 & -35151 \\
        GAM-S & -777822 & -390867 & -391461 & -68889 & -34839 & -34840 \\
        GAMLSS & -769269 & -387376 & -387433 & -65818 & -33319 & -33341 \\
        GAMLSS-S & \textit{-766609} & \textit{-386190} & \textit{-387617} & \textit{-65203} & \textit{-33065} & \textit{-33130} \\
        CDR & -778839 & -391217 & -391891 & -69559 & -35066 & -35118 \\
        \hline
        CDRNN base & -759754 & -385571 & \textbf{-386209} & -62253 & -31809 & \textbf{-31832}\\
        --Nonlinear & -764022 & -385926 & --- & -63625 & -32488 & ---\\
        --Nonstationary & -760876 & -384995 & --- & -62414 & -31858 & ---\\
        --Heteroscedastic & -777779 & -390825 & --- & -68305 & -34588 & ---\\
        +RNN & -756290 & -385808 & --- & -61663 & -31765 & ---\\
        Units $\div$ 2 (16) & -762452 & -384804 & --- & -63172 & -32168 & ---\\
        Units $\times$ 2 (64) & \textit{\textbf{-755227}} & -386670 & --- & -62006 & -31770 & ---\\
        Layers - 1 (1) & -760365 & -384948 & --- & -62841 & -32178 & ---\\
        Layers + 1 (3) & -756745 & -385449 & --- & -62689 & -31963 & ---\\
        Weight Reg $\div$ 5 (1) & -756150 & -386446 & --- & -61631 & -31900 & ---\\
        Weight Reg $\times$ 5 (25) & -764699 & -384783 & --- & -63320 & -32141 & ---\\
        Ranef Reg $\div$ 10 (1) & -758189 & -386038 & --- & -61740 & -31896 & ---\\
        Ranef Reg $\times$ 10 (100) & -760369 & -385477 & --- & -62212 & \textit{\textbf{-31750}} & ---\\
        Dropout $\div$ 2 (0.05) & -755535 & -387555 & --- & \textit{\textbf{-61508}} & -31826 & ---\\
        Dropout $\times$ 2 (0.2) & -766793 & -386002 & --- & -64406 & -32649 & ---\\
        Learning Rate $\div$ 3 (0.001) & -758376 & -386336 & --- & -62048 & -31998 & ---\\
        Learning Rate $\times$ 3 (0.009) & -762212 & \textit{\textbf{-384560}} & --- & -62661 & -31898 & ---\\
        Batch Size $\div$ 2 (512) & -760408 & -384876 & --- & -62282 & -31881 & ---\\
        Batch Size $\times$ 2 (2048) & -758487 & -384933 & --- & -62168 & -31879 & ---\\
    \end{tabular}
    \caption{\textbf{Dundee (scan path duration likelihood).} Log likelihood from CDRNN vs.\ linear mixed-effects (LME), generalized additive model (GAM), and generalized additive model for location, scale, and shape (GAMLSS) baselines with and without three additional spillover positions (-S) to help capture delayed effects, as well as kernel-based (non-neural) CDR (LME, GAM, and CDR performance as reported in \textit{\citen{shainschuler19}}).
    LME baselines show the marginal likelihood for the training set (the default likelihood implemented by the \texttt{lme4} package).
    All other likelihoods are conditional on the fitted model.
    Estimates from LME-S are omitted because training exceeded the two-week maximum runtime permitted by our compute resource.
    CDRNN variants add recurrence (+RNN) and modify the number of Units (Units), number of hidden layers (Layers), weight regularization strength (Reg), random effects regularization strength (RanReg), dropout rate (Dropout), learning rate (LR), and batch size (Batch).
    Of the CDRNN models, only CDRNN base is evaluated on the test set.
    Best-performing models within the sets of baseline and CDRNN models are shown in \textit{italics}. Best-performing overall models are shown in \textbf{bold}. Daggers (\textdagger) indicate convergence failures.}
    \label{tab:app-dundee-sp-ll-bakeoff}
\end{table}

\begin{table}
    \footnotesize
    \sffamily
    \centering
    \begin{tabular}{r|ccc|ccc}
        & \multicolumn{3}{|c}{Dundee First Pass Duration (ms)} & \multicolumn{3}{|c}{Dundee First Pass Duration (log ms)}\\
        Model & Train & Expl & Test & Train & Expl & Test\\
        \hline
        LME & -581489\textsuperscript{\textdagger} & -292478\textsuperscript{\textdagger} & -292359\textsuperscript{\textdagger} & -45022 & -22847 & -22689 \\
        LME-S & -581426\textsuperscript{\textdagger} & -292408\textsuperscript{\textdagger} & -292357\textsuperscript{\textdagger} & -44707\textsuperscript{\textdagger} & -22694\textsuperscript{\textdagger} & -22603\textsuperscript{\textdagger} \\
        GAM & -580851 & -292246 & -292097 & -44706 & -22744 & -22601 \\
        GAM-S & -580394 & -292048 & -291926 & -44115 & -22479 & -22334 \\
        GAMLSS & -572205 & -288244 & \textit{-288091} & -42622 & -21720 & -21588 \\
        GAMLSS-S & \textit{-571176} & \textit{-288228} & -288270 & \textit{-42117} & \textit{-21529} & \textit{-21416} \\
        CDR & -581093 & -292368 & -292287 & -44560 & -22805 & -22514\\
        \hline
        CDRNN base & -568093 & -287574 & \textbf{-287588} & -40389 & -20736 & \textbf{-20655}\\
        --Nonlinear & -571742 & -288394 & --- & -41358 & -21139 & ---\\
        --Nonstationary & -567573 & -287672 & --- & -40313 & -20726 & ---\\
        --Heteroscedastic & -579919 & -292049 & --- & -43886 & -22348 & ---\\
        +RNN & -565905 & -288309 & --- & -39750 & \textit{\textbf{-20604}} & ---\\
        Units $\div$ 2 (16) & -569042 & -288025 & --- & -40743 & -20838 & ---\\
        Units $\times$ 2 (64) & -565389 & -287831 & --- & -40387 & -20732 & ---\\
        Layers - 1 (1) & -568460 & -287740 & --- & -40335 & -20863 & ---\\
        Layers + 1 (3) & -568386 & -287866 & --- & -40414 & -20730 & ---\\
        Weight Reg $\div$ 5 (1) & -564244 & -288287 & --- & \textit{\textbf{-39712}} & -20722 & ---\\
        Weight Reg $\times$ 5 (25) & -569491 & \textit{\textbf{-287459}} & --- & -40913 & -20901 & ---\\
        Ranef Reg $\div$ 10 (1) & \textbf{\textit{-564069}} & -288306 & --- & -39734 & -20706 & ---\\
        Ranef Reg $\times$ 10 (100) & -566034 & -288112 & --- & -40328 & -20747 & ---\\
        Dropout $\div$ 2 (0.05) & -565890 & -287747 & --- & -39924 & -20707 & ---\\
        Dropout $\times$ 2 (0.2) & -570617 & -287467 & --- & -41311 & -21007 & ---\\
        Learning Rate $\div$ 3 (0.001) & -568584 & -287635 & --- & -40174 & -20715 & ---\\
        Learning Rate $\times$ 3 (0.009) & -569433 & -287480 & --- & -40800 & -20852 & ---\\
        Batch Size $\div$ 2 (512) & -569016 & -287829 & --- & -40497 & -20788 & ---\\
        Batch Size $\times$ 2 (2048) & -567679 & -288099 & --- & -40083 & -20685 & ---\\
    \end{tabular}
    \caption{\textbf{Dundee (first pass duration likelihood).} Log likelihood from CDRNN vs.\ linear mixed-effects (LME), generalized additive model (GAM), and generalized additive model for location, scale, and shape (GAMLSS) baselines with and without three additional spillover positions (-S) to help capture delayed effects, as well as kernel-based (non-neural) CDR (LME, GAM, and CDR performance as reported in \textit{\citen{shainschuler19}}).
    LME baselines show the marginal likelihood for the training set (the default likelihood implemented by the \texttt{lme4} package).
    All other likelihoods are conditional on the fitted model.
    CDRNN variants add recurrence (+RNN) and modify the number of Units (Units), number of hidden layers (Layers), weight regularization strength (Reg), random effects regularization strength (RanReg), dropout rate (Dropout), learning rate (LR), and batch size (Batch).
    Of the CDRNN models, only CDRNN base is evaluated on the test set.
    Best-performing models within the sets of baseline and CDRNN models are shown in \textit{italics}. Best-performing overall models are shown in \textbf{bold}. Daggers (\textdagger) indicate convergence failures.}
    \label{tab:app-dundee-fp-ll-bakeoff}
\end{table}

\begin{table}
    \footnotesize
    \sffamily
    \centering
    \begin{tabular}{r|ccc|ccc}
        & \multicolumn{3}{|c}{Dundee Go-Past Duration (ms)} & \multicolumn{3}{|c}{Dundee Go-Past Duration (log ms)}\\
        Model & Train & Expl & Test & Train & Expl & Test\\
        \hline
        LME & -636316 & -315308\textsuperscript{\textdagger} & -317511\textsuperscript{\textdagger} & -59556\textsuperscript{\textdagger} & -29690\textsuperscript{\textdagger} & -29689\textsuperscript{\textdagger} \\
        LME-S & -636199\textsuperscript{\textdagger} & -315280\textsuperscript{\textdagger} & -317546\textsuperscript{\textdagger} & -59220\textsuperscript{\textdagger} & -29522\textsuperscript{\textdagger} & -29557\textsuperscript{\textdagger} \\
        GAM & -635991 & -315169 & -317376 & -59212 & -29562 & -29573 \\
        GAM-S &  -635454 & -315285 & -317086 & -58549 & -29277 & -29287 \\
        GAMLSS & -624191 & -309765 & -311138 & -57494 & -28715 & -28728 \\
        GAMLSS-S & \textit{-623189} & \textit{-309534} & \textit{-310788} & \textit{-56876} & \textit{-28476} & \textit{-28533} \\
        CDR & -635180 & -316198 & -315934 & -59006 & -29447 & -29484 \\
        \hline
        CDRNN base & -608836 & -306662 & \textbf{-308434} & -54552 & -27509 & \textbf{-27551}\\
        --Nonlinear & -615024 & -306674 & --- & -55768 & -27947 & ---\\
        --Nonstationary & -608761 & -305773 & --- & -54602 & \textit{\textbf{-27468}} & ---\\
        --Heteroscedastic & -634122 & -313797 & --- & -58199 & -29067 & ---\\
        +RNN & -597059 & -307338 & --- & -54365 & -27494 & ---\\
        Units $\div$ 2 (16) & -609577 & -306199 & --- & -55064 & -27576 & ---\\
        Units $\times$ 2 (64) & -599103 & -308324 & --- & -54837 & -27556 & ---\\
        Layers - 1 (1) & -609947 & -306182 & --- & -54984 & -27718 & ---\\
        Layers + 1 (3) & -608017 & -305609 & --- & -54953 & -27578 & ---\\
        Weight Reg $\div$ 5 (1) & \textit{\textbf{-600313}} & -308170 & --- & \textit{\textbf{-53776}} & -27514 & ---\\
        Weight Reg $\times$ 5 (25) & -610347 & -305665 & --- & -55389 & -27752 & ---\\
        Ranef Reg $\div$ 10 (1) & -605305 & -306491 & --- & -54180 & -27540 & ---\\
        Ranef Reg $\times$ 10 (100) & -606665 & -305867 & --- & -54793 & -27600 & ---\\
        Dropout $\div$ 2 (0.05) & -600600 & -308092 & --- & -54111 & -27503 & ---\\
        Dropout $\times$ 2 (0.2) & -612388 & -305737 & --- & -55724 & -27842 & ---\\
        Learning Rate $\div$ 3 (0.001) & -605416 & -308270 & --- & -54534 & -27526 & ---\\
        Learning Rate $\times$ 3 (0.009) & -609569 & \textit{\textbf{-305508}} & --- & -55052 & -27638 & ---\\
        Batch Size $\div$ 2 (512) & -606452 & -306474 & --- & -54796 & -27547 & ---\\
        Batch Size $\times$ 2 (2048) & -605107 & -306633 & --- & -54729 & -27597 & ---\\
    \end{tabular}
    \caption{\textbf{Dundee (go-past duration likelihood).} Log likelihood from CDRNN vs.\ linear mixed-effects (LME), generalized additive model (GAM), and generalized additive model for location, scale, and shape (GAMLSS) baselines with and without three additional spillover positions (-S) to help capture delayed effects, as well as kernel-based (non-neural) CDR (LME, GAM, and CDR performance as reported in \textit{\citen{shainschuler19}}).
    LME baselines show the marginal likelihood for the training set (the default likelihood implemented by the \texttt{lme4} package).
    All other likelihoods are conditional on the fitted model.
    CDRNN variants add recurrence (+RNN) and modify the number of Units (Units), number of hidden layers (Layers), weight regularization strength (Reg), random effects regularization strength (RanReg), dropout rate (Dropout), learning rate (LR), and batch size (Batch).
    Of the CDRNN models, only CDRNN base is evaluated on the test set.
    Best-performing models within the sets of baseline and CDRNN models are shown in \textit{italics}. Best-performing overall models are shown in \textbf{bold}. Daggers (\textdagger) indicate convergence failures.}
    \label{tab:app-dundee-gp-ll-bakeoff}
\end{table}

\begin{figure}

    \footnotesize
    \sffamily
    \centering
    
    \textbf{\Large Dundee (Scan Path Duration)}
    
    \vspace{1em}
    
    \begin{subfigure}[t]{0.49\textwidth}
        \centering
        \makebox[0.49\textwidth]{\centering Base}%
        \makebox[0.49\textwidth]{\centering + RNN}
        \begin{overpic}[width=0.49\textwidth]{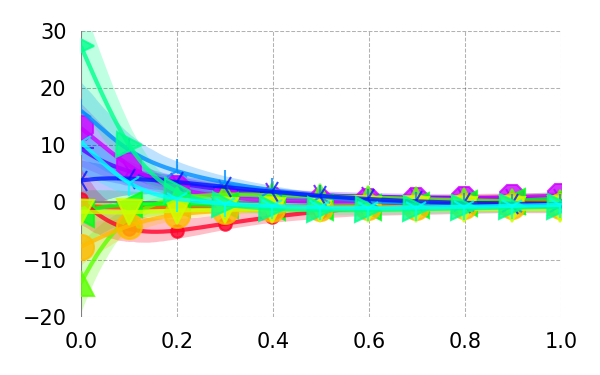}
            \put (-15,-80) {\rotatebox[origin=c]{90}{\large Change in Scan Path Duration (ms)}}
        \end{overpic}%
        \includegraphics[width=0.49\textwidth]{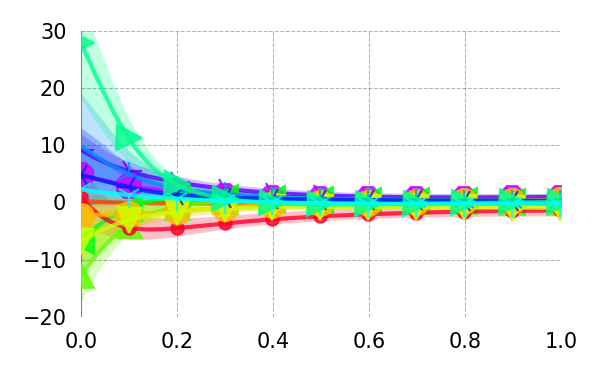}
    \end{subfigure}
    \begin{subfigure}[t]{0.49\textwidth}
        \centering
        \makebox[0.49\textwidth]{\centering Hidden Units $\div$ 2 (16)}%
        \makebox[0.49\textwidth]{\centering Hidden Units $\times$ 2 (64)}
        \includegraphics[width=0.49\textwidth]{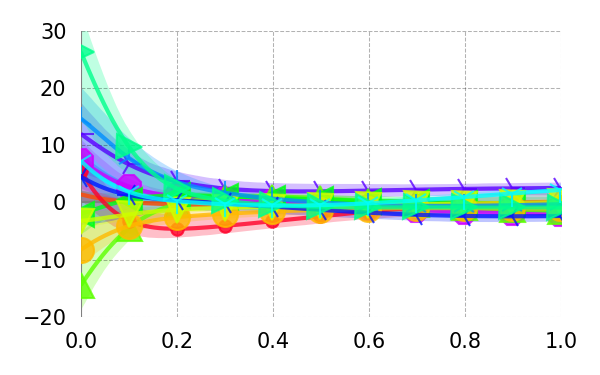}
        \includegraphics[width=0.49\textwidth]{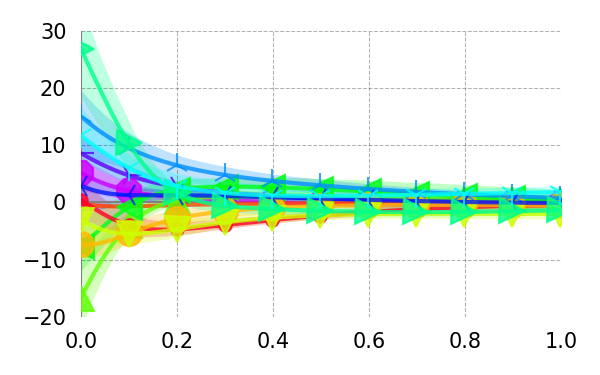}
    \end{subfigure}
    
    \begin{subfigure}[t]{0.49\textwidth}
        \centering
        \makebox[0.49\textwidth]{\centering Hidden Layers - 1 (1)}%
        \makebox[0.49\textwidth]{\centering Hidden Layers + 1 (3)}
        \includegraphics[width=0.49\textwidth]{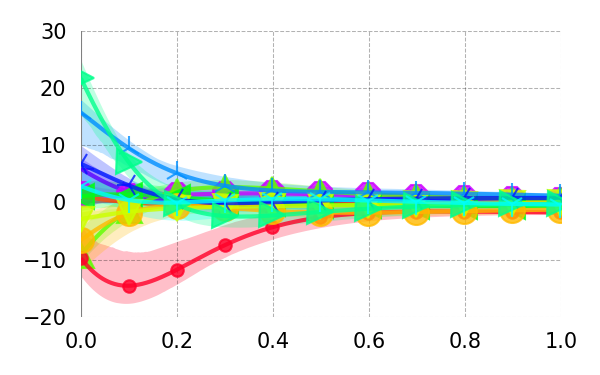}
        \includegraphics[width=0.49\textwidth]{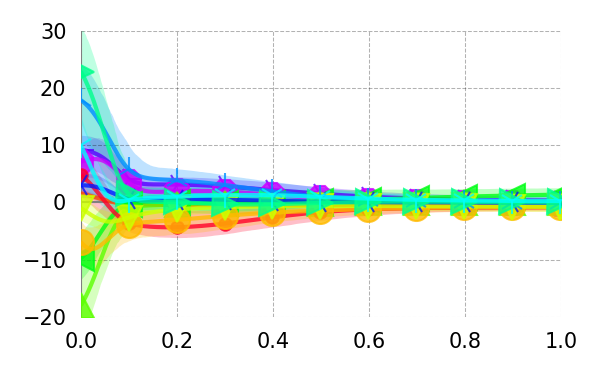}
    \end{subfigure}
    \begin{subfigure}[t]{0.49\textwidth}
        \centering
        \makebox[0.49\textwidth]{\centering Weight Reg $\div$ 5 (1)}%
        \makebox[0.49\textwidth]{\centering Weight Reg $\times$ 5 (5)}
        \includegraphics[width=0.49\textwidth]{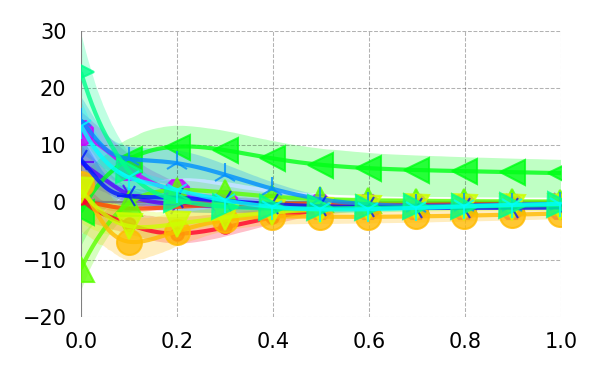}
        \includegraphics[width=0.49\textwidth]{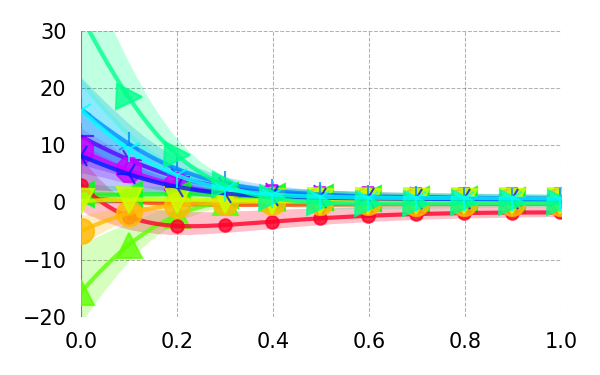}
    \end{subfigure}
    
    \begin{subfigure}[t]{0.49\textwidth}
        \centering
        \makebox[0.49\textwidth]{\centering Ranef Reg $\div$ 10 (1)}%
        \makebox[0.49\textwidth]{\centering Ranef Reg $\times$ 10 (100)}
        \includegraphics[width=0.49\textwidth]{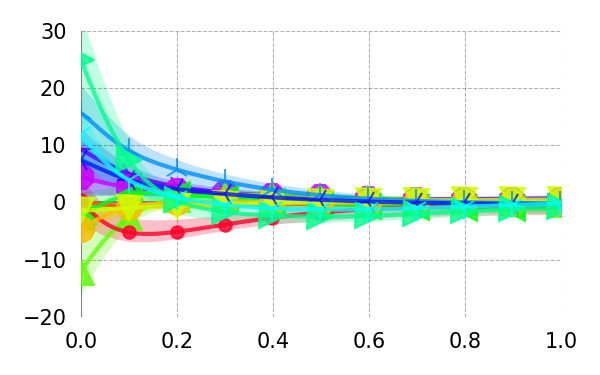}
        \includegraphics[width=0.49\textwidth]{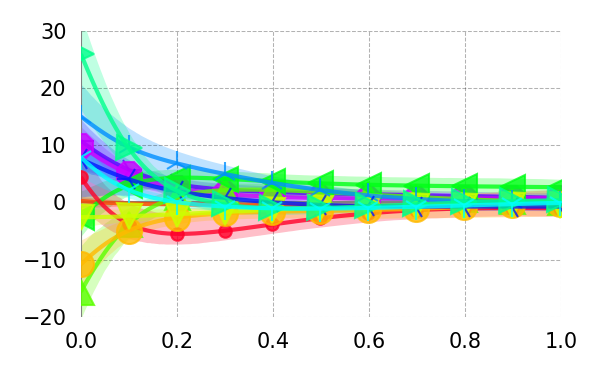}
    \end{subfigure}
    \begin{subfigure}[t]{0.49\textwidth}
        \centering
        \makebox[0.49\textwidth]{\centering Dropout $\div$ 2 (0.05)}%
        \makebox[0.49\textwidth]{\centering Dropout $\times$ 2 (0.2)}
        \includegraphics[width=0.49\textwidth]{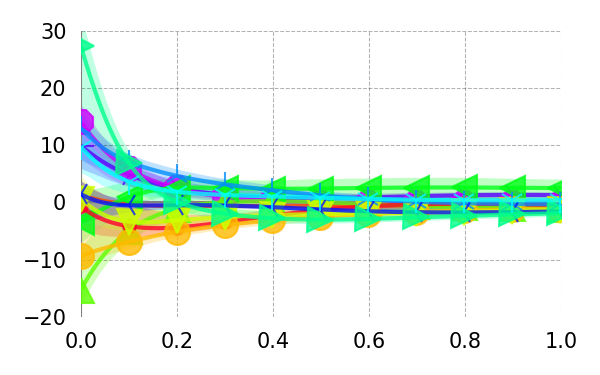}
        \includegraphics[width=0.49\textwidth]{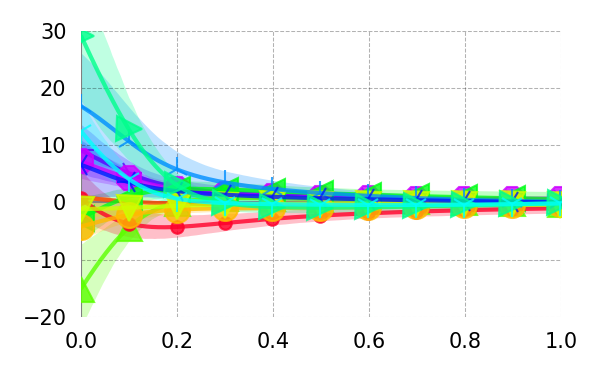}
    \end{subfigure}
    
    \begin{subfigure}[t]{0.49\textwidth}
        \centering
        \makebox[0.49\textwidth]{\centering Learning Rate $\div$ 3 (0.001)}%
        \makebox[0.49\textwidth]{\centering Learning Rate $\times$ 3 (0.009)}
        \includegraphics[width=0.49\textwidth]{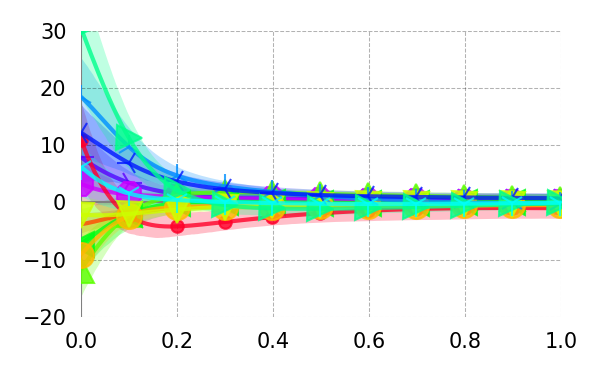}
        \includegraphics[width=0.49\textwidth]{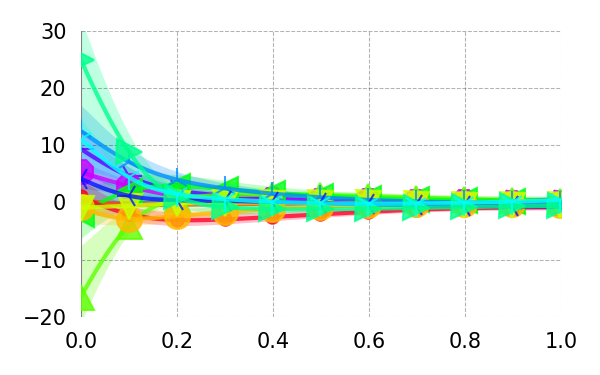}
    \end{subfigure}
    \begin{subfigure}[t]{0.49\textwidth}
        \centering
        \makebox[0.49\textwidth]{\centering Batch Size $\div$ 2 (512)}%
        \makebox[0.49\textwidth]{\centering Batch Size $\times$ 2 (2048)}
        \includegraphics[width=0.49\textwidth]{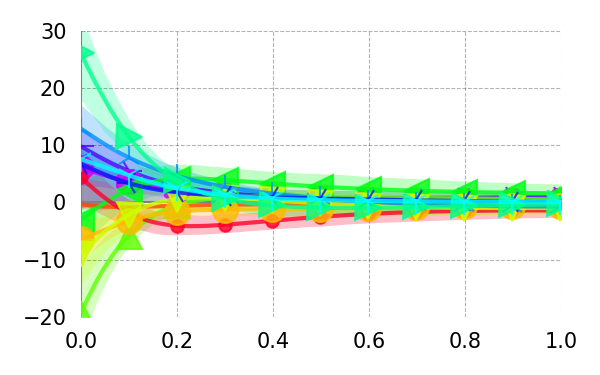}
        \includegraphics[width=0.49\textwidth]{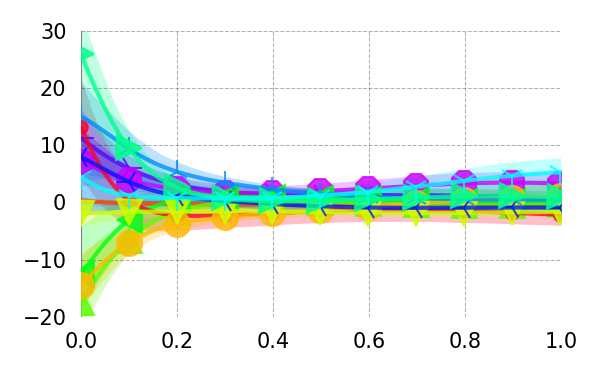}
    \end{subfigure}
    
    \vspace{1em}
    
    \textbf{\large Consistency}
    
    Standard deviation of exploratory set log-likelihood: 246
    
    \vspace{0.5em}
    
    \begin{subfigure}[t]{0.19\textwidth}
        \centering
        \makebox[0.9\textwidth]{\centering Rep 1}
        Expl LL: -385571
        \includegraphics[width=\textwidth]{{results_cdrnn_journal_dundee_raw_sp_CDR_main_irf_univariate_fdurSPsummed_mean_mc}.png}
    \end{subfigure}
    \begin{subfigure}[t]{0.19\textwidth}
        \centering
        \makebox[0.9\textwidth]{\centering Rep 2}
        Expl LL: -384988
        \includegraphics[width=\textwidth]{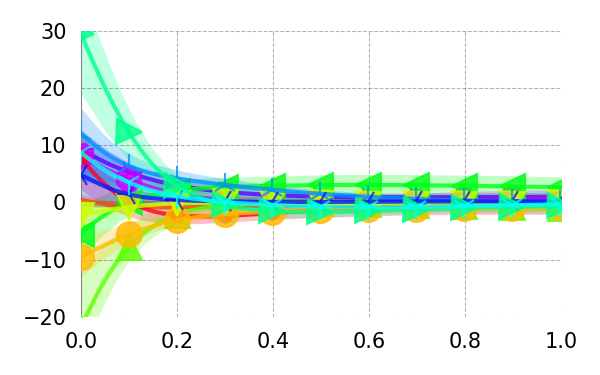}
    \end{subfigure}
    \begin{subfigure}[t]{0.19\textwidth}
        \centering
        \makebox[0.9\textwidth]{\centering Rep 3}
        Expl LL: -385681
        \includegraphics[width=\textwidth]{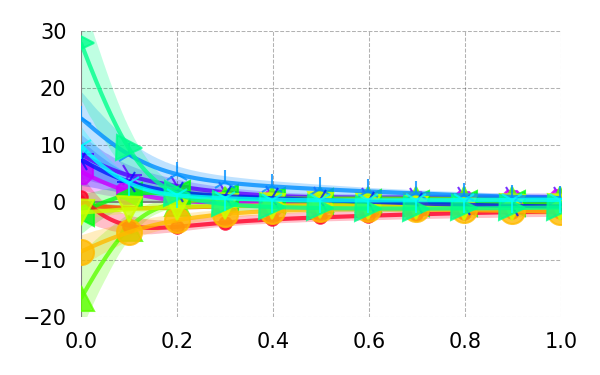}
    \end{subfigure}
    \begin{subfigure}[t]{0.19\textwidth}
        \centering
        \makebox[0.9\textwidth]{\centering Rep 4}
        Expl LL: -385253
        \includegraphics[width=\textwidth]{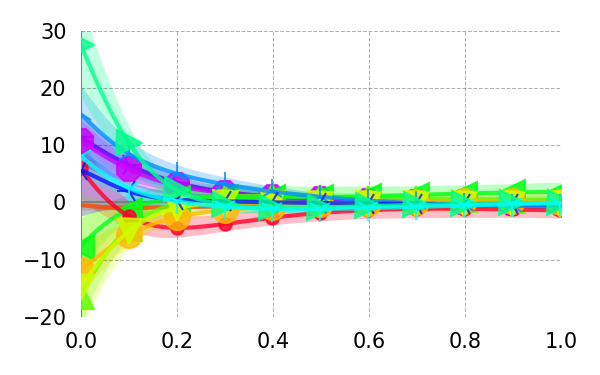}
    \end{subfigure}
    \begin{subfigure}[t]{0.19\textwidth}
        \centering
        \makebox[0.9\textwidth]{\centering Rep 5}
        Expl LL: -385458
        \includegraphics[width=\textwidth]{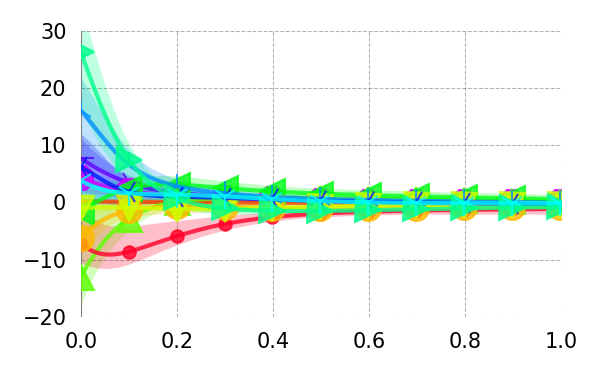}
    \end{subfigure}
    
    {\large Delay (s)}
    
    \vspace{0.5em}
    
    {
    {\kone~rate \hspace{1em}}
    {\ktwo~notregression}\\
    {\kthree~saccade~length \hspace{1em}}
    {\kfive~previous~was~fixated \hspace{1em}}
    {\kseven~word~length \hspace{1em}}
    {\knine~unigram~surprisal \hspace{1em}}\\
    {\keleven~5-gram~surprisal}\\
    {\kfour~saccade~length~(+reg) \hspace{1em}}
    {\ksix~previous~was~fixated~(+reg) \hspace{1em}}
    {\keight~word~length~(+reg) \hspace{1em}}
    {\kten~unigram~surprisal~(+reg) \hspace{1em}}\\
    {\ktwelve~5-gram~surprisal~(+reg)}
    }
    
    \vspace{0.5em}
    
    \caption{\textbf{Dundee (scan path duration):} Univariate CDRNN IRF estimates from the Dundee eye-tracking corpus (scan path duration). Results using base hyperparameters are compared to estimates from models that deviate from the base in some dimension. Plots under ``Consistency'' show estimates from five replicates of the ``base'' configuration, where ``Rep 1'' is the same model as ``base'' above, replotted for ease of comparison.}
    \label{fig:app-dundee-raw-sp}
    
\end{figure}

\begin{figure}

    \footnotesize
    \sffamily
    \centering
    
    \textbf{\Large Dundee (Log Scan Path Duration)}
    
    \vspace{1em}

    \begin{subfigure}[t]{0.49\textwidth}
        \centering
        \makebox[0.49\textwidth]{\centering Base}%
        \makebox[0.49\textwidth]{\centering + RNN}
        \begin{overpic}[width=0.49\textwidth]{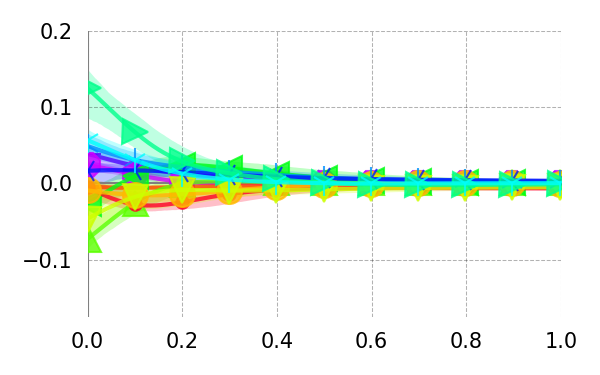}
            \put (-15,-80) {\rotatebox[origin=c]{90}{\large Change in Scan Path Duration (log-ms)}}
        \end{overpic}%
        \includegraphics[width=0.49\textwidth]{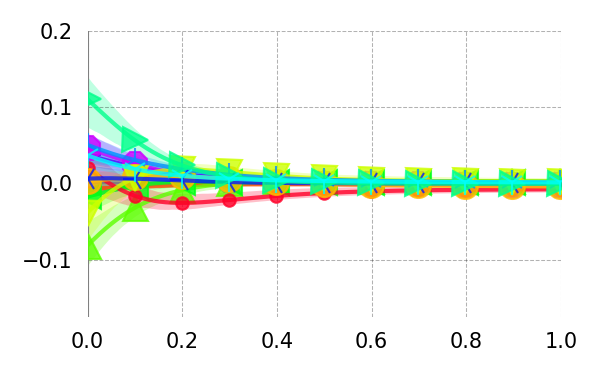}
    \end{subfigure}
    \begin{subfigure}[t]{0.49\textwidth}
        \centering
        \makebox[0.49\textwidth]{\centering Hidden Units $\div$ 2 (16)}%
        \makebox[0.49\textwidth]{\centering Hidden Units $\times$ 2 (64)}
        \includegraphics[width=0.49\textwidth]{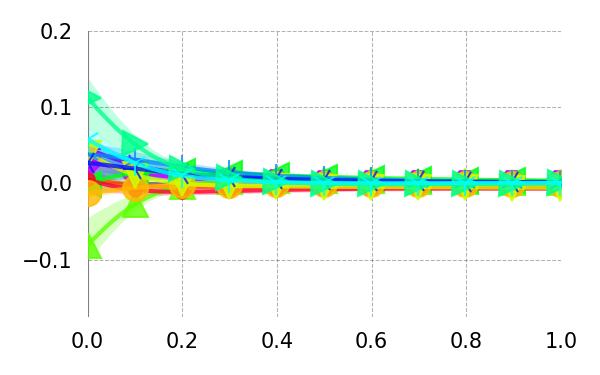}
        \includegraphics[width=0.49\textwidth]{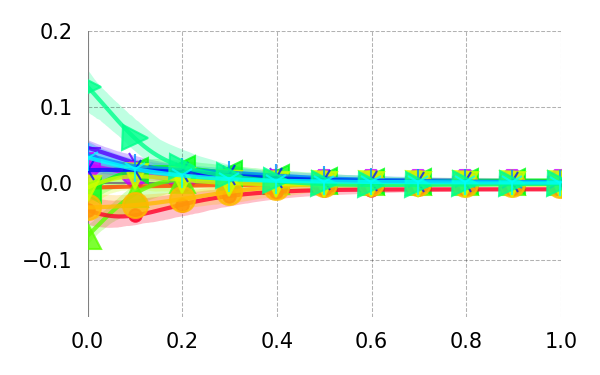}
    \end{subfigure}
    
    \begin{subfigure}[t]{0.49\textwidth}
        \centering
        \makebox[0.49\textwidth]{\centering Hidden Layers - 1 (1)}%
        \makebox[0.49\textwidth]{\centering Hidden Layers + 1 (3)}
        \includegraphics[width=0.49\textwidth]{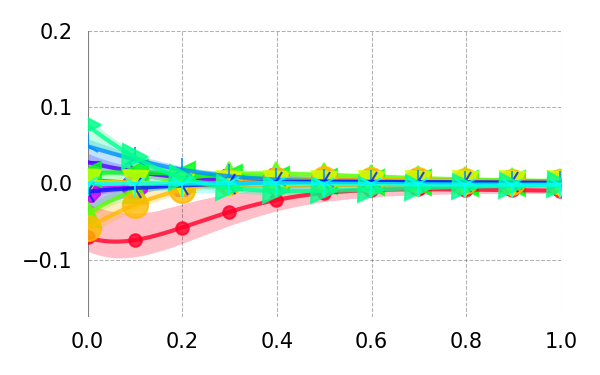}
        \includegraphics[width=0.49\textwidth]{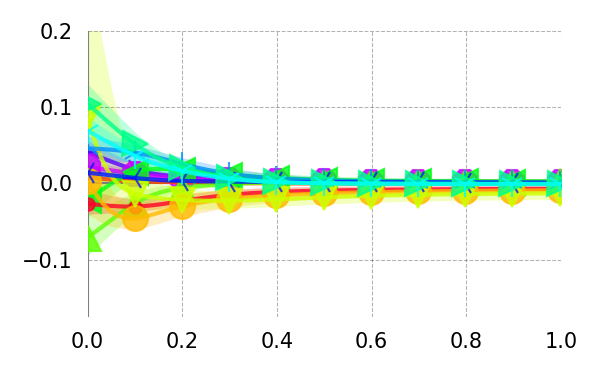}
    \end{subfigure}
    \begin{subfigure}[t]{0.49\textwidth}
        \centering
        \makebox[0.49\textwidth]{\centering Weight Reg $\div$ 5 (1)}%
        \makebox[0.49\textwidth]{\centering Weight Reg $\times$ 5 (5)}
        \includegraphics[width=0.49\textwidth]{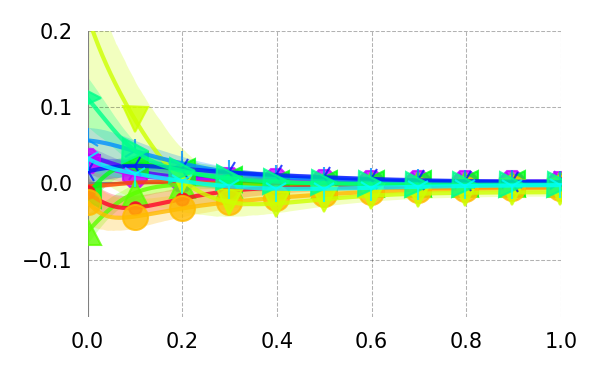}
        \includegraphics[width=0.49\textwidth]{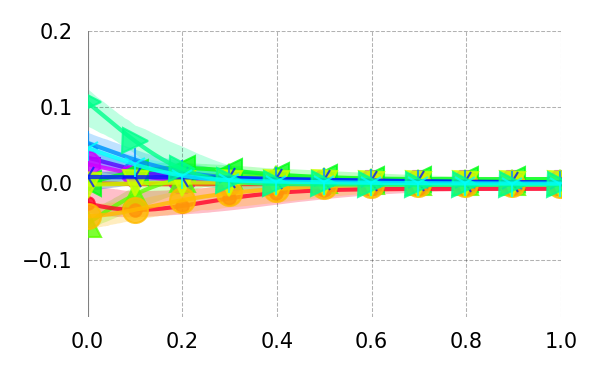}
    \end{subfigure}
    
    \begin{subfigure}[t]{0.49\textwidth}
        \centering
        \makebox[0.49\textwidth]{\centering Ranef Reg $\div$ 10 (1)}%
        \makebox[0.49\textwidth]{\centering Ranef Reg $\times$ 10 (100)}
        \includegraphics[width=0.49\textwidth]{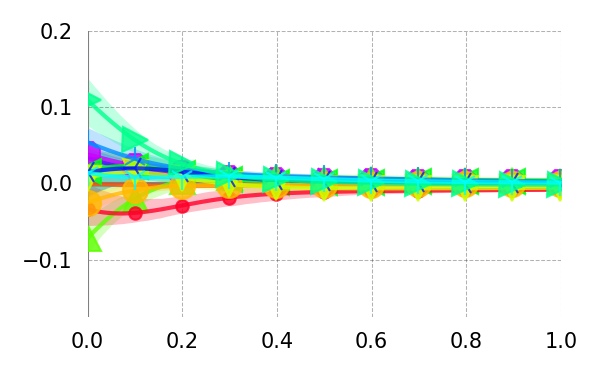}
        \includegraphics[width=0.49\textwidth]{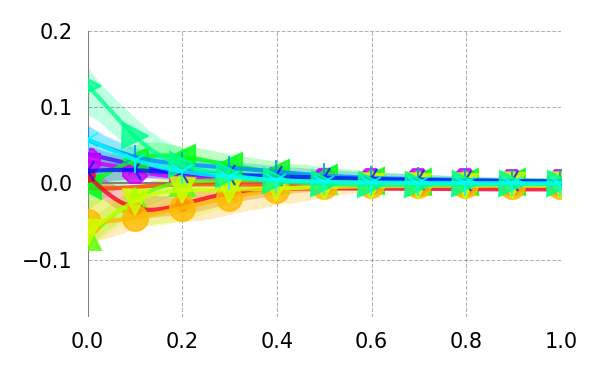}
    \end{subfigure}
    \begin{subfigure}[t]{0.49\textwidth}
        \centering
        \makebox[0.49\textwidth]{\centering Dropout $\div$ 2 (0.05)}%
        \makebox[0.49\textwidth]{\centering Dropout $\times$ 2 (0.2)}
        \includegraphics[width=0.49\textwidth]{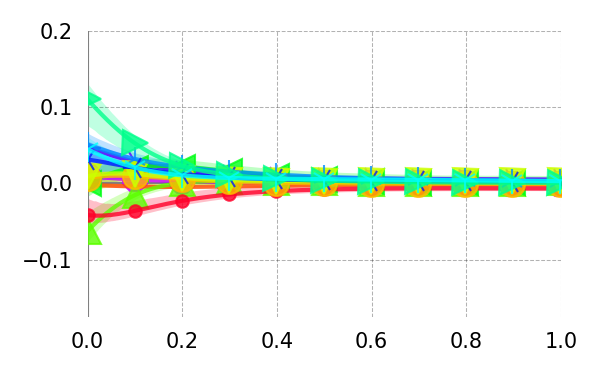}
        \includegraphics[width=0.49\textwidth]{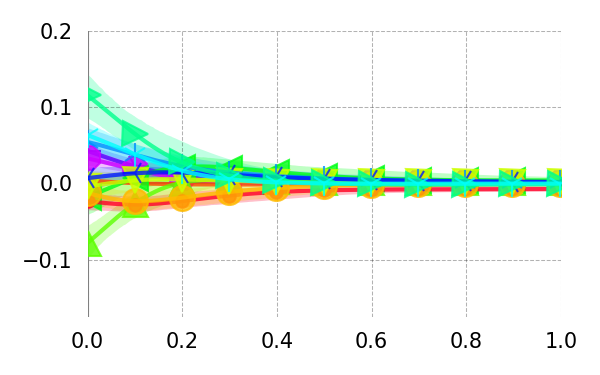}
    \end{subfigure}
    
    \begin{subfigure}[t]{0.49\textwidth}
        \centering
        \makebox[0.49\textwidth]{\centering Learning Rate $\div$ 3 (0.001)}%
        \makebox[0.49\textwidth]{\centering Learning Rate $\times$ 3 (0.009)}
        \includegraphics[width=0.49\textwidth]{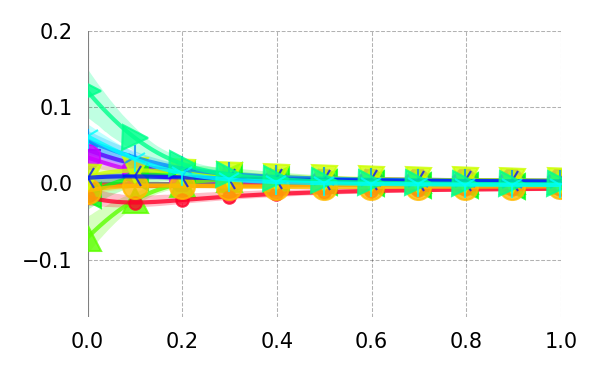}
        \includegraphics[width=0.49\textwidth]{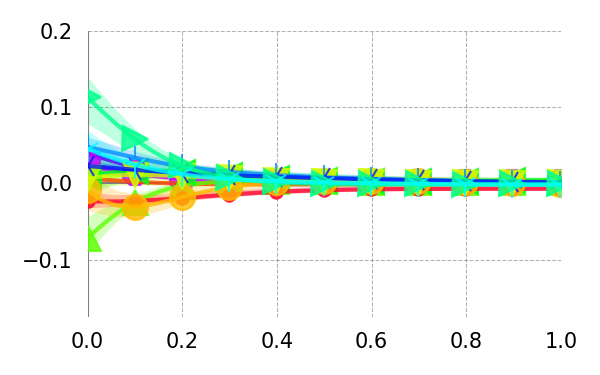}
    \end{subfigure}
    \begin{subfigure}[t]{0.49\textwidth}
        \centering
        \makebox[0.49\textwidth]{\centering Batch Size $\div$ 2 (512)}%
        \makebox[0.49\textwidth]{\centering Batch Size $\times$ 2 (2048)}
        \includegraphics[width=0.49\textwidth]{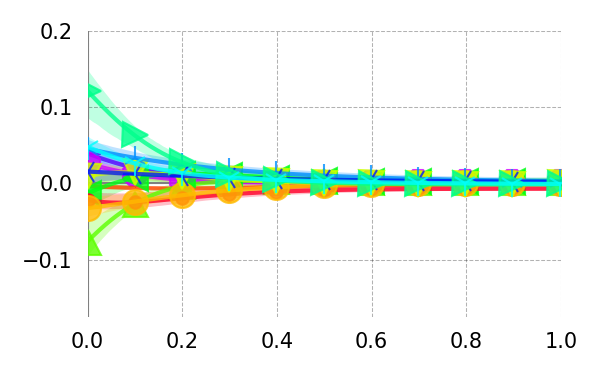}
        \includegraphics[width=0.49\textwidth]{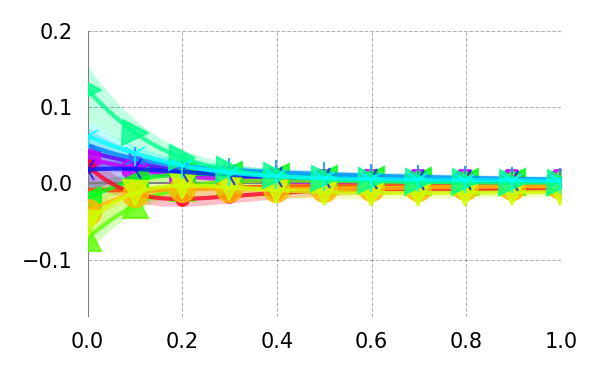}
    \end{subfigure}
    
    \vspace{1em}
    
    \textbf{\large Consistency}
    
    Standard deviation of exploratory set log-likelihood: 56
    
    \vspace{0.5em}
    
    \begin{subfigure}[t]{0.19\textwidth}
        \centering
        \makebox[0.9\textwidth]{\centering Rep 1}
        Expl LL: -31809
        \includegraphics[width=\textwidth]{{results_cdrnn_journal_dundee_log_sp_CDR_main_irf_univariate_log.fdurSPsummed._mean_mc}.png}
    \end{subfigure}
    \begin{subfigure}[t]{0.19\textwidth}
        \centering
        \makebox[0.9\textwidth]{\centering Rep 2}
        Expl LL: -31821
        \includegraphics[width=\textwidth]{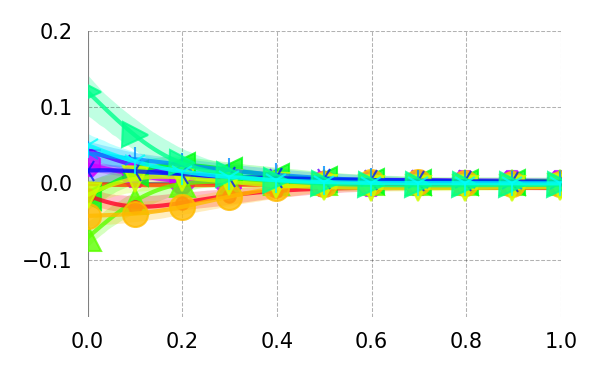}
    \end{subfigure}
    \begin{subfigure}[t]{0.19\textwidth}
        \centering
        \makebox[0.9\textwidth]{\centering Rep 3}
        Expl LL: -31922
        \includegraphics[width=\textwidth]{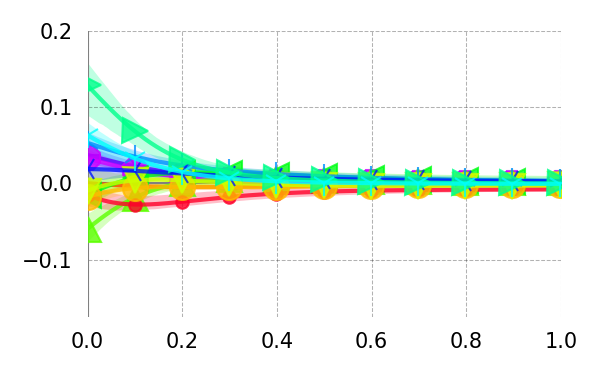}
    \end{subfigure}
    \begin{subfigure}[t]{0.19\textwidth}
        \centering
        \makebox[0.9\textwidth]{\centering Rep 4}
        Expl LL: -31761
        \includegraphics[width=\textwidth]{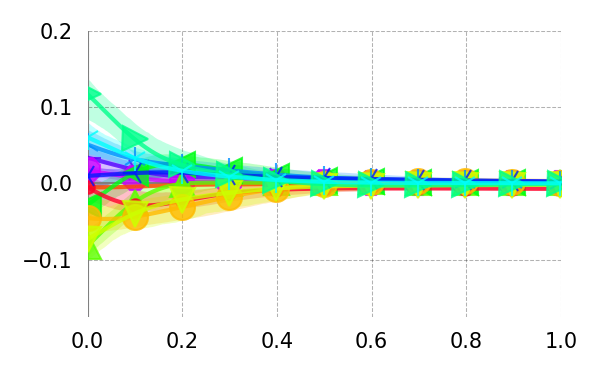}
    \end{subfigure}
    \begin{subfigure}[t]{0.19\textwidth}
        \centering
        \makebox[0.9\textwidth]{\centering Rep 5}
        Expl LL: -31781
        \includegraphics[width=\textwidth]{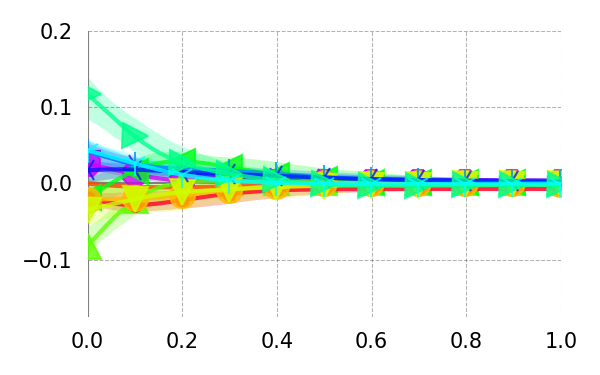}
    \end{subfigure}
    
    {\large Delay (s)}
    
    \vspace{0.5em}
    
    {
    {\kone~rate \hspace{1em}}
    {\ktwo~notregression}\\
    {\kthree~saccade~length \hspace{1em}}
    {\kfive~previous~was~fixated \hspace{1em}}
    {\kseven~word~length \hspace{1em}}
    {\knine~unigram~surprisal \hspace{1em}}\\
    {\keleven~5-gram~surprisal}\\
    {\kfour~saccade~length~(+reg) \hspace{1em}}
    {\ksix~previous~was~fixated~(+reg) \hspace{1em}}
    {\keight~word~length~(+reg) \hspace{1em}}
    {\kten~unigram~surprisal~(+reg) \hspace{1em}}\\
    {\ktwelve~5-gram~surprisal~(+reg)}
    }
    
    \vspace{0.5em}
    
    \caption{\textbf{Dundee (log scan path duration):} Univariate CDRNN IRF estimates from the Dundee eye-tracking corpus (log-transformed scan path duration). Results using base hyperparameters are compared to estimates from models that deviate from the base in some dimension. Plots under ``Consistency'' show estimates from five replicates of the ``base'' configuration, where ``Rep 1'' is the same model as ``base'' above, replotted for ease of comparison.}
    \label{fig:app-dundee-log-sp}
    
\end{figure}

\begin{figure}

    \footnotesize
    \sffamily
    \centering
    
    \textbf{\Large Dundee (First Pass Duration)}
    
    \vspace{1em}
    
    \begin{subfigure}[t]{0.49\textwidth}
        \centering
        \makebox[0.49\textwidth]{\centering Base}%
        \makebox[0.49\textwidth]{\centering + RNN}
        \begin{overpic}[width=0.49\textwidth]{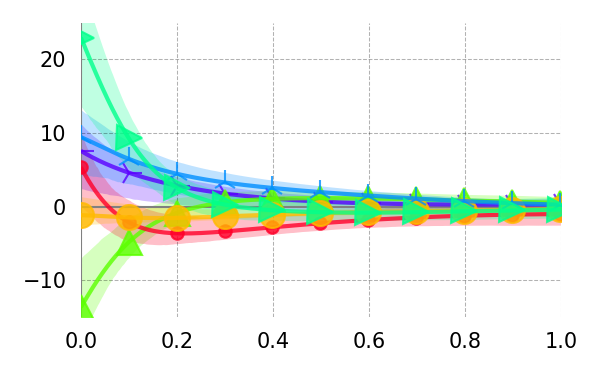}
            \put (-15,-80) {\rotatebox[origin=c]{90}{\large Change in First Pass Duration (ms)}}
        \end{overpic}%
        \includegraphics[width=0.49\textwidth]{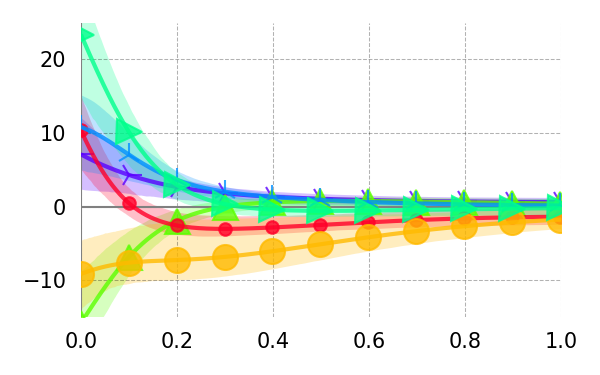}
    \end{subfigure}
    \begin{subfigure}[t]{0.49\textwidth}
        \centering
        \makebox[0.49\textwidth]{\centering Hidden Units $\div$ 2 (16)}%
        \makebox[0.49\textwidth]{\centering Hidden Units $\times$ 2 (64)}
        \includegraphics[width=0.49\textwidth]{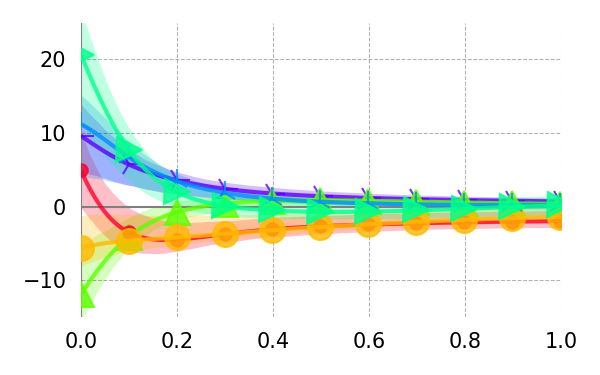}
        \includegraphics[width=0.49\textwidth]{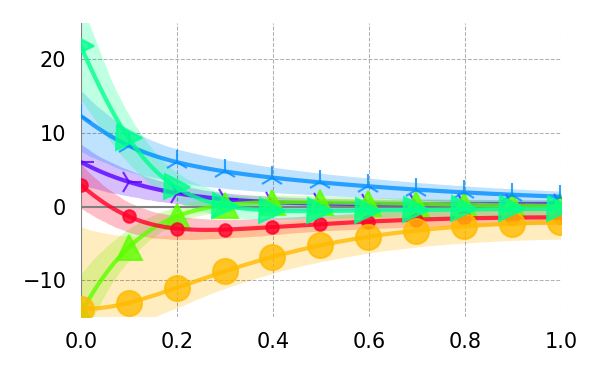}
    \end{subfigure}
    
    \begin{subfigure}[t]{0.49\textwidth}
        \centering
        \makebox[0.49\textwidth]{\centering Hidden Layers - 1 (1)}%
        \makebox[0.49\textwidth]{\centering Hidden Layers + 1 (3)}
        \includegraphics[width=0.49\textwidth]{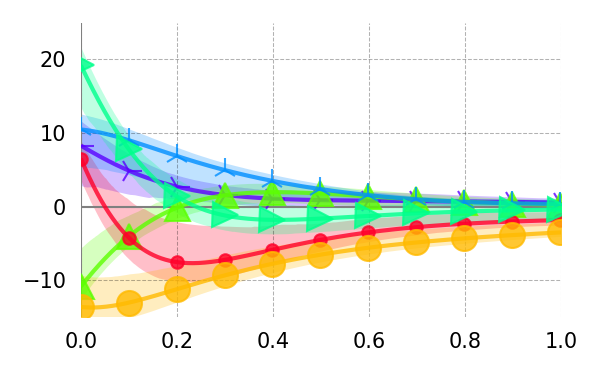}
        \includegraphics[width=0.49\textwidth]{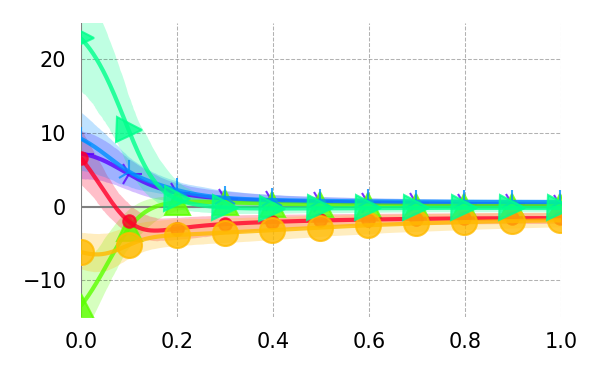}
    \end{subfigure}
    \begin{subfigure}[t]{0.49\textwidth}
        \centering
        \makebox[0.49\textwidth]{\centering Weight Reg $\div$ 5 (1)}%
        \makebox[0.49\textwidth]{\centering Weight Reg $\times$ 5 (5)}
        \includegraphics[width=0.49\textwidth]{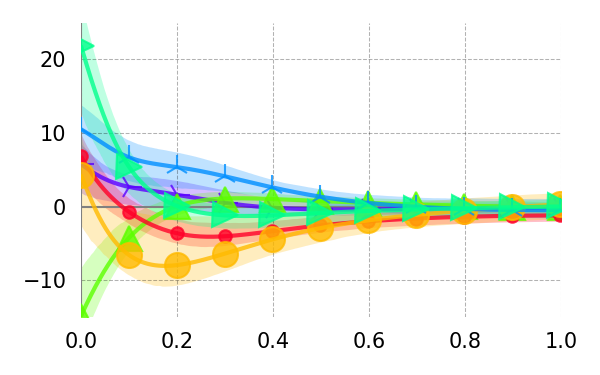}
        \includegraphics[width=0.49\textwidth]{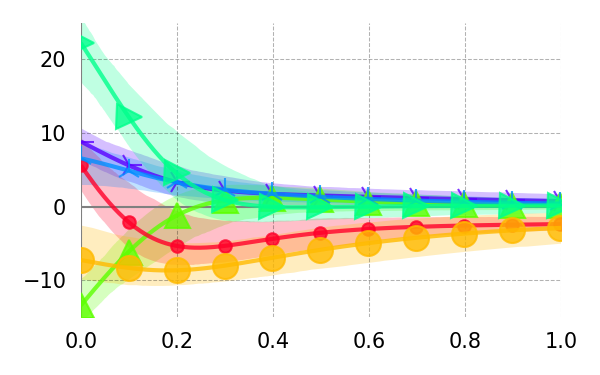}
    \end{subfigure}
    
    \begin{subfigure}[t]{0.49\textwidth}
        \centering
        \makebox[0.49\textwidth]{\centering Ranef Reg $\div$ 10 (1)}%
        \makebox[0.49\textwidth]{\centering Ranef Reg $\times$ 10 (100)}
        \includegraphics[width=0.49\textwidth]{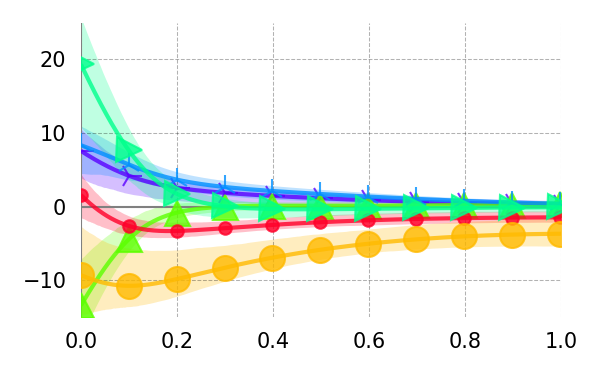}
        \includegraphics[width=0.49\textwidth]{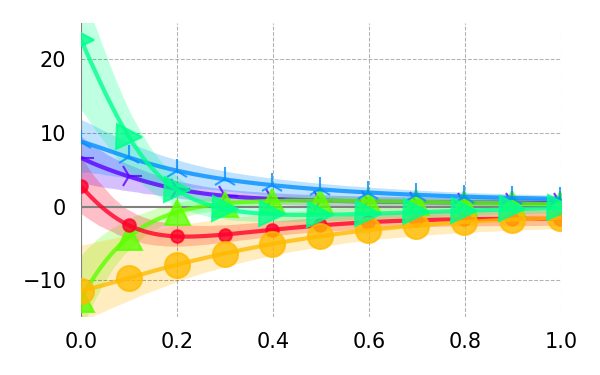}
    \end{subfigure}
    \begin{subfigure}[t]{0.49\textwidth}
        \centering
        \makebox[0.49\textwidth]{\centering Dropout $\div$ 2 (0.05)}%
        \makebox[0.49\textwidth]{\centering Dropout $\times$ 2 (0.2)}
        \includegraphics[width=0.49\textwidth]{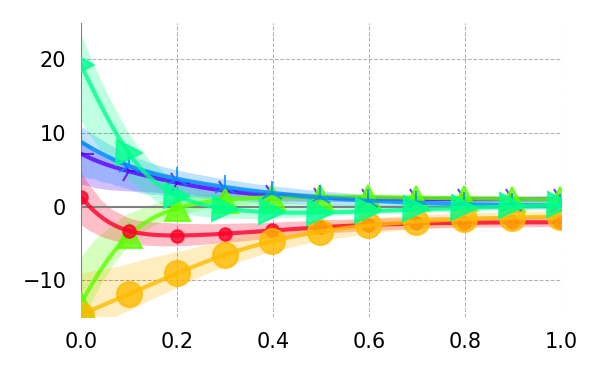}
        \includegraphics[width=0.49\textwidth]{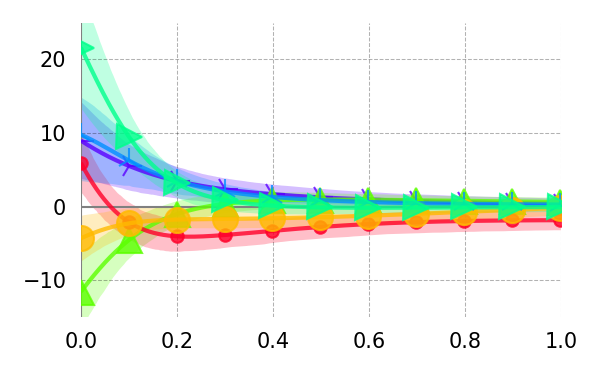}
    \end{subfigure}
    
    \begin{subfigure}[t]{0.49\textwidth}
        \centering
        \makebox[0.49\textwidth]{\centering Learning Rate $\div$ 3 (0.001)}%
        \makebox[0.49\textwidth]{\centering Learning Rate $\times$ 3 (0.009)}
        \includegraphics[width=0.49\textwidth]{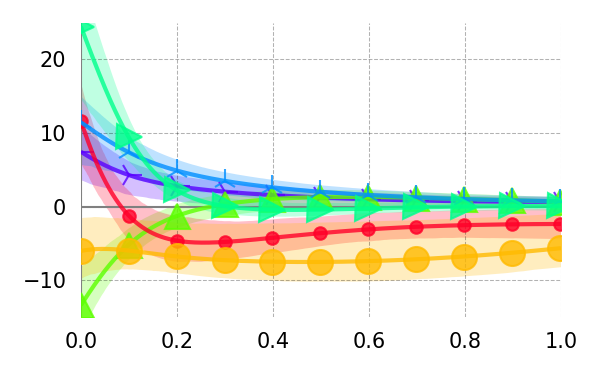}
        \includegraphics[width=0.49\textwidth]{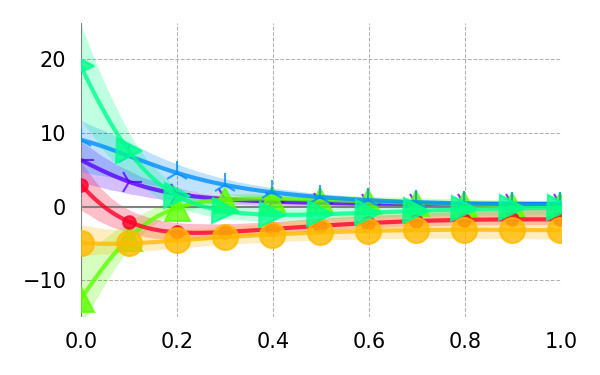}
    \end{subfigure}
    \begin{subfigure}[t]{0.49\textwidth}
        \centering
        \makebox[0.49\textwidth]{\centering Batch Size $\div$ 2 (512)}%
        \makebox[0.49\textwidth]{\centering Batch Size $\times$ 2 (2048)}
        \includegraphics[width=0.49\textwidth]{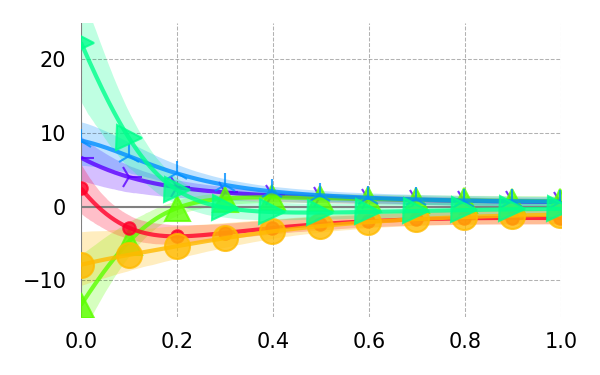}
        \includegraphics[width=0.49\textwidth]{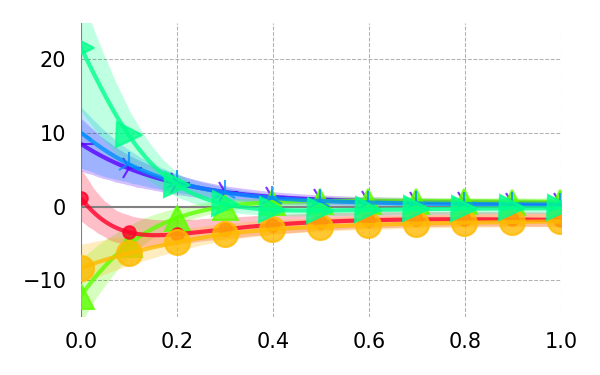}
    \end{subfigure}
    
    \vspace{1em}
    
    \textbf{\large Consistency}
    
    Standard deviation of exploratory set log-likelihood: 137
    
    \vspace{0.5em}
    
    \begin{subfigure}[t]{0.19\textwidth}
        \centering
        \makebox[0.9\textwidth]{\centering Rep 1}
        Expl LL: -287574
        \includegraphics[width=\textwidth]{{results_cdrnn_journal_dundee_raw_fp_CDR_main_irf_univariate_fdurFP_mean_mc}.png}
    \end{subfigure}
    \begin{subfigure}[t]{0.19\textwidth}
        \centering
        \makebox[0.9\textwidth]{\centering Rep 2}
        Expl LL: -287674
        \includegraphics[width=\textwidth]{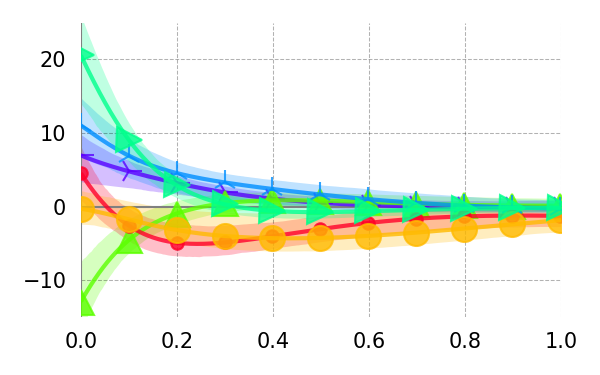}
    \end{subfigure}
    \begin{subfigure}[t]{0.19\textwidth}
        \centering
        \makebox[0.9\textwidth]{\centering Rep 3}
        Expl LL: -287641
        \includegraphics[width=\textwidth]{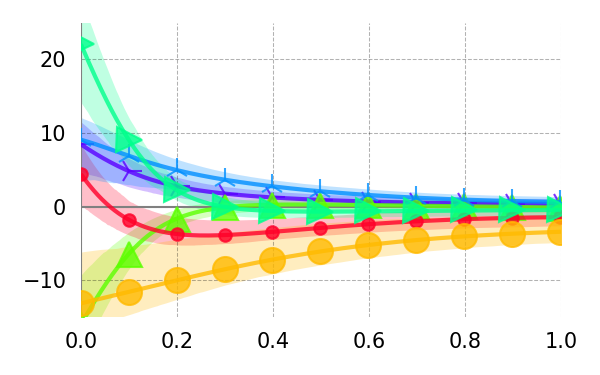}
    \end{subfigure}
    \begin{subfigure}[t]{0.19\textwidth}
        \centering
        \makebox[0.9\textwidth]{\centering Rep 4}
        Expl LL: -287942
        \includegraphics[width=\textwidth]{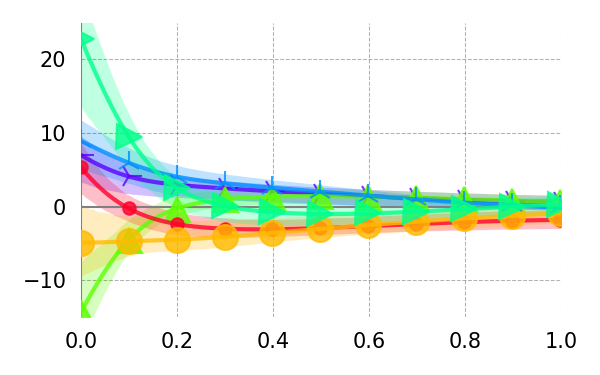}
    \end{subfigure}
    \begin{subfigure}[t]{0.19\textwidth}
        \centering
        \makebox[0.9\textwidth]{\centering Rep 5}
        Expl LL: -287849
        \includegraphics[width=\textwidth]{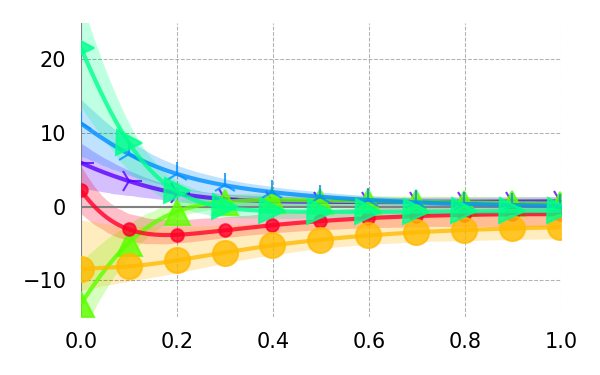}
    \end{subfigure}
    
    {\large Delay (s)}
    
    \vspace{0.5em}
    
    {
    {\kone~rate \hspace{1em}}
    {\kthree~saccade~length \hspace{1em}}
    {\kfive~previous~was~fixated \hspace{1em}}
    {\kseven~word~length \hspace{1em}}
    {\knine~unigram~surprisal \hspace{1em}}
    {\keleven~5-gram~surprisal}
    }
    
    \vspace{0.5em}
    
    \caption{\textbf{Dundee (first pass duration):} Univariate CDRNN IRF estimates from the Dundee eye-tracking corpus (first pass duration). Results using base hyperparameters are compared to estimates from models that deviate from the base in some dimension. Plots under ``Consistency'' show estimates from five replicates of the ``base'' configuration, where ``Rep 1'' is the same model as ``base'' above, replotted for ease of comparison.}
    \label{fig:app-dundee-raw-fp}
    
\end{figure}

\begin{figure}

    \footnotesize
    \sffamily    
    \centering
    
    \textbf{\Large Dundee (Log First Pass Duration)}
    
    \vspace{1em}
    
    \begin{subfigure}[t]{0.49\textwidth}
        \centering
        \makebox[0.49\textwidth]{\centering Base}%
        \makebox[0.49\textwidth]{\centering + RNN}
        \begin{overpic}[width=0.49\textwidth]{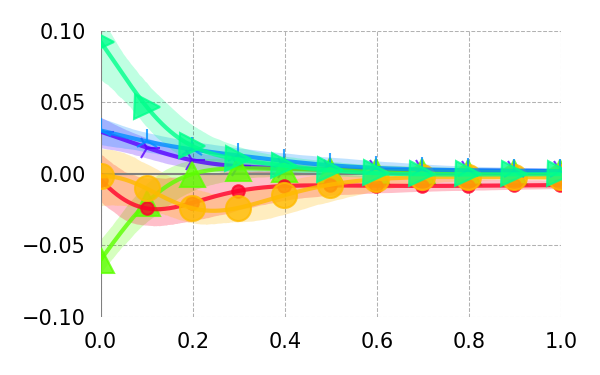}
            \put (-15,-80) {\rotatebox[origin=c]{90}{\large Change in First Pass Duration (log-ms)}}
        \end{overpic}%
        \includegraphics[width=0.49\textwidth]{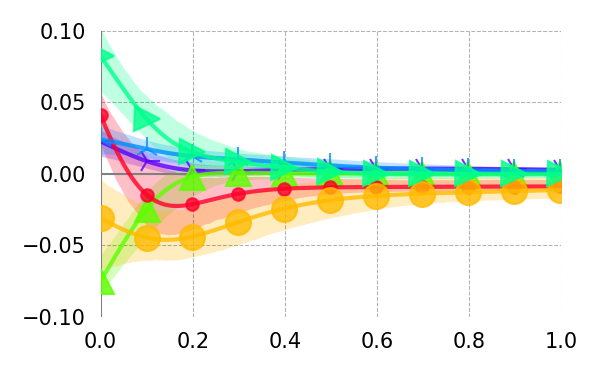}
    \end{subfigure}
    \begin{subfigure}[t]{0.49\textwidth}
        \centering
        \makebox[0.49\textwidth]{\centering Hidden Units $\div$ 2 (16)}%
        \makebox[0.49\textwidth]{\centering Hidden Units $\times$ 2 (64)}
        \includegraphics[width=0.49\textwidth]{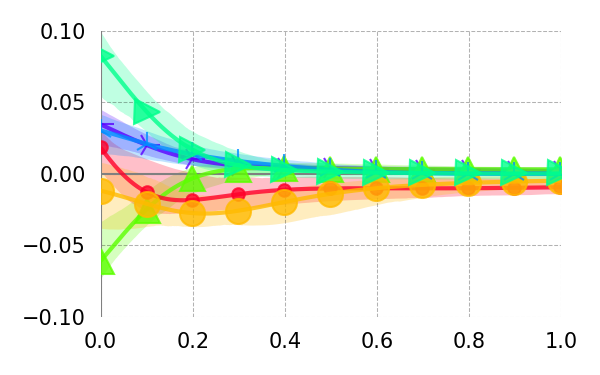}
        \includegraphics[width=0.49\textwidth]{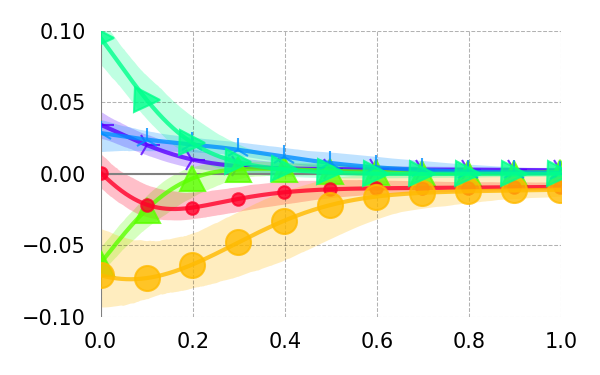}
    \end{subfigure}
    
    \begin{subfigure}[t]{0.49\textwidth}
        \centering
        \makebox[0.49\textwidth]{\centering Hidden Layers - 1 (1)}%
        \makebox[0.49\textwidth]{\centering Hidden Layers + 1 (3)}
        \includegraphics[width=0.49\textwidth]{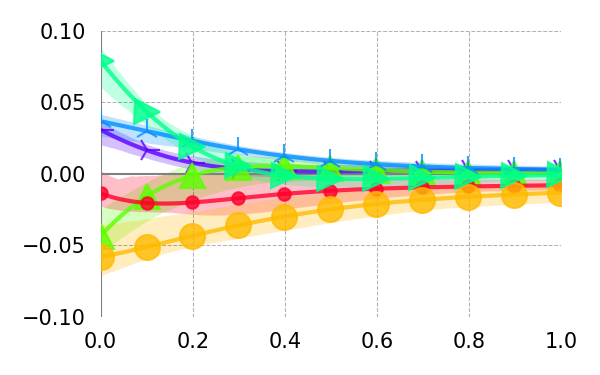}
        \includegraphics[width=0.49\textwidth]{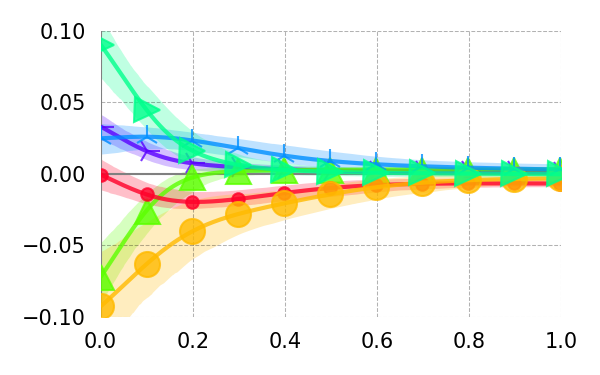}
    \end{subfigure}
    \begin{subfigure}[t]{0.49\textwidth}
        \centering
        \makebox[0.49\textwidth]{\centering Weight Reg $\div$ 5 (1)}%
        \makebox[0.49\textwidth]{\centering Weight Reg $\times$ 5 (5)}
        \includegraphics[width=0.49\textwidth]{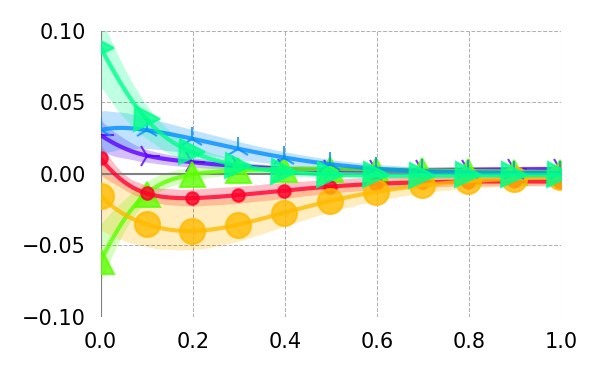}
        \includegraphics[width=0.49\textwidth]{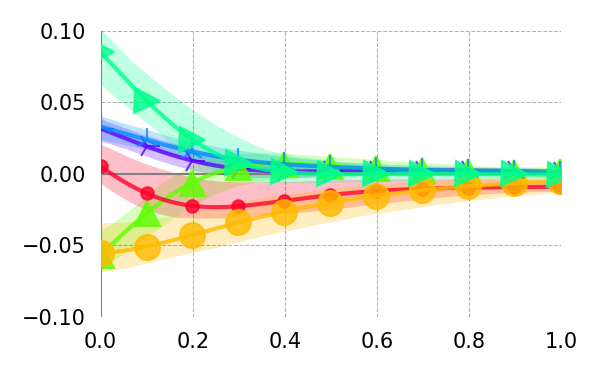}
    \end{subfigure}
    
    \begin{subfigure}[t]{0.49\textwidth}
        \centering
        \makebox[0.49\textwidth]{\centering Ranef Reg $\div$ 10 (1)}%
        \makebox[0.49\textwidth]{\centering Ranef Reg $\times$ 10 (100)}
        \includegraphics[width=0.49\textwidth]{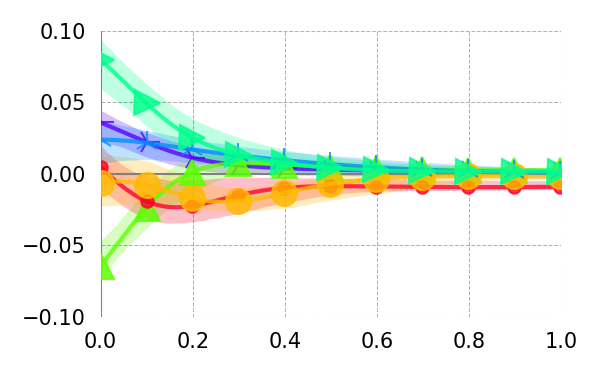}
        \includegraphics[width=0.49\textwidth]{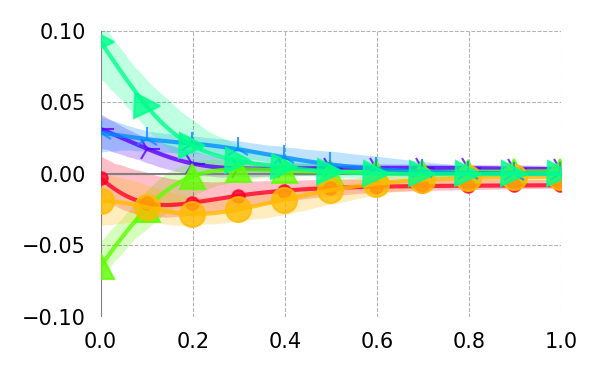}
    \end{subfigure}
    \begin{subfigure}[t]{0.49\textwidth}
        \centering
        \makebox[0.49\textwidth]{\centering Dropout $\div$ 2 (0.05)}%
        \makebox[0.49\textwidth]{\centering Dropout $\times$ 2 (0.2)}
        \includegraphics[width=0.49\textwidth]{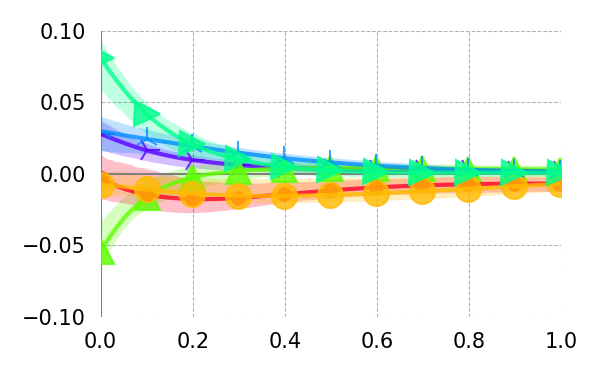}
        \includegraphics[width=0.49\textwidth]{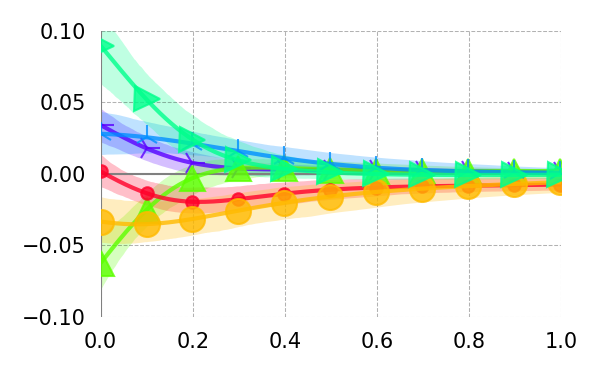}
    \end{subfigure}
    
    \begin{subfigure}[t]{0.49\textwidth}
        \centering
        \makebox[0.49\textwidth]{\centering Learning Rate $\div$ 3 (0.001)}%
        \makebox[0.49\textwidth]{\centering Learning Rate $\times$ 3 (0.009)}
        \includegraphics[width=0.49\textwidth]{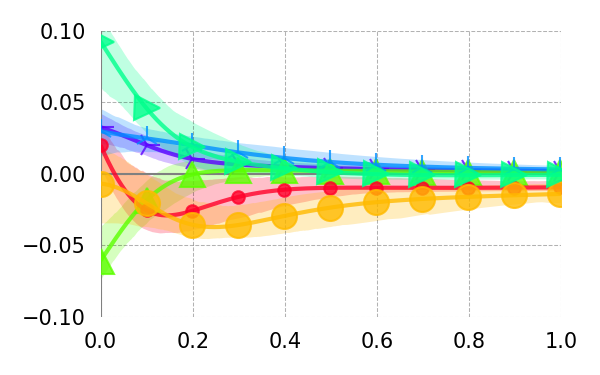}
        \includegraphics[width=0.49\textwidth]{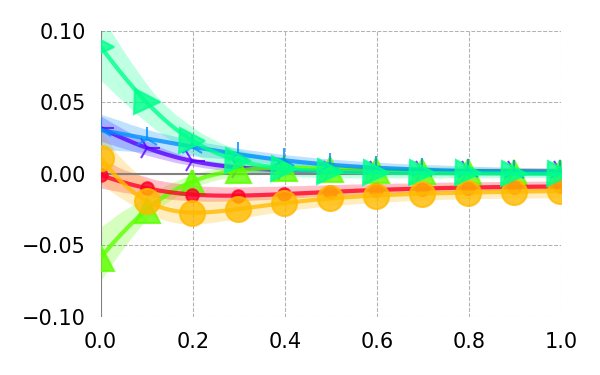}
    \end{subfigure}
    \begin{subfigure}[t]{0.49\textwidth}
        \centering
        \makebox[0.49\textwidth]{\centering Batch Size $\div$ 2 (512)}%
        \makebox[0.49\textwidth]{\centering Batch Size $\times$ 2 (2048)}
        \includegraphics[width=0.49\textwidth]{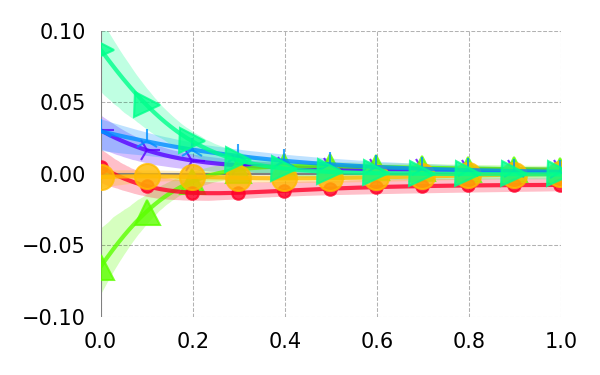}
        \includegraphics[width=0.49\textwidth]{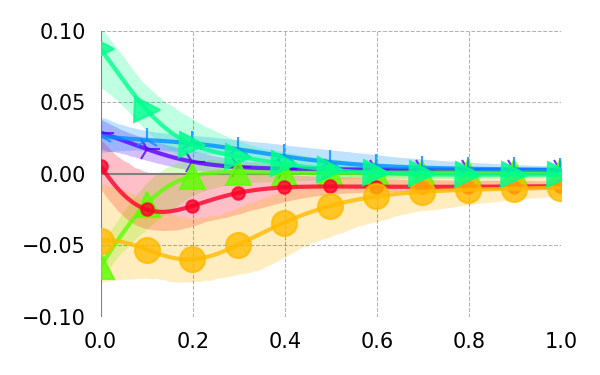}
    \end{subfigure}
    
    \vspace{1em}
    
    \textbf{\large Consistency}
    
    Standard deviation of exploratory set log-likelihood: 17
    
    \vspace{0.5em}
    
    \begin{subfigure}[t]{0.19\textwidth}
        \centering
        \makebox[0.9\textwidth]{\centering Rep 1}
        Expl LL: -20736
        \includegraphics[width=\textwidth]{{results_cdrnn_journal_dundee_log_fp_CDR_main_irf_univariate_log.fdurFP._mean_mc}.png}
    \end{subfigure}
    \begin{subfigure}[t]{0.19\textwidth}
        \centering
        \makebox[0.9\textwidth]{\centering Rep 2}
        Expl LL: -20735
        \includegraphics[width=\textwidth]{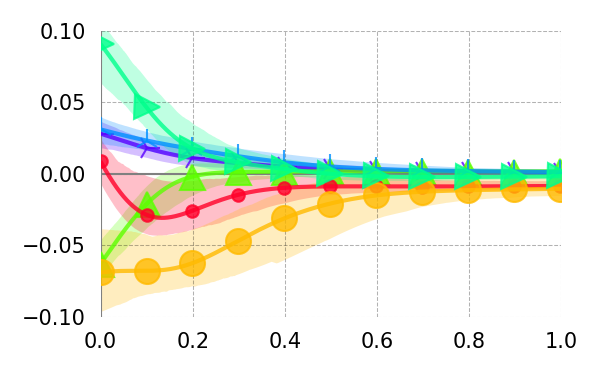}
    \end{subfigure}
    \begin{subfigure}[t]{0.19\textwidth}
        \centering
        \makebox[0.9\textwidth]{\centering Rep 3}
        Expl LL: -20751
        \includegraphics[width=\textwidth]{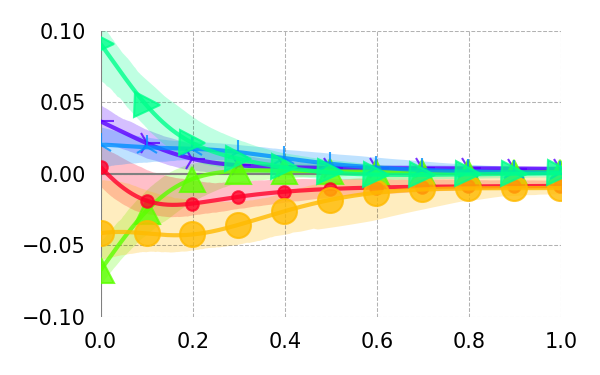}
    \end{subfigure}
    \begin{subfigure}[t]{0.19\textwidth}
        \centering
        \makebox[0.9\textwidth]{\centering Rep 4}
        Expl LL: -20702
        \includegraphics[width=\textwidth]{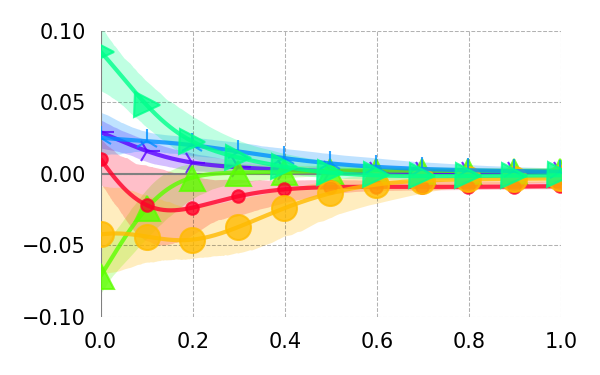}
    \end{subfigure}
    \begin{subfigure}[t]{0.19\textwidth}
        \centering
        \makebox[0.9\textwidth]{\centering Rep 5}
        Expl LL: -20720
        \includegraphics[width=\textwidth]{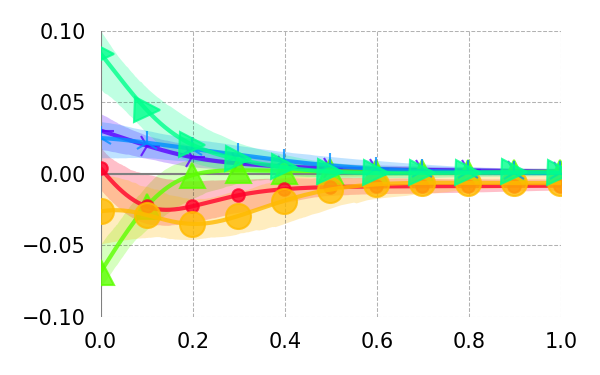}
    \end{subfigure}
    
    {\large Delay (s)}
    
    \vspace{0.5em}
    
    {
    {\kone~rate \hspace{1em}}
    {\kthree~saccade~length \hspace{1em}}
    {\kfive~previous~was~fixated \hspace{1em}}
    {\kseven~word~length \hspace{1em}}
    {\knine~unigram~surprisal \hspace{1em}}
    {\keleven~5-gram~surprisal}
    }
    
    \vspace{0.5em}
    
    \caption{\textbf{Dundee (log first pass duration):} Univariate CDRNN IRF estimates from the Dundee eye-tracking corpus (log-transformed first pass duration). Results using base hyperparameters are compared to estimates from models that deviate from the base in some dimension. Plots under ``Consistency'' show estimates from five replicates of the ``base'' configuration, where ``Rep 1'' is the same model as ``base'' above, replotted for ease of comparison.}
    \label{fig:app-dundee-log-fp}
    
\end{figure}

\begin{figure}

    \footnotesize
    \sffamily
    \centering
    
    \textbf{\Large Dundee (Go-Past Duration)}
    
    \vspace{1em}
    
    \begin{subfigure}[t]{0.49\textwidth}
        \centering
        \makebox[0.49\textwidth]{\centering Base}%
        \makebox[0.49\textwidth]{\centering + RNN}
        \begin{overpic}[width=0.49\textwidth]{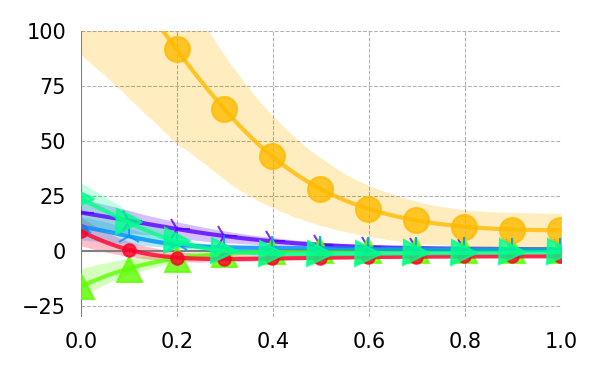}
            \put (-15,-80) {\rotatebox[origin=c]{90}{\large Change in Go-Past Duration (ms)}}
        \end{overpic}%
        \includegraphics[width=0.49\textwidth]{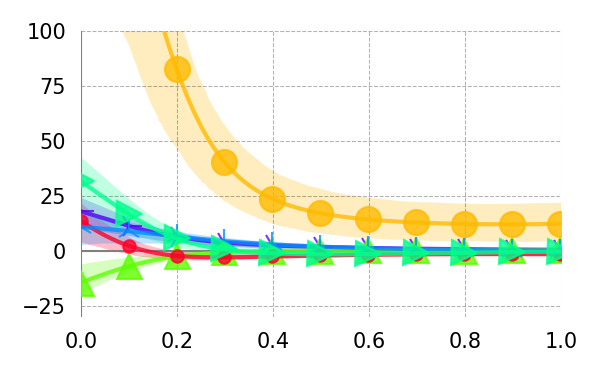}
    \end{subfigure}
    \begin{subfigure}[t]{0.49\textwidth}
        \centering
        \makebox[0.49\textwidth]{\centering Hidden Units $\div$ 2 (16)}%
        \makebox[0.49\textwidth]{\centering Hidden Units $\times$ 2 (64)}
        \includegraphics[width=0.49\textwidth]{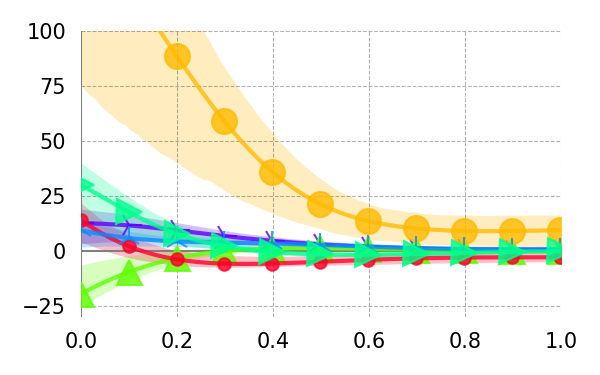}
        \includegraphics[width=0.49\textwidth]{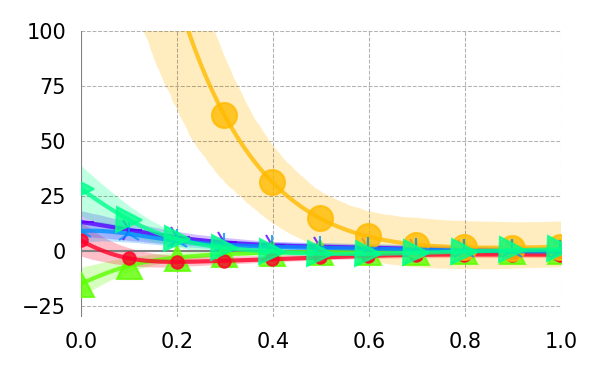}
    \end{subfigure}
    
    \begin{subfigure}[t]{0.49\textwidth}
        \centering
        \makebox[0.49\textwidth]{\centering Hidden Layers - 1 (1)}%
        \makebox[0.49\textwidth]{\centering Hidden Layers + 1 (3)}
        \includegraphics[width=0.49\textwidth]{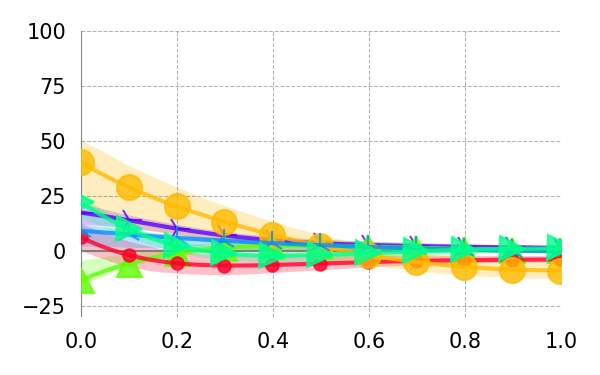}
        \includegraphics[width=0.49\textwidth]{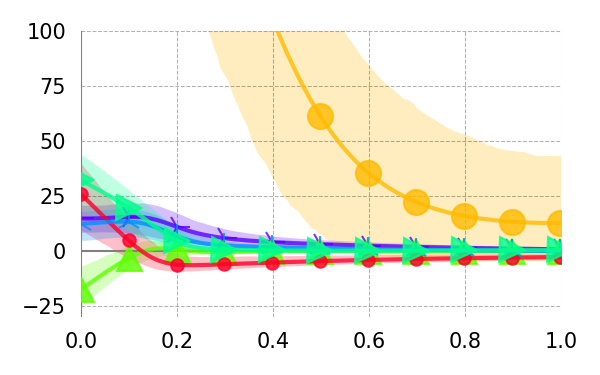}
    \end{subfigure}
    \begin{subfigure}[t]{0.49\textwidth}
        \centering
        \makebox[0.49\textwidth]{\centering Weight Reg $\div$ 5 (1)}%
        \makebox[0.49\textwidth]{\centering Weight Reg $\times$ 5 (5)}
        \includegraphics[width=0.49\textwidth]{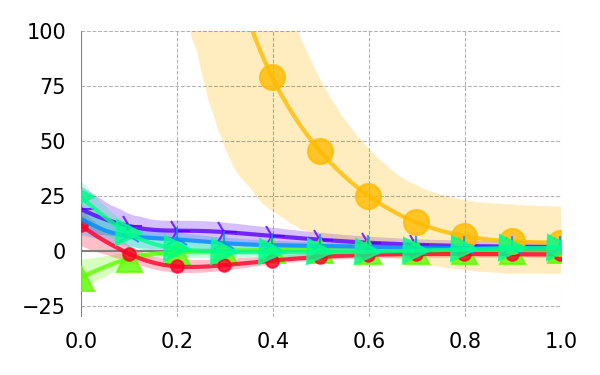}
        \includegraphics[width=0.49\textwidth]{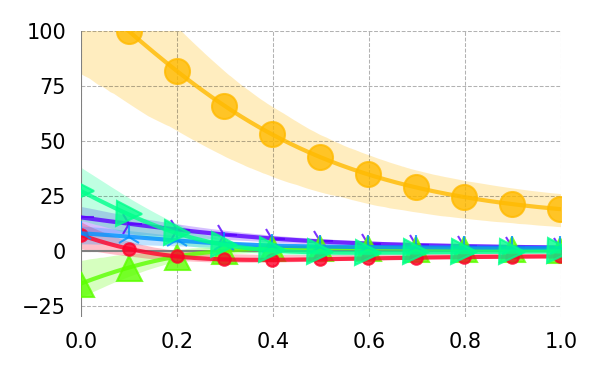}
    \end{subfigure}
    
    \begin{subfigure}[t]{0.49\textwidth}
        \centering
        \makebox[0.49\textwidth]{\centering Ranef Reg $\div$ 10 (1)}%
        \makebox[0.49\textwidth]{\centering Ranef Reg $\times$ 10 (100)}
        \includegraphics[width=0.49\textwidth]{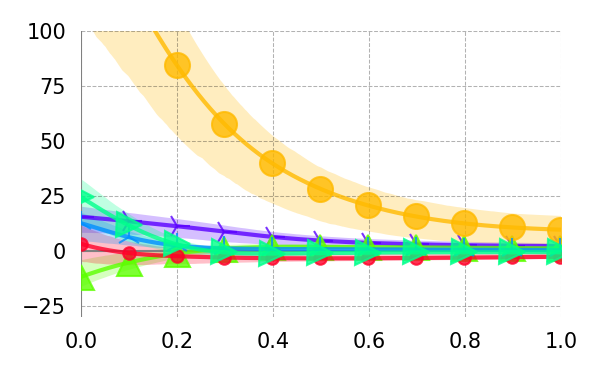}
        \includegraphics[width=0.49\textwidth]{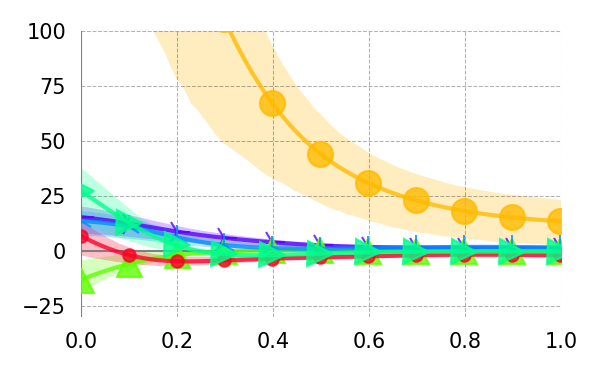}
    \end{subfigure}
    \begin{subfigure}[t]{0.49\textwidth}
        \centering
        \makebox[0.49\textwidth]{\centering Dropout $\div$ 2 (0.05)}%
        \makebox[0.49\textwidth]{\centering Dropout $\times$ 2 (0.2)}
        \includegraphics[width=0.49\textwidth]{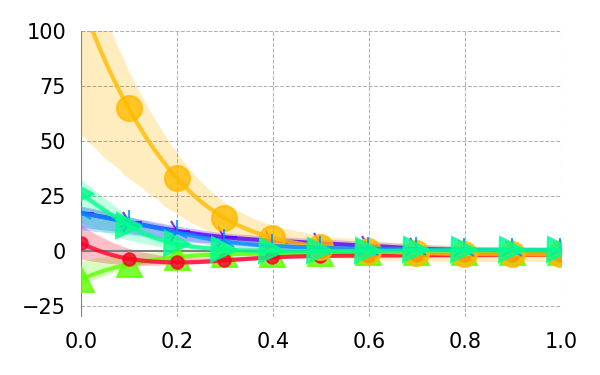}
        \includegraphics[width=0.49\textwidth]{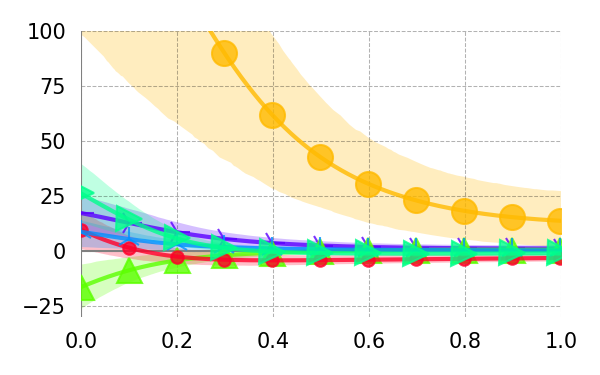}
    \end{subfigure}
    
    \begin{subfigure}[t]{0.49\textwidth}
        \centering
        \makebox[0.49\textwidth]{\centering Learning Rate $\div$ 3 (0.001)}%
        \makebox[0.49\textwidth]{\centering Learning Rate $\times$ 3 (0.009)}
        \includegraphics[width=0.49\textwidth]{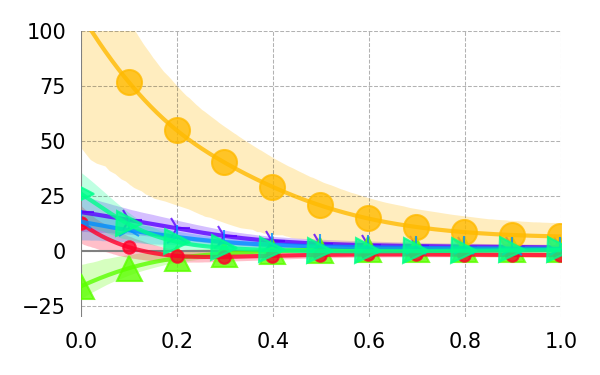}
        \includegraphics[width=0.49\textwidth]{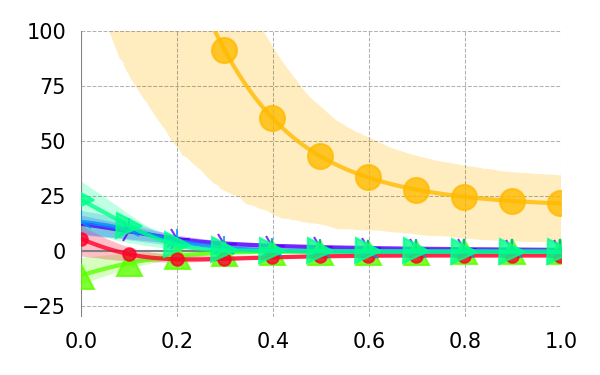}
    \end{subfigure}
    \begin{subfigure}[t]{0.49\textwidth}
        \centering
        \makebox[0.49\textwidth]{\centering Batch Size $\div$ 2 (512)}%
        \makebox[0.49\textwidth]{\centering Batch Size $\times$ 2 (2048)}
        \includegraphics[width=0.49\textwidth]{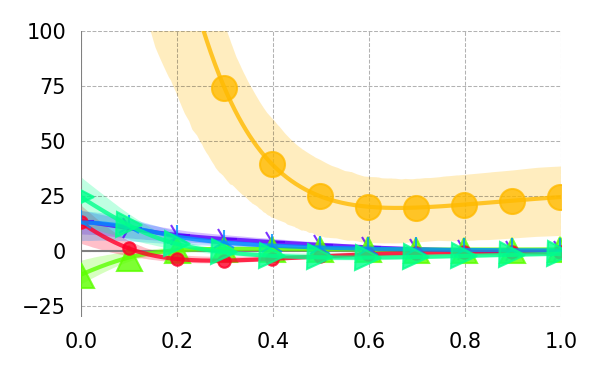}
        \includegraphics[width=0.49\textwidth]{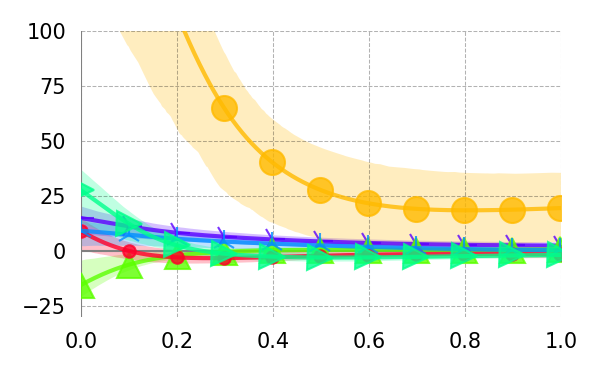}
    \end{subfigure}
    
    \vspace{1em}    
    
    \textbf{\large Consistency}
    
    Standard deviation of exploratory set log-likelihood: 370
    
    \vspace{0.5em}
    
    \begin{subfigure}[t]{0.19\textwidth}
        \centering
        \makebox[0.9\textwidth]{\centering Rep 1}
        Expl LL: -306662
        \includegraphics[width=\textwidth]{{results_cdrnn_journal_dundee_raw_gp_CDR_main_irf_univariate_fdurGP_mean_mc}.png}
    \end{subfigure}
    \begin{subfigure}[t]{0.19\textwidth}
        \centering
        \makebox[0.9\textwidth]{\centering Rep 2}
        Expl LL: -306772
        \includegraphics[width=\textwidth]{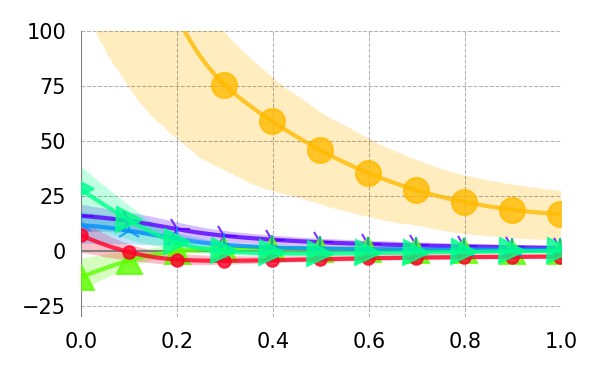}
    \end{subfigure}
    \begin{subfigure}[t]{0.19\textwidth}
        \centering
        \makebox[0.9\textwidth]{\centering Rep 3}
        Expl LL: -306319
        \includegraphics[width=\textwidth]{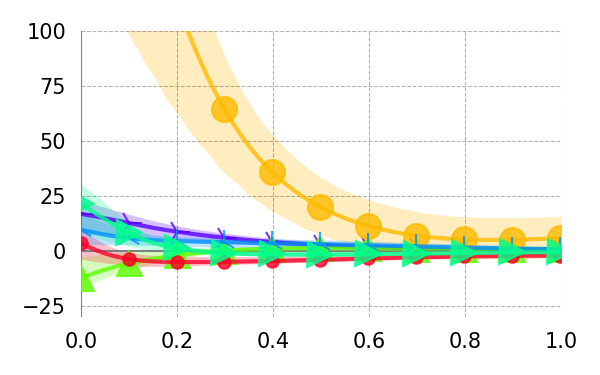}
    \end{subfigure}
    \begin{subfigure}[t]{0.19\textwidth}
        \centering
        \makebox[0.9\textwidth]{\centering Rep 4}
        Expl LL: -307026
        \includegraphics[width=\textwidth]{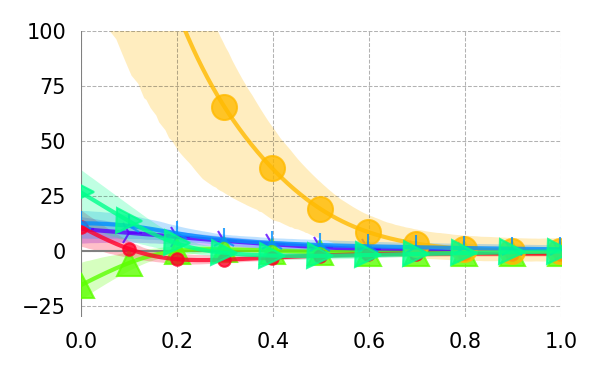}
    \end{subfigure}
    \begin{subfigure}[t]{0.19\textwidth}
        \centering
        \makebox[0.9\textwidth]{\centering Rep 5}
        Expl LL: -305966
        \includegraphics[width=\textwidth]{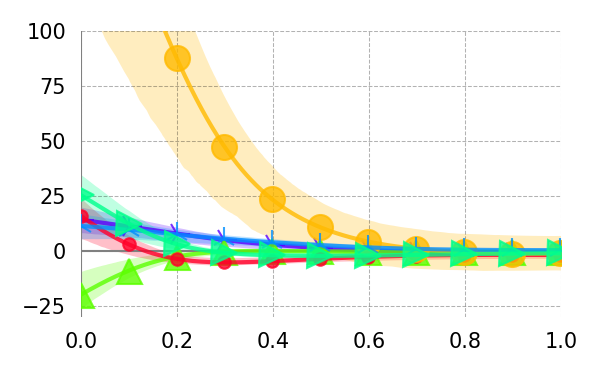}
    \end{subfigure}
    
    {\large Delay (s)}
    
    \vspace{0.5em}
    
    {
    {\kone~rate \hspace{1em}}
    {\kthree~saccade~length \hspace{1em}}
    {\kfive~previous~was~fixated \hspace{1em}}
    {\kseven~word~length \hspace{1em}}
    {\knine~unigram~surprisal \hspace{1em}}
    {\keleven~5-gram~surprisal}
    }
    
    \vspace{0.5em}
    
    \caption{\textbf{Dundee (go-past duration):} Univariate CDRNN IRF estimates from the Dundee eye-tracking corpus (go-past duration). Results using base hyperparameters are compared to estimates from models that deviate from the base in some dimension. Plots under ``Consistency'' show estimates from five replicates of the ``base'' configuration, where ``Rep 1'' is the same model as ``base'' above, replotted for ease of comparison. Some large estimates are clipped to preserve readability.}
    \label{fig:app-dundee-raw-gp}
    
\end{figure}

\begin{figure}

    \footnotesize
    \sffamily
    \centering
    
    \textbf{\Large Dundee (Log Go-Past Duration)}
    
    \vspace{1em}
    
    \begin{subfigure}[t]{0.49\textwidth}
        \centering
        \makebox[0.49\textwidth]{\centering Base}%
        \makebox[0.49\textwidth]{\centering + RNN}
        \begin{overpic}[width=0.49\textwidth]{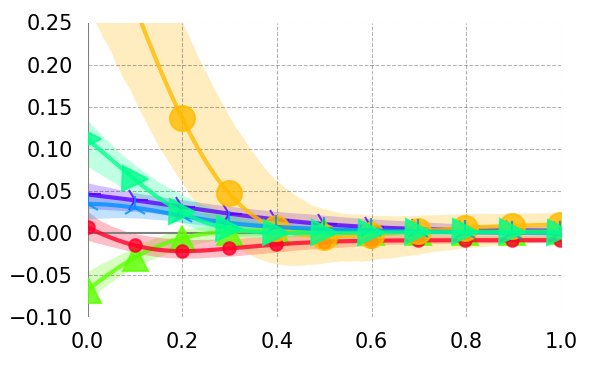}
            \put (-15,-80) {\rotatebox[origin=c]{90}{\large Change in Go-Past Duration (log-ms)}}
        \end{overpic}%
        \includegraphics[width=0.49\textwidth]{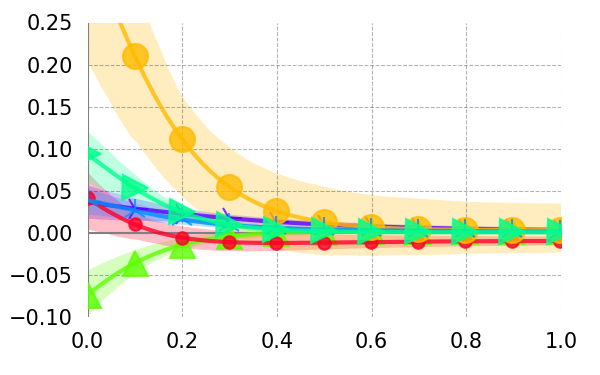}
    \end{subfigure}
    \begin{subfigure}[t]{0.49\textwidth}
        \centering
        \makebox[0.49\textwidth]{\centering Hidden Units $\div$ 2 (16)}%
        \makebox[0.49\textwidth]{\centering Hidden Units $\times$ 2 (64)}
        \includegraphics[width=0.49\textwidth]{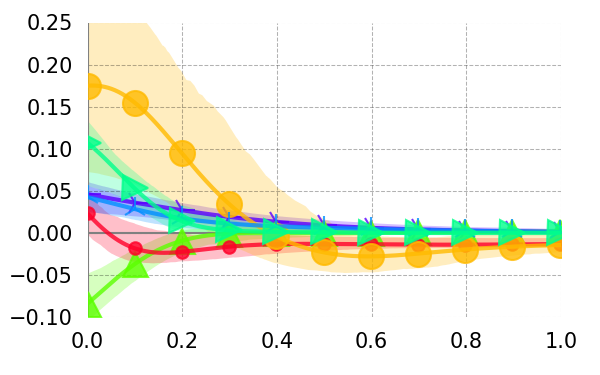}
        \includegraphics[width=0.49\textwidth]{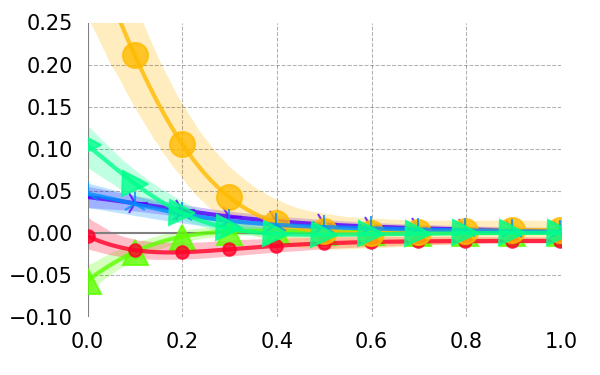}
    \end{subfigure}
    
    \begin{subfigure}[t]{0.49\textwidth}
        \centering
        \makebox[0.49\textwidth]{\centering Hidden Layers - 1 (1)}%
        \makebox[0.49\textwidth]{\centering Hidden Layers + 1 (3)}
        \includegraphics[width=0.49\textwidth]{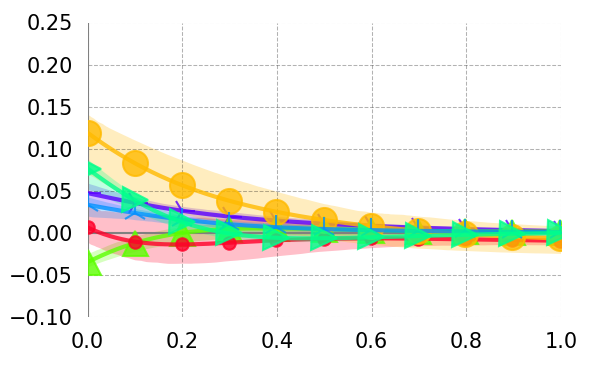}
        \includegraphics[width=0.49\textwidth]{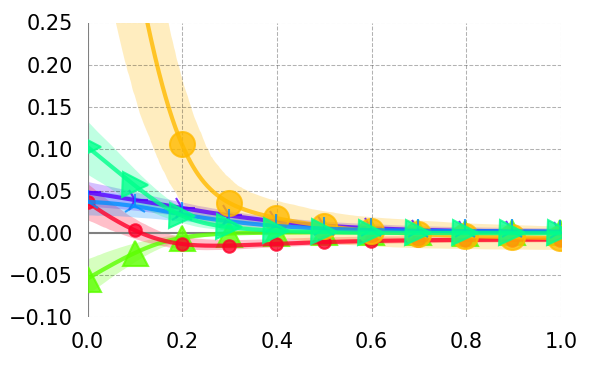}
    \end{subfigure}
    \begin{subfigure}[t]{0.49\textwidth}
        \centering
        \makebox[0.49\textwidth]{\centering Weight Reg $\div$ 5 (1)}%
        \makebox[0.49\textwidth]{\centering Weight Reg $\times$ 5 (5)}
        \includegraphics[width=0.49\textwidth]{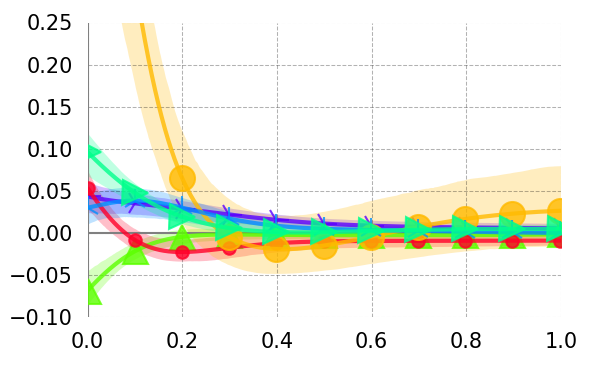}
        \includegraphics[width=0.49\textwidth]{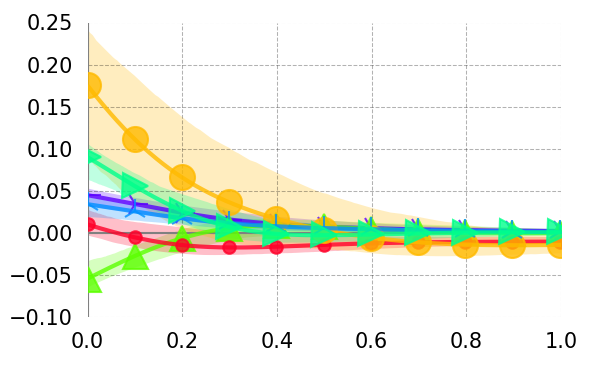}
    \end{subfigure}
    
    \begin{subfigure}[t]{0.49\textwidth}
        \centering
        \makebox[0.49\textwidth]{\centering Ranef Reg $\div$ 10 (1)}%
        \makebox[0.49\textwidth]{\centering Ranef Reg $\times$ 10 (100)}
        \includegraphics[width=0.49\textwidth]{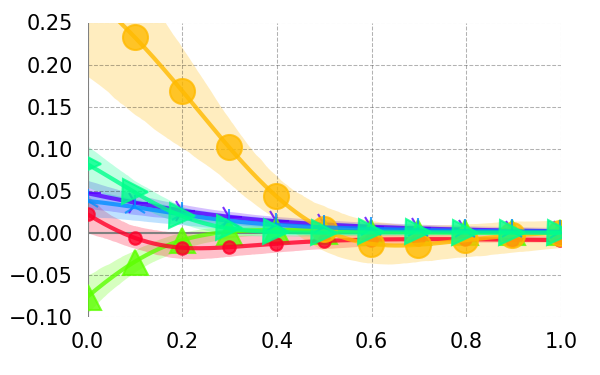}
        \includegraphics[width=0.49\textwidth]{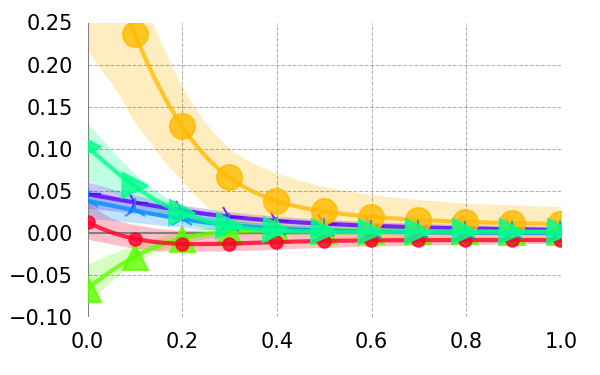}
    \end{subfigure}
    \begin{subfigure}[t]{0.49\textwidth}
        \centering
        \makebox[0.49\textwidth]{\centering Dropout $\div$ 2 (0.05)}%
        \makebox[0.49\textwidth]{\centering Dropout $\times$ 2 (0.2)}
        \includegraphics[width=0.49\textwidth]{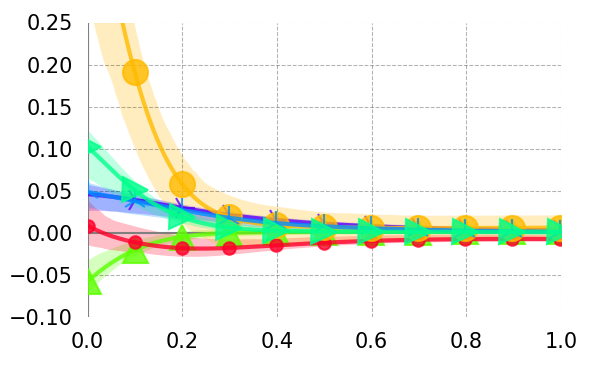}
        \includegraphics[width=0.49\textwidth]{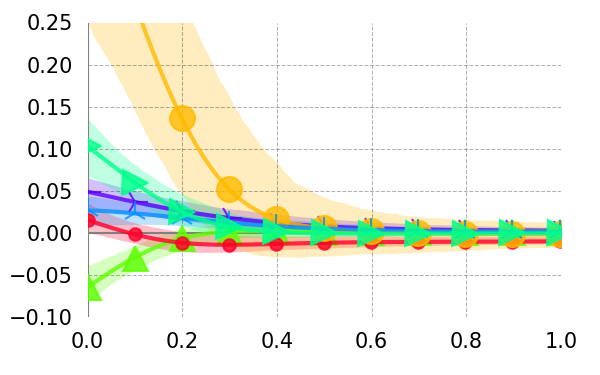}
    \end{subfigure}
    
    \begin{subfigure}[t]{0.49\textwidth}
        \centering
        \makebox[0.49\textwidth]{\centering Learning Rate $\div$ 3 (0.001)}%
        \makebox[0.49\textwidth]{\centering Learning Rate $\times$ 3 (0.009)}
        \includegraphics[width=0.49\textwidth]{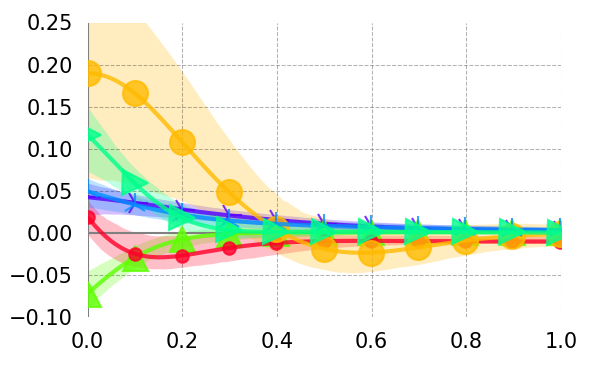}
        \includegraphics[width=0.49\textwidth]{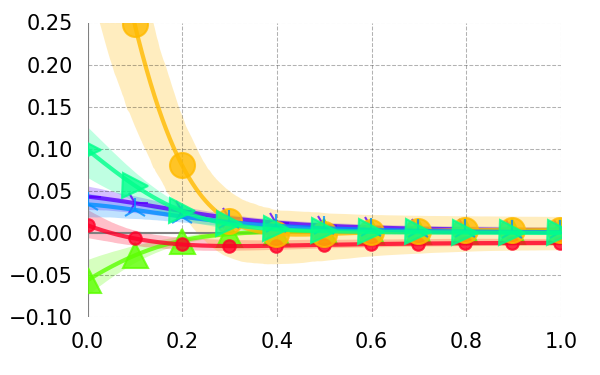}
    \end{subfigure}
    \begin{subfigure}[t]{0.49\textwidth}
        \centering
        \makebox[0.49\textwidth]{\centering Batch Size $\div$ 2 (512)}%
        \makebox[0.49\textwidth]{\centering Batch Size $\times$ 2 (2048)}
        \includegraphics[width=0.49\textwidth]{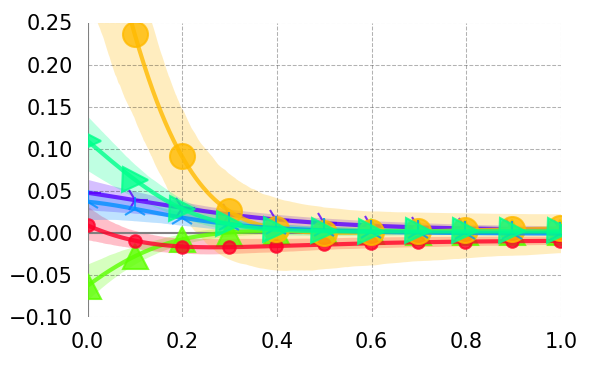}
        \includegraphics[width=0.49\textwidth]{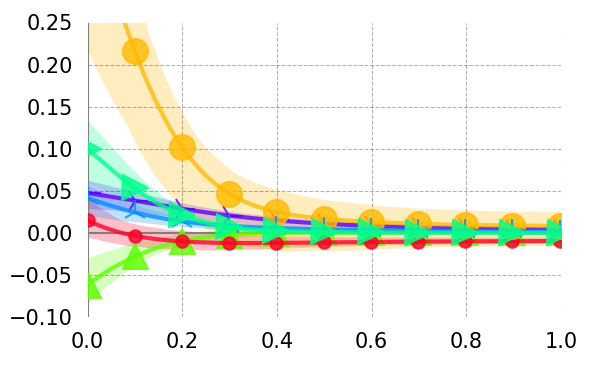}
    \end{subfigure}
    
    \vspace{1em}
    
    \textbf{\large Consistency}
    
    Standard deviation of exploratory set log-likelihood: 26
    
    \vspace{0.5em}
    
    \begin{subfigure}[t]{0.19\textwidth}
        \centering
        \makebox[0.9\textwidth]{\centering Rep 1}
        Expl LL: -27509
        \includegraphics[width=\textwidth]{{results_cdrnn_journal_dundee_log_gp_CDR_main_irf_univariate_log.fdurGP._mean_mc}.png}
    \end{subfigure}
    \begin{subfigure}[t]{0.19\textwidth}
        \centering
        \makebox[0.9\textwidth]{\centering Rep 2}
        Expl LL: -27561
        \includegraphics[width=\textwidth]{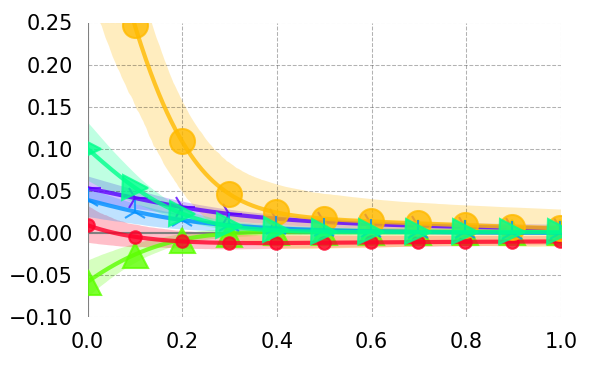}
    \end{subfigure}
    \begin{subfigure}[t]{0.19\textwidth}
        \centering
        \makebox[0.9\textwidth]{\centering Rep 3}
        Expl LL: -27532
        \includegraphics[width=\textwidth]{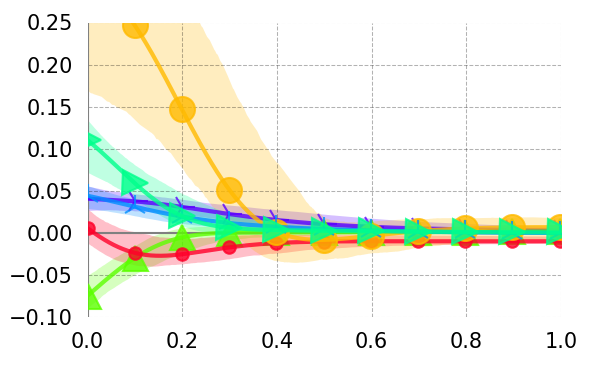}
    \end{subfigure}
    \begin{subfigure}[t]{0.19\textwidth}
        \centering
        \makebox[0.9\textwidth]{\centering Rep 4}
        Expl LL: -27583
        \includegraphics[width=\textwidth]{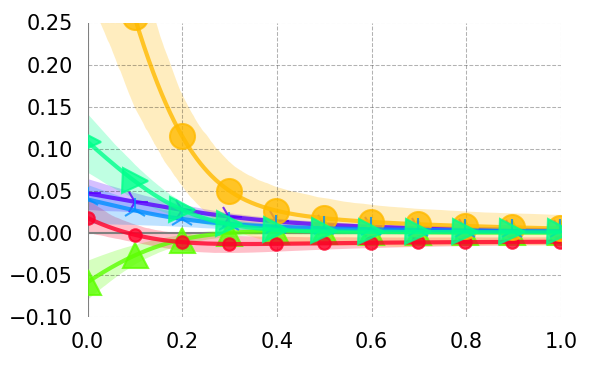}
    \end{subfigure}
    \begin{subfigure}[t]{0.19\textwidth}
        \centering
        \makebox[0.9\textwidth]{\centering Rep 5}
        Expl LL: -27567
        \includegraphics[width=\textwidth]{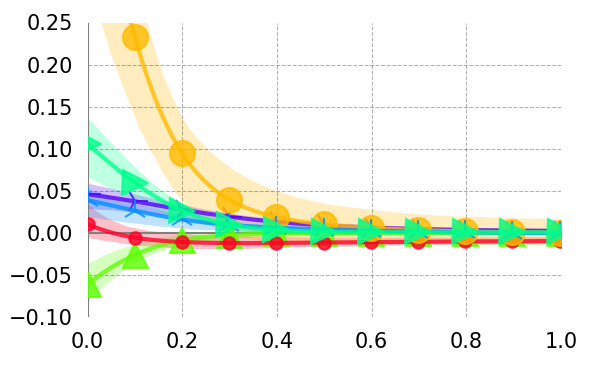}
    \end{subfigure}
    
    {\large Delay (s)}
    
    \vspace{0.5em}
    
    {
    {\kone~rate \hspace{1em}}
    {\kthree~saccade~length \hspace{1em}}
    {\kfive~previous~was~fixated \hspace{1em}}
    {\kseven~word~length \hspace{1em}}
    {\knine~unigram~surprisal \hspace{1em}}
    {\keleven~5-gram~surprisal}
    }
    
    \vspace{0.5em}
    
    \caption{\textbf{Dundee (log go-past duration):} Univariate CDRNN IRF estimates from the Dundee eye-tracking corpus (log-transformed go-past duration). Results using base hyperparameters are compared to estimates from models that deviate from the base in some dimension. Plots under ``Consistency'' show estimates from five replicates of the ``base'' configuration, where ``Rep 1'' is the same model as ``base'' above, replotted for ease of comparison. Some large estimates are clipped to preserve readability.}
    \label{fig:app-dundee-log-gp}
    
\end{figure}

\FloatBarrier


\subsection{Natural Stories (Self-Paced Reading)}

In-sample and out-of-sample predictive performance on the Natural Stories SPR dataset is reported in \textbf{Supplementary Table~\ref{tab:app-natstor-ll-bakeoff}}.
Effect estimates in Natural Stories SPR are plotted in \textbf{Supplementary Figures~\ref{fig:app-natstor-raw}--\ref{fig:app-natstor-log}}.
Estimates are plausible and highly consistent across model variants.

\begin{table}
    \footnotesize
    \sffamily
    \centering
    \begin{tabular}{r|ccc|ccc}
        & \multicolumn{3}{|c}{Natural Stories Reading Time (ms)} & \multicolumn{3}{|c}{Natural Stories Reading Time (log ms)}\\
        Model & Train & Expl & Test & Train & Expl & Test\\
        \hline
        LME & -2444089 & -1231466 & -1224073 & -62248 & -31964 & -31579 \\
        LME-S & -2442652\textsuperscript{\textdagger} & -1230711\textsuperscript{\textdagger} & -1223393\textsuperscript{\textdagger} & -60232\textsuperscript{\textdagger} & -30596\textsuperscript{\textdagger} & -30220\textsuperscript{\textdagger} \\
        GAM & -2441347 & -1230885 & -1223532 & -59305 & -31454 & -31004 \\
        GAM-S & -2439456 & -1230163 & -1222835 & -55926 & -30069 & -29660 \\
        GAMLSS & -2386588 & -1203461 & -1196525 & -39605 & -21024 & -20608 \\
        GAMLSS-S & \textit{-2383131} & \textit{-1202004} & \textit{-1195295} & -37320 & -19867 & -19539 \\
        CDR & -2421724 & -1220612 & -1213359 & \textit{-18839} & \textit{-10430} & \textit{-10641} \\
        \hline
        CDRNN base & -2263597 & -1224555 & \textbf{-1191747} & 40237 & 10986 & \textbf{12521}\\
        --Nonlinear & -2275008 & -1224562 & --- & 36589 & 10515 & ---\\
        --Nonstationary & -2266085 & -1220449 & --- & 36023 & 11530 & ---\\
        --Heteroscedastic & -2414607 & -1217627 & --- & -10311 & -6838 & ---\\
        +RNN & \textit{\textbf{-2206627}} & -1364111 & --- & \textit{\textbf{49360}} & -166401 & ---\\
        Units $\div$ 2 (16) & -2294417 & -1208977 & --- & 35453 & 12122 & ---\\
        Units $\times$ 2 (64) & -2235490 & -1264557 & --- & 48125 & 8478 & ---\\
        Layers - 1 (1) & -2304313 & -1201513 & --- & 30965 & 10535 & ---\\
        Layers + 1 (3) & -2245651 & -1237794 & --- & 24978 & 220 & ---\\
        Weight Reg $\div$ 5 (1) & -2250353 & -1209774 & --- & 44075 & 12403 & ---\\
        Weight Reg $\times$ 5 (25) & -2292479 & -1226248 & --- & 25585 & 7387 & ---\\
        Ranef Reg $\div$ 10 (1) & -2240188 & -1247420 & --- & 58557 & 7333 & ---\\
        Ranef Reg $\times$ 10 (100) & -2260072 & -1256309 & --- & 40365 & 10835 & ---\\
        Dropout $\div$ 2 (0.05) & -2242987 & -1240444 & --- & 52666 & 10747 & ---\\
        Dropout $\times$ 2 (0.2) & -2328802 & -1190181 & --- & 19053 & 7226 & ---\\
        Learning Rate $\div$ 3 (0.001) & -2261631 & -1288272 & --- & 44836 & \textit{\textbf{12769}} & ---\\
        Learning Rate $\times$ 3 (0.009) & -2288092 & \textit{\textbf{-1189849}} & --- & 32165 & 12346 & ---\\
        Batch Size $\div$ 2 (512) & -2276237 & -1206397 & --- & 37294 & 11336 & ---\\
        Batch Size $\times$ 2 (2048) & -2256670 & -1237186 & --- & 42204 & 11307 & ---\\
    \end{tabular}
    \caption{\textbf{Natural Stories (reading time likelihood).} Log likelihood from CDRNN vs.\ linear mixed-effects (LME), generalized additive model (GAM), and generalized additive model for location, scale, and shape (GAMLSS) baselines with and without three additional spillover positions (-S) to help capture delayed effects, as well as kernel-based (non-neural) CDR (LME, GAM, and CDR performance as reported in \textit{\citen{shainschuler19}}).
    LME baselines show the marginal likelihood for the training set (the default likelihood implemented by the \texttt{lme4} package).
    All other likelihoods are conditional on the fitted model.
    CDRNN variants add recurrence (+RNN) and modify the number of Units (Units), number of hidden layers (Layers), weight regularization strength (Reg), random effects regularization strength (RanReg), dropout rate (Dropout), learning rate (LR), and batch size (Batch).
    Of the CDRNN models, only CDRNN base is evaluated on the test set.
    Best-performing models within the sets of baseline and CDRNN models are shown in \textit{italics}. Best-performing overall models are shown in \textbf{bold}. Daggers (\textdagger) indicate convergence failures.}
    \label{tab:app-natstor-ll-bakeoff}
\end{table}

\begin{figure}

    \footnotesize
    \sffamily
    \centering
    
    \textbf{\Large Natural Stories (Self-Paced Reading Time)}
    
    \vspace{1em}
    
    \begin{subfigure}[t]{0.49\textwidth}
        \centering
        \makebox[0.49\textwidth]{\centering Base}%
        \makebox[0.49\textwidth]{\centering + RNN}
        \begin{overpic}[width=0.49\textwidth]{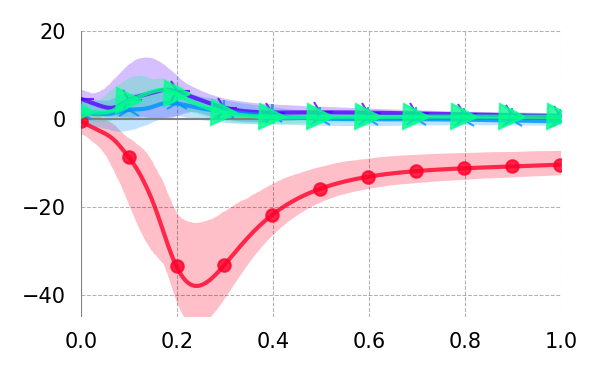}
            \put (-15,-80) {\rotatebox[origin=c]{90}{\large Change in Reading Time (ms)}}
        \end{overpic}%
        \includegraphics[width=0.49\textwidth]{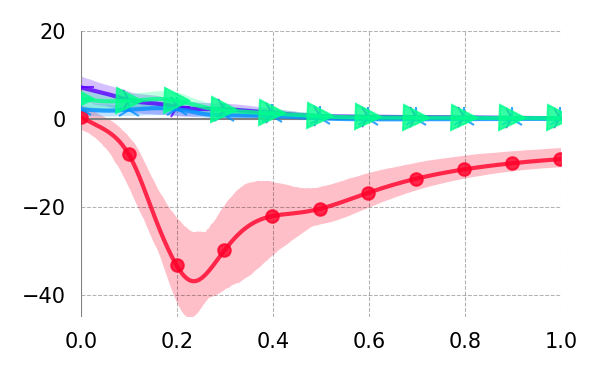}
    \end{subfigure}
    \begin{subfigure}[t]{0.49\textwidth}
        \centering
        \makebox[0.49\textwidth]{\centering Hidden Units $\div$ 2 (16)}%
        \makebox[0.49\textwidth]{\centering Hidden Units $\times$ 2 (64)}
        \includegraphics[width=0.49\textwidth]{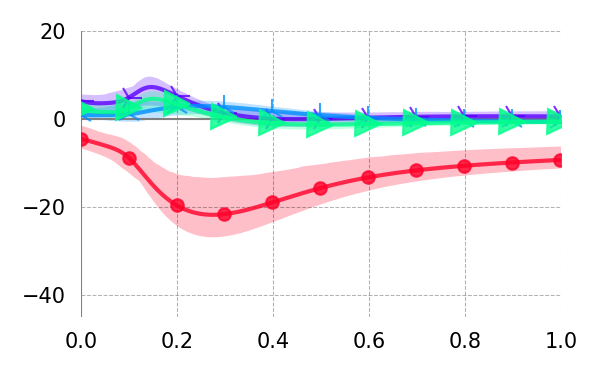}
        \includegraphics[width=0.49\textwidth]{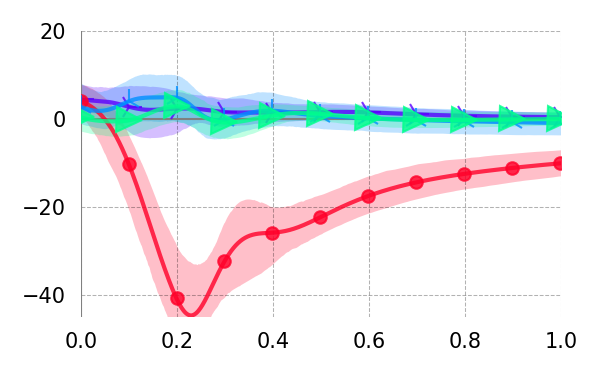}
    \end{subfigure}
    
    \begin{subfigure}[t]{0.49\textwidth}
        \centering
        \makebox[0.49\textwidth]{\centering Hidden Layers - 1 (1)}%
        \makebox[0.49\textwidth]{\centering Hidden Layers + 1 (3)}
        \includegraphics[width=0.49\textwidth]{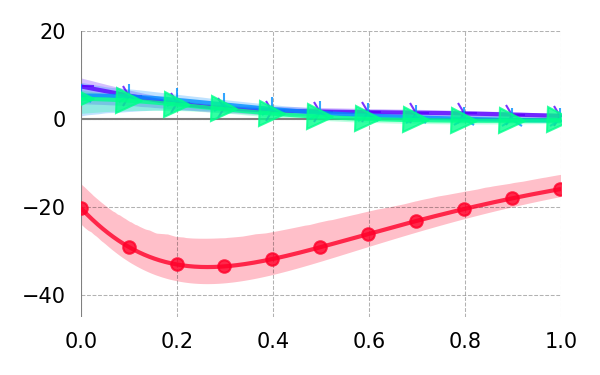}
        \includegraphics[width=0.49\textwidth]{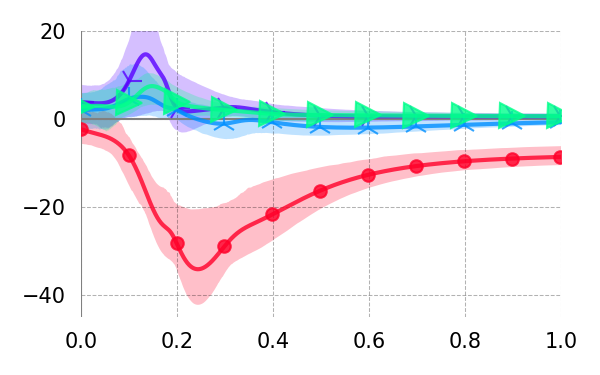}
    \end{subfigure}
    \begin{subfigure}[t]{0.49\textwidth}
        \centering
        \makebox[0.49\textwidth]{\centering Weight Reg $\div$ 5 (1)}%
        \makebox[0.49\textwidth]{\centering Weight Reg $\times$ 5 (5)}
        \includegraphics[width=0.49\textwidth]{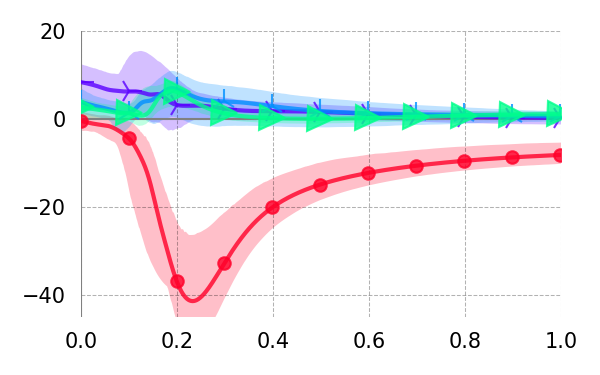}
        \includegraphics[width=0.49\textwidth]{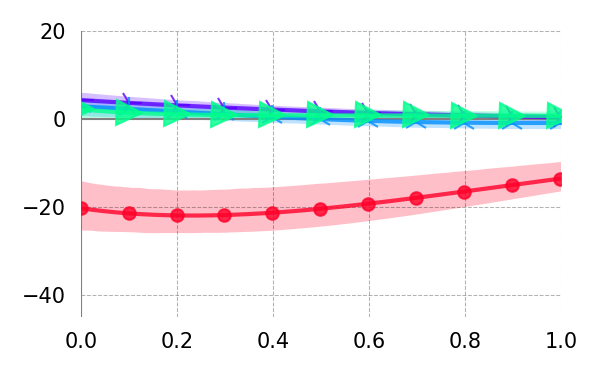}
    \end{subfigure}
    
    \begin{subfigure}[t]{0.49\textwidth}
        \centering
        \makebox[0.49\textwidth]{\centering Ranef Reg $\div$ 10 (1)}%
        \makebox[0.49\textwidth]{\centering Ranef Reg $\times$ 10 (100)}
        \includegraphics[width=0.49\textwidth]{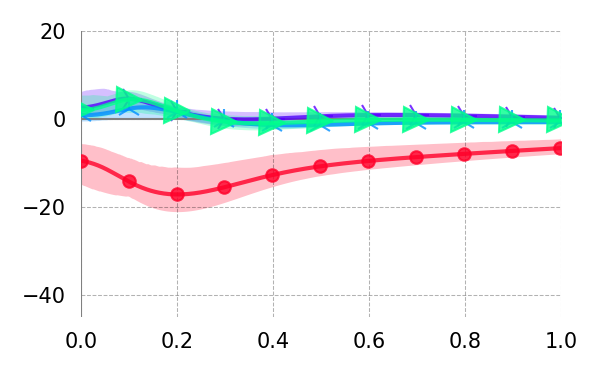}
        \includegraphics[width=0.49\textwidth]{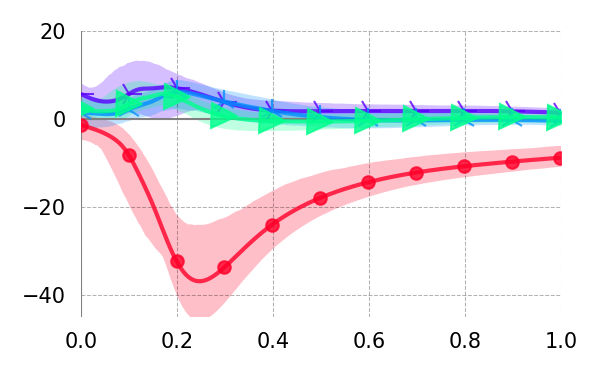}
    \end{subfigure}
    \begin{subfigure}[t]{0.49\textwidth}
        \centering
        \makebox[0.49\textwidth]{\centering Dropout $\div$ 2 (0.05)}%
        \makebox[0.49\textwidth]{\centering Dropout $\times$ 2 (0.2)}
        \includegraphics[width=0.49\textwidth]{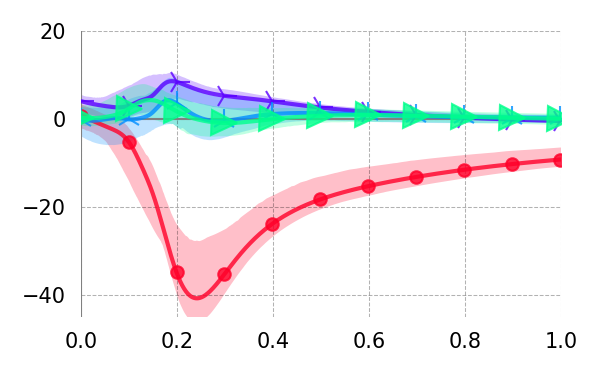}
        \includegraphics[width=0.49\textwidth]{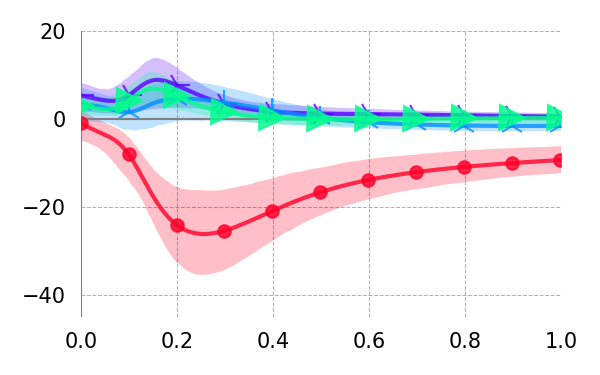}
    \end{subfigure}
    
    \begin{subfigure}[t]{0.49\textwidth}
        \centering
        \makebox[0.49\textwidth]{\centering Learning Rate $\div$ 3 (0.001)}%
        \makebox[0.49\textwidth]{\centering Learning Rate $\times$ 3 (0.009)}
        \includegraphics[width=0.49\textwidth]{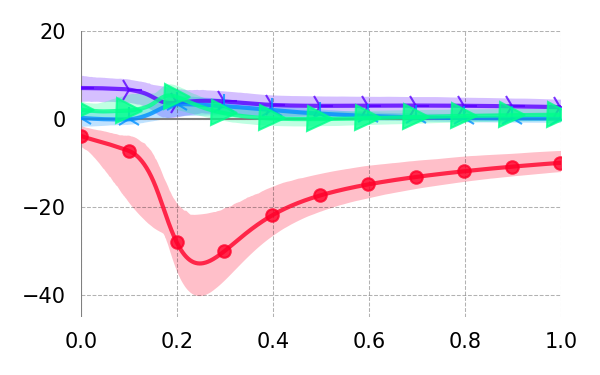}
        \includegraphics[width=0.49\textwidth]{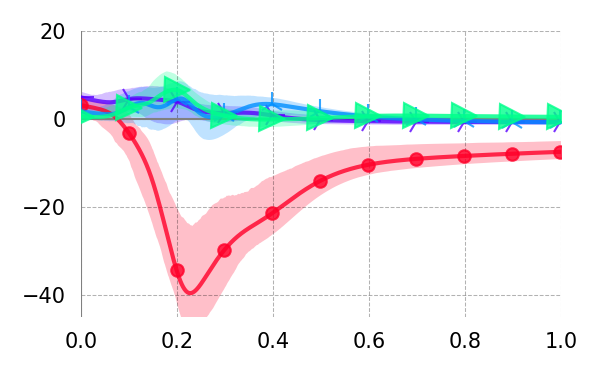}
    \end{subfigure}
    \begin{subfigure}[t]{0.49\textwidth}
        \centering
        \makebox[0.49\textwidth]{\centering Batch Size $\div$ 2 (512)}%
        \makebox[0.49\textwidth]{\centering Batch Size $\times$ 2 (2048)}
        \includegraphics[width=0.49\textwidth]{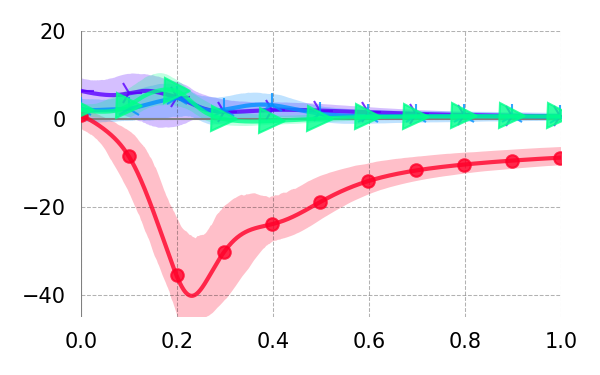}
        \includegraphics[width=0.49\textwidth]{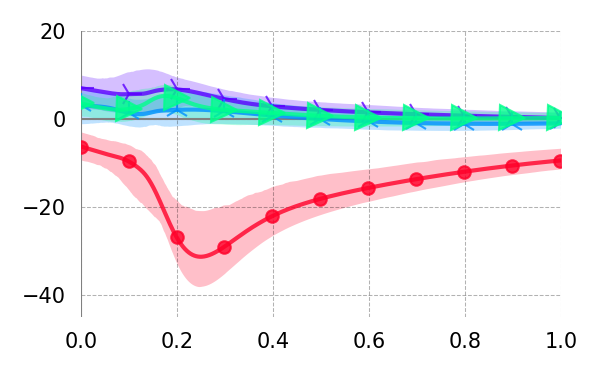}
    \end{subfigure}
    
    \vspace{1em}    
    
    \textbf{\large Consistency}
    
    Standard deviation of exploratory set log-likelihood: 12619
    
    \vspace{0.5em}
    
    \begin{subfigure}[t]{0.19\textwidth}
        \centering
        \makebox[0.9\textwidth]{\centering Rep 1}
        Expl LL: -1224555
        \includegraphics[width=\textwidth]{{results_cdrnn_journal_natstor_raw_CDR_main_irf_univariate_fdur_mean_mc}.png}
    \end{subfigure}
    \begin{subfigure}[t]{0.19\textwidth}
        \centering
        \makebox[0.9\textwidth]{\centering Rep 2}
        Expl LL: -1203388
        \includegraphics[width=\textwidth]{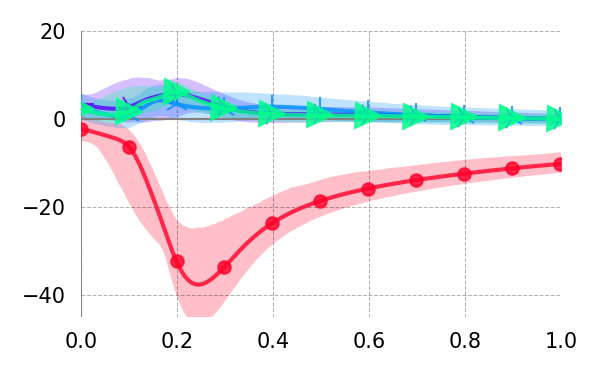}
    \end{subfigure}
    \begin{subfigure}[t]{0.19\textwidth}
        \centering
        \makebox[0.9\textwidth]{\centering Rep 3}
        Expl LL: -1234128
        \includegraphics[width=\textwidth]{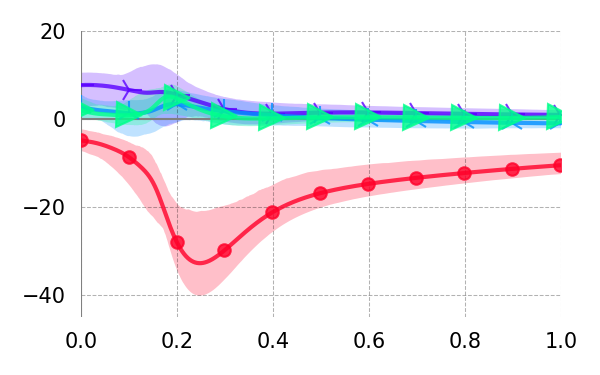}
    \end{subfigure}
    \begin{subfigure}[t]{0.19\textwidth}
        \centering
        \makebox[0.9\textwidth]{\centering Rep 4}
        Expl LL: -1211429
        \includegraphics[width=\textwidth]{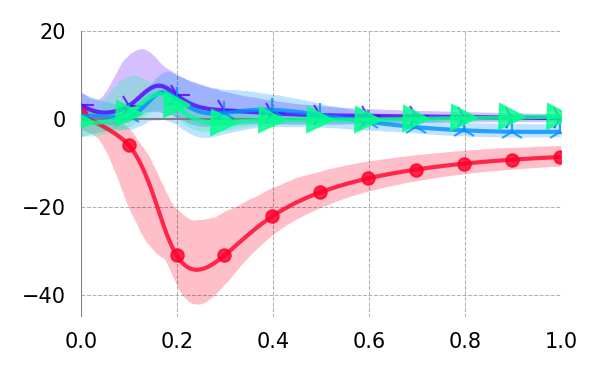}
    \end{subfigure}
    \begin{subfigure}[t]{0.19\textwidth}
        \centering
        \makebox[0.9\textwidth]{\centering Rep 5}
        Expl LL: -1235588
        \includegraphics[width=\textwidth]{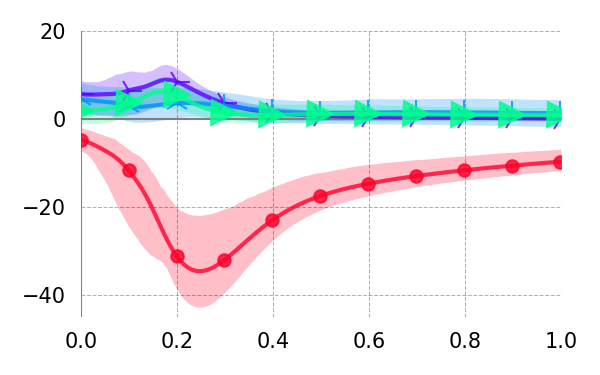}
    \end{subfigure}
    
    {\large Delay (s)}
    
    \vspace{0.5em}
    
    {
    {\kone~rate \hspace{1em}}
    {\kseven~word~length \hspace{1em}}
    {\knine~unigram~surprisal \hspace{1em}}
    {\keleven~5-gram~surprisal}
    }
    
    \vspace{0.5em}
    
    \caption{\textbf{Natural Stories (reading time):} Univariate CDRNN IRF estimates from the Natural Stories self-paced reading corpus. Results using base hyperparameters are compared to estimates from models that deviate from the base in some dimension. Plots under ``Consistency'' show estimates from five replicates of the ``base'' configuration, where ``Rep 1'' is the same model as ``base'' above, replotted for ease of comparison.}
    \label{fig:app-natstor-raw}
    
\end{figure}

\begin{figure}

    \footnotesize
    \sffamily
    \centering
    
    \textbf{\Large Natural Stories (Log Self-Paced Reading Time)}
    
    \vspace{1em}

    \begin{subfigure}[t]{0.49\textwidth}
        \centering
        \makebox[0.49\textwidth]{\centering Base}%
        \makebox[0.49\textwidth]{\centering + RNN}
        \begin{overpic}[width=0.49\textwidth]{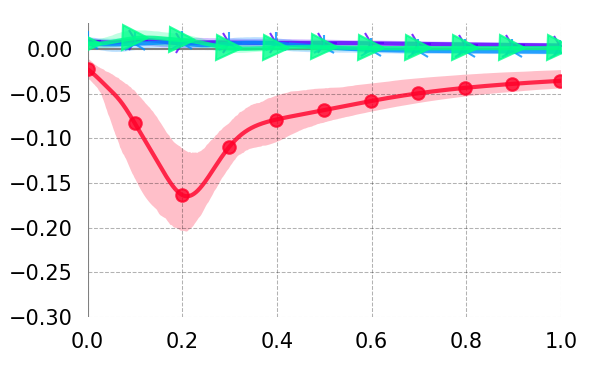}
            \put (-15,-80) {\rotatebox[origin=c]{90}{\large Change in Reading Time (log-ms)}}
        \end{overpic}%
        \includegraphics[width=0.49\textwidth]{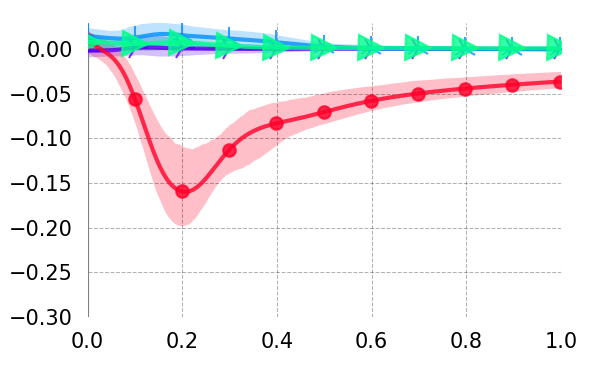}
    \end{subfigure}
    \begin{subfigure}[t]{0.49\textwidth}
        \centering
        \makebox[0.49\textwidth]{\centering Hidden Units $\div$ 2 (16)}%
        \makebox[0.49\textwidth]{\centering Hidden Units $\times$ 2 (64)}
        \includegraphics[width=0.49\textwidth]{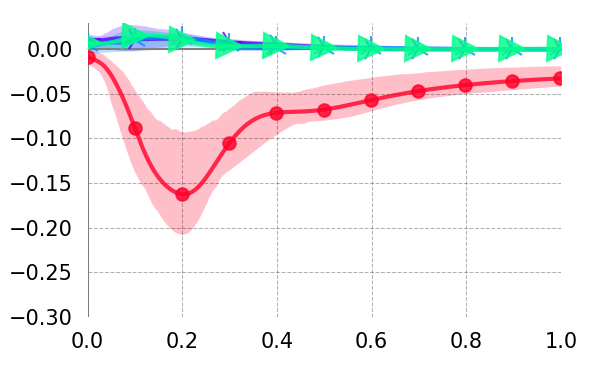}
        \includegraphics[width=0.49\textwidth]{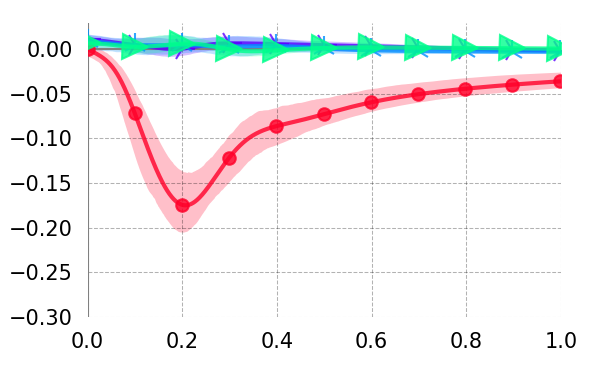}
    \end{subfigure}
    
    \begin{subfigure}[t]{0.49\textwidth}
        \centering
        \makebox[0.49\textwidth]{\centering Hidden Layers - 1 (1)}%
        \makebox[0.49\textwidth]{\centering Hidden Layers + 1 (3)}
        \includegraphics[width=0.49\textwidth]{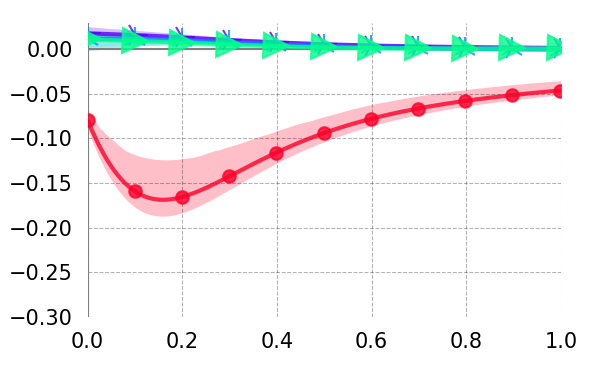}
        \includegraphics[width=0.49\textwidth]{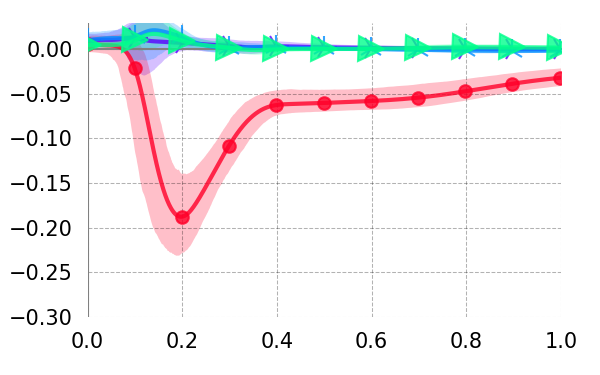}
    \end{subfigure}
    \begin{subfigure}[t]{0.49\textwidth}
        \centering
        \makebox[0.49\textwidth]{\centering Weight Reg $\div$ 5 (1)}%
        \makebox[0.49\textwidth]{\centering Weight Reg $\times$ 5 (5)}
        \includegraphics[width=0.49\textwidth]{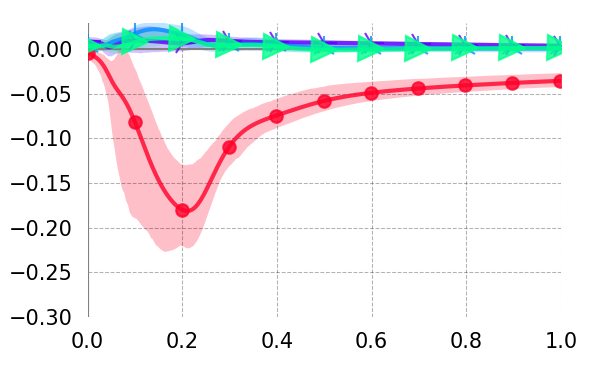}
        \includegraphics[width=0.49\textwidth]{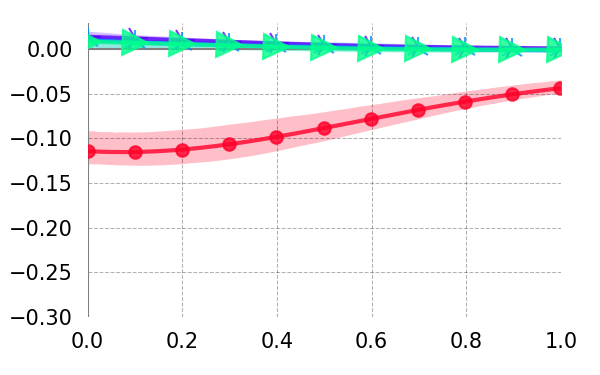}
    \end{subfigure}
    
    \begin{subfigure}[t]{0.49\textwidth}
        \centering
        \makebox[0.49\textwidth]{\centering Ranef Reg $\div$ 10 (1)}%
        \makebox[0.49\textwidth]{\centering Ranef Reg $\times$ 10 (100)}
        \includegraphics[width=0.49\textwidth]{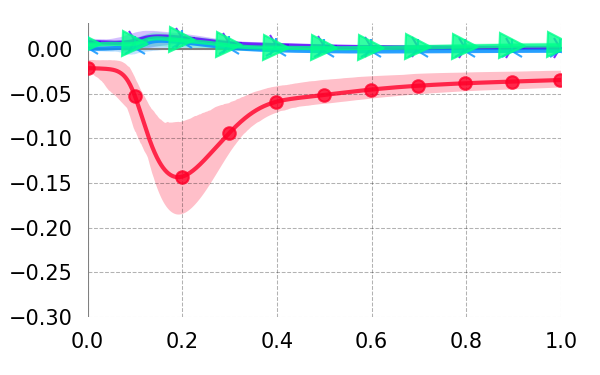}
        \includegraphics[width=0.49\textwidth]{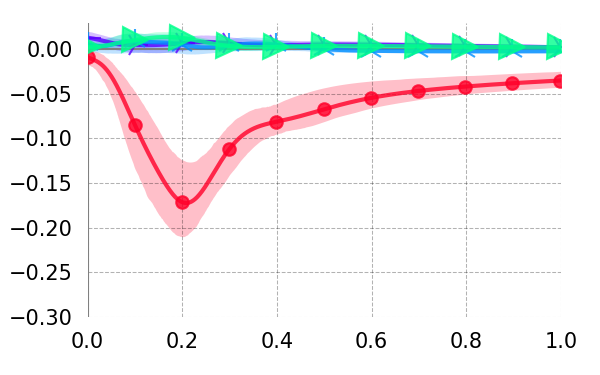}
    \end{subfigure}
    \begin{subfigure}[t]{0.49\textwidth}
        \centering
        \makebox[0.49\textwidth]{\centering Dropout $\div$ 2 (0.05)}%
        \makebox[0.49\textwidth]{\centering Dropout $\times$ 2 (0.2)}
        \includegraphics[width=0.49\textwidth]{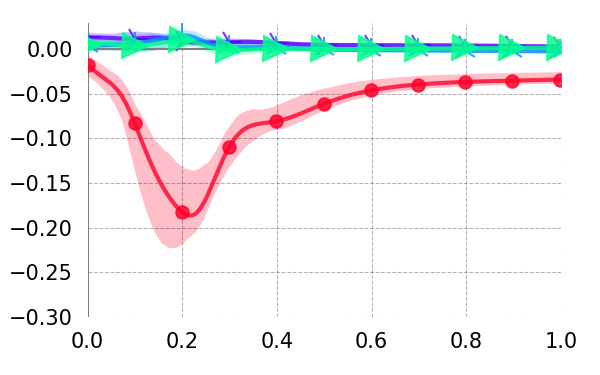}
        \includegraphics[width=0.49\textwidth]{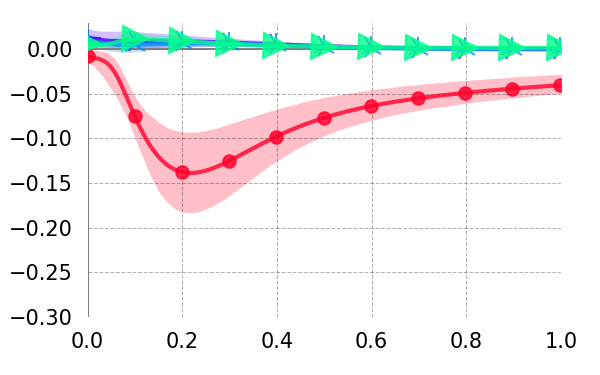}
    \end{subfigure}
    
    \begin{subfigure}[t]{0.49\textwidth}
        \centering
        \makebox[0.49\textwidth]{\centering Learning Rate $\div$ 3 (0.001)}%
        \makebox[0.49\textwidth]{\centering Learning Rate $\times$ 3 (0.009)}
        \includegraphics[width=0.49\textwidth]{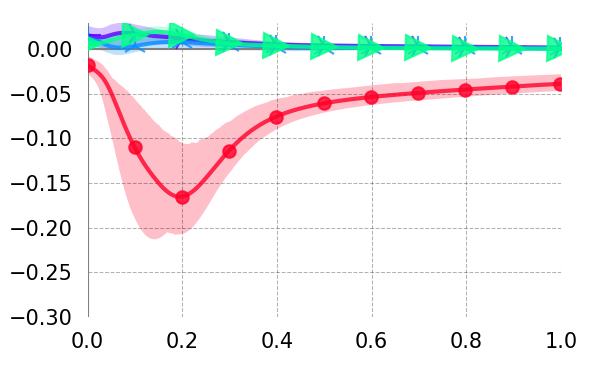}
        \includegraphics[width=0.49\textwidth]{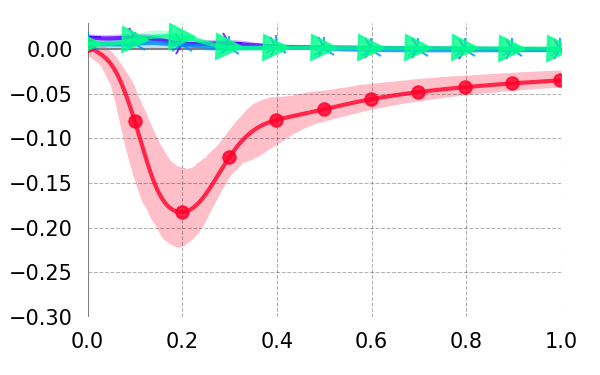}
    \end{subfigure}
    \begin{subfigure}[t]{0.49\textwidth}
        \centering
        \makebox[0.49\textwidth]{\centering Batch Size $\div$ 2 (512)}%
        \makebox[0.49\textwidth]{\centering Batch Size $\times$ 2 (2048)}
        \includegraphics[width=0.49\textwidth]{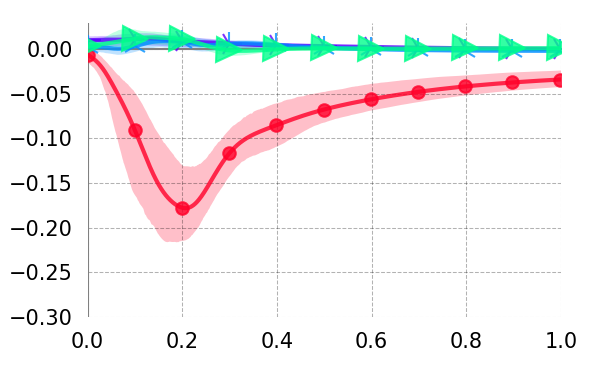}
        \includegraphics[width=0.49\textwidth]{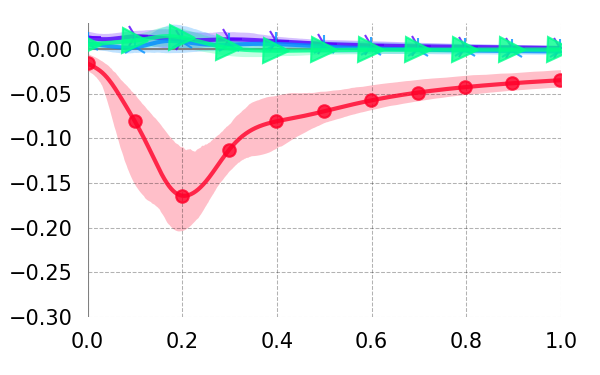}
    \end{subfigure}
    
    \vspace{1em}
    
    \textbf{\large Consistency}
    
    Standard deviation of exploratory set log-likelihood: 992
    
    \vspace{0.5em}
    
    \begin{subfigure}[t]{0.19\textwidth}
        \centering
        \makebox[0.9\textwidth]{\centering Rep 1}
        Expl LL: 10986
        \includegraphics[width=\textwidth]{{results_cdrnn_journal_natstor_log_CDR_main_irf_univariate_log.fdur._mean_mc}.png}
    \end{subfigure}
    \begin{subfigure}[t]{0.19\textwidth}
        \centering
        \makebox[0.9\textwidth]{\centering Rep 2}
        Expl LL: 12370
        \includegraphics[width=\textwidth]{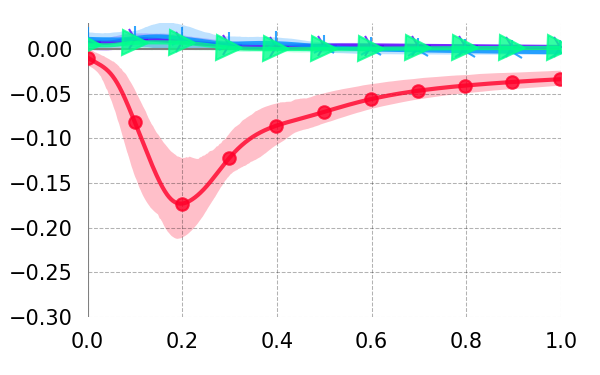}
    \end{subfigure}
    \begin{subfigure}[t]{0.19\textwidth}
        \centering
        \makebox[0.9\textwidth]{\centering Rep 3}
        Expl LL: 12197
        \includegraphics[width=\textwidth]{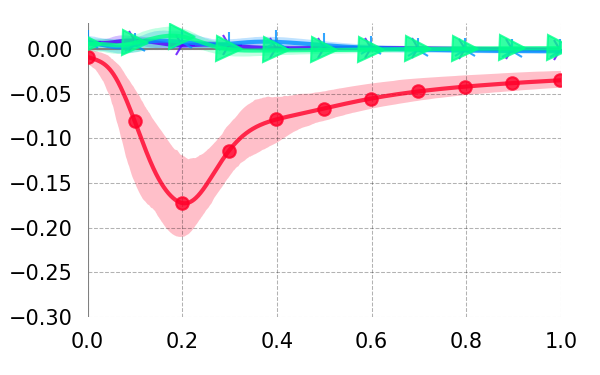}
    \end{subfigure}
    \begin{subfigure}[t]{0.19\textwidth}
        \centering
        \makebox[0.9\textwidth]{\centering Rep 4}
        Expl LL: 10323
        \includegraphics[width=\textwidth]{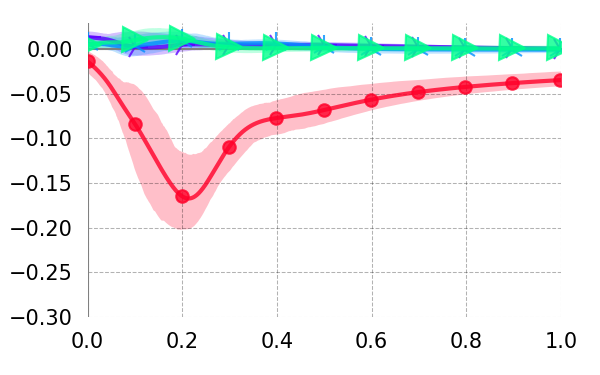}
    \end{subfigure}
    \begin{subfigure}[t]{0.19\textwidth}
        \centering
        \makebox[0.9\textwidth]{\centering Rep 5}
        Expl LL: 10323
        \includegraphics[width=\textwidth]{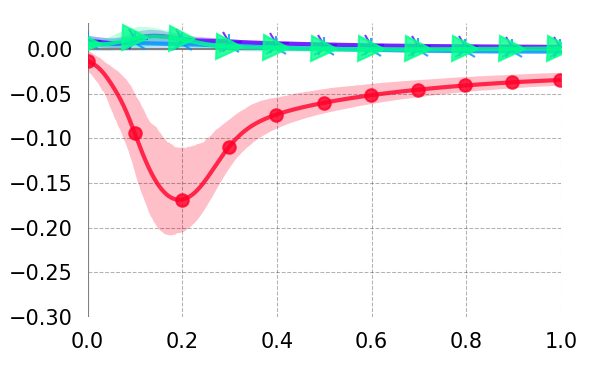}
    \end{subfigure}
    
    {\large Delay (s)}
    
    \vspace{0.5em}
    
    {
    {\kone~rate \hspace{1em}}
    {\kseven~word~length \hspace{1em}}
    {\knine~unigram~surprisal \hspace{1em}}
    {\keleven~5-gram~surprisal}
    }
    
    \vspace{0.5em}
    
    \caption{\textbf{Natural Stories (log reading time):} Univariate CDRNN IRF estimates from the Natural Stories self-paced reading corpus (log transformed). Results using base hyperparameters are compared to estimates from models that deviate from the base in some dimension. Plots under ``Consistency'' show estimates from five replicates of the ``base'' configuration, where ``Rep 1'' is the same model as ``base'' above, replotted for ease of comparison.}
    \label{fig:app-natstor-log}
    
\end{figure}

\FloatBarrier


\subsection{Natural Stories (fMRI)}
\label{app:fmri}

In-sample and out-of-sample predictive performance on the Natural Stories fMRI dataset is reported in \textbf{Supplementary Table~\ref{tab:app-fmri-ll-bakeoff}}.
Effect estimates in Natural Stories fMRI are plotted in \textbf{Supplementary Figure~\ref{fig:app-fmri}}.
Estimates are plausible and broadly similar across model variants, although they vary in their details substantially more than the estimates for the reading datasets, likely due to higher levels of intrinsic noise in this domain.

\begin{table}
    \footnotesize
    \sffamily
    \centering
    \begin{tabular}{r|ccc}
        & \multicolumn{3}{|c}{Natural Stories fMRI BOLD} \\
        Model & Train & Expl & Test\\
        \hline
        Canonical HRF (LME) & -264358\textsuperscript{\textdagger} & -137291\textsuperscript{\textdagger} & -132788\textsuperscript{\textdagger} \\
        Interpolated (LME) & -265012\textsuperscript{\textdagger} & -137644\textsuperscript{\textdagger} & -133003\textsuperscript{\textdagger} \\
        Averaged (LME) & -264489\textsuperscript{\textdagger} & -137513\textsuperscript{\textdagger} & -132881\textsuperscript{\textdagger} \\
        Lanczos (LME) & -264581\textsuperscript{\textdagger} & -137465\textsuperscript{\textdagger} & -132834\textsuperscript{\textdagger} \\
        GAM & -263474 & -137209 & -132495\\
        GAMLSS & \textit{-253869} & \textit{-131987} & \textit{-127839}\\
        CDR & -263889 & -137012 & -132730\\
        \hline
        CDRNN base & -250013 & -130579 & \textbf{-126590}\\
        --Nonlinear & -249531 & -130140 & ---\\
        --Nonstationary & -248835 & \textit{\textbf{-129948}} & ---\\
        --Heteroscedastic & -261503 & -136442 & ---\\
        +RNN & \textit{\textbf{-237930}} & -160340 & ---\\
        Units $\div$ 2 (16) & -250772 & -130873 & ---\\
        Units $\times$ 2 (64) & -247813 & -129789 & ---\\
        Layers - 1 (1) & -252764 & -131610 & ---\\
        Layers + 1 (3) & -249815 & -130801 & ---\\
        Weight Reg $\div$ 5 (1) & -249345 & -130486 & ---\\
        Weight Reg $\times$ 5 (25) & -253495 & -131802 & ---\\
        Ranef Reg $\times$ 10 (100) & -244538 & -130107 & ---\\
        Ranef Reg $\times$ 10 (100)00 & -252097 & -131434 & ---\\
        Dropout $\div$ 2 (0.05) & -248638 & -130129 & ---\\
        Dropout $\times$ 2 (0.2) & -253291 & -131724 & ---\\
        Learning Rate $\div$ 3 (0.001) & -249397 & -130629 & ---\\
        Learning Rate $\times$ 3 (0.009) & -249432 & -130120 & ---\\
        Batch Size $\div$ 2 (512) & -248414 & -129862 & ---\\
        Batch Size $\times$ 2 (2048) & -251670 & -131464 & ---\\
    \end{tabular}
    \caption{\textbf{Natural Stories (fMRI BOLD response in language-selective regions).} Log likelihood from CDRNN vs.\ linear mixed-effects (LME) fitted to fMRI data treated with different established preprocessing techniques, including pre-convolution with the canonical HRF, linear interpolation, averaging predictor values between acquisition times, and Lanczos interpolation, as well as generalized additive models (GAMs) and generalized additive models for location, scale, and shape (GAMLSS) fitted to data pre-convolved with the canonical HRF (LME and CDR performance as reported in \textit{\citen{shainschuler19}}).
    LME baselines show the marginal likelihood for the training set (the default likelihood implemented by the \texttt{lme4} package).
    All other likelihoods are conditional on the fitted model.
    CDRNN variants add recurrence (+RNN) and modify the number of Units (Units), number of hidden layers (Layers), weight regularization strength (Reg), random effects regularization strength (RanReg), dropout rate (Dropout), learning rate (LR), and batch size (Batch).
    Of the CDRNN models, only CDRNN base is evaluated on the test set.
    Best-performing models within the sets of baseline and CDRNN models are shown in \textit{italics}. Best-performing overall models are shown in \textbf{bold}. Daggers (\textdagger) indicate convergence failures.}
    \label{tab:app-fmri-ll-bakeoff}
\end{table}
\begin{figure}

    \footnotesize
    \sffamily
    \centering
    
    \textbf{\Large Natural Stories (fMRI)}
    
    \vspace{1em}
    
    \begin{subfigure}[t]{0.49\textwidth}
        \centering
        \makebox[0.49\textwidth]{\centering Base}%
        \makebox[0.49\textwidth]{\centering + RNN}
        \begin{overpic}[width=0.49\textwidth]{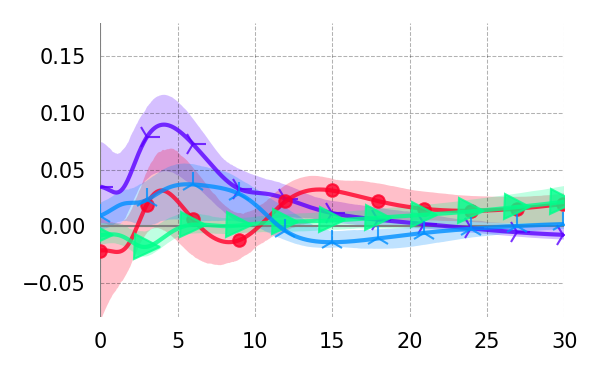}
            \put (-15,-80) {\rotatebox[origin=c]{90}{\large Change in BOLD}}
        \end{overpic}%
        \includegraphics[width=0.49\textwidth]{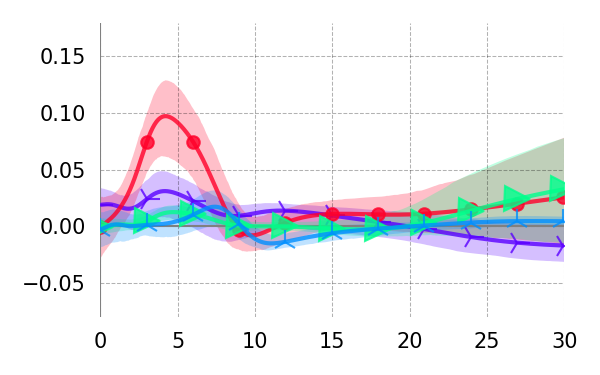}
    \end{subfigure}
    \begin{subfigure}[t]{0.49\textwidth}
        \centering
        \makebox[0.49\textwidth]{\centering Hidden Units $\div$ 2 (16)}%
        \makebox[0.49\textwidth]{\centering Hidden Units $\times$ 2 (64)}
        \includegraphics[width=0.49\textwidth]{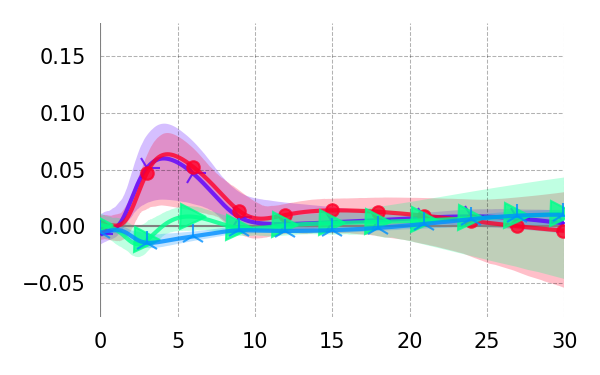}
        \includegraphics[width=0.49\textwidth]{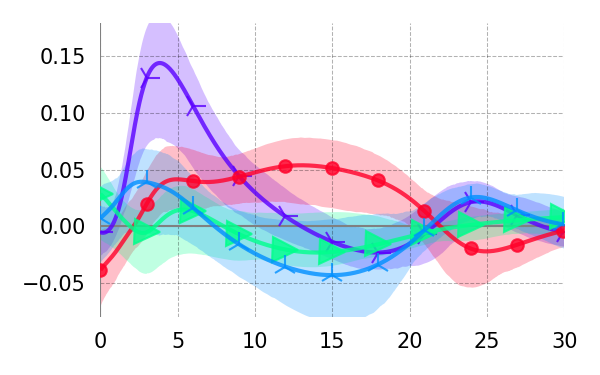}
    \end{subfigure}
    
    \begin{subfigure}[t]{0.49\textwidth}
        \centering
        \makebox[0.49\textwidth]{\centering Hidden Layers - 1 (1)}%
        \makebox[0.49\textwidth]{\centering Hidden Layers + 1 (3)}
        \includegraphics[width=0.49\textwidth]{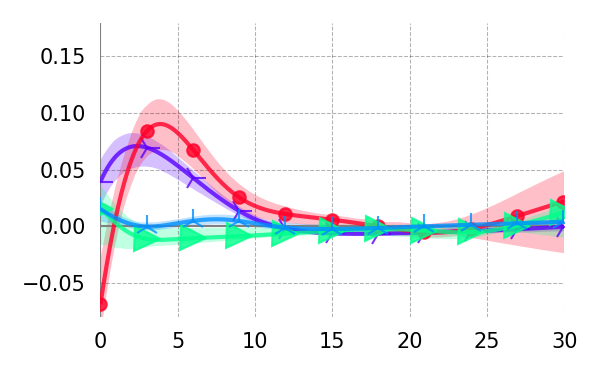}
        \includegraphics[width=0.49\textwidth]{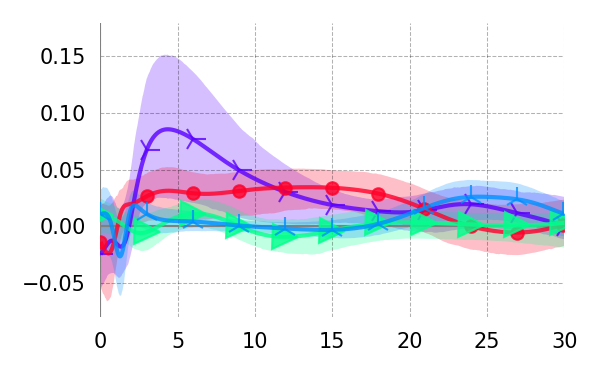}
    \end{subfigure}
    \begin{subfigure}[t]{0.49\textwidth}
        \centering
        \makebox[0.49\textwidth]{\centering Weight Reg $\div$ 5 (1)}%
        \makebox[0.49\textwidth]{\centering Weight Reg $\times$ 5 (5)}
        \includegraphics[width=0.49\textwidth]{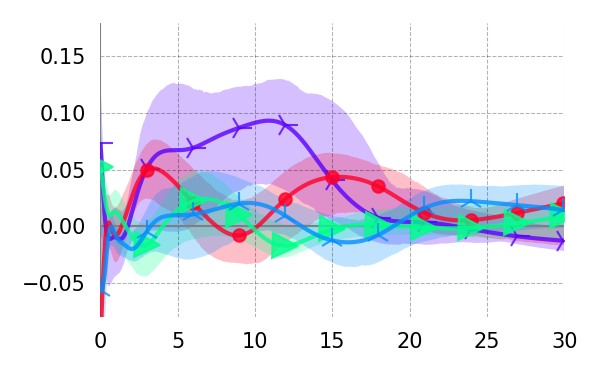}
        \includegraphics[width=0.49\textwidth]{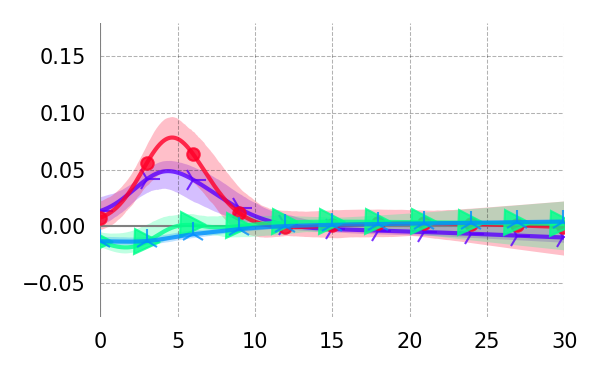}
    \end{subfigure}
    
    \begin{subfigure}[t]{0.49\textwidth}
        \centering
        \makebox[0.49\textwidth]{\centering Ranef Reg $\div$ 10 (100)}%
        \makebox[0.49\textwidth]{\centering Ranef Reg $\times$ 10 (10000)}
        \includegraphics[width=0.49\textwidth]{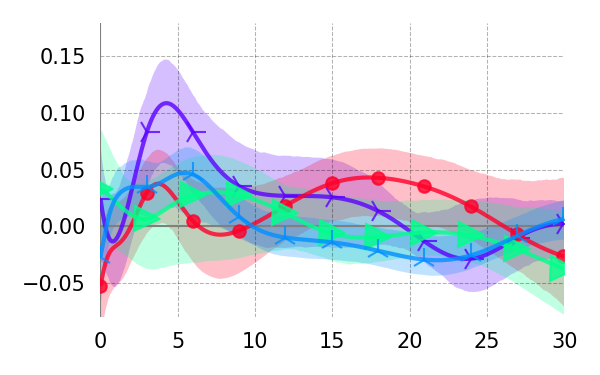}
        \includegraphics[width=0.49\textwidth]{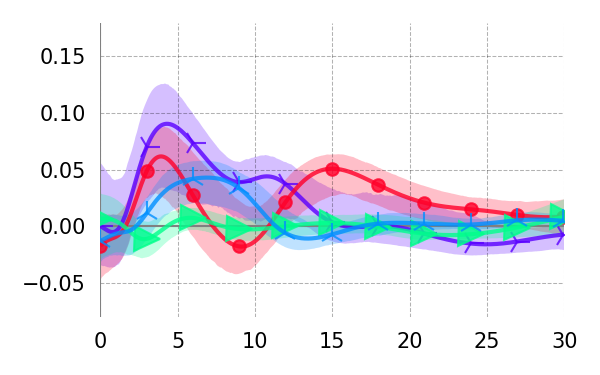}
    \end{subfigure}
    \begin{subfigure}[t]{0.49\textwidth}
        \centering
        \makebox[0.49\textwidth]{\centering Dropout $\div$ 2 (0.05)}%
        \makebox[0.49\textwidth]{\centering Dropout $\times$ 2 (0.2)}
        \includegraphics[width=0.49\textwidth]{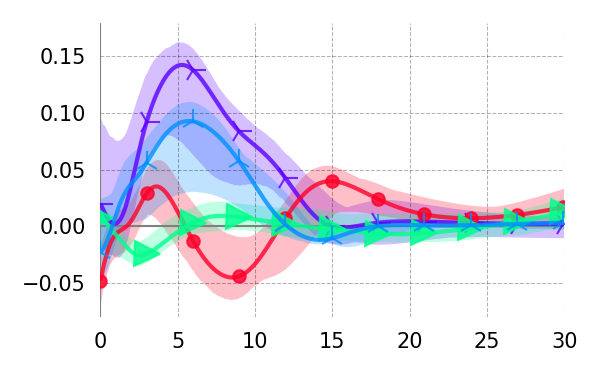}
        \includegraphics[width=0.49\textwidth]{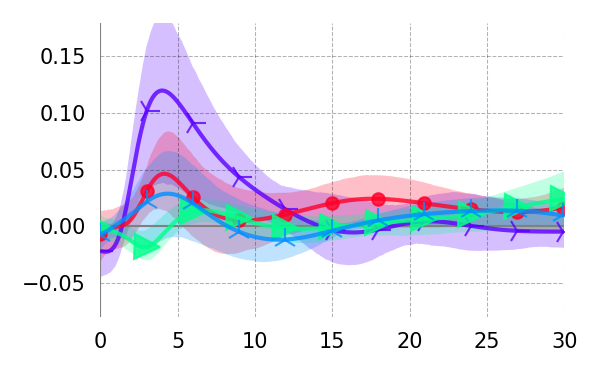}
    \end{subfigure}
    
    \begin{subfigure}[t]{0.49\textwidth}
        \centering
        \makebox[0.49\textwidth]{\centering Learning Rate $\div$ 3 (0.001)}%
        \makebox[0.49\textwidth]{\centering Learning Rate $\times$ 3 (0.009)}
        \includegraphics[width=0.49\textwidth]{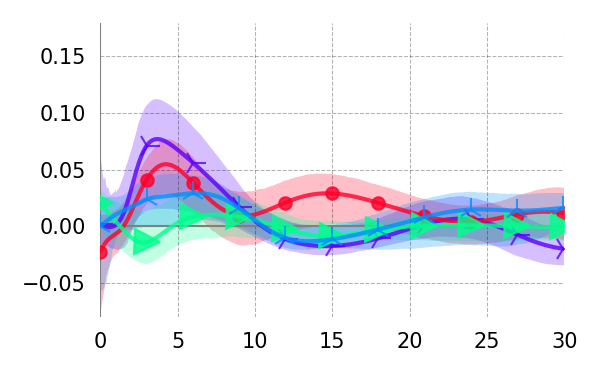}
        \includegraphics[width=0.49\textwidth]{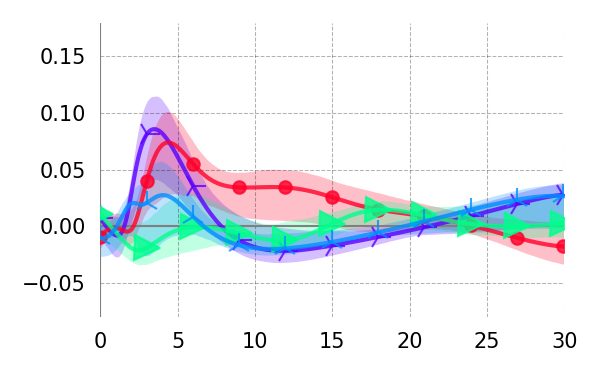}
    \end{subfigure}
    \begin{subfigure}[t]{0.49\textwidth}
        \centering
        \makebox[0.49\textwidth]{\centering Batch Size $\div$ 2 (512)}%
        \makebox[0.49\textwidth]{\centering Batch Size $\times$ 2 (2048)}
        \includegraphics[width=0.49\textwidth]{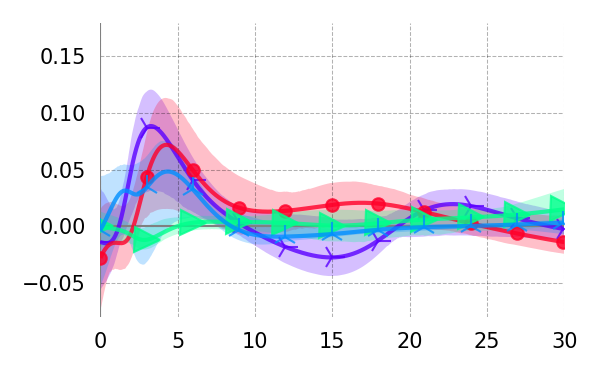}
        \includegraphics[width=0.49\textwidth]{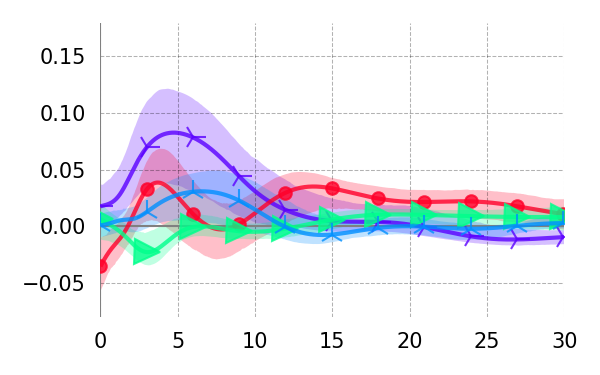}
    \end{subfigure}
    
    \vspace{1em}    
    
    \textbf{\large Consistency}
    
    Standard deviation of exploratory set log-likelihood: 327
    
    \vspace{0.5em}
    
    \begin{subfigure}[t]{0.19\textwidth}
        \centering
        \makebox[0.9\textwidth]{\centering Rep 1}
        Expl LL: -130579
        \includegraphics[width=\textwidth]{{results_cdrnn_journal_fmri_CDR_main_irf_univariate_BOLD_mean_mc}.png}
    \end{subfigure}
    \begin{subfigure}[t]{0.19\textwidth}
        \centering
        \makebox[0.9\textwidth]{\centering Rep 2}
        Expl LL: -129823
        \includegraphics[width=\textwidth]{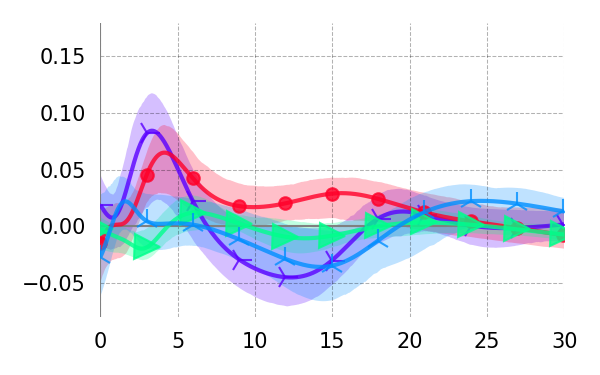}
    \end{subfigure}
    \begin{subfigure}[t]{0.19\textwidth}
        \centering
        \makebox[0.9\textwidth]{\centering Rep 3}
        Expl LL: -129813
        \includegraphics[width=\textwidth]{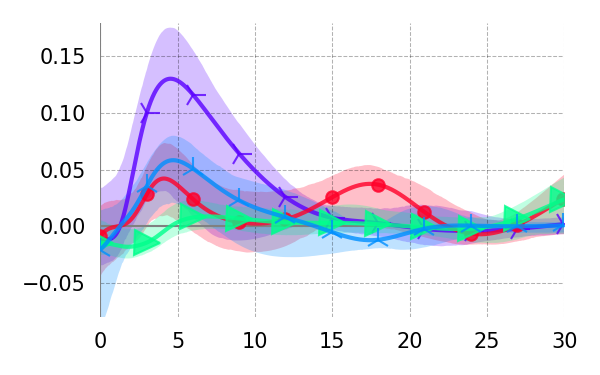}
    \end{subfigure}
    \begin{subfigure}[t]{0.19\textwidth}
        \centering
        \makebox[0.9\textwidth]{\centering Rep 4}
        Expl LL: -129902
        \includegraphics[width=\textwidth]{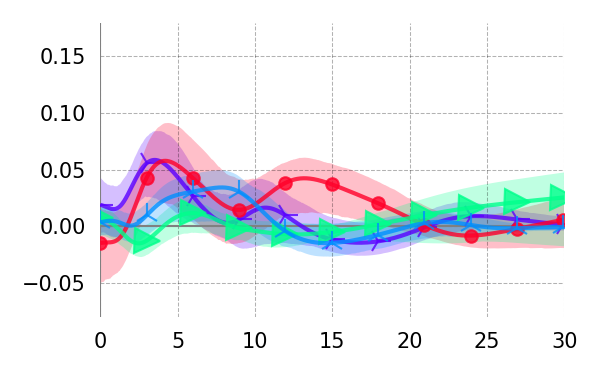}
    \end{subfigure}
    \begin{subfigure}[t]{0.19\textwidth}
        \centering
        \makebox[0.9\textwidth]{\centering Rep 5}
        Expl LL: -130428
        \includegraphics[width=\textwidth]{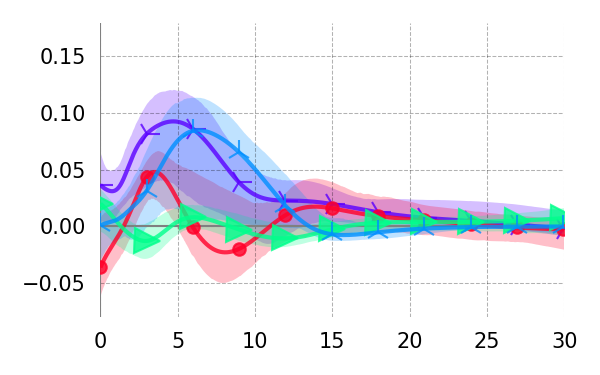}
    \end{subfigure}
    
    {\large Delay (s)}
    
    \vspace{0.5em}
    
    {
    {\kone~rate \hspace{1em}}
    {\kseven~word~length \hspace{1em}}
    {\knine~unigram~surprisal \hspace{1em}}
    {\keleven~5-gram~surprisal}
    }
    
    \vspace{0.5em}
    
    \caption{\textbf{Natural Stories (fMRI):} Univariate CDRNN IRF estimates from the language-selective brain regions in the Natural Stories fMRI dataset. Results using base hyperparameters are compared to estimates from models that deviate from the base in some dimension. Plots under ``Consistency'' show estimates from five replicates of the ``base'' configuration, where ``Rep 1'' is the same model as ``base'' above, replotted for ease of comparison.}
    \label{fig:app-fmri}
    
\end{figure}

\FloatBarrier


\section{Full Results: Consistency}

The full results for the model consistency analysis (variability of predictive performance across replicates of the CDRNN base model definition) are presented in Table~\ref{tab:app-consistency}.

\begin{table}
    \footnotesize
    \sffamily
    \centering
    \begin{tabular}{r|cccc|cccc}
         & \multicolumn{8}{|c}{Loglik}\\
         & \multicolumn{4}{|c}{Train} & \multicolumn{4}{|c}{Expl}\\
        Dataset & Median & Min & Max & Spread & Median & Min & Max & Spread\\
        \hline
        Synth E0 & -47624 & -47671 & -47532 & 139 & -48869 & -49000 & -48792 & 208\\
        Synth E1 & -47609 & -47673 & -47470 & 203 & -48867 & -48963 & -48791 & 172\\
        Synth E10 & -48076 & -48133 & -48023 & 110 & -49385 & -49512 & -49311 & 201\\
        Synth E100 & -59908 & -59968 & -59750 & 218 & -61233 & -61333 & -61148 & 185\\
        Synth FSS & -52739 & -52751 & -52561 & 189 & -53254 & -53424 & -53231 & 194\\
        Synth FSL & -45813 & -45888 & -45720 & 168 & -48691 & -48830 & -48615 & 215\\
        Synth RSS & -52422 & -52602 & -52285 & 317 & -54500 & -54717 & -54337 & 380\\
        Synth RSL & -46954 & -46991 & -46707 & 285 & -49574 & -49817 & -49415 & 402\\
        Synth RAS & -51955 & -51979 & -51846 & 133 & -53033 & -53085 & -52937 & 148\\
        Synth RAL & -43984 & -44023 & -43821 & 202 & -46732 & -46855 & -46657 & 198\\
        Synth R0.00 & -50621 & -50695 & -50540 & 155 & -52434 & -52640 & -52396 & 243\\
        Synth R0.25 & -50100 & -50192 & -49999 & 192 & -51456 & -51647 & -51403 & 244\\
        Synth R0.50 & -49548 & -49597 & -49470 & 126 & -50183 & -50214 & -50110 & 104\\
        Synth R0.75 & -48820 & -48969 & -48741 & 228 & -49447 & -49461 & -49352 & 109\\
        Synth R0.90 & -48613 & -48640 & -48463 & 178 & -49033 & -49085 & -48949 & 136\\
        Synth R0.95 & -48487 & -48492 & -48422 & 71 & -49230 & -49248 & -49221 & 27\\
        Synth Exp & -56491 & -56553 & -56414 & 138 & -57847 & -57962 & -57742 & 220\\
        Synth Normal & -53332 & -53374 & -53160 & 214 & -55208 & -55366 & -55090 & 277\\
        Synth Gamma & -53004 & -53147 & -52939 & 2084 & -54443 & -54532 & -54313 & 219\\
        \hline
        NatStor (RT) & -2257085 & -2260266 & -2254447 & 5819 & -1215293 & -1228864 & -1201714 & 27150\\
        NatStor (log RT) & 44814 & 42910 & 46898 & 3988 & 11662 & 11021 & 13561 & 2540\\
        Dundee (FP) & -564252 & -565470 & -563804 & 1666 & -288257 & -288433 & -288071 & 362\\
        Dundee (log FP) & -39696 & -39740 & -39438 & 302 & -20735 & -20750 & -20709 & 41\\
        Dundee (GP) & -602841 & -603463 & -601880 & 1583 & -307875 & -307936 & -307280 & 656\\
        Dundee (log GP) & -53825 & -53889 & -53628 & 261 & -27556 & -27586 & -27474 & 112\\
        Dundee (SP) & -756298 & -756572 & -755303 & 1269 & -386310 & -386896 & -385588 & 1307\\
        Dundee (log SP) & -61205 & -61381 & -61038 & 343 & -31834 & -31914 & -31740 & 174\\
        fMRI & -248687 & -249617 & -247269 & 2348 & -130094 & -130470 & -129692 & 777\\
    \end{tabular}
    \caption{\textbf{Consistency of Performance Across Replicates.} Training and exploratory set log likelihoods across 5 replicates of the base CDRNN model for each response variable in the main analyses. The log likelihood statistic is high variance across replicates of the same model (typically spread over hundreds or even thousands of likelihood points), leading to the possibility that statistical differences in performance between models may be driven by optimization noise. For this reason, it is recommended (a) to statistically compare ensembles of models and (b) to use early stopping on a validation set to prevent overfitting.}
    \label{tab:app-consistency}
\end{table}

\FloatBarrier

\section{LME, GAM, GAMLSS, and CDR Baseline Model Specifications}
\label{app:si-formulae}

For ease of reference, here we present the model formulae used to define the LME (\texttt{lme4} package), GAM (\texttt{mgcv} package), and GAMLSS (\texttt{gamlss} package) baseline models in R as well as the non-neural CDR (\texttt{cdr} package) baseline models in Python.
For full software implementation details, see the public code repository at \url{https://github.com/coryshain/cdr}.
Variable names below are modified from those used in our codebase for readability.
Variable names suffixed with S$n$ (e.g., S2) are ``lagged'', representing that variable's value from $n$ timesteps into the past.
In all GAMLSS models, the same formula was used simultaneously for the $\mu$ and $\sigma$ parameters.

{\raggedright 
\subsection{Dundee (Scan Path)}

\subsubsection{LME (0 lags)}

y $\sim$ Trial + SentencePosition + NotARegression + SaccadeLength + SaccadeLengthInRegression + PreviousWasFixated + PreviousWasFixatedInRegression + WordLength + WordLengthInRegression + UnigramSurprisal + UnigramSurprisalInRegression + 5GramSurprisal + 5GramSurprisalInRegression + (1 + Trial + SentencePosition + NotARegression + SaccadeLength + SaccadeLengthInRegression + PreviousWasFixated + PreviousWasFixatedInRegression + WordLength + WordLengthInRegression + UnigramSurprisal + UnigramSurprisalInRegression + 5GramSurprisal + 5GramSurprisalInRegression $\mid$ Participant)

\subsubsection{LME (3 lags)}

y $\sim$ Trial + SentencePosition + NotARegression + NotARegressionS1 + NotARegressionS2 + NotARegressionS3 + SaccadeLength + SaccadeLengthInRegression + SaccadeLengthS1 + SaccadeLengthInRegressionS1 + SaccadeLengthS2 + SaccadeLengthInRegressionS2 + SaccadeLengthS3 + SaccadeLengthInRegressionS3 + PreviousWasFixated + PreviousWasFixatedInRegression + PreviousWasFixatedS1 + PreviousWasFixatedInRegressionS1 + PreviousWasFixatedS2 + PreviousWasFixatedInRegressionS2 + PreviousWasFixatedS3 + PreviousWasFixatedInRegressionS3 + WordLength + WordLengthInRegression + WordLengthS1 + WordLengthInRegressionS1 + WordLengthS2 + WordLengthInRegressionS2 + WordLengthS3 + WordLengthInRegressionS3 + UnigramSurprisal + UnigramSurprisalInRegression + UnigramSurprisalS1 + UnigramSurprisalInRegressionS1 + UnigramSurprisalS2 + UnigramSurprisalInRegressionS2 + UnigramSurprisalS3 + UnigramSurprisalInRegressionS3 + 5GramSurprisal + 5GramSurprisalInRegression + 5GramSurprisalS1 + 5GramSurprisalInRegressionS1 + 5GramSurprisalS2 + 5GramSurprisalInRegressionS2 + 5GramSurprisalS3 + 5GramSurprisalInRegressionS3 + (1 + Trial + SentencePosition + NotARegression + NotARegressionS1 + NotARegressionS2 + NotARegressionS3 + SaccadeLength + SaccadeLengthInRegression + SaccadeLengthS1 + SaccadeLengthInRegressionS1 + SaccadeLengthS2 + SaccadeLengthInRegressionS2 + SaccadeLengthS3 + SaccadeLengthInRegressionS3 + PreviousWasFixated + PreviousWasFixatedInRegression + PreviousWasFixatedS1 + PreviousWasFixatedInRegressionS1 + PreviousWasFixatedS2 + PreviousWasFixatedInRegressionS2 + PreviousWasFixatedS3 + PreviousWasFixatedInRegressionS3 + WordLength + WordLengthInRegression + WordLengthS1 + WordLengthInRegressionS1 + WordLengthS2 + WordLengthInRegressionS2 + WordLengthS3 + WordLengthInRegressionS3 + UnigramSurprisal + UnigramSurprisalInRegression + UnigramSurprisalS1 + UnigramSurprisalInRegressionS1 + UnigramSurprisalS2 + UnigramSurprisalInRegressionS2 + UnigramSurprisalS3 + UnigramSurprisalInRegressionS3 + 5GramSurprisal + 5GramSurprisalInRegression + 5GramSurprisalS1 + 5GramSurprisalInRegressionS1 + 5GramSurprisalS2 + 5GramSurprisalInRegressionS2 + 5GramSurprisalS3 + 5GramSurprisalInRegressionS3 $\mid$ Participant)

\subsubsection{GAM (0 lags)}

y $\sim$ s(Trial) + s(SentencePosition) + NotARegression + s(SaccadeLength, k=4) + s(SaccadeLengthInRegression, k=4) + PreviousWasFixated + PreviousWasFixatedInRegression + s(WordLength) + s(WordLengthInRegression) + s(UnigramSurprisal) + s(UnigramSurprisalInRegression) + s(5GramSurprisal) + s(5GramSurprisalInRegression) + s(Participant, bs=``re'') + s(Trial, Participant, bs=``re'') + s(SentencePosition, Participant, bs=``re'') + s(NotARegression, Participant, bs=``re'') + s(SaccadeLength, Participant, bs=``re'') + s(SaccadeLengthInRegression, Participant, bs=``re'') + s(PreviousWasFixated, Participant, bs=``re'') + s(PreviousWasFixatedInRegression, Participant, bs=``re'') + s(WordLength, Participant, bs=``re'') + s(WordLengthInRegression, Participant, bs=``re'') + s(UnigramSurprisal, Participant, bs=``re'') + s(UnigramSurprisalInRegression, Participant, bs=``re'') + s(5GramSurprisal, Participant, bs=``re'') + s(5GramSurprisalInRegression, Participant, bs=``re'')

\subsubsection{GAM (3 lags)}

y $\sim$ s(Trial, bs=`fs') + s(SentencePosition, bs=`fs') + NotARegression + NotARegressionS1 + NotARegressionS2 + NotARegressionS3 + s(SaccadeLength, k=4) +  s(SaccadeLengthInRegression, k=4) + s(SaccadeLengthS1, k=4) + s(SaccadeLengthInRegressionS1, k=4) + s(SaccadeLengthS2, k=4) + s(SaccadeLengthInRegressionS2, k=4) + s(SaccadeLengthS3, k=4) + s(SaccadeLengthInRegressionS3, k=4) + PreviousWasFixated + PreviousWasFixatedInRegression + PreviousWasFixatedS1 + PreviousWasFixatedInRegressionS1 + PreviousWasFixatedS2 + PreviousWasFixatedInRegressionS2 + PreviousWasFixatedS3 + PreviousWasFixatedInRegressionS3 + s(WordLength) + s(WordLengthInRegression) + s(WordLengthS1) + s(WordLengthInRegressionS1) + s(WordLengthS2) + s(WordLengthInRegressionS2) + s(WordLengthS3) + s(WordLengthInRegressionS3) + s(UnigramSurprisal) + s(UnigramSurprisalInRegression) + s(UnigramSurprisalS1) + s(UnigramSurprisalInRegressionS1) + s(UnigramSurprisalS2) + s(UnigramSurprisalInRegressionS2) + s(UnigramSurprisalS3) + s(UnigramSurprisalInRegressionS3) + s(5GramSurprisal) + s(5GramSurprisalInRegression) + s(5GramSurprisalS1) + s(5GramSurprisalInRegressionS1) + s(5GramSurprisalS2) + s(5GramSurprisalInRegressionS2) + s(5GramSurprisalS3) + s(5GramSurprisalInRegressionS3) + s(Participant, bs=``re'') + s(Trial, Participant, bs=``re'') + s(SentencePosition, Participant, bs=``re'') + s(NotARegression, Participant, bs=``re'') + s(NotARegressionS1, Participant, bs=``re'') + s(NotARegressionS2, Participant, bs=``re'') + s(NotARegressionS3, Participant, bs=``re'') + s(SaccadeLength, Participant, bs=``re'') + s(SaccadeLengthInRegression, Participant, bs=``re'') + s(SaccadeLengthS1, Participant, bs=``re'') + s(SaccadeLengthInRegressionS1, Participant, bs=``re'') + s(SaccadeLengthS2, Participant, bs=``re'') + s(SaccadeLengthInRegressionS2, Participant, bs=``re'') + s(SaccadeLengthS3, Participant, bs=``re'') + s(SaccadeLengthInRegressionS3, Participant, bs=``re'') + s(PreviousWasFixated, Participant, bs=``re'') + s(PreviousWasFixatedInRegression, Participant, bs=``re'') + s(PreviousWasFixatedS1, Participant, bs=``re'') + s(PreviousWasFixatedInRegressionS1, Participant, bs=``re'') + s(PreviousWasFixatedS2, Participant, bs=``re'') + s(PreviousWasFixatedInRegressionS2, Participant, bs=``re'') + s(PreviousWasFixatedS3, Participant, bs=``re'') + s(PreviousWasFixatedInRegressionS3, Participant, bs=``re'') + s(WordLength, Participant, bs=``re'') + s(WordLengthInRegression, Participant, bs=``re'') + s(WordLengthS1, Participant, bs=``re'') + s(WordLengthInRegressionS1, Participant, bs=``re'') + s(WordLengthS2, Participant, bs=``re'') + s(WordLengthInRegressionS2, Participant, bs=``re'') + s(WordLengthS3, Participant, bs=``re'') + s(WordLengthInRegressionS3, Participant, bs=``re'') + s(UnigramSurprisal, Participant, bs=``re'') + s(UnigramSurprisalInRegression, Participant, bs=``re'') + s(UnigramSurprisalS1, Participant, bs=``re'') + s(UnigramSurprisalInRegressionS1, Participant, bs=``re'') + s(UnigramSurprisalS2, Participant, bs=``re'') + s(UnigramSurprisalInRegressionS2, Participant, bs=``re'') + s(UnigramSurprisalS3, Participant, bs=``re'') + s(UnigramSurprisalInRegressionS3, Participant, bs=``re'') + s(5GramSurprisal, Participant, bs=``re'') + s(5GramSurprisalInRegression, Participant, bs=``re'') + s(5GramSurprisalS1, Participant, bs=``re'') + s(5GramSurprisalInRegressionS1, Participant, bs=``re'') + s(5GramSurprisalS2, Participant, bs=``re'') + s(5GramSurprisalInRegressionS2, Participant, bs=``re'') + s(5GramSurprisalS3, Participant, bs=``re'') + s(5GramSurprisalInRegressionS3, Participant, bs=``re'')

\subsubsection{GAMLSS (0 lags)}

y $\sim$ pb(Trial) + pb(SentencePosition) + NotARegression + SaccadeLength + SaccadeLengthInRegression + PreviousWasFixated + PreviousWasFixatedInRegression + pb(WordLength) + pb(WordLengthInRegression) + pb(UnigramSurprisal) + pb(UnigramSurprisalInRegression) + pb(5GramSurprisal) + pb(5GramSurprisalInRegression) + random(Participant)

\subsubsection{GAMLSS (3 lags)}

y $\sim$ pb(Trial) + pb(SentencePosition) + NotARegression + NotARegressionS1 + NotARegressionS2 + NotARegressionS3 + SaccadeLength +  SaccadeLengthInRegression + SaccadeLengthS1 + SaccadeLengthInRegressionS1 + SaccadeLengthS2 + SaccadeLengthInRegressionS2 + SaccadeLengthS3 + SaccadeLengthInRegressionS3 + PreviousWasFixated + PreviousWasFixatedInRegression + PreviousWasFixatedS1 + PreviousWasFixatedInRegressionS1 + PreviousWasFixatedS2 + PreviousWasFixatedInRegressionS2 + PreviousWasFixatedS3 + PreviousWasFixatedInRegressionS3 + pb(WordLength) + pb(WordLengthInRegression) + pb(WordLengthS1) + pb(WordLengthInRegressionS1) + pb(WordLengthS2) + pb(WordLengthInRegressionS2) + pb(WordLengthS3) + pb(WordLengthInRegressionS3) + pb(UnigramSurprisal) + pb(UnigramSurprisalInRegression) + pb(UnigramSurprisalS1) + pb(UnigramSurprisalInRegressionS1) + pb(UnigramSurprisalS2) + pb(UnigramSurprisalInRegressionS2) + pb(UnigramSurprisalS3) + pb(UnigramSurprisalInRegressionS3) + pb(5GramSurprisal) + pb(5GramSurprisalInRegression) + pb(5GramSurprisalS1) + pb(5GramSurprisalInRegressionS1) + pb(5GramSurprisalS2) + pb(5GramSurprisalInRegressionS2) + pb(5GramSurprisalS3) + pb(5GramSurprisalInRegressionS3) + random(Participant)

\subsubsection{CDR}

y $\sim$ Trial + SentencePosition + C(Rate + NotARegression + SaccadeLength + SaccadeLengthInRegression + PreviousWasFixated + PreviousWasFixatedInRegression + WordLength + WordLengthInRegression + UnigramSurprisal + UnigramSurprisalInRegression + 5GramSurprisal + 5GramSurprisalInRegression, ShiftedGammaShapeGT1(alpha=2, beta=2, delta=-0.5)) + (Trial + SentencePosition + C(Rate + NotARegression + SaccadeLength + SaccadeLengthInRegression + PreviousWasFixated + PreviousWasFixatedInRegression + WordLength + WordLengthInRegression + UnigramSurprisal + UnigramSurprisalInRegression + 5GramSurprisal + 5GramSurprisalInRegression, ShiftedGammaShapeGT1(alpha=2, beta=2, delta=-0.5, ran=T)) $\mid$ Participant)

\subsection{Dundee (First Pass and Go-Past)}

\subsubsection{LME (0 lags)}

y $\sim$ Trial + SentencePosition + SaccadeLength + PreviousWasFixated + WordLength + UnigramSurprisal + 5GramSurprisal + (1 + Trial + SentencePosition + SaccadeLength + PreviousWasFixated + WordLength + UnigramSurprisal + 5GramSurprisal $\mid$ Participant)

\subsubsection{LME (3 lags)}

y $\sim$ Trial + SentencePosition + SaccadeLength + SaccadeLengthS1 + SaccadeLengthS2 + SaccadeLengthS3 + PreviousWasFixated + PreviousWasFixatedS1 + PreviousWasFixatedS2 + PreviousWasFixatedS3 + WordLength + WordLengthS1 + WordLengthS2 + WordLengthS3 + UnigramSurprisal + UnigramSurprisalS1 + UnigramSurprisalS2 + UnigramSurprisalS3 + 5GramSurprisal + 5GramSurprisalS1 + 5GramSurprisalS2 + 5GramSurprisalS3 + (1 + Trial + SentencePosition + SaccadeLength + SaccadeLengthS1 + SaccadeLengthS2 + SaccadeLengthS3 + PreviousWasFixated + PreviousWasFixatedS1 + PreviousWasFixatedS2 + PreviousWasFixatedS3 + WordLength + WordLengthS1 + WordLengthS2 + WordLengthS3 + UnigramSurprisal + UnigramSurprisalS1 + UnigramSurprisalS2 + UnigramSurprisalS3 + 5GramSurprisal + 5GramSurprisalS1 + 5GramSurprisalS2 + 5GramSurprisalS3 $\mid$ Participant)

\subsubsection{GAM (0 lags)}

y $\sim$ s(Trial) + s(SentencePosition) + s(SaccadeLength, k=4) + PreviousWasFixated + s(WordLength) + s(UnigramSurprisal) + s(5GramSurprisal) + s(Participant, bs=``re'') + s(Trial, Participant, bs=``re'') + s(SentencePosition, Participant, bs=``re'') + s(SaccadeLength, Participant, bs=``re'') + s(PreviousWasFixated, Participant, bs=``re'') + s(WordLength, Participant, bs=``re'') + s(UnigramSurprisal, Participant, bs=``re'') + s(5GramSurprisal, Participant, bs=``re'')

\subsubsection{GAM (3 lags)}

y $\sim$ s(Trial) + s(SentencePosition) + s(SaccadeLength, k=4) + s(SaccadeLengthS1, k=4) + s(SaccadeLengthS2, k=4) + s(SaccadeLengthS3, k=4) + PreviousWasFixated + PreviousWasFixatedS1 + PreviousWasFixatedS2 + PreviousWasFixatedS3 + s(WordLength) + s(WordLengthS1) + s(WordLengthS2) + s(WordLengthS3) + s(UnigramSurprisal) + s(UnigramSurprisalS1) + s(UnigramSurprisalS2) + s(UnigramSurprisalS3) + s(5GramSurprisal) + s(5GramSurprisalS1) + s(5GramSurprisalS2) + s(5GramSurprisalS3) + s(Participant, bs=``re'') + s(Trial, Participant, bs=``re'') + s(SentencePosition, Participant, bs=``re'') + s(SaccadeLength, Participant, bs=``re'') + s(SaccadeLengthS1, Participant, bs=``re'') + s(SaccadeLengthS2, Participant, bs=``re'') + s(SaccadeLengthS3, Participant, bs=``re'') + s(PreviousWasFixated, Participant, bs=``re'') + s(PreviousWasFixatedS1, Participant, bs=``re'') + s(PreviousWasFixatedS2, Participant, bs=``re'') + s(PreviousWasFixatedS3, Participant, bs=``re'') + s(WordLength, Participant, bs=``re'') + s(WordLengthS1, Participant, bs=``re'') + s(WordLengthS2, Participant, bs=``re'') + s(WordLengthS3, Participant, bs=``re'') + s(UnigramSurprisal, Participant, bs=``re'') + s(UnigramSurprisalS1, Participant, bs=``re'') + s(UnigramSurprisalS2, Participant, bs=``re'') + s(UnigramSurprisalS3, Participant, bs=``re'') + s(5GramSurprisal, Participant, bs=``re'') + s(5GramSurprisalS1, Participant, bs=``re'') + s(5GramSurprisalS2, Participant, bs=``re'') + s(5GramSurprisalS3, Participant, bs=``re'')

\subsubsection{GAMLSS (0 lags)}

y $\sim$ pb(Trial) + pb(SentencePosition) + SaccadeLength + PreviousWasFixated + pb(WordLength) + pb(UnigramSurprisal) + pb(5GramSurprisal) + random(Participant)

\subsubsection{GAMLSS (3 lags)}

y $\sim$ pb(Trial) + pb(SentencePosition) + SaccadeLength + SaccadeLengthS1 + SaccadeLengthS2 + SaccadeLengthS3 + PreviousWasFixated + PreviousWasFixatedS1 + PreviousWasFixatedS2 + PreviousWasFixatedS3 + pb(WordLength) + pb(WordLengthS1) + pb(WordLengthS2) + pb(WordLengthS3) + pb(UnigramSurprisal) + pb(UnigramSurprisalS1) + pb(UnigramSurprisalS2) + pb(UnigramSurprisalS3) + pb(5GramSurprisal) + pb(5GramSurprisalS1) + pb(5GramSurprisalS2) + pb(5GramSurprisalS3) + random(Participant)

\subsubsection{CDR}

y $\sim$ Trial + SentencePosition + C(Rate + SaccadeLength + PreviousWasFixated + WordLength + UnigramSurprisal + 5GramSurprisal, ShiftedGammaShapeGT1(alpha=2, beta=2, delta=-0.5)) + (Trial + SentencePosition + C(Rate + SaccadeLength + PreviousWasFixated + WordLength + UnigramSurprisal + 5GramSurprisal, ShiftedGammaShapeGT1(ran=T)) $\mid$ Participant)

\subsection{Natural Stories Self-Paced Reading (Reading Time)}

\subsubsection{LME (0 lags)}

y $\sim$ Trial + SentencePosition + WordLength + UnigramSurprisal + 5GramSurprisal + (1 + Trial + SentencePosition + WordLength + UnigramSurprisal + 5GramSurprisal $\mid$ Participant)

\subsubsection{LME (3 lags)}

y $\sim$ Trial + SentencePosition + WordLength + WordLengthS1 + WordLengthS2 + WordLengthS3 + + UnigramSurprisal + UnigramSurprisalS1 + UnigramSurprisalS2 + UnigramSurprisalS3 + 5GramSurprisal + 5GramSurprisalS1 + 5GramSurprisalS2 + 5GramSurprisalS3 + (1 + Trial + SentencePosition + WordLength + WordLengthS1 + WordLengthS2 + WordLengthS3 + UnigramSurprisal + UnigramSurprisalS1 + UnigramSurprisalS2 + UnigramSurprisalS3 + 5GramSurprisal + 5GramSurprisalS1 + 5GramSurprisalS2 + 5GramSurprisalS3 $\mid$ Participant)

\subsubsection{GAM (0 lags)}

y $\sim$ s(Trial) + s(SentencePosition) + s(WordLength) + s(UnigramSurprisal) + s(5GramSurprisal) + s(Participant, bs=``re'') + s(Trial, Participant, bs=``re'') + s(SentencePosition, Participant, bs=``re'') + s(WordLength, Participant, bs=``re'') + s(UnigramSurprisal, Participant, bs=``re'') + s(5GramSurprisal, Participant, bs=``re'')

\subsubsection{GAM (3 lags)}

y $\sim$ s(Trial) + s(Trial) + s(SentencePosition) + s(WordLength) + s(WordLengthS1) + s(WordLengthS2) + s(WordLengthS3) + s(UnigramSurprisal) + s(UnigramSurprisalS1) + s(UnigramSurprisalS2) + s(UnigramSurprisalS3) + s(5GramSurprisal) + s(5GramSurprisalS1) + s(5GramSurprisalS2) + s(5GramSurprisalS3) + s(Participant, bs=``re'') + s(Trial, Participant, bs=``re'') + s(SentencePosition, Participant, bs=``re'') + s(WordLength, Participant, bs=``re'') + s(WordLengthS1, Participant, bs=``re'') + s(WordLengthS2, Participant, bs=``re'') + s(WordLengthS3, Participant, bs=``re'') + s(UnigramSurprisal, Participant, bs=``re'') + s(UnigramSurprisalS1, Participant, bs=``re'') + s(UnigramSurprisalS2, Participant, bs=``re'') + s(UnigramSurprisalS3, Participant, bs=``re'') + s(5GramSurprisal, Participant, bs=``re'') + s(5GramSurprisalS1, Participant, bs=``re'') + s(5GramSurprisalS2, Participant, bs=``re'') + s(5GramSurprisalS3, Participant, bs=``re'')

\subsubsection{GAMLSS (0 lags)}

y $\sim$ pb(Trial) + pb(SentencePosition) + pb(WordLength) + pb(UnigramSurprisal) + pb(5GramSurprisal) + re(random=~1$\mid$Participant) + random(Participant)

\subsubsection{GAMLSS (3 lags)}

y $\sim$ pb(Trial) + pb(SentencePosition) + pb(WordLength) + pb(WordLengthS1) + pb(WordLengthS2) + pb(WordLengthS3) + pb(UnigramSurprisal) + pb(UnigramSurprisalS1) + pb(UnigramSurprisalS2) + pb(UnigramSurprisalS3) + pb(5GramSurprisal) + pb(5GramSurprisalS1) + pb(5GramSurprisalS2) + pb(5GramSurprisalS3) + random(Participant)

\subsubsection{CDR}

y $\sim$ Trial + SentencePosition + C(Rate + WordLength + UnigramSurprisal + 5GramSurprisal, ShiftedGammaShapeGT1()) + (Trial + SentencePosition + C(Rate + WordLength + UnigramSurprisal + 5GramSurprisal, ShiftedGammaShapeGT1(ran=T)) $\mid$ Participant)

\subsection{Natural Stories fMRI (BOLD)}

\subsubsection{LME}

y $\sim$ TR + Rate + SoundPower + UnigramSurprisal + 5GramSurprisal + (TR + Rate + SoundPower + UnigramSurprisal + 5GramSurprisal $\mid$ | fROI) + (1 $\mid$ Participant)

\subsubsection{GAM}

y $\sim$ s(TR) + s(Rate) + s(SoundPower) + s(UnigramSurprisal) + s(5GramSurprisal) + s(Participant, bs=``re'') + s(fROI, bs=``re'') + s(TR, fROI, bs=``re'') + s(Rate, fROI, bs=``re'') + s(SoundPower, fROI, bs=``re'') + s(UnigramSurprisal, fROI, bs=``re'') + s(5GramSurprisal, fROI, bs=``re'')

\subsubsection{GAMLSS}

y $\sim$ pb(TR) + pb(Rate) + pb(SoundPower) + pb(UnigramSurprisal) + pb(5GramSurprisal) + random(Participant) + random(fROI)

\subsubsection{CDR}

y $\sim$ TR + C(Rate + SoundPower + UnigramSurprisal + 5GramSurprisal, HRFDoubleGamma5(irf\textunderscore id=HRF)) + (TR + C(Rate + SoundPower + UnigramSurprisal + 5GramSurprisal, HRFDoubleGamma5(irf\textunderscore id=HRF, ran=T)) $\mid$ fROI) + (1 $\mid$ Participant)

}

\end{appendices}

\end{document}